\documentclass{article} 
\usepackage{conference,times}

\usepackage[most]{tcolorbox}
\usepackage{amsmath}
\usepackage{amsfonts}
\usepackage{graphicx}
\usepackage{xcolor}
\usepackage{booktabs}
\usepackage{multirow}
\usepackage{amssymb}
\usepackage{float}

\usepackage{amsmath,amsfonts,bm}

\def\eqref#1{equation~\ref{#1}}

\def\1{\bm{1}}

\DeclareMathAlphabet{\mathsfit}{\encodingdefault}{\sfdefault}{m}{sl}
\SetMathAlphabet{\mathsfit}{bold}{\encodingdefault}{\sfdefault}{bx}{n}

\DeclareMathOperator*{\argmin}{arg\,min}

\usepackage[utf8]{inputenc} 
\usepackage[T1]{fontenc}    
\usepackage{hyperref}       
\usepackage{url}            
\usepackage{cleveref}
\usepackage{enumitem}
\usepackage{tcolorbox}
\usepackage{mathtools}
\usepackage{wrapfig}
\usepackage{graphicx}
\usepackage{caption}
\usepackage[toc, title]{appendix}
\usepackage{minitoc}
\usepackage{xcolor}
\usepackage{subcaption}
\usepackage{graphicx}

\usepackage{multibib}
\newcites{main,app}{{},{}} 

\makeatletter
\DeclareRobustCommand\onedot{\futurelet\@let@token\@onedot}
\def\@onedot{\ifx\@let@token.\else.\null\fi}

\newcommand{\eg}{\emph{e.g\@\onedot}}

\newcommand{\Ie}{\emph{I.e\@\onedot}}
\newcommand{\ie}{\emph{i.e\@\onedot}}

\definecolor{cs1}{HTML}{c7e9b4}
\definecolor{cs2}{HTML}{7fcdbb}
\definecolor{cs3}{HTML}{41b6c4}
\definecolor{cs4}{HTML}{1d91c0}
\definecolor{cs5}{HTML}{225ea8}
\definecolor{cs6}{HTML}{253494}
\definecolor{cs7}{HTML}{081d58}
\definecolor{cd2}{HTML}{addd8e}
\definecolor{cd3}{HTML}{78c679}
\definecolor{cd4}{HTML}{41ab5d}
\definecolor{cd5}{HTML}{238443}
\definecolor{cd6}{HTML}{006837}
\definecolor{cd7}{HTML}{004529}
\definecolor{ForestGray}{HTML}{004529}
\definecolor{codegreen}{rgb}{0,0.6,0}
\definecolor{codegray}{rgb}{0.5,0.5,0.5}
\definecolor{codepurple}{rgb}{0.58,0,0.82}
\definecolor{backcolour}{rgb}{0.95,0.95,0.92}
\definecolor{jcred}{HTML}{e31a1c}
\definecolor{jcgreen}{HTML}{33a02c}
\definecolor{jcblue}{HTML}{1f78b4}
\definecolor{jcorange}{HTML}{ff7f00}
\definecolor{jcpurple}{HTML}{6a3d9a}
\definecolor{jclightred}{HTML}{fb8072}
\definecolor{jclightgreen}{HTML}{b3de69}
\definecolor{jclightblue}{HTML}{80b1d3}
\definecolor{jclightorange}{HTML}{fdb462}
\definecolor{jclightpurple}{HTML}{bebada}
\definecolor{jcredl}{HTML}{fb8072}
\definecolor{jcgreenl}{HTML}{b3de69}
\definecolor{jcbluel}{HTML}{80b1d3}
\definecolor{jcorangel}{HTML}{fdb462}
\definecolor{jcpurplel}{HTML}{bebada}
\definecolor{jcyellow}{HTML}{fdb462}
\definecolor{cadmiumorange}{rgb}{0.93, 0.53, 0.18}
\definecolor{emerald}{rgb}{0.31, 0.78, 0.47}
\definecolor{amaranth}{rgb}{0.9, 0.17, 0.31}
\definecolor{candypink}{rgb}{0.89, 0.44, 0.48}
\definecolor{caribbeangreen}{rgb}{0.0, 0.8, 0.6}
\definecolor{cornflowerblue}{rgb}{0.39, 0.58, 0.93}
\definecolor{limegreen}{rgb}{0.2, 0.8, 0.2}
\definecolor{mayablue}{rgb}{0.21,0.49,0.74}
\definecolor{lightblue}{RGB}{248,252,255}
\definecolor{framegray}{RGB}{60,76,80}

\newcommand{\introtakeawaybox}[2][]{%
  \begin{tcolorbox}[%
    colback=cornflowerblue!10,
    colframe=cornflowerblue!90,
    boxrule=0.5pt,
    arc=0mm,
    left=5pt,
    right=5pt,
    top=2pt,
    bottom=2pt,
    fontupper={\fontsize{9.5pt}{11.75pt}\selectfont}
  ]
    \IfNoValueTF{#1}{\textbf{Contributions:} #2}{\textbf{Contributions #1:} #2}%
  \end{tcolorbox}%
  \vspace*{-0.1cm}
}

\newcommand{\empiricaltakeawaybox}[2][]{%
  \begin{tcolorbox}[
    colback=cornflowerblue!10, %
    colframe=cornflowerblue!90,
    float=t,
    boxrule=0.5pt,
    arc=0mm,
    left=5pt,
    right=5pt,
    top=2pt,
    bottom=2pt,
    fontupper={\fontsize{9.5pt}{11.75pt}\selectfont}
  ]
    \IfNoValueTF{#1}{\textbf{Empirical Takeaway:} #2}{\textbf{Empirical Takeaways #1:} #2} %
  \end{tcolorbox}%
  \vspace*{-0.1cm}
}

\newcommand{\takeawaybox}[2][]{%
  \begin{tcolorbox}[
    colback=cornflowerblue!10, %
    colframe=cornflowerblue!90,
    boxrule=0.5pt,
    arc=0mm,
    left=5pt,
    right=5pt,
    top=2pt,
    bottom=2pt,
    fontupper={\fontsize{9.5pt}{11.75pt}\selectfont}
  ]
    \IfNoValueTF{#1}{\textbf{Takeaway:} #2}{\textbf{#1} #2} %
  \end{tcolorbox}%
  \vspace*{-0.1cm}
}

\newcommand{\observationbox}[2][]{%
  \begin{tcolorbox}[
    colback=emerald!10, %
    colframe=emerald!90,
    boxrule=0.5pt,
    arc=0mm,
    left=5pt,
    right=5pt,
    top=2pt,
    bottom=2pt,
    fontupper={\fontsize{9.5pt}{11.75pt}\selectfont}
  ]
    \IfNoValueTF{#1}{\textbf{Observation:} #2}{\textbf{#1} #2} %
  \end{tcolorbox}%
  \vspace*{-0.1cm}
}

\newtcolorbox{custombox}{
  colback=jcorange!10,
  colframe=jcorange!90,
  boxrule=0.8pt,
  arc=0mm,
  left=8pt,
  right=8pt,
  top=8pt,
  bottom=8pt
}

\hypersetup{
  linkcolor  = amaranth,
    filecolor=magenta,      
    urlcolor=teal,
    citecolor=mayablue, %
    colorlinks = true,
    breaklinks = true,
    bookmarksopen=true
}

\DeclarePairedDelimiter{\braces}{\{}{\}}

\usepackage{fancyhdr}
\title{Scaling Laws for Mixed Quantization}

\author{%
    Zeyu Cao\( ^{1}\)\thanks{Equal Contribution.} \quad 
    Boyang Gu \(^{2}\)\footnotemark[1] \quad 
    Cheng Zhang\( ^2 \) \quad 
    Pedro Gimenes\( ^2 \) \quad 
    Jianqiao Lu\( ^3 \) \\
    \textbf{Jianyi Cheng\( ^4 \) \quad 
    Xitong Gao\( ^{5,6} \) \quad 
    Yiren Zhao\( ^{2} \)}\thanks{Corresponding author.}
    \\
    \( ^1 \)Department of Computer Science and Technology, University of Cambridge \\
    \( ^2 \)Department of Electrical \& Electronic Engineering,
    Imperial College London \\
    \( ^3 \)Department of Computer Science, University of Hong Kong \\
    \( ^4 \)School of Informatics, University of Edinburgh
    \\
    \texttt{zeyu.cao@cl.cam.ac.uk}, \texttt{jqlu@cs.hku.hk}, \texttt{jianyi.cheng@ed.ac.uk} \\
    \texttt{\{boyang.gu19, cheng.zhang122, pedro.gimenes19, a.zhao\}@ic.ac.uk} \\
    \texttt{xt.gao@siat.ac.cn}
}

\iclrfinalcopy
\begin{document}

\maketitle

\begin{abstract}
Post-training quantization of Large Language Models (LLMs) has proven effective in reducing the memory and computational requirements for inference. In this study, we focus on a straightforward question: When aiming for a target accuracy or perplexity with low-precision quantization, how much high-precision computation needs to be preserved and how fine-grained this quantization would need to be as we scale LLMs to larger sizes? We first introduce two critical metrics named the quantization ratio ($Q_r$) and quantization block size ($Q_b$). The former measures the number of parameters quantized to low-precision arithmetic normalized by the total parameter count, whereas the latter defines the number of values within a block that share a scaling factor, akin to the block size concept introduced in the FP4 format in NVIDIA's Blackwell architecture. Through extensive and carefully controlled experiments across different model and quantization methods, we propose a unified scaling law on post-training quantization (PTQ) that can predict loss degeneration for varying $Q_r$ and $Q_b$. For $Q_r$, our scaling law implies that parameter scaling and ratio scaling have a multiplicative relationship. Consequently, larger models are more amenable to a higher quantization ratio $Q_r$, thus supporting an increase in the adoption of mixed quantization for inference. Regarding $Q_b$, our findings indicate that a small block size, similar to that used in Blackwell, is not essential for large models. Employing a small $Q_b$ can instead unnecessarily complicate the design of the hardware circuit.
\end{abstract}

\section{Introduction}
\label{sec:intro}
Large language models (LLMs) have demonstrated remarkable performance in a range of natural language processing (NLP) tasks \citep{brown2020language}, and state-of-the-art models now contain billions of parameters \citep{anthropic2025claude37card,openai2025gpt45card,deepmind2025gemini25}. As such, researchers have attempted to understand the scaling laws of LLMs by characterizing how the required number of training tokens scales with parameter count to train compute-optimal models under a fixed compute budget \citep{kaplan2020scaling,chincilla}. These works provide insight into how to best allocate resources in training increasingly large LLMs.

Despite these training scaling laws, the substantial size of LLMs and their computational demands require significant hardware resources. As such, quantization has become a promising solution to increase the compute and memory efficiency of LLM inference \citep{lin2024awq,chee2024quip,ashkboos2024quarot}. The recently proposed Scaling Laws for Precision by \citeauthor{kumar2024scaling} further examined the interplay among precision levels, model parameters, and data -- they suggested a revised version of the Chinchilla form \citep{kaplan2020scaling}, with an additional error term to capture the impacts of post-training quantization (PTQ).

Meanwhile, there is also an active research thread, which has shown that weights and activations in pre-trained transformer blocks often yield magnitude outliers \citep{gu2024attention}. This issue has been tackled by assigning higher precision to outliers while putting the remainder of the network at lower precision \citep{dettmers2022gpt3, zhang2023revisiting,dettmers2023spqr}. With the introduction of more compact arithmetic types in NVIDIA's Blackwell GPUs, including the support for 4-bit and 6-bit MXFP formats \citep{tirumala2024nvidia}. These arithmetic types, such as MX arithmetics, incorporate a block size $Q_b$, in which a set of values shares the same scaling factor. The first iteration of this concept, known as block floating point, dates back to the 1990s and was used in digital signal processing \citep{kobayashi1999new}. This can be seen as a granularity of quantization where, at $Q_b=1$, the arithmetic resembles the traditional floating-point number format, comprising both a mantissa and an exponent for scaling purposes. Subsequently, this concept of blockwise scaling factors was adopted in the domain of low bitwidth quantization for neural networks \citep{dai2021vs, lingle2023transformer, darvish2020pushing}, including the recently available MX-format on the Blackwell GPUs. These hardware advancements enable both high- and low-precision computations on silicon, making \textbf{mixed-precision inference} a compelling approach to maintain model performance while reducing compute and memory costs.

Driven by the significance of developing systematic scaling laws to direct future research in mixed quantization, we aim to address a largely under-explored question: How does the optimal ratio of low-precision elements and the optimal granularity in a mixed quantization mapping change as the model size enlarges? Put another way, \textit{what are the scaling laws that govern mixed quantization?}

For a pair of low- and high-precision parameters $(W_l, W_h)$, we define the mixed-quantization ratio $Q_r$ as the ratio of parameters using low-precision arithmetic to the total number of parameters, and consider the scenario where this allocation happens normally only at the post-training stage. We consider block size granularities as $Q_b$ for the block-based quantization methods. Then our search space over the respective random variable of low-precision and high-precision parameters $(\mathcal{W}_l, \mathcal{W}_h)$ for a given pair of $Q_r$ and $Q_b$ is as follows. 
\begin{equation}
    (\mathcal{W}_l, \mathcal{W}_h) \sim \braces*{(W_l, W_h): \frac{\lVert W_l \rVert_0}{\lVert W_h \rVert_0 + \lVert W_l \rVert_0} = Q_r \land \mathcal{Q}_b(W_l, W_h) = Q_b}\,,
\end{equation}
where $\lVert \cdot \rVert_0$ refers to the $l_0$ norm, and $\mathcal{Q}_b(W_l, W_h)$ is the function that calculates the block size used for the searched quantization. We experimented with 3 different LLM model families and 4 different post-training quantization methods, covering weight-only and weight-activation (W-A) quantizations. In total, we applied mixed quantization to 17 models, with model sizes ranging from 60M to 14B, resulting in a total of $54,600$ quantized model checkpoints.

For a language model with $N$ parameters, trained on $D$ tokens with a fixed quantization ratio $Q_r$ and fixed block size $Q_b$, the loss can be considered a discrete random variable $\mathcal{L}$. The variation in loss is due to the combinatorial space among all possible combinations of low- and high-precision parameter pairs for the given $Q_r$ and $Q_b$. Note that in our post-training quantization space, $D$ is fixed for the given model. We describe a unified scaling law that formulates the model inference loss as the sum of the usual Chinchilla form with a post-training quantization loss degeneration term ($\Delta$), which itself is also a discrete random variable defined similarly on $(\mathcal{W}_l, \mathcal{W}_h)$.
\begin{equation}
\label{eq:intro-law}
    \mathcal{L}(N, D, Q_r, Q_b) = \underbrace{\underbrace{aN^{-\alpha}}_{\text{Training-time Effects}} + b D^{-\beta} + 
E}_\text{Usual Chinchilla form} + \underbrace{\Delta(N, Q_b, Q_r)}_{\text{Loss Degeneration}}\,,
\end{equation}
where $a$, $b$, $\alpha$, $\beta$, and $E$ are constants. 

We claim that the optimal (minimum) and expectation of $\Delta$ are the overall effect of ratio, parameter, and granularity scaling effects. \Ie{},
\begin{equation}
\label{eq:full-unified-law}
    \delta^{\text{opt}}(N, Q_r, Q_b) = \min(\Delta(N, Q_r, Q_b)) =C \cdot 
    \underbrace{ e^{AQ_r}}_{\text{ratio scaling}}
    \cdot \underbrace{N^{-\gamma_N}}_{\text{parameter scaling}} \cdot \underbrace{(Q_b  + d)^{\gamma_c}}_{\text{granularity scaling}},
\end{equation}
\begin{equation}
    \mathbb{E}[\delta(N, Q_r, Q_b)] = C^\prime e^{A^\prime Q_r} N^{-\gamma_N^{\prime}}(Q_b  + d^\prime)^{\gamma_c^\prime}\,,
\end{equation}
respectively, where $A$, $A^\prime$, $C$, $C^\prime$, $\gamma_N$, $\gamma_N^{\prime}$, $d$, $d^\prime$, $\gamma_c$, and $\gamma_c^{\prime}$ are constant coefficients. This finding also suggests a strong link to the prior work conducted by \citeauthor{kumar2024scaling}. If $Q_r = 1$ and $Q_b$ are fixed as constants, \Cref{eq:full-unified-law} agrees asymptotically with the Precision Scaling Law proposed by~\citeauthor{kumar2024scaling} (\Cref{sec:laws}). We make the following contributions:

\begin{enumerate}[leftmargin=*]
    \item \textbf{Unified Scaling Law on PTQ with ratio and granularity scaling.} We propose a unified scaling law that considers both the mixed quantization ratio $Q_r$ and quantization block size granularity $Q_b$, and its expectation and minimum forms reach asymptotic agreement with existing Precision Scaling Laws shown by~\citeauthor{kumar2024scaling}. 
    Our scaling laws integrate model sizes, mixed quantization ratios, and quantization granularities, thereby extending the scope beyond existing scaling laws that focus solely on precision levels.
    
    \item \textbf{Ratio matters more.} Our scaling law indicates that, since $\mathcal{O} (e^{AQ_r})$ grows faster than $\mathcal{O}(N^{\gamma_N})$, the growth of $Q_r$ dominates the quantization loss. However, although $e^{AQ_r}$ growth faster than $N^{\gamma_N}$ asymptotically, $Q_r$ can only be $1$ at most while $N$ can grow without a limit, showing that larger models can accommodate a progressively larger quantization ratio. This effectively shows that mixed quantization is a promising future direction for further reducing model sizes and model computation complexity.
    
    \item \textbf{The quantization ``strength'' provides diminishing returns at large model sizes.} Since $\gamma_c$ is within $0$ and $1$ with our scaling law (See~\Cref{sec:laws}), lowering $Q_r$ is more effective than lowering $Q_b$ related terms, allowing better quantization performance. 
\end{enumerate}

\section{Scaling Laws for Mixed Quantization}
\label{sec:laws}
Our scaling law can be considered a further development of the previous Precision Scaling Law \citep{kumar2024scaling}. The usual Chinchilla form~\citep{chincilla} suggests that for a language model with a total number of parameters $N$, trained on a number of tokens $D$, its loss $L$ should scale as follows.
\begin{equation}
\label{eq:Chinchilla}
    L(N, D) = aN^{-\alpha} + bD^{-\beta} + E\,,
\end{equation}
where $a$, $b$, $\alpha$, $\beta$ and $E$ are positive constants. The Precision Scaling Law~\citep{kumar2024scaling} further investigates the case where post-training quantization is applied. In detail, following previous notations, if the model is trained and tested (inferenced) at different precisions $P_{\text{train}}$ and $P_{\text{post}}$, \Cref{eq:Chinchilla} becomes 
\begin{equation}
\label{eq:precision-scaling-law}
    L(N, D, P_{\text{train}}, P_{\text{post}}) = aN_{\text{eff}}^{-\alpha} + bD^{-\beta} + E + \delta_{\text{PTQ}}(N_{\text{eff}}, D, P_{\text{train}}, P_{\text{post}})\,,
\end{equation}
where $\delta$ is the post-training quantization loss degeneration. It has the following form:
\begin{equation}
\label{eq:original-ptq-loss}
    \delta_{\text{PTQ}}(N, D, P_{\text{train}}, P_{\text{post}}) = C_T e^{- P_{\text{post}} / \gamma_{\text{post}}} \left( \frac{D^{\gamma_D}}{N^{\gamma_N}} \right) \prod_{x \in \{\text{w,a,kv}\}} [ 1 - e^{-C_x (P_x - P_{\text{post}})} ]\,,
\end{equation}
where $P_{\text{w}}$, $P_{\text{a}}$, $P_{\text{kv}}$ are the weight, activation, and kv-cache precision bit-width at training, and $C_T$, $\gamma_{\text{post}}$, $\gamma_D$, $\gamma_N$, and $C_x$-s are constants. Such a scaling law for the loss degeneration was developed under the consideration that the quantization setting is applied to all weights. \Cref{eq:original-ptq-loss}, proposed by \citet{kumar2024scaling}, only allows the degree of freedom to quantize the combination of weight, activation, and KV-cache. However, different quantization configurations should also be considered for mixed quantization inference. To investigate this aspect, we introduce the quantization ratio $Q_r$ and the quantization block size $Q_b$. 

Consider a model $M$ with size $N$, with a block size $Q_b$, we define the block number as $N_b = N / Q_b$. As $M=\{m_i\}_{i=1}^N$, we define a partition of $M$ into $N_b$ blocks, \ie{} $\mathcal{B} = \{b_j = m_{(j-1)\cdot Q_b + 1:j\cdot Q_b+1}\}_{j=1}^{N_b}$. Given a codebook collection $\mathcal{C} = \{C_k\}_{k=1}^{N_c}$ of size $N_c$ where each codebook $C_k$ is a binary partition function over $\mathcal{B}$ to a pair of low- and high-precision components, \ie{}, $C_k(b_i) = (W_l^{(i,k)}, W_h^{(i,k)})$, we further define a mapping $f: \mathcal{B} \to \{j\}_{j=1}^{N_c}$ according to the quantization setting. Then we denote $\mathcal{Q}_b(W_l, W_h) = Q_b$ as saying that the block size is $Q_b$ for 
\begin{equation}
    (W_l, W_h) = \left(\bigcup_{j=1}^{N_b} C_{f(b_j)}^0(b_j), \bigcup_{j=1}^{N_b} C_{f(b_j)}^1(b_j)\right)\,.
\end{equation}

For a given quantization ratio ($Q_r$) and quantization block size ($Q_b$), there are different pairs of $(W_l, W_h)$. The set of all possible pairs forms the binary discrete random variable $(\mathcal{W}_l, \mathcal{W}_h)$. For each pair of $(W_l, W_h)$, we could calculate the corresponding quantized model loss and post-quantization loss degeneration, with a bit of abuse of notation, \Cref{eq:precision-scaling-law} turns into its random variable form
\begin{equation}
\label{eq:overall-random-variable-law}
    \mathcal{L}(N, D, Q_r, Q_b) = aN^{-\alpha} + b D^{-\beta} + E + \Delta(N, Q_b, Q_r)\,,
\end{equation}
where $\mathcal{L}$ and $\Delta$ are the random variable forms of $L$ and $\delta$ respectively.

We define the optimal allocation of low-precision parameters for a model under $Q_r$ and $Q_b$ as $(W_l^{\text{opt}}, W_h^{\text{opt}})$, where it minimizes the loss $L$. \Ie{}, 
\begin{equation}
\label{equ:opt}
    (W_l^{\text{opt}}, W_h^{\text{opt}}) = \argmin_{(W_l, W_h)} L(W_l, W_h) \quad\text{s.t.}\quad \frac{\lVert W_l \rVert_0}{\lVert W_h \rVert_0 + \lVert W_l \rVert_0} = Q_r \land \mathcal{Q}_b(W_l, W_h) = Q_b\,.
\end{equation}
It is easily seen that the pair $(W_l^{\text{opt}}, W_h^{\text{opt}})$ minimizes the loss degeneration $\delta$ too. \Cref{equ:opt} outlines the optimization problem used to evaluate the hypothesized scaling laws. Note that the number of candidate pairs $(W_l, W_h)$ explodes with the weight numbers. In this work, we find an approximate solution to the problem using a random search algorithm to allocate a numerical precision to each network component (\ie{}, layer or matrix multiply operation, according to the quantization method). We estimate the expectation via its unbiased estimator. For $n$ observed loss degeneration $(\delta_i)_{i=1}^n$, $\mathbb{E}(\delta)$ is approximated by
\begin{equation}
\label{eq:mean-estimator}
    \mu_{\delta} ((\delta_i)_{i=1}^n) = \frac{\sum_{i=1}^n \delta_i}{n}\,.
\end{equation}
With a larger number of random search trials, we get a more precise estimate of loss degeneration. However, we discovered that even a small amount of random sampling of $\Delta$ could reveal its distribution quite successfully. Detailed discussion of this phenomenon can be seen in Appendix~\ref{app:search-space-distribution}.

Recall \Cref{eq:overall-random-variable-law}, since $(W_l^{\text{opt}}, W_h^{\text{opt}})$ is obtained from a group of candidates, we rewrite \Cref{eq:overall-random-variable-law} to the form of its best realization 
\begin{align}
    L^{\text{opt}}(N, D, Q_r, Q_b) = aN^{-\alpha} + bD^{-\beta} + E + \delta^{\text{opt}}(N, Q_r, Q_b)\,, 
\label{eq:opt-scaling-law} \\
    \delta^{\text{opt}}(N, Q_r, Q_b) = \delta(W_l^{\text{opt}}, W_h^{\text{opt}})\,.
\end{align}
Similarly, one may be interested in the expectation loss degeneration $\mathbb{E}(\Delta)$ rather than the optimal value. The expectation counterpart of \Cref{eq:opt-scaling-law} is
\begin{equation}
    \mathbb{E}[\mathcal{L}(N, D, Q_r, Q_b)] = aN^{-\alpha} + bD^{-\beta} + E + \mathbb{E}[\Delta(N, Q_r, Q_b)]\,.
\end{equation}
Instead of studying the scaling law of one model, we focus on the \textbf{minimum loss a group of models can achieve optimally}. Furthermore, no weight training is performed after quantization to observe immediate performance degradation. We claim that such optimality also follows its own scaling. We describe the observed scaling laws and present empirical evidence to support them in \Cref{sec:experiments}.

\paragraph{Parameter scaling in \Cref{eq:intro-law}}
Given a fixed $Q_r$, the optimal and expected loss degeneration ($\delta^{\text{opt}}$, $\mathbb{E}(\Delta)$) decreases polynomially as the model size $N = \lVert W_h^{\text{opt}} \rVert_0 + \lVert W_l^{\text{opt}} \rVert_0$ increases.

\paragraph{Ratio scaling in \Cref{eq:intro-law}}
Given a fixed model size ($N$), the optimal and expected loss degeneration ($\delta^{\text{opt}}$, $\mathbb{E}(\Delta)$) increases exponentially as the quantization ratio $Q_r$ increases.

Those two scaling laws posit one of our central hypotheses: the loss degeneration is affected by both the model size and the quantization ratio. We further discovered that their effects are independent. Therefore, we claim the following weak law of loss degeneration:
\begin{custombox}
\textbf{Weak Law of Loss Degeneration.} Given a fixed quantization block size $Q_b$, a fixed quantization method, a model architecture that can scale its parameters $N$ and change the quantization ratio $Q_r$. Following previous definitions, the optimal (minimum) and expectation of the loss degeneration have the following form:
\begin{equation}
\label{eq:our-ptq}
    \delta^{\text{opt}}(N, Q_r) = C e^{AQ_r} N^{-\gamma_N}\,,
\end{equation}
\begin{equation}
\label{eq:our-ptq-expectation}
    \mathbb{E}[\Delta(N, Q_r)] = C^\prime e^{A^\prime Q_r} N^{-\gamma_N^{\prime}}\,,
\end{equation}
where $C$, $C^\prime$, $A$, $A^\prime$, $\gamma_N$, and $\gamma_N^\prime$ are constant coefficients that vary with different quantization block sizes and methods.
\end{custombox}
We could interpret $e^{AQ_r}$ and $N^{-\gamma_N}$ as the quantization scaling effect and parameter scaling effect, respectively. Note that our scaling law is an extension of the precision scaling law shown in \Cref{eq:original-ptq-loss}. For $Q_r = 1$, the only possible pair is $(W_l, W_h) = (\emptyset, M)$. In such a case, our random variable $\Delta$ degenerates into its only realization, and the quantization scaling effect is absorbed into the coefficient $C$. On the other hand, if we fix $D$, $P_\text{train}$, and $P_{\text{post}}$ in \Cref{eq:original-ptq-loss}, it leaves us with only a term of model size $N$ that aligns with our parameter scaling effect.

In the weak law of loss degeneration, while $\delta^{\text{opt}}$ increases with $Q_r$ exponentially, a larger model size $N$ could compensate for such degeneration. Quantitatively, by fixing the loss budget $\delta^{\text{opt}} = l$, we have the following:
\begin{equation}
\label{eq:weak-compensation-rate}
    \log(N) = A^{\prime\prime} Q_r + C^{\prime\prime}\,,
\end{equation}
where $A^{\prime\prime} = A / \gamma_N$ and $C^{\prime\prime} = (\log(C) - \log(l)) / \gamma_N$. This shows that given a fixed loss budget ($\delta=l$), the maximum achievable mixed precision quantization ratio $Q_r$ increases as the model size $N$ increases. This aligns with findings from related research, such as AWQ \citep{lin2024awq}, Quip \citep{chee2024quip}, and LQER \citep{zhang2024lqer}, which empirically demonstrated that larger models can accommodate more aggressive quantization levels. An alternative view, also reflected in related work, is that for a fixed quantization ratio, task loss decreases when the model size becomes larger.

The weak scaling law is sufficient for the case where only the model size ($N$) and the quantization ratio $Q_r$ are taken into consideration. If we want to estimate the loss degeneration with respect to the quantization block size $Q_b$, a stronger scaling law is needed. In fact, the scaling effect of granularity (block size) is also independent from the model size $N$ and quantization ratio $Q_r$.

\begin{custombox}
\textbf{Strong Law of Loss Degeneration.} Given a fixed quantization method, a model architecture that can scale its parameters $N$, change the quantization ratio $Q_r$, and vary its quantization block size $Q_b$. Following previous definitions, the optimal (minimum) and expectation of the loss degeneration has the following form:
\begin{equation}
    \delta^{\text{opt}}(N, Q_r, Q_b) = C e^{AQ_r} N^{-\gamma_N} (Q_b + d)^{\gamma_c}\,,
\end{equation}
\begin{equation}
    \mathbb{E}[\Delta(N, Q_r, Q_b)] = C^\prime e^{A^\prime Q_r} N^{-\gamma_N^{\prime}} (Q_b  + d^\prime)^{\gamma_c^\prime}\,,
\end{equation}
where $A$, $A^\prime$, $C$, $C^\prime$, $\gamma_N$, $\gamma_N^{\prime}$, $d$, $d^\prime$, $\gamma_c$, and $\gamma_c^{\prime}$ are constant coefficients that depend on the quantization method.
\end{custombox}
For the strong scaling law, when fixing the quantization block size $Q_b$, the granularity scaling effect $(Q_b + d)^{\gamma_c}$ can be absorbed into the constant $C$ in the weak law, indicating the effectiveness of the weak law for the case where $Q_b$ is fixed.

\section{Experiments}
\label{sec:experiments}
\subsection{Setup}
\label{sec:exp:setup}

\paragraph{Models and benchmarks}
We evaluate a range of model families, including LLaMA \cite{touvron2023llama2, dubey2024llama}, Qwen-1.5 \cite{bai2023qwen} and Qwen-3~\cite{qwen3report2025}, at sizes ranging from 60M to 14B. Besides these models, to collect the mixed quantization results at small-scale models for extrapolation, we followed the Chinchilla scaling law and pretrained a series of LLaMA-like models consisting of $N \in \{60M, 200M, 400M, 600M, 1.1B\}$ parameters on FineWeb~\citep{lozhkov2024fineweb-edu}, a high-quality pretraining dataset released by HuggingFace. We refer to these models as \textsc{CLM} series models to differentiate them from the vanilla Meta LLaMA models. More details are available in~\Cref{app:clm-pretrain}. We choose Qwen-1.5 and Qwen-3 to validate our experimental results, enabling detailed analysis of the proposed scaling laws. We subsample a set of 1000 entries from the SlimPajama \cite{cerebras2023slimpajama} dataset for evaluation, which is another open-sourced pretraining dataset, as the perplexity/pretraining loss on SlimPajama better captures the performance of these base models compared with downstream tasks~\citep{dubey2024llama3}. 

\paragraph{Quantization methods}
We mainly use MXINT~\cite{rouhani2023mxformat}, and HQQ~\cite{badri2023hqq} at various bit-widths and block sizes for low-precision formats, and BF16 as the high-precision format. Specifically, if not mentioned specifically, by default, all our experiments are performed under weight-and-activation quantization with MXINT-4. We also include weight-and-activation quantized MXINT-2, weight-only quantized MXINT-4, and weight-only HQQ with 4 bit-widths and 64 block sizes to verify that our scaling laws apply to other data formats. We choose HQQ because it is a calibration-free method that achieves state-of-the-art performance, simplifying our search loop. 

\paragraph{Mixed quantization strategy}
For our primary experiments, we perform quantization at two granularities: layer-wise and matmul-wise. In the former, the quantization ratio is determined by the number of transformer layers cast to low precision. In the latter, we consider the precision for each individual matrix multiplication. We find a solution for \Cref{eq:full-unified-law} by searching through random trials of 100 from each quantization configuration. We justify these choices in Appendix~\Cref{app:search-space-distribution}. For clarity, our experiments include a quantization ratio $Q_r\in \{0.5, 0.6, 0.7, 0.8, 0.9, 0.95, 0.975\}$. Note that, in each trial, the inner loop of the search conducts post-training quantization (PTQ), and the entire search process involves no training.

\paragraph{Platform and GPU hours}
We perform experiments on a cluster of DGX A100 eight-GPU pods, each with 40GB VRAM, with roughly 15k A100 GPU hours in total. We also spend around 1k GPU hours tuning search hyper-parameters, such as determining the number of trials and searching the loss landscape to determine an appropriate quantization ratio. The pretraining of CLM models takes around 1k GPU hours on another DGX H100 eight-GPU pod.
\subsection{Insights from the Unified Scaling Law}
\label{sec:experiments:main-results}
We mainly show results of the layer-wise loss landscape to verify our claims, but the same laws can also be applied to matrix multiplication-wise (matmul-wise). We present CLM, Llama, and QWen-3 in the main texts. All additional results, \textit{e.g.}, matmul-wise and QWen-1.5, are in \Cref{app:matmul-result}.

\subsubsection{Model Size ($N$) and quantization ratio ($Q_r$)}
\begin{figure}[!t]
\centering
\begin{subfigure}[b]{0.3\textwidth} \centering
\includegraphics[width=\textwidth]{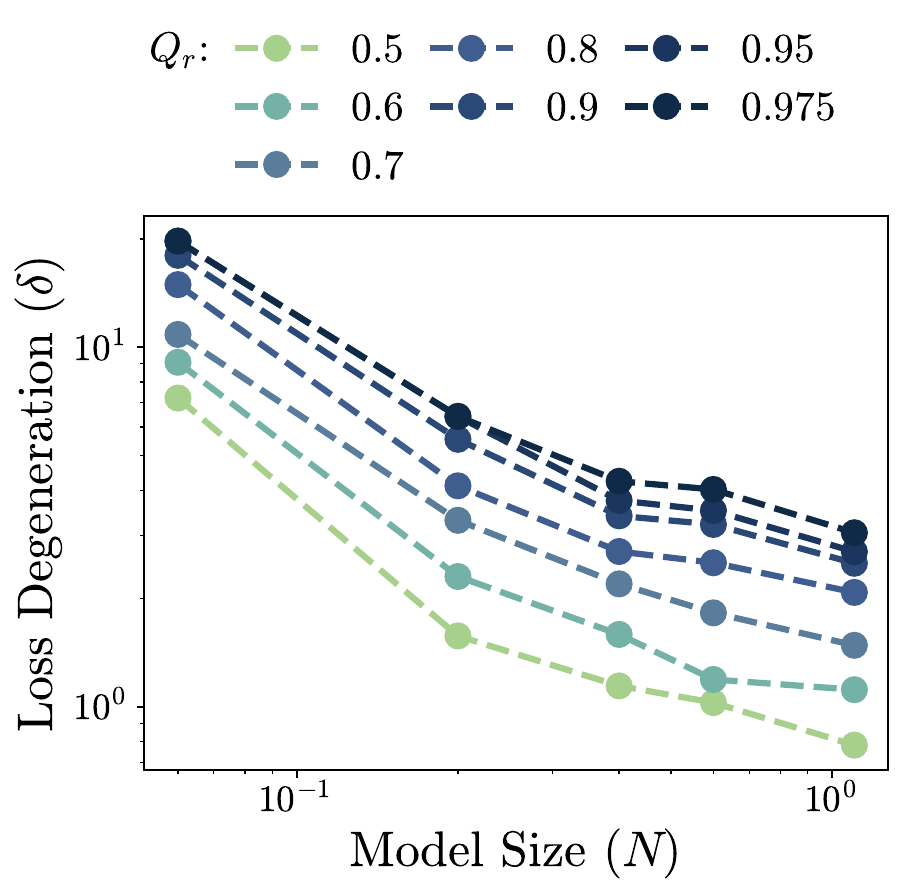}
\captionsetup{font=scriptsize}
\caption{CLM Actual Loss}
\end{subfigure}
\hfill
\begin{subfigure}[b]{0.3\textwidth} \centering
\includegraphics[width=\textwidth]{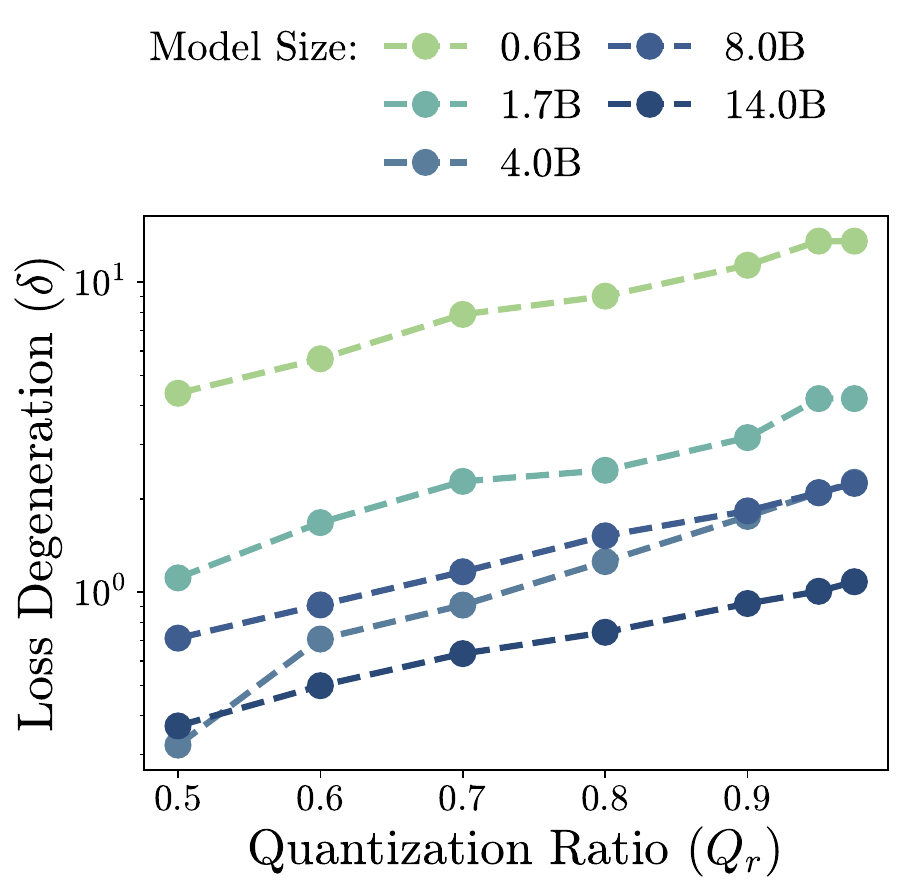}
\captionsetup{font=scriptsize}
\caption{Qwen-3 Actual Loss}
\end{subfigure}
\hfill
\begin{subfigure}[b]{0.3\textwidth} \centering
\centering
\includegraphics[width=\textwidth]{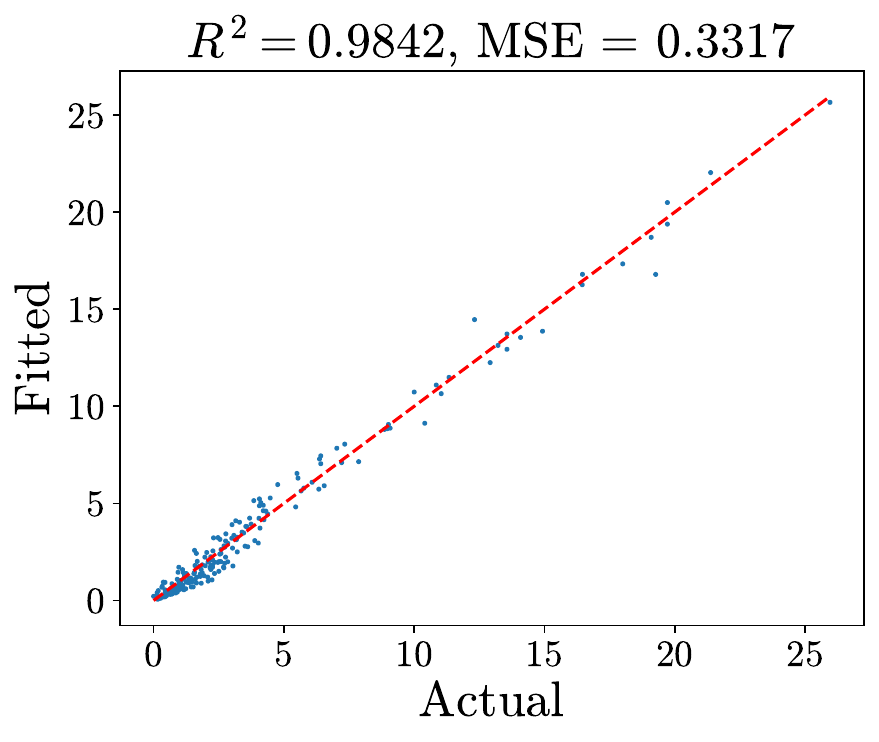}
\captionsetup{font=scriptsize}
\caption{Weak Law $\delta^{\text{opt}}$}
\end{subfigure}

\begin{subfigure}[b]{0.3\textwidth} \centering
\includegraphics[width=\textwidth]{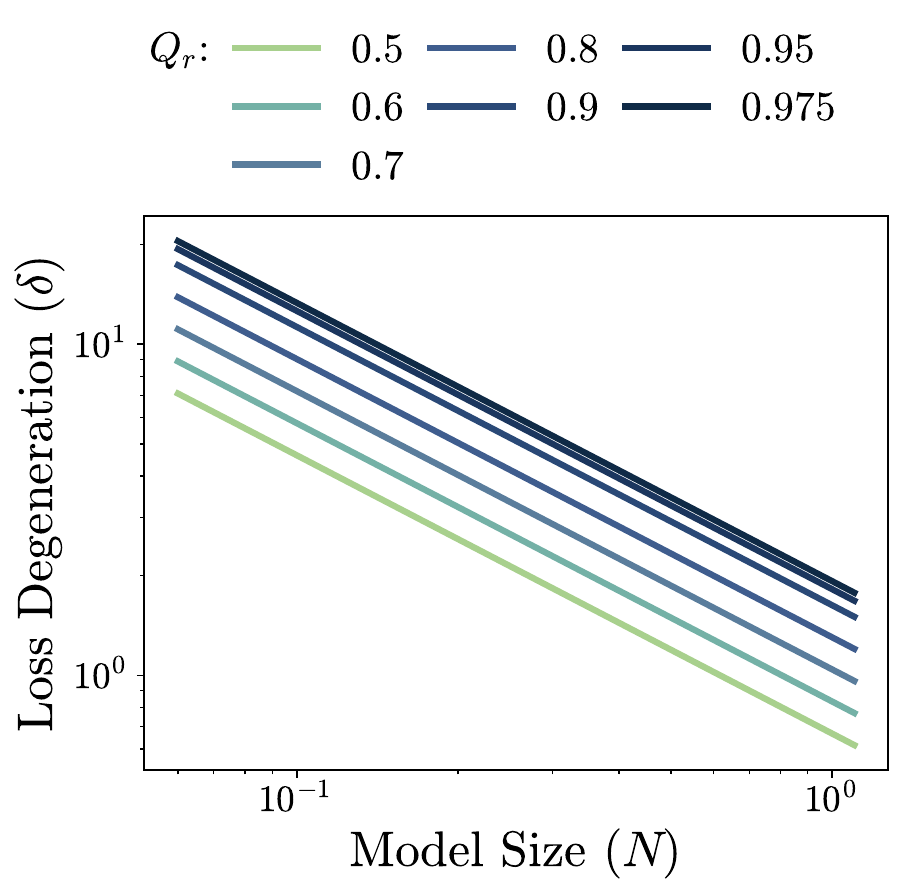}
\captionsetup{font=scriptsize}
\caption{CLM Predicted Loss}
\end{subfigure}
\hfill
\begin{subfigure}[b]{0.3\textwidth} \centering
\includegraphics[width=\textwidth]{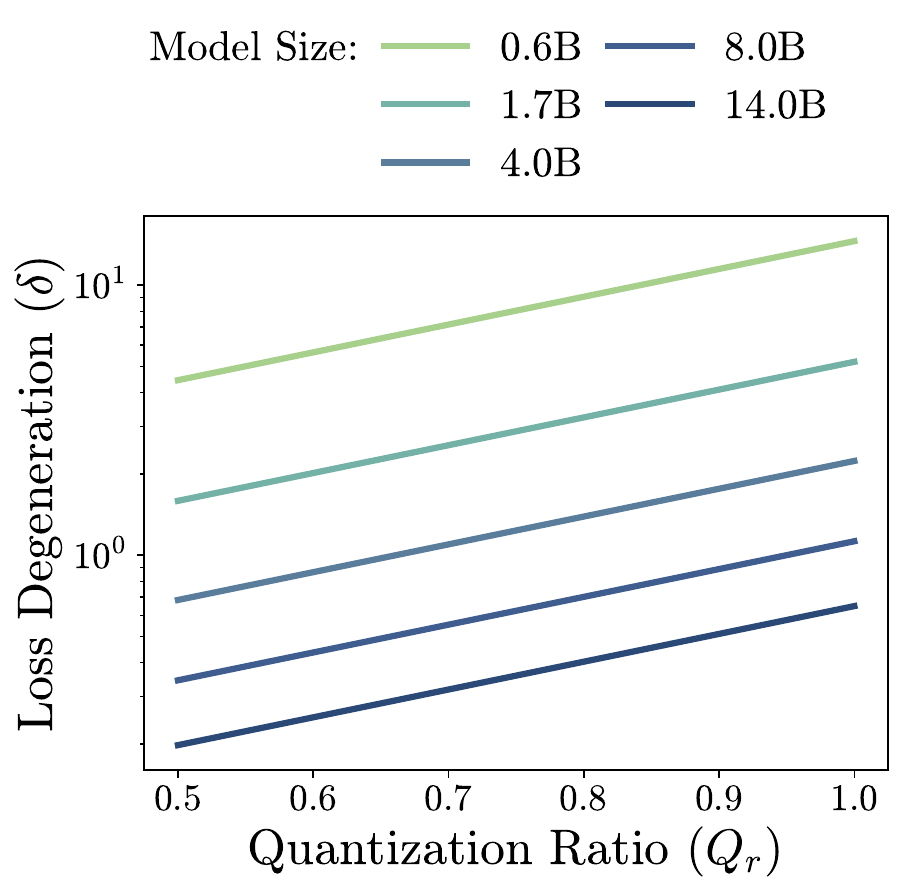}
\captionsetup{font=scriptsize}
\caption{Qwen-3 Predicted Loss}
\end{subfigure}
\hfill
\begin{subfigure}[b]{0.3\textwidth} \centering
\centering
\includegraphics[width=\textwidth]{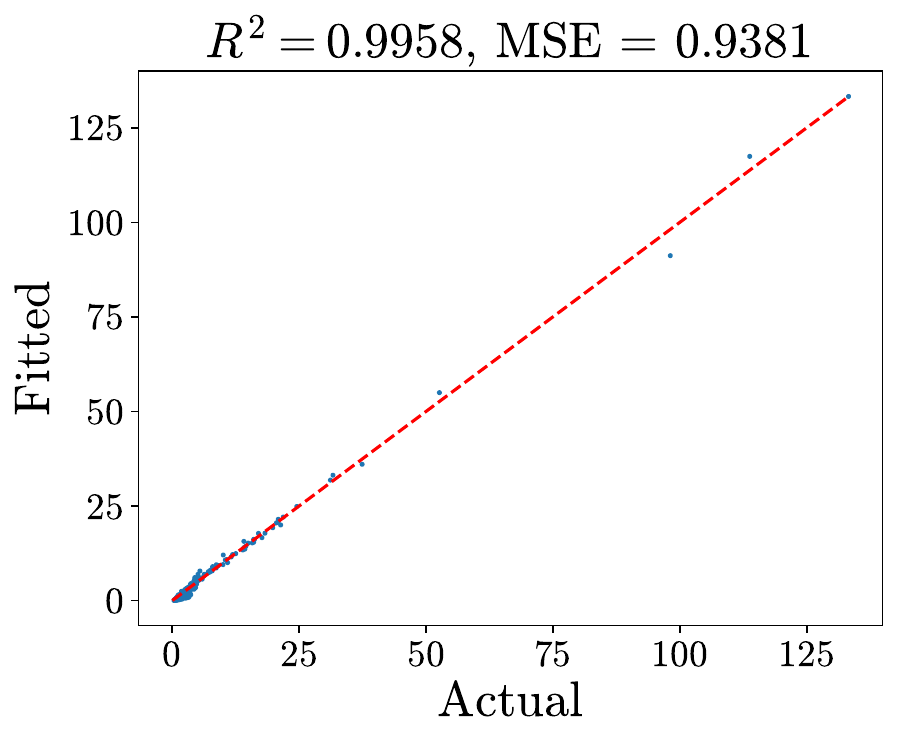}
\captionsetup{font=scriptsize}
\caption{Weak Law $\delta_\mu$}
\label{fig:scaling1-layer-f}
\end{subfigure}
\caption{\textbf{Weak law experiment results.} (a,b,d,e): The actual and fitted loss degeneration contour for the layer-wise MXINT4 quantization for CLM and Qwen3 architecture. (a,d): Loss contour w.r.t. model size $N$ for the CLM model family. (b,e): Loss contour w.r.t. quantization ratio $Q_r$ for the Qwen3 model family. (e, f): The fitted versus actual $\delta^{\text{opt}}$ and $\delta_{\mu}$ for all MXINT-4 quantization (layer- and matrix multiplication-wise) for CLM, Qwen1.5, and Qwen3. Due to space constraint, we only present $N$ v.s. $\delta$ for CLM and $Q_r$ v.s. $\delta$ in the figure,  see~\Cref{app:layer-result,app:matmul-result} for full figures of all results.}
  \label{fig:scaling1-layer}
\end{figure}

Figure~\ref{fig:scaling1-layer} shows the actual and fitted $\delta^{\text{opt}}$ for layer-wise MXINT-4 quantization for CLM series and Qwen3. The overall shape of the contour is the same for actual and fitted losses, with the statistics of $R^2 = 0.98$. As introduced in~\Cref{sec:laws}, the loss degeneration scaling law also works for the expectation value $\mathbb{E}(\delta)$,with a $R^2 = 0.99$, indicating the effectiveness of our law for $\mathbb{E}(\Delta)$. The corresponding figures and fitted parameters can be seen in \Cref{app:formula-fitting}. From the results in \Cref{fig:scaling1-layer}, we can make the following observations that are in line with our weak law:
\observationbox[Observation 1:]{
With an increasing $N$, we can further increase $Q_r$, which is in agreement with our combined ratio scaling and parameter scaling terms: $e^{A Q_r} N^{-\gamma_N}$.}

\subsubsection{Quantization ratio ($Q_r$) and quantization block size ($Q_b$)}
\begin{figure}[!t]
\centering
\begin{subfigure}[b]{0.315\textwidth} \centering
\includegraphics[width=\textwidth]{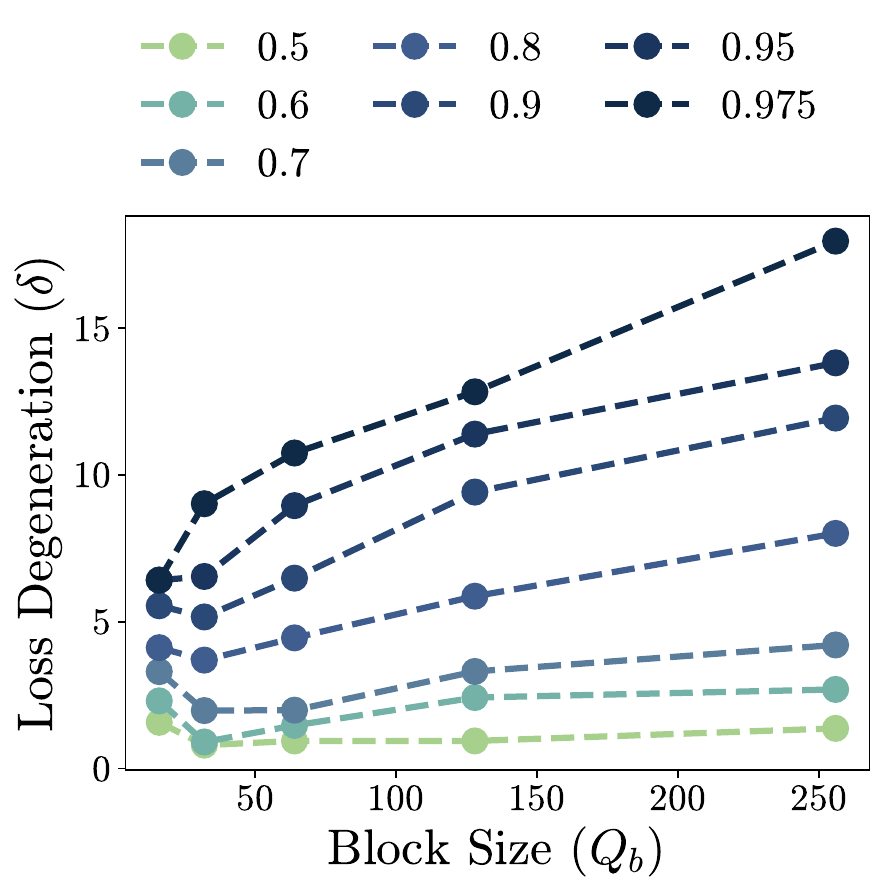}
\captionsetup{font=scriptsize}
\caption{200M Actual Loss}
\end{subfigure}
\hfill
\begin{subfigure}[b]{0.315\textwidth} \centering
\includegraphics[width=\textwidth]{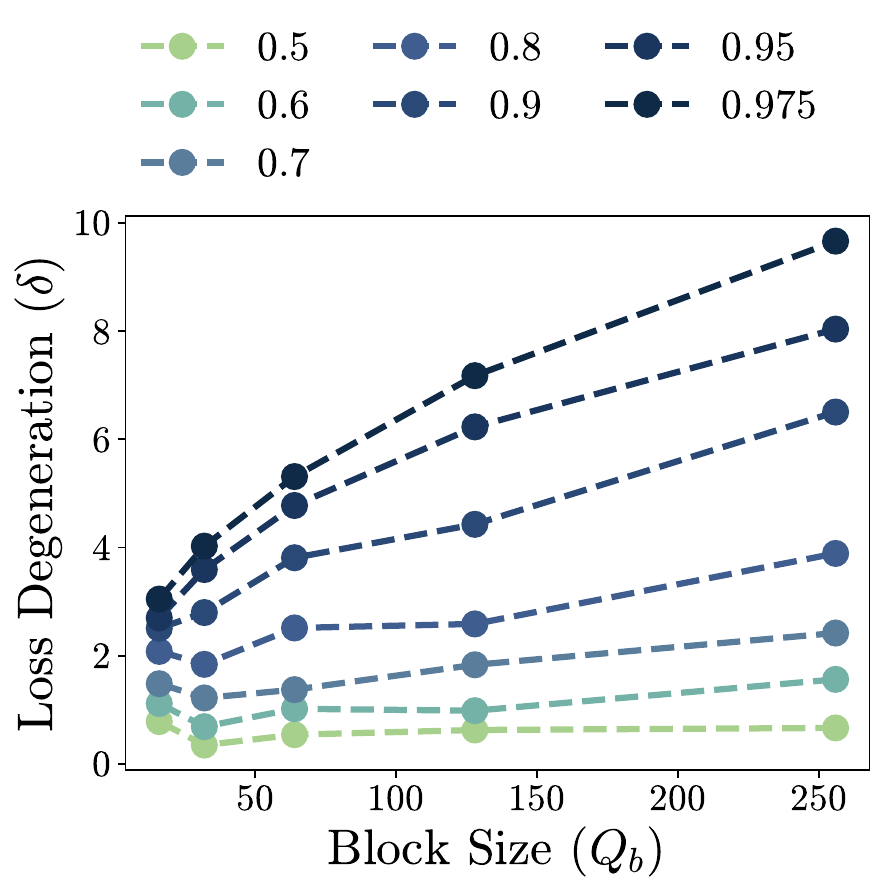}
\captionsetup{font=scriptsize}
\caption{1.1B Actual Loss}
\end{subfigure}
\hfill
\begin{subfigure}[b]{0.315\textwidth}
\centering
\includegraphics[width=\textwidth]{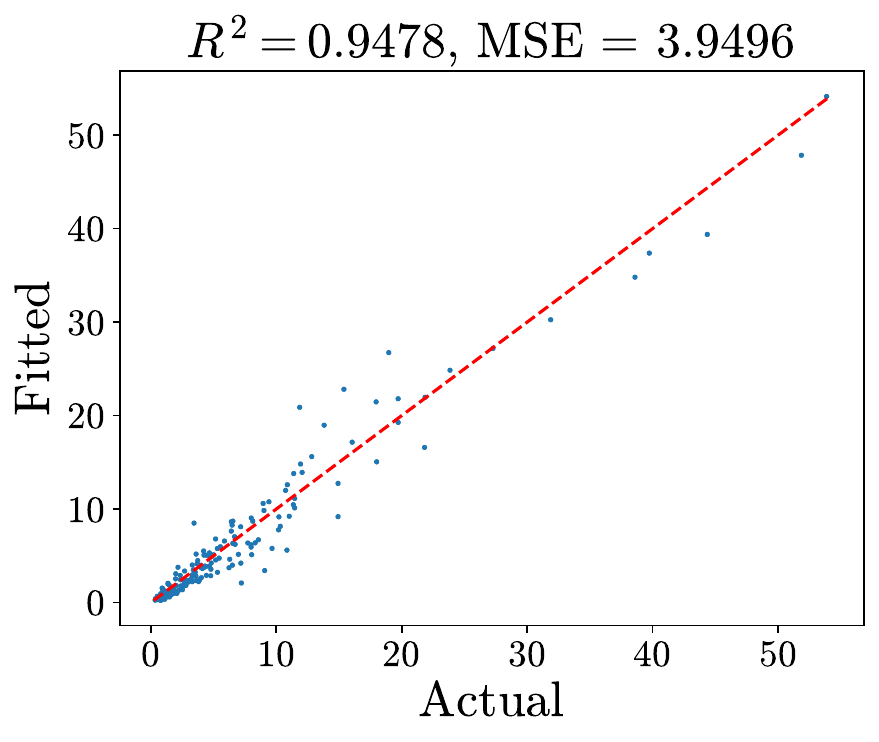}
\caption{Strong law $\delta^{\text{opt}}$}
\end{subfigure}
\vfill
\begin{subfigure}[b]{0.315\textwidth} \centering
\includegraphics[width=\textwidth]{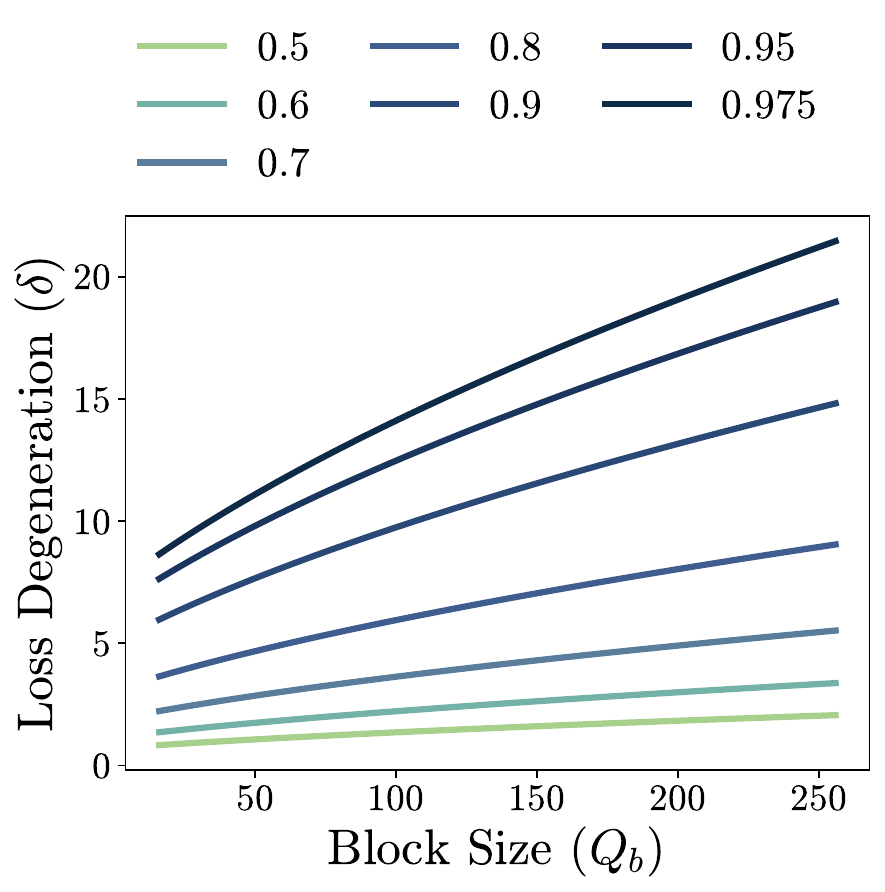}
\captionsetup{font=scriptsize}
\caption{200M Predicted Loss}
\end{subfigure}
\hfill
\begin{subfigure}[b]{0.315\textwidth} \centering
\includegraphics[width=\textwidth]{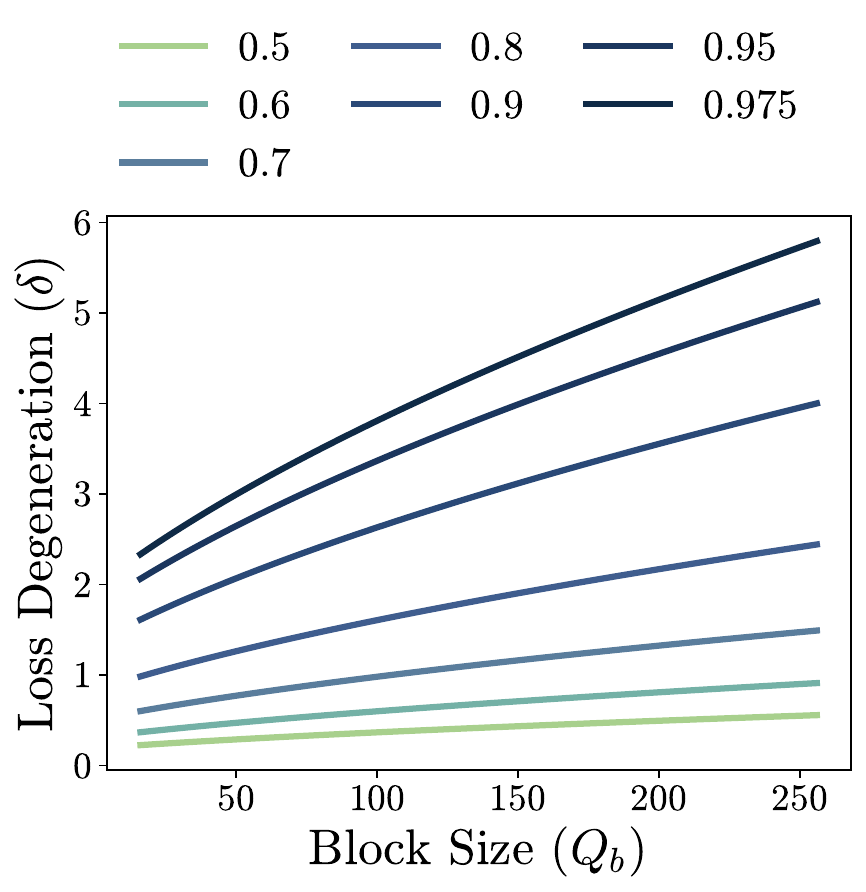}
\captionsetup{font=scriptsize}
\caption{1.1B Predicted Loss}
\end{subfigure}
\hfill
\begin{subfigure}[b]{0.315\textwidth}
\centering
\includegraphics[width=\textwidth]{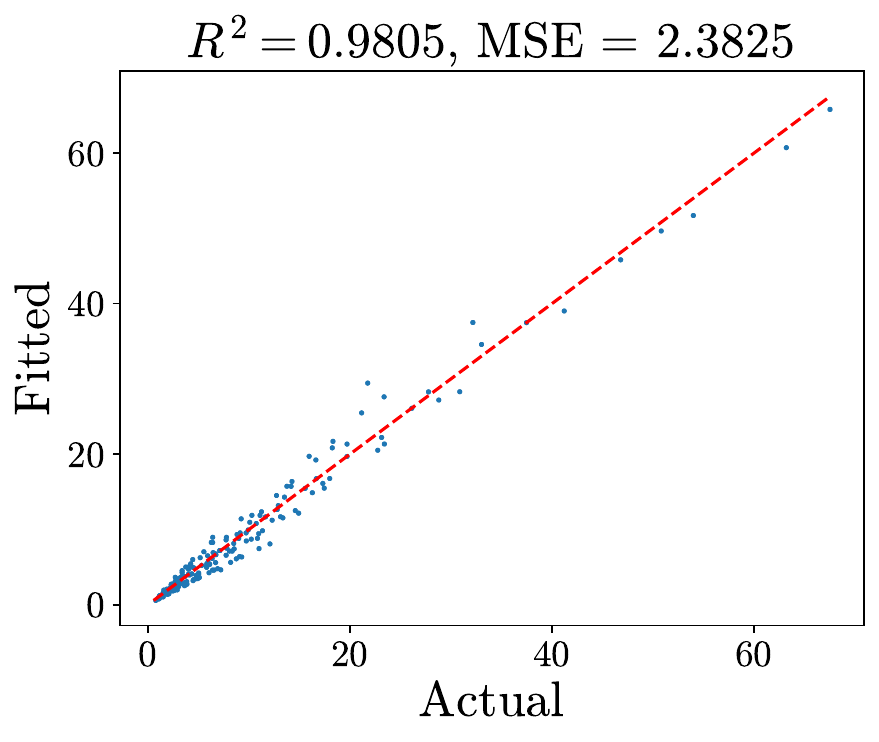}
\captionsetup{font=scriptsize}
\caption{Strong law $\delta_\mu$}
\end{subfigure}
  \caption{\textbf{Strong law experiment results}. (a,b,c,d): The actual and fitted loss degeneration contour for CLM-200M and CLM-1.1B w.r.t. block size $Q_b$. Each line corresponds to a fixed quantization ratio ($Q_r$). (e,f): The fitted versus actual $\delta^{\text{opt}}$ and $\delta_{\mu}$. Due to space constraint, we only present $N=200M,1.1B$,  see~\Cref{app:layer-result} for full figures of all $N$ blocksize results.}
  \label{fig:scaling2-layer}
\end{figure}

As stated in~\Cref{sec:background:quantization}, recent LLM quantization methods adopt fine-grained quantization, meaning tensors are split into small blocks, quantized and then scaled individually. In this part, we empirically verify our unified scaling law by performing mixed-precision quantization search at blocksize $Q_b \in \{16, 32, 64, 128, 256\}$. We present the experiments on CLM series models with matrix-multiplication-wise for finer granularities. As shown in~\Cref{fig:scaling2-layer}, the overall shape of the contour is the same for actual and fitted losses on $\delta^{opt}$, with the statistics $R^2=0.95$. From~\Cref{sec:laws}, the loss degeneration scaling law also works for the expectation value $\mathbb{E}(\delta)$,with a $R^2 = 0.98$, indicating the effectiveness of our law for $\mathbb{E}(\Delta)$. The corresponding figures and fitted parameters are in \Cref{app:formula-fitting}. From the results in \Cref{fig:scaling1-layer}, we can make the following observations that are in line with our strong law:
\observationbox[Observation 2:]{
When the quantization ratio $Q_r$ is low, employing a smaller block size $Q_b$ is more effective in minimizing performance degradation, which agrees with our combined ratio scaling and granularity scaling terms: $e^{AQ_r} (Q_b + d)^{\gamma_c}$.}
\subsection{Extending to other quantization methods}
\begin{figure}[!t]
\centering
\begin{subfigure}[b]{0.315\textwidth} \centering
\includegraphics[width=\textwidth]{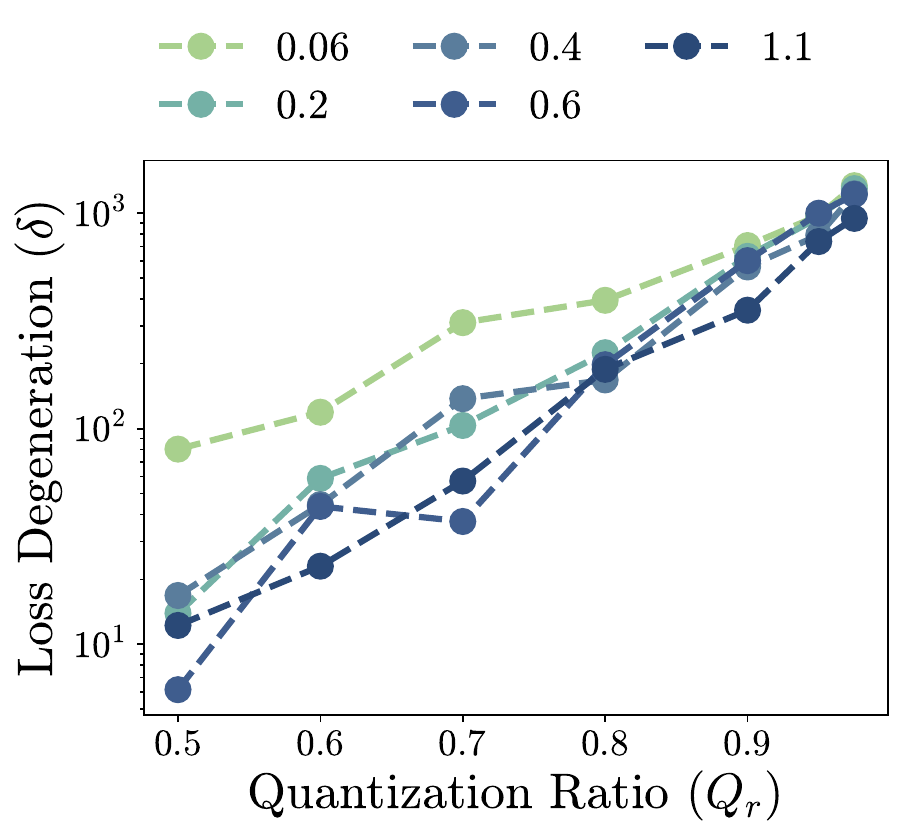}
\captionsetup{font=scriptsize}
\caption{WA-MXINT-2 Actual Loss}
\end{subfigure}
\begin{subfigure}[b]{0.01\textwidth}
~
\end{subfigure}
\begin{subfigure}[b]{0.315\textwidth} \centering
\includegraphics[width=\textwidth]{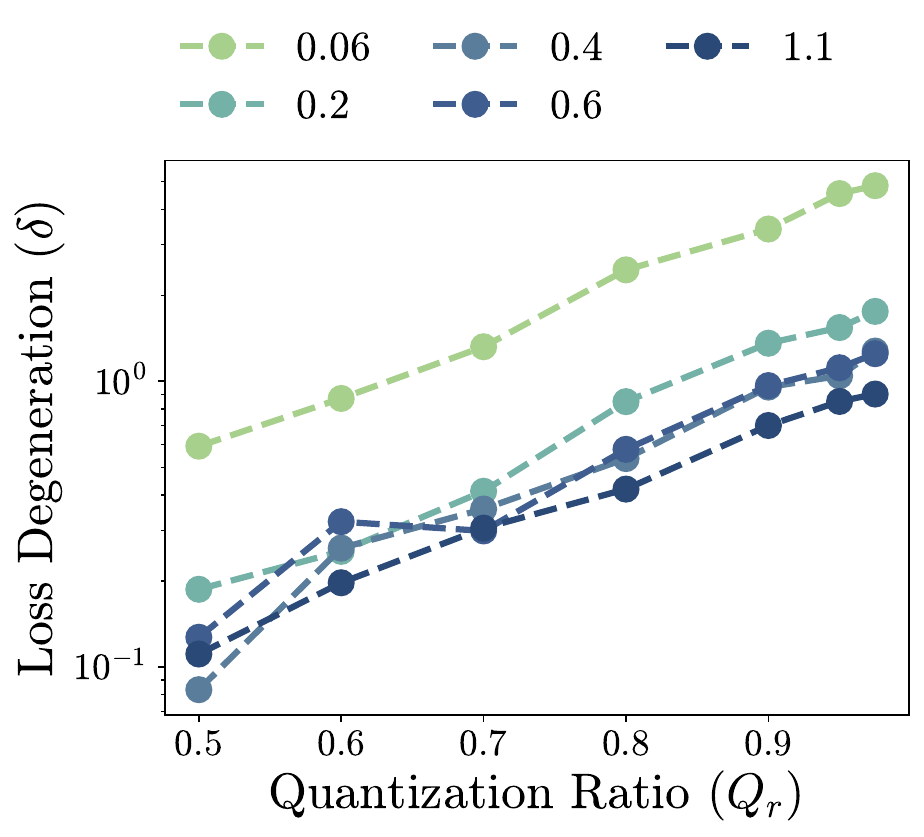}
\captionsetup{font=scriptsize}
\caption{W-MXINT-4 Actual Loss}
\end{subfigure}
\begin{subfigure}[b]{0.01\textwidth}
~
\end{subfigure}
\begin{subfigure}[b]{0.315\textwidth} \centering
\includegraphics[width=\textwidth]{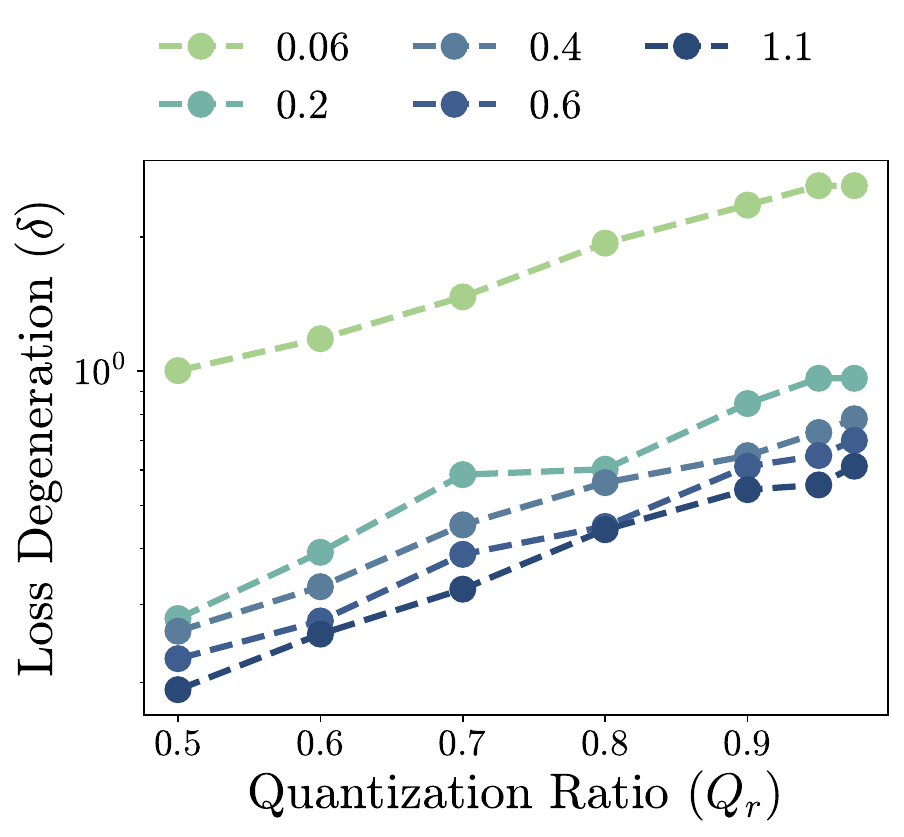}
\captionsetup{font=scriptsize}
\caption{W-HQQ 4 bit Actual Loss}
\end{subfigure}
\begin{subfigure}[b]{0.315\textwidth} \centering
\includegraphics[width=\textwidth]{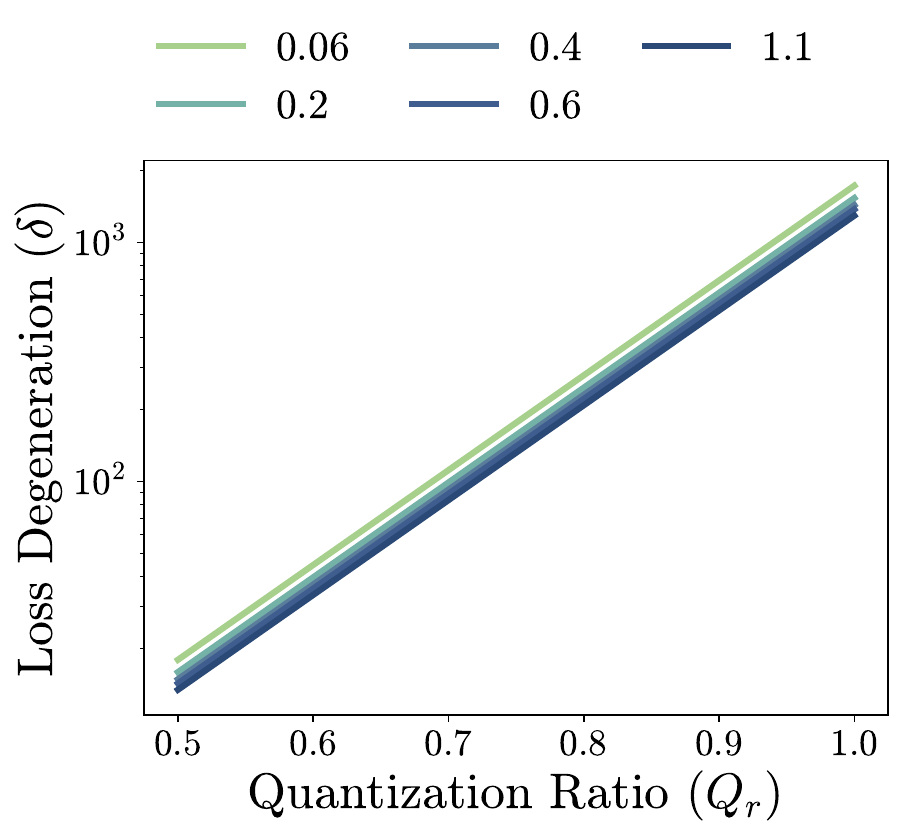}
\captionsetup{font=scriptsize}
\caption{WA-MXINT-2 Predicted Loss}
\end{subfigure}
\begin{subfigure}[b]{0.01\textwidth}
~
\end{subfigure}
\begin{subfigure}[b]{0.315\textwidth} \centering
\includegraphics[width=\textwidth]{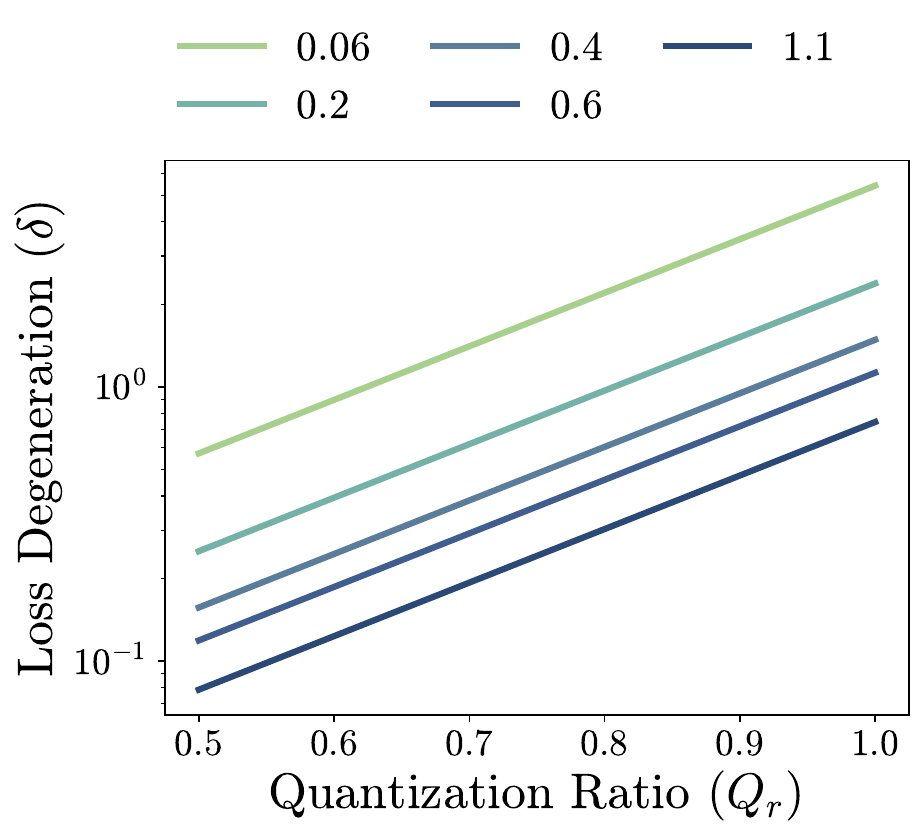}
\captionsetup{font=scriptsize}
\caption{W-MXINT-4 Predicted Loss}
\end{subfigure}
\begin{subfigure}[b]{0.01\textwidth}
~
\end{subfigure}
\begin{subfigure}[b]{0.315\textwidth} \centering
\includegraphics[width=\textwidth]{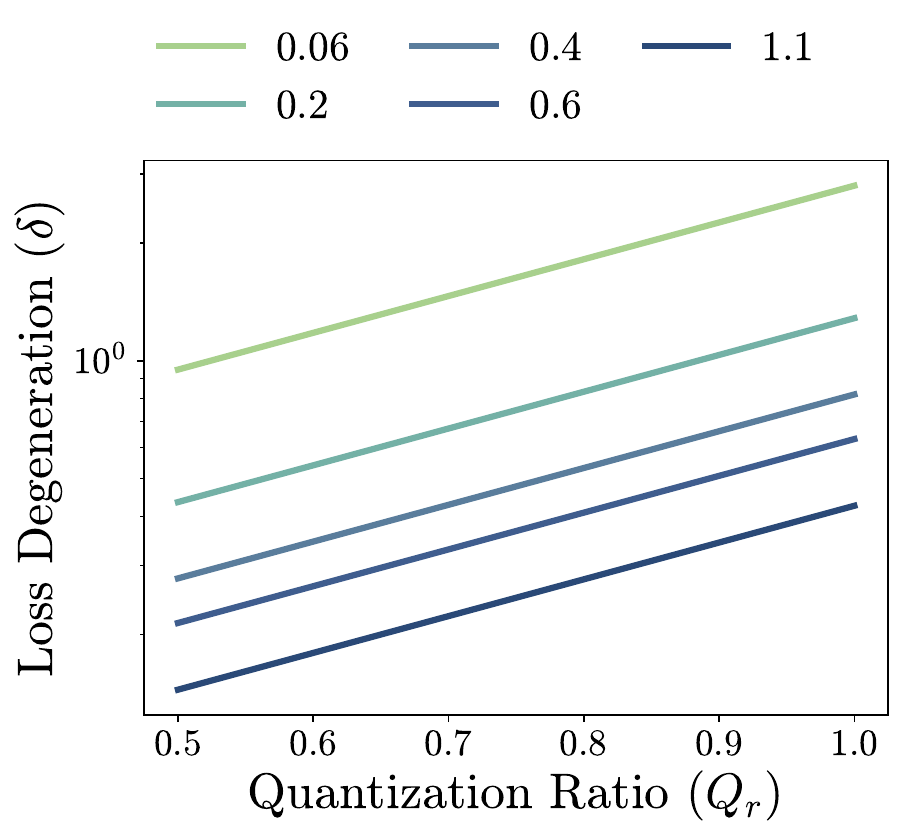}
\captionsetup{font=scriptsize}
\caption{W-HQQ 4 bit Predicted Loss}
\end{subfigure}
  \caption{\textbf{Other Arithmetic Formats Results.} (a,d) The actual and fitted loss degeneration contour for layer-wise Weight-Activation-MXINT-2 (WA-MXINT-2) on CLM architectures. 
  (b,e) The actual and fitted loss degeneration contour for layer-wise Weight-only MXINT-4 (W-MXINT-4) on CLM architectures. (c,f) The actual and fitted loss degeneration contour for layer-wise Weight-only Weight-only 4bit-HQQ (HQQ-4) on CLM architectures. In the figure, each line corresponds to a fixed model size $N$ in billions (B). The fitted statistics are presented in~\Cref{app:formula-fitting} due to space constraint.}
  \label{fig:scaling-other-format-layer}
\end{figure}

All our previous experiments consider W-A quantization with MXINT-4 as mentioned in \Cref{sec:exp:setup}. We also experimented with different quantization settings. For comparison, we showed that the loss degeneration scaling law still holds for other arithmetic formats and quantization methods. Moreover, as proposed by~\citet{dotzel2024learningfp4}, we consider weight-only quantization (with activations kept at 16-bit) for the precision allocation search. The results are presented in~\Cref{fig:scaling-other-format-layer}. All three quantization settings follow our scaling law with different fitted parameters. Note that changing the precision (MXINT-4 and MXINT-2) will shift the parameters completely, which is different from the previous precision scaling law in \Cref{eq:original-ptq-loss}. Additional results are in~\Cref{app:extending-other-format}. 
\observationbox[Observation 3:]{
Our unified scaling law can be applied to other arithmetic formats.}
\subsection{Lessons Learned}
\begin{figure}[htbp]
\centering
\begin{subfigure}[b]{0.48\textwidth} \centering
\includegraphics[width=\textwidth]{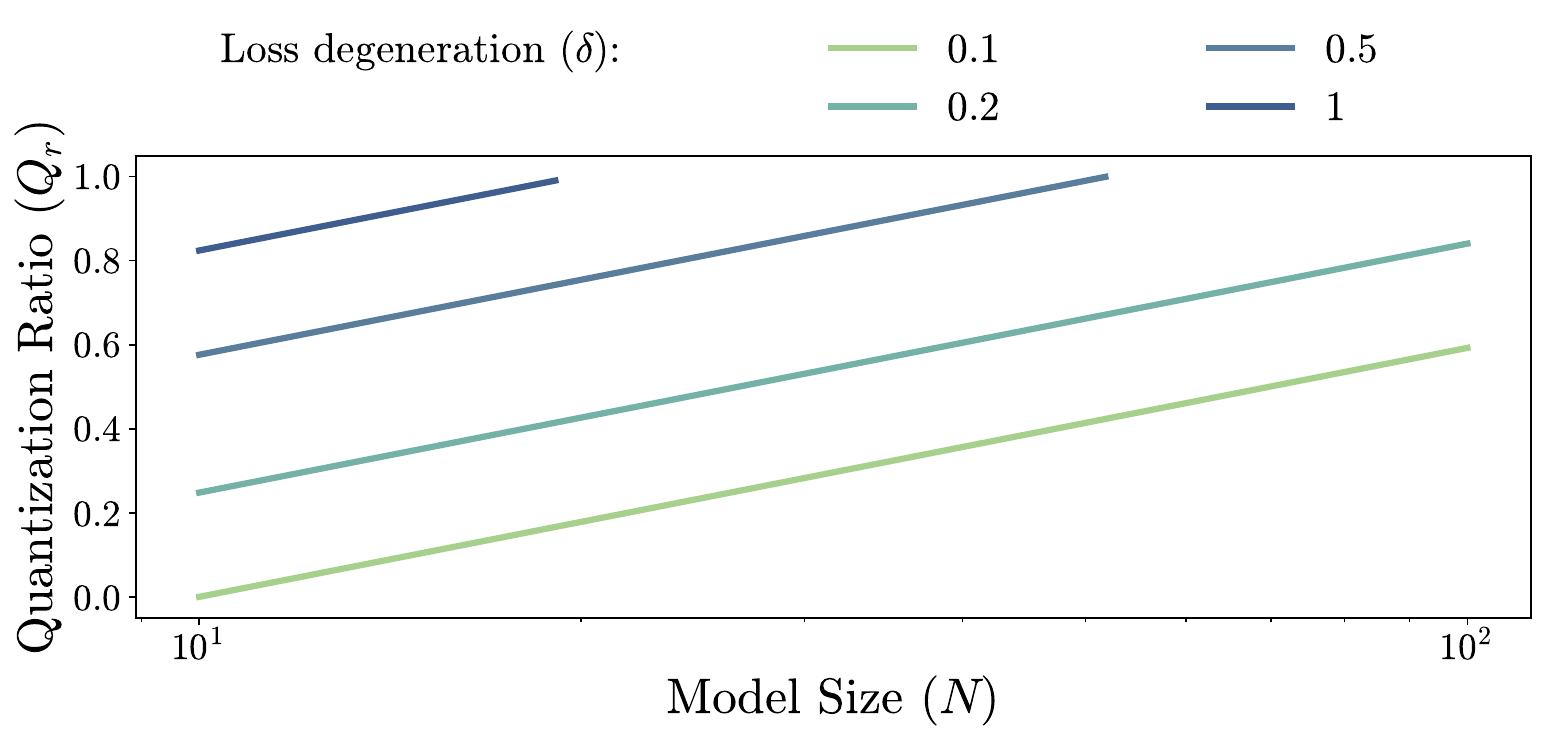}
    \captionsetup{font=scriptsize}
    \caption{$N$ compare to $Q_r$ under dfferent fixed $\delta$ budget}
    \label{fig:wrapped-take-2}
\end{subfigure}
\hfill
\begin{subfigure}[b]{0.48\textwidth} \centering
\includegraphics[width=\textwidth]{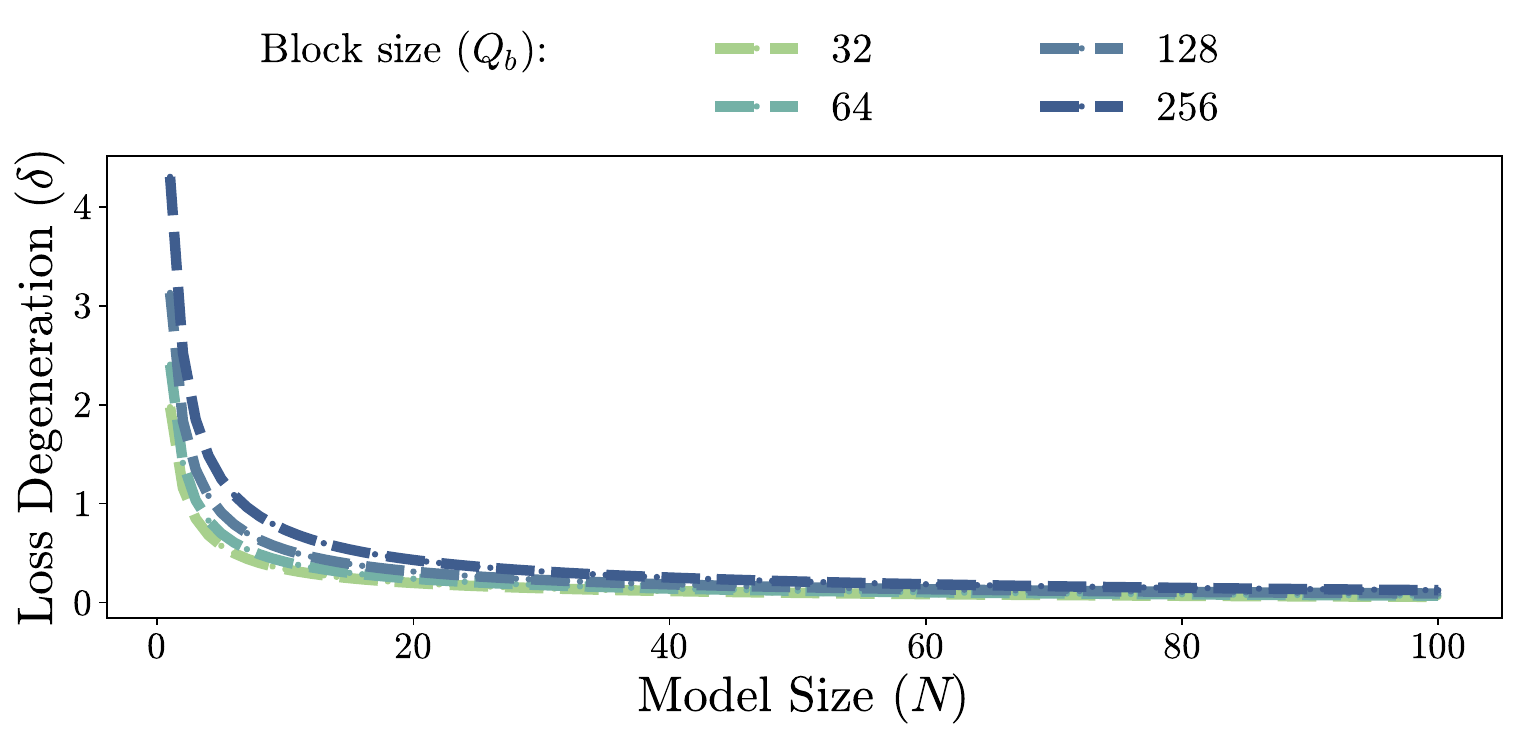}
    \captionsetup{font=scriptsize}
    \caption{$N$ compare to $\delta$ under different fixed $Q_b$.}
    \label{fig:wrapped-take-3}
\end{subfigure}
\caption{\textbf{Effects of our scaling laws.} The effect of our scaling law with (a) effect of quantization ratio $Q_r$  when changing model size $N$ under a fixed loss degeneration budget $\delta$ and (b) effect of blocksize granularities $Q_b$ on loss degeneration $\delta$ when changing $N$. Both are based on the fitted parameters for CLM series models.}
\end{figure}

As discussed in Section~\ref{sec:laws}, our weak scaling law aligns with the precision scaling law when $Q_r = 1$. Also, the strong scaling law that introduced the quasi-polynomial term did not deviate our scaling law away from the precision scaling law.

\takeawaybox[Takeaway 1:]{
Across different $Q_r$ values, our unified scaling law agrees asymptotically with previously proposed Precision Scaling Laws proposed by \citet{kumar2024scaling}.
}

To observe the effect of different quantization ratios $Q_r$ when varying model size $N$, we plot~\Cref{fig:wrapped-take-2} based on our unified scaling law on the CLM series model by fixing the $\delta$ budget.\textbf{ When models are beyond 70B, $Q_r$ can be set to beyond $0.9$ with less than $\delta=0.5$ loss degeneration}.

\takeawaybox[Takeaway 2:]{
With an increase in model size, the quantization ratio $Q_r$ can be set to a high value. This is due to the parameter scaling term in \Cref{eq:intro-law} accelerating at a much faster rate compared to the ratio scaling.}

For different block size granularities $Q_b$ when varying model size $N$, we plot~\Cref{fig:wrapped-take-3} based on our unified scaling law on CLM series model. \textbf{This demonstrates that when the models are larger than $50$B, the difference between $Q_b=128$ and $Q_b=32$ is less than $0.5$ in terms of loss degeneration $\delta$}. This makes us challenge the design choice of the current 128 block size utilized in NVIDIA's Blackwell GPUs for Mixed-Precision Floating-Point (MXFP) arithmetic might be unnecessarily large for PTQ LLM inference tasks.

\takeawaybox[Takeaway 3:]{
As the model size $N$ becomes sufficiently large, changing the block size $Q_b$ may yield diminished returns.
}

\section{Background and Related Work}
\label{sec:background}
\paragraph{Quantization outliers and mixed quantization}
\label{sec:background:quantization}
A weight or activation value is considered an outlier when there is a significant deviation from its mean distribution. Activation outliers have been observed more frequently in large models \citep{wei2022outlier,zhang2023revisiting} as cascaded layers accumulate quantization errors. In weight-only quantization, weights are mapped to low precision~\citep{frantar2022gptq,lin2024awq}. Recent weight-only quantization works focus on efficient vector quantization that maps high-precision weight tensors into indices and codebooks, such as QuiP~\citep{chee2024quip}, AQLM~\citep{egiazarian2024aqlm}, and QTIP~\cite{tseng2024qtip}. Meanwhile, weight-activation quantization usually transfers activation magnitudes to weights using invertible scale matrices \citep{xiao2023smoothquant} before quantizing both weights and activations \citep{wei2023outlier,xiao2023smoothquant,shao2023omniquant}. Recent works explore incoherence processing to achieve this, such as SpinQuant~\citep{liu2024spinquant} and Quarot~\citep{ashkboos2024quarot}. There are also works that solve the outlier problem via new number formats to accommodate the dynamic range of outliers \citep{zhang2023revisiting,rouhani2023mxformat,zou2024bie}. For example, MXINT/MXFP \citep{darvish2020pushing} is a recent standard for hardware-efficient numerical formats. MXINT shares an exponent across a block of mantissas \citep{rouhani2023mxformat}, and MXFP shares an exponent across a block of MiniFloats, which is already supported in NVIDIA Blackwell. The hardware efficiency of these methods often outperforms standard low-precision floating-point computation, although custom hardware support is required.

\paragraph{Scaling laws of LLM training}
\citet{kaplan2020scaling} showed, through empirical analysis, that Transformer performance follows a power law trend. In contrast, Chinchilla \cite{chincilla} argues that existing LLMs are under-trained relative to their size, and parameter count should be increased in line with the number of training tokens. The findings from \cite{reconciling} later explained the discrepancy between Kaplan and Hoffman, reaffirming the validity of the Chinchilla scaling laws. A highly relevant study is the recently proposed Scaling Laws for Precisions by \cite{kumar2024scaling}. This work focuses on both pre-training and post-training quantization in which all parameters within the models are uniformly quantized to a single precision level. We discuss more in \Cref{sec:laws} into the relevance of our mixed quantization scaling laws, showing that in extreme cases, our unified mixed quantization law simplifies to adhere to the Scaling Laws for Precisions \citep{kumar2024scaling}.

\section{Discussion, Limitation and Conclusion}
\label{sec:discussion}
\paragraph{Implications on AI inference hardware and systems}
We show that larger models can accommodate increasingly more low-precision components without performance degradation. This validates the recent trend of increasing support for low-precision arithmetic computation in hardware such as GPUs and TPUs \cite{tirumala2024nvidia,choquette2022nvidia}. The insight from our unified scaling law highlights the \textbf{need for increased low-precision resources in future hardware devices}.

\paragraph{Extension to further architectures and arithmetic formats}
It is natural to consider whether the observed findings in this work extend to larger LLMs, such as 400B ones. Additionally, the same trends could be explored in more architectures including MoE models such as DeepSeek~\citep{liu2024deepseek} and Mixtral \citep{jiang2024mixtral}. Finally, further arithmetic formats such as ternary \citep{chen2024ternaryllm} and additional configurations from the MXINT \citep{rouhani2023mxformat} standard offer opportunities for further exploration. One specific challenge is the quantization approach used in this paper is emulated following \citet{zhangmase}, where it incurs more computation, hence impedes the evaluation of larger models (\eg{}, 400B). A possible future direction would be to test these scaling laws on large models using actual MXINT4 and MXFP4 quantization
upon the availability of compatible hardware.

\paragraph{Hypotheses on other efficient AI methods}
While we focused primarily on quantization, a clear direction for future research involves examining scaling trends for other Efficient AI methods, such as sparsity. We hypothesize that the scaling laws for such methods will closely resemble the scaling laws for quantization introduced in this work. More broadly, we hypothesize the existence of \textbf{a broader scaling law governing how the ratio of approximate compute to exact compute scales with model sizes}, and the granularity at which approximate compute is applied.

\paragraph{Conclusion}
In this paper, we introduce a unified scaling law for PTQ of LLMs, supported by thorough experiments. We explore the implications of this law on mixed quantization strategies and the selection of quantization block sizes. Our findings provide direction for future LLM quantization research and suggest a potential for mixed-quantization LLM inference accelerators.

\newpage
\bibliography{refs}
\bibliographystyle{conference}

\newpage
\appendix

\section{Pre-training CLM series Model}
\label{app:clm-pretrain}
As discussed in~\Cref{sec:exp:setup}, we pre-trained a series of causal language models (CLMs) to facilitate our derivation of mixed quantization scaling laws and corresponding experiments, before we cross-validate our scaling laws on other open-source models.

\begin{table}[h]
    \centering
    \caption{Model architecture of CLMs.}
    \label{tab:app:clm-arch}
    \begin{small}
        \begin{tabular}{@{}lcccccc@{}}
            \toprule
            Model & Vocab size             & Model dim & FFN dim & \#Hidden layers & \#Attention heads & \#Key/value heads \\ \midrule
            60M   & \multirow{5}{*}{49152} & 384       & 1408    & 22              & 4                 & 2                 \\
            200M  &                        & 768       & 2688    & 24              & 12                & 4                 \\
            400M  &                        & 960       & 3328    & 30              & 15                & 5                 \\
            600M  &                        & 1152      & 4096    & 32              & 18                & 6                 \\
            1.1B  &                        & 1536      & 5376    & 32              & 24                & 8                 \\ \bottomrule
        \end{tabular}
    \end{small}
\end{table}

\paragraph{Model architecture and tokenizer}
We adopt the Llama-3 architecture, which incorporates group query attention (GQA) and rotary positional embedding (RoPE)~\citep{dubey2024llama3}. \Cref{tab:app:clm-arch} summarizes the model architecture of our CLM series. We use \texttt{HuggingFaceTB/cosmo2-tokenizer}
\footnote{\texttt{cosmo2-tokenizer}: \url{https://huggingface.co/HuggingFaceTB/cosmo2-tokenizer}},
an open-source tokenizer trained on 1M tokens and released by HuggingFace.

\begin{figure}[h]
    \centering
    \includegraphics[width=0.5\textwidth]{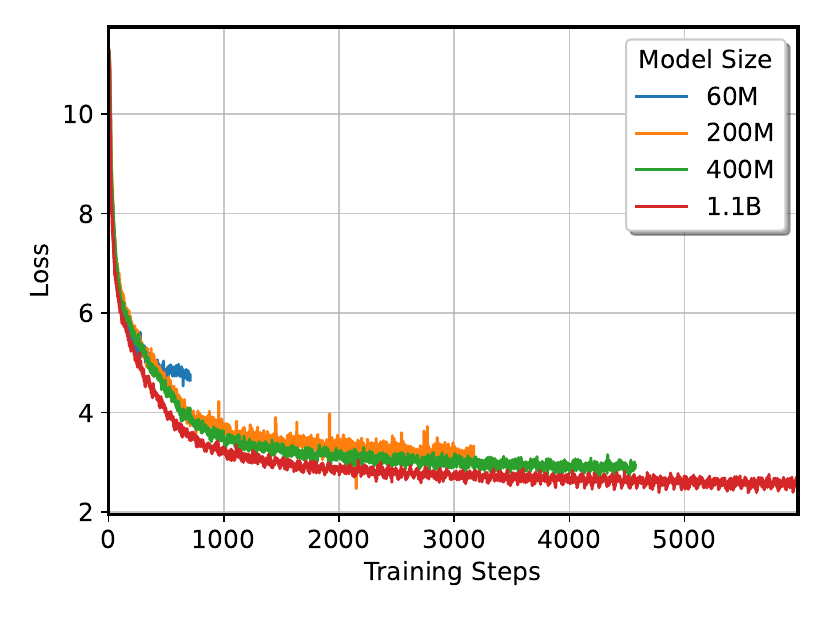}
    \caption{Pre-training loss of CLM series.}
    \label{fig:app:clm-pretrain-loss}
\end{figure}

\paragraph{Pre-training setup}
We pre-train our models on FineWeb-Edu~\citep{lozhkov2024fineweb-edu}, a pre-training dataset of cleaned and deduplicated english web data from CommonCrawl. We follow Chinchilla~\citep{chincilla} to determine the number of tokens for pre-training, \ie{}, multiplying the number of parameters (without embedding layers) by 22. We use the AdamW optimizer with a linear learning rate schedule with 20\% warmup. The initial learning rate is set to 1e-4 and the sequence length is set to 2048. We use various batch sizes for different models, \ie{}, 96 for 60M, 64 for 200M, 96 for 400M, 96 for 600M, and 192 for 1.1B. \Cref{fig:app:clm-pretrain-loss} shows the pre-training loss of our CLM series. We will open-source our pre-trained models and tokenizer once the paper is accepted.

\paragraph{Pretraining framework}
We use TorchTitan~\citep{liang2025torchtitan}, a PyTorch-based distributed training framework to pretrain our models. We apply fully sharded data parallel (FSDP) and gradient checkpointing to save GPU memory. Eight NVIDIA H200 GPUs were used for pre-training 600M and 1.1B models, and eight NVIDIA A100 GPUs were used for pre-training 60M, 200M, and 400M models. Before conducting the mixed quantization experiments, we convert the pre-trained models from torch distributed format to HuggingFace format.

\section{Distribution for Search Space}
\label{app:search-space-distribution}
\begin{figure}[htbp]
\centering
\begin{subfigure}[b]{0.24\textwidth}
    \centering
    \includegraphics[width=\textwidth]{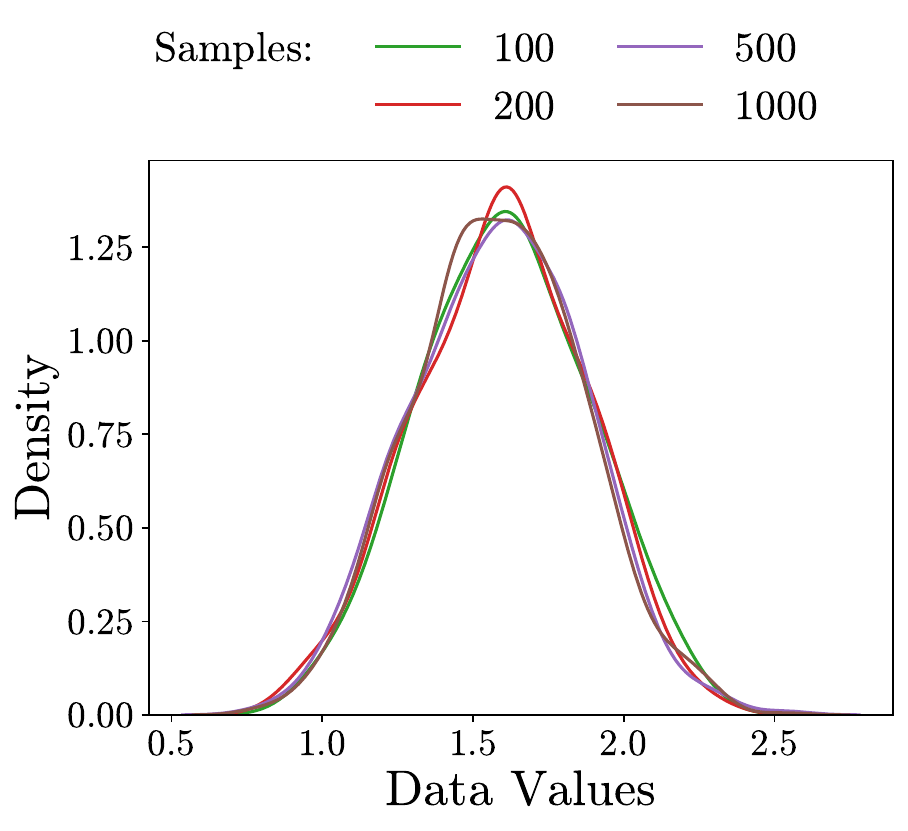}
    \captionsetup{font=scriptsize}
    \caption{CLM-1.1B, $Q_r = 0.5$}
    \label{fig:vary-ratio-start}
\end{subfigure}\hfill
\begin{subfigure}[b]{0.24\textwidth}
    \centering
    \includegraphics[width=\textwidth]{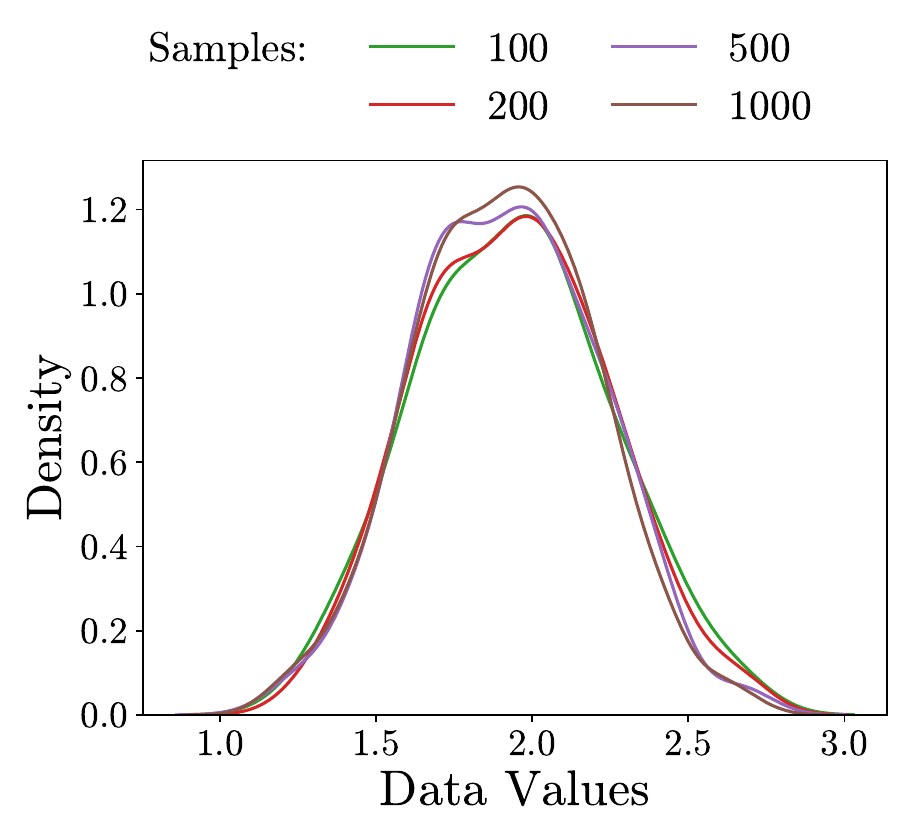}
    \captionsetup{font=scriptsize}
    \caption{CLM-1.1B, $Q_r = 0.6$}
\end{subfigure}\hfill
\begin{subfigure}[b]{0.24\textwidth}
    \centering
    \includegraphics[width=\textwidth]{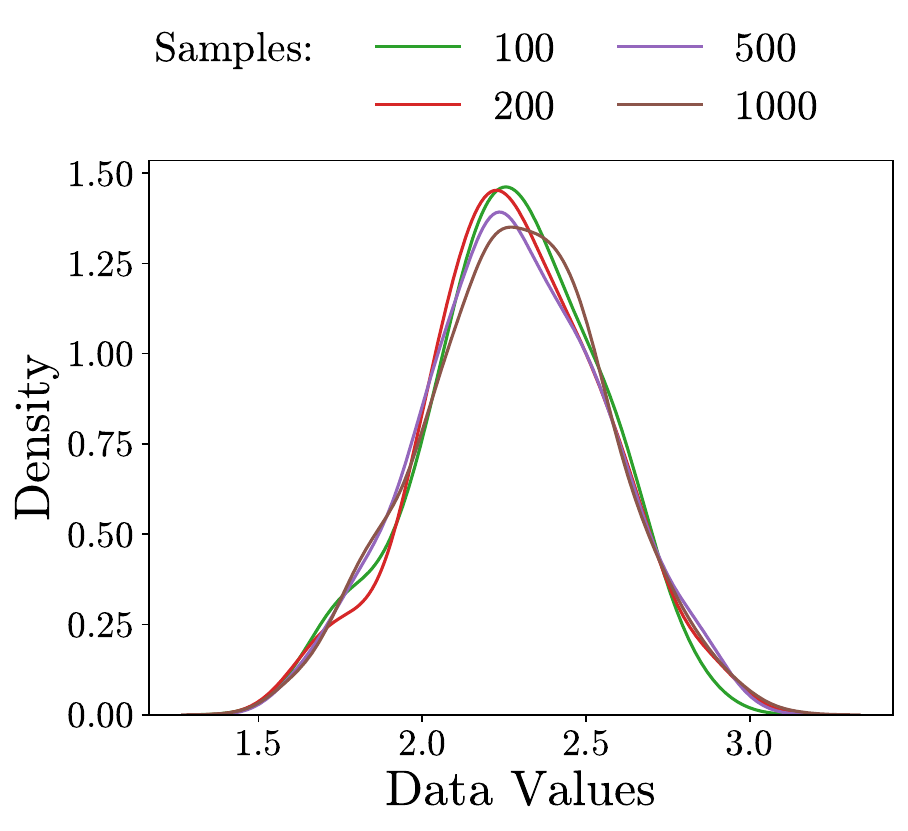}
    \captionsetup{font=scriptsize}
    \caption{CLM-1.1B, $Q_r = 0.7$}
\end{subfigure}\hfill
\begin{subfigure}[b]{0.24\textwidth}
    \centering
    \includegraphics[width=\textwidth]{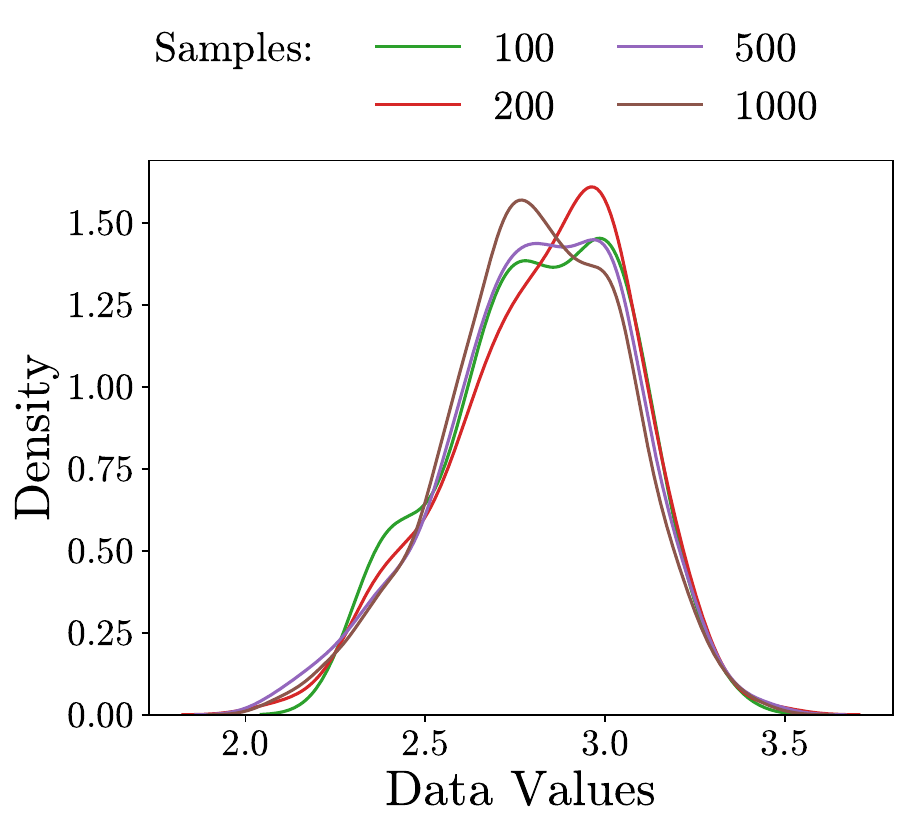}
    \captionsetup{font=scriptsize}
    \caption{CLM-1.1B, $Q_r = 0.8$}

\end{subfigure}

\begin{subfigure}[b]{0.24\textwidth}
    \centering
    \includegraphics[width=\textwidth]{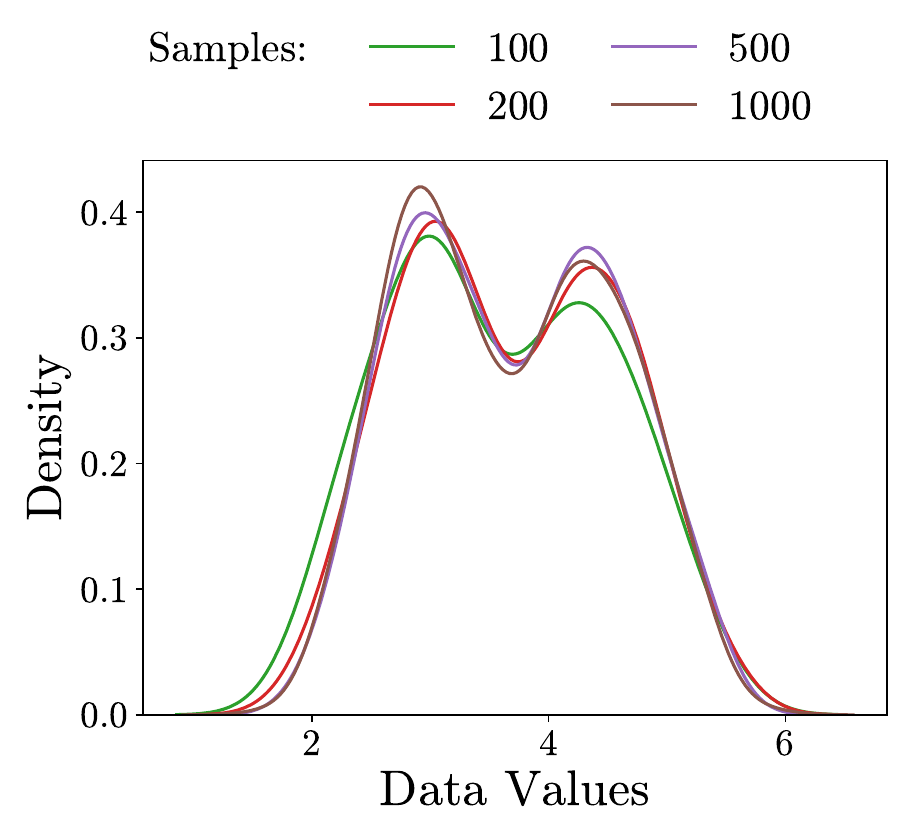}
    \captionsetup{font=scriptsize}
    \caption{CLM-200M, $Q_r = 0.5$}
\end{subfigure}\hfill
\begin{subfigure}[b]{0.24\textwidth}
    \centering
    \includegraphics[width=\textwidth]{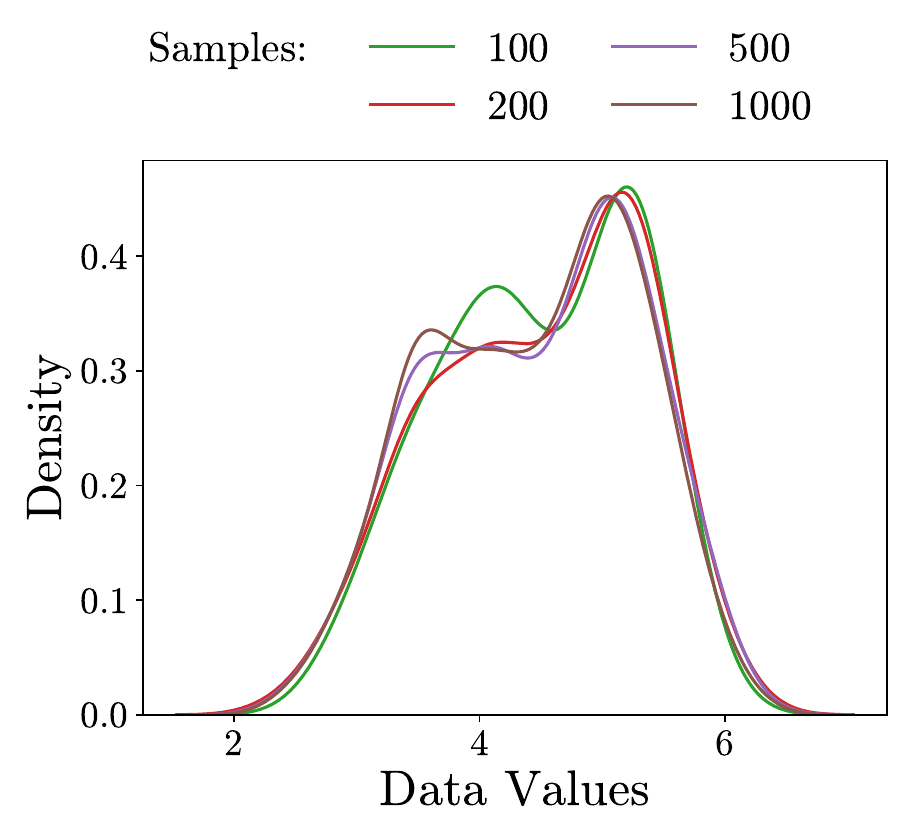}
    \captionsetup{font=scriptsize}
    \caption{CLM-200M, $Q_r = 0.6$}
\end{subfigure}\hfill
\begin{subfigure}[b]{0.24\textwidth}
    \centering
    \includegraphics[width=\textwidth]{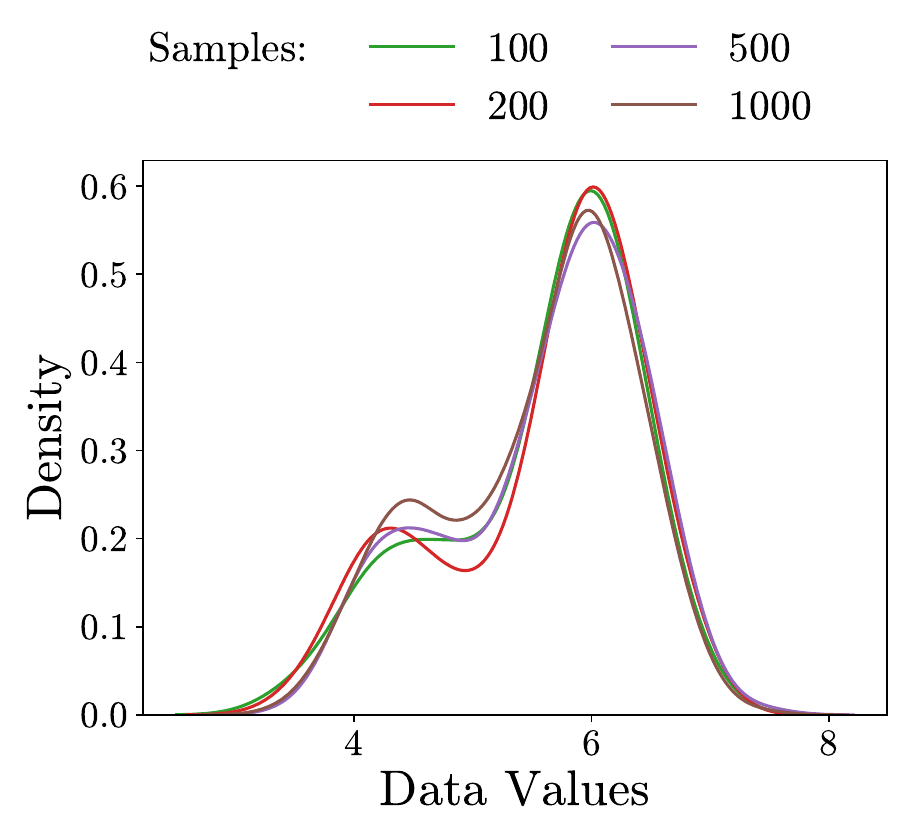}
    \captionsetup{font=scriptsize}
    \caption{CLM-200M, $Q_r = 0.7$}
\end{subfigure}\hfill
\begin{subfigure}[b]{0.24\textwidth}
    \centering
    \includegraphics[width=\textwidth]{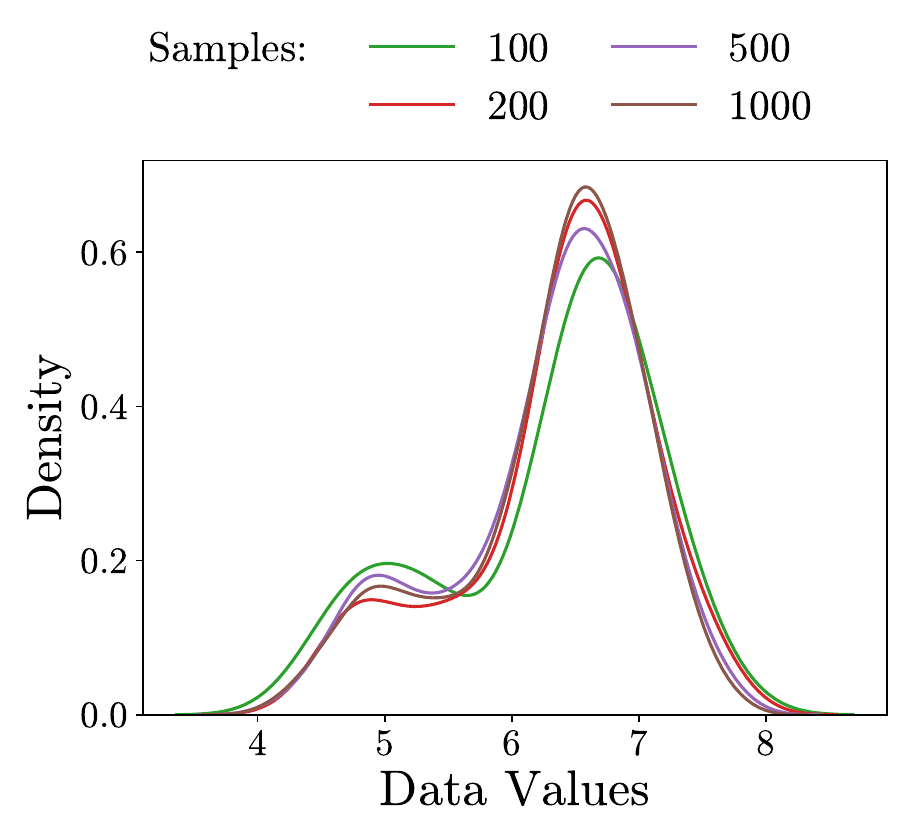}
    \captionsetup{font=scriptsize}
    \caption{CLM-200M, $Q_r = 0.8$}
    \label{fig:vary-ratio-end}    
\end{subfigure}

\begin{subfigure}[b]{0.24\textwidth}
    \centering
    \includegraphics[width=\textwidth]{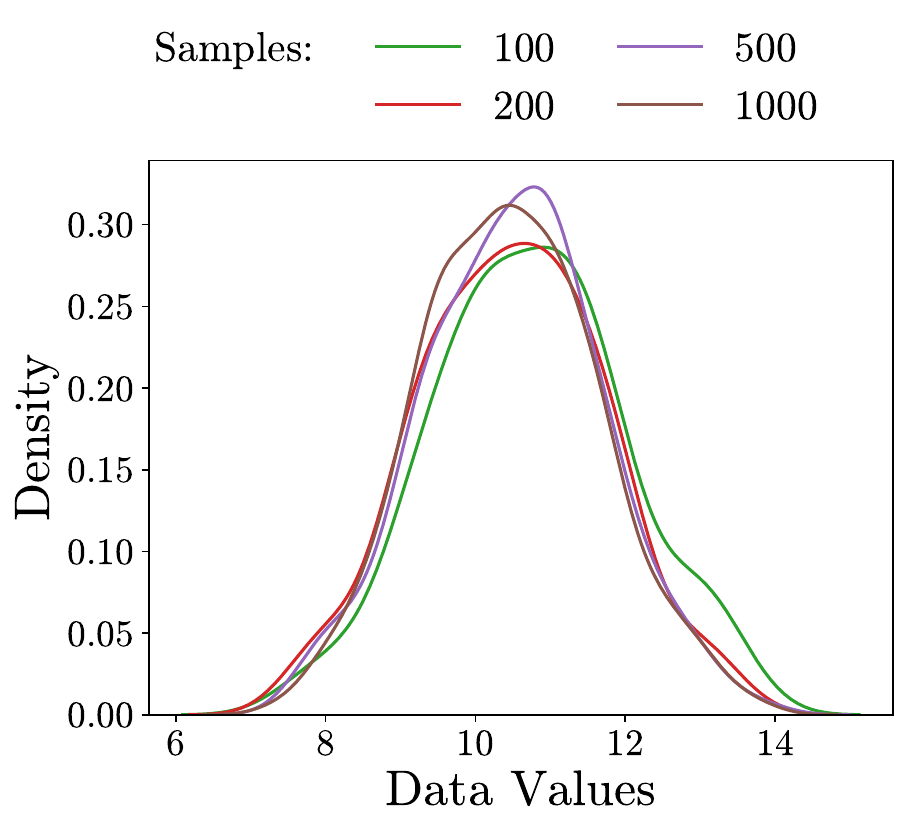}
    \captionsetup{font=scriptsize}
    \caption{CLM-60M, $Q_r = 0.5$}
    \label{fig:vary-size-start}  
\end{subfigure}\hfill
\begin{subfigure}[b]{0.24\textwidth}
    \centering
    \includegraphics[width=\textwidth]{figures/distribution/clm-200m-0.5.pdf}
    \captionsetup{font=scriptsize}
    \caption{CLM-200M, $Q_r = 0.5$}
\end{subfigure}\hfill
\begin{subfigure}[b]{0.24\textwidth}
    \centering
    \includegraphics[width=\textwidth]{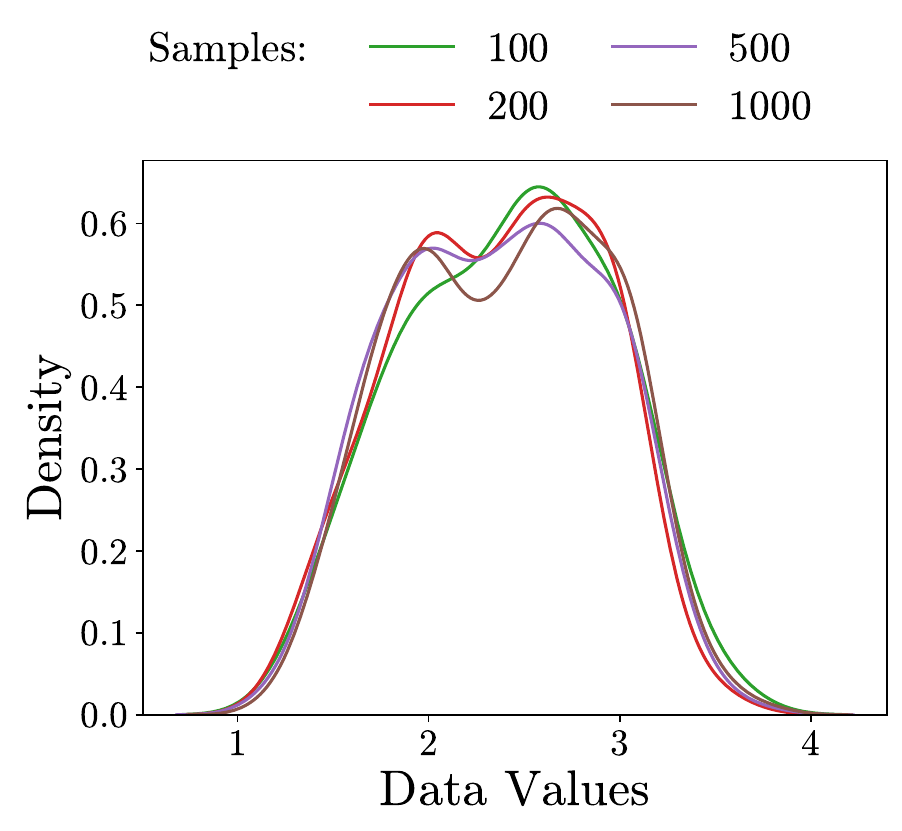}
    \captionsetup{font=scriptsize}
    \caption{CLM-400M, $Q_r = 0.5$}
\end{subfigure}\hfill
\begin{subfigure}[b]{0.24\textwidth}
    \centering
    \includegraphics[width=\textwidth]{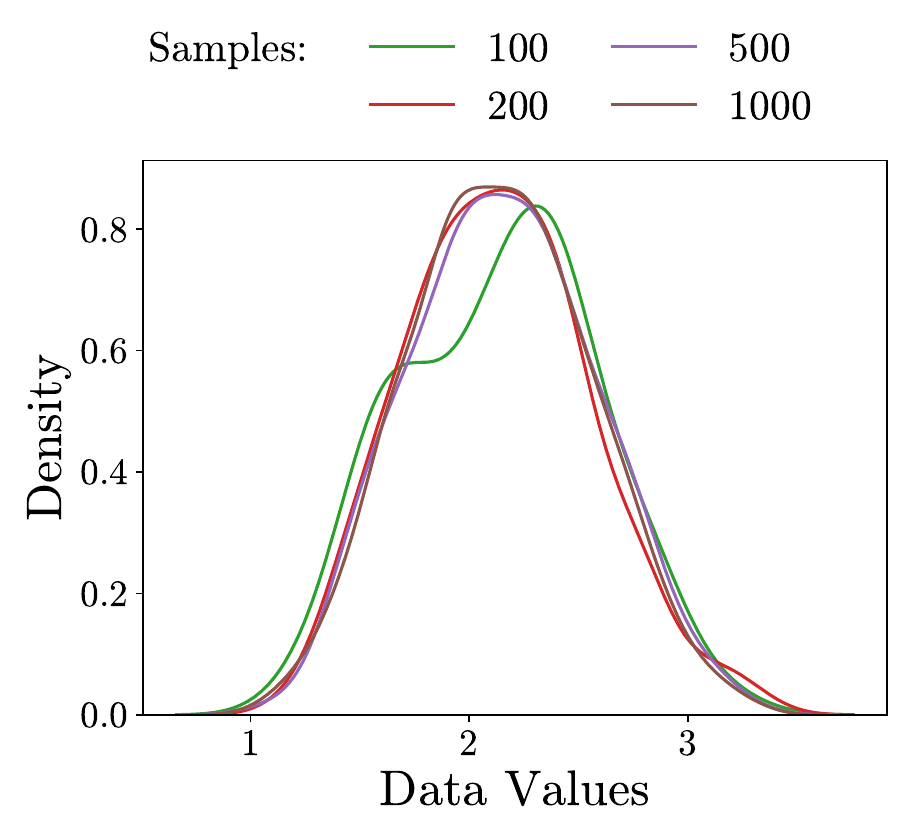}
    \captionsetup{font=scriptsize}
    \caption{CLM-600M, $Q_r = 0.5$}
\end{subfigure}

\begin{subfigure}[b]{0.24\textwidth}
    \centering
    \includegraphics[width=\textwidth]{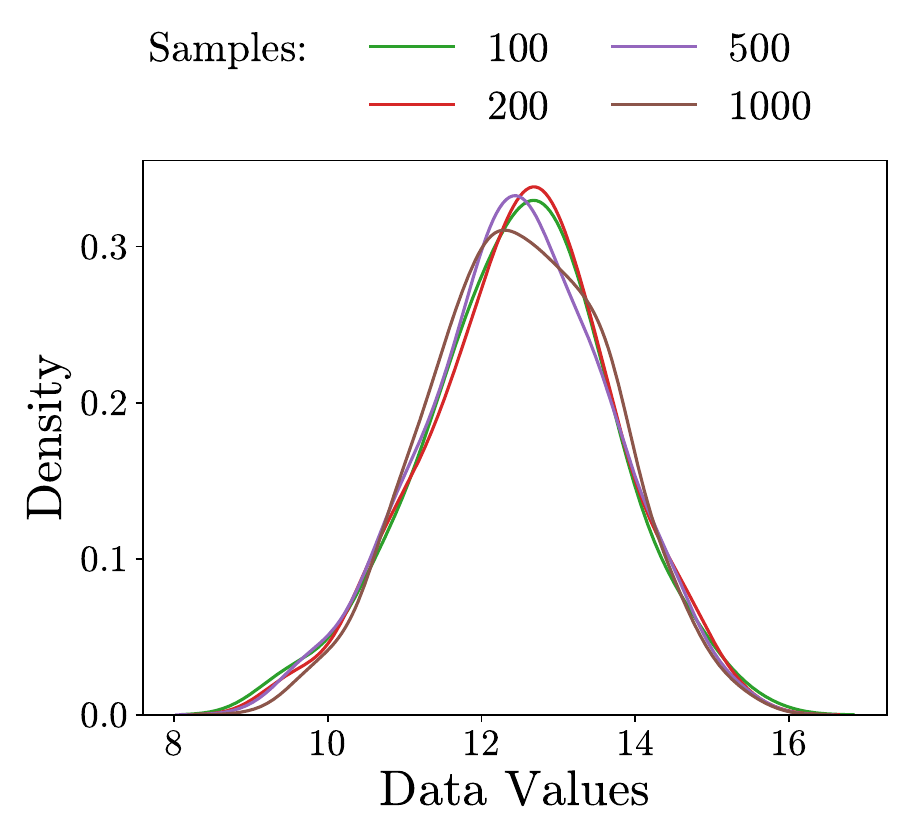}
    \captionsetup{font=scriptsize}
    \caption{CLM-60M, $Q_r = 0.6$}
\end{subfigure}\hfill
\begin{subfigure}[b]{0.24\textwidth}
    \centering
    \includegraphics[width=\textwidth]{figures/distribution/clm-200m-0.6.pdf}
    \captionsetup{font=scriptsize}
    \caption{CLM-200M, $Q_r = 0.6$}
\end{subfigure}\hfill
\begin{subfigure}[b]{0.24\textwidth}
    \centering
    \includegraphics[width=\textwidth]{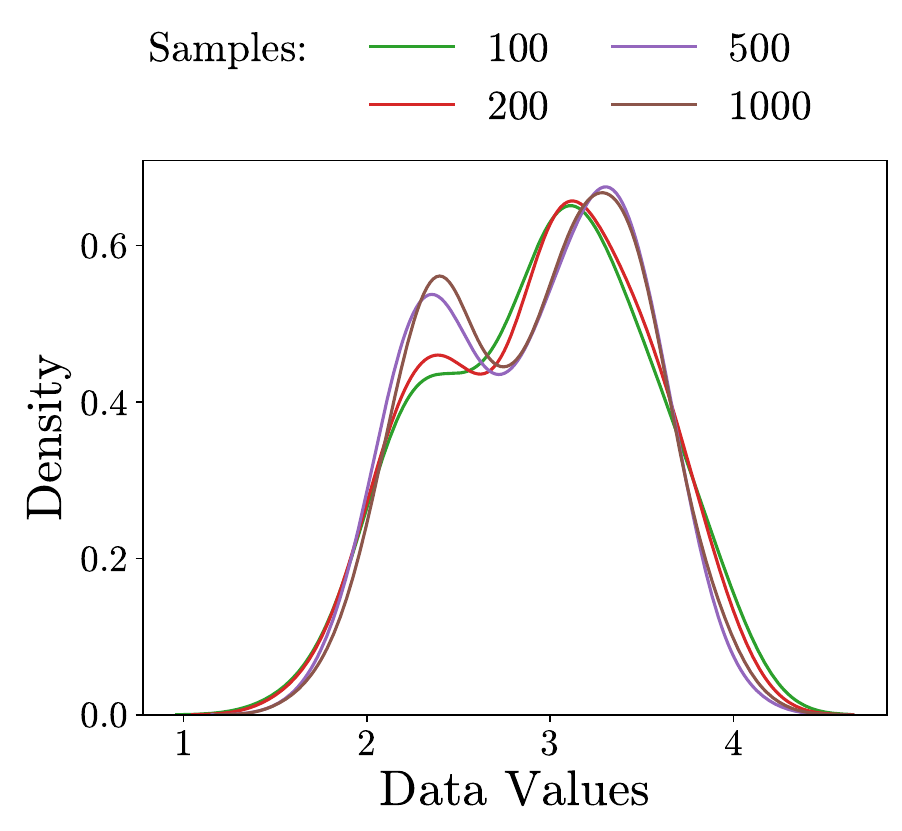}
    \captionsetup{font=scriptsize}
    \caption{CLM-400M, $Q_r = 0.6$}
\end{subfigure}\hfill
\begin{subfigure}[b]{0.24\textwidth}
    \centering
    \includegraphics[width=\textwidth]{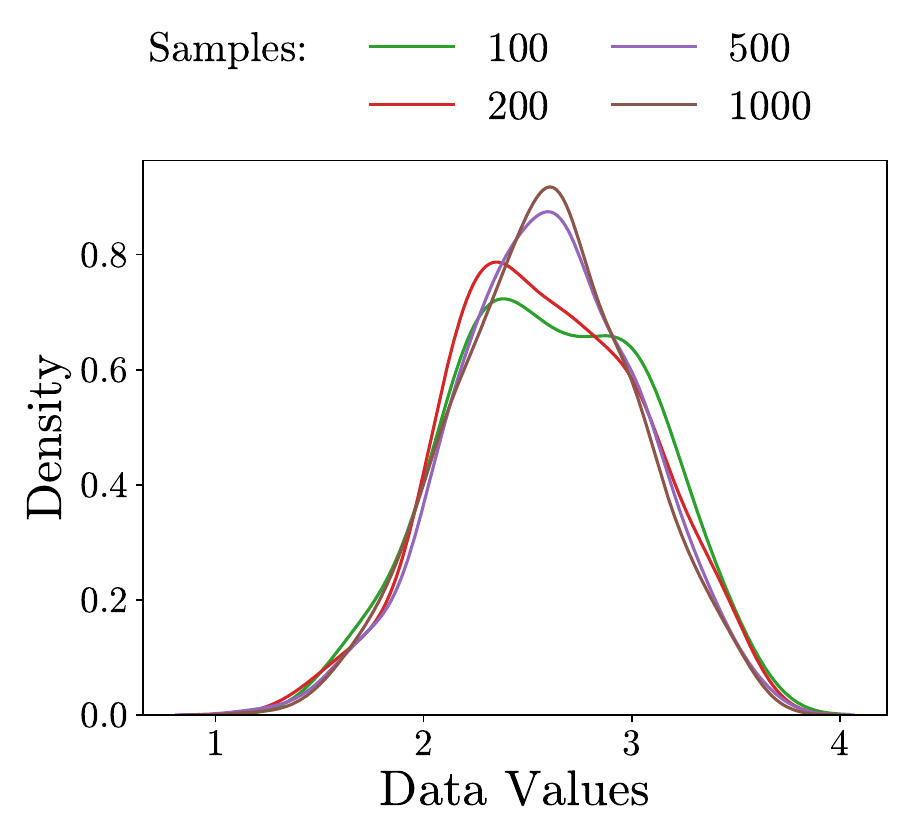}
    \captionsetup{font=scriptsize}
    \caption{CLM-600M, $Q_r = 0.6$}
    \label{fig:vary-size-end}   
\end{subfigure}

\caption{\textbf{KDE of distribution $\Delta$.} Different KDE for different $Q_r$ under a fixed models: (a,b,c,d) Changing $Q_r$ for CLM-1.1B; (e,f,g,h) Changing $Q_r$ for CLM-200M. KDE for different model under fixed $Q_r$; (i,j,k,l) Changing $N$ for $Q_r=0.5$; (m,n,o,p) Changing $N$ for $Q_r=0.6$.}
\label{fig:appendix-distribution}
\end{figure}

\Cref{fig:appendix-distribution} present the kernel density estimates (KDEs) of the distribution $\Delta$ under MXINT-4  layerwise quantization, evaluated across varying numbers of trials: 100, 200, 500, and 1000. These KDEs are generated using a Gaussian kernel, which may not be the optimal kernel, providing a consistent and smooth estimate that is sufficient for comparative analysis.

We present \Crefrange{fig:vary-ratio-start}{fig:vary-ratio-end}, which shows the KDEs of different quantization ratios $Q_r$ (from 0.5 to 0.8) for a fixed model (CLM-1.1B and CLM-200M), whereas \Crefrange{fig:vary-size-start}{fig:vary-size-end} shows the KDEs of different model sizes (from 60M to 600M) under fixed $Q_r$(0.5 and 0.6). Across all figures, as the sample size increases from 100 to 1000, the estimated distribution tends to remain the same. The KDEs generated appear nearly indistinguishable in shape from that of 100. This consistency holds throughout different $Q_r$ and model sizes.

\textbf{These figures suggest that the empirical distribution of $\Delta$ can be reliably estimated with as few as 100 samples}, as presented in~\Cref{sec:exp:setup}. That is, even with relatively small sample sizes, the mean ($\mathbb{E}(\Delta)$) and minimum ($\delta^{\text{opt}}$) values of $\Delta$ can still be well captured. Therefore, from both a computational and statistical efficiency perspective, excessive sampling provides diminishing returns. This insight is particularly valuable since large-scale sampling is computationally expensive.

\section{Layer-wise Results}
\label{app:layer-result}
\begin{figure}[htbp]
\centering
\begin{subfigure}[b]{0.3\textwidth} \centering
\includegraphics[width=\textwidth]{figures/figures-main/llama-layer/logloss_vs_logpara-size.pdf}
\captionsetup{font=scriptsize}
\caption{CLM Actual Loss}
\end{subfigure}
\hfill
\begin{subfigure}[b]{0.3\textwidth} \centering
\includegraphics[width=\textwidth]{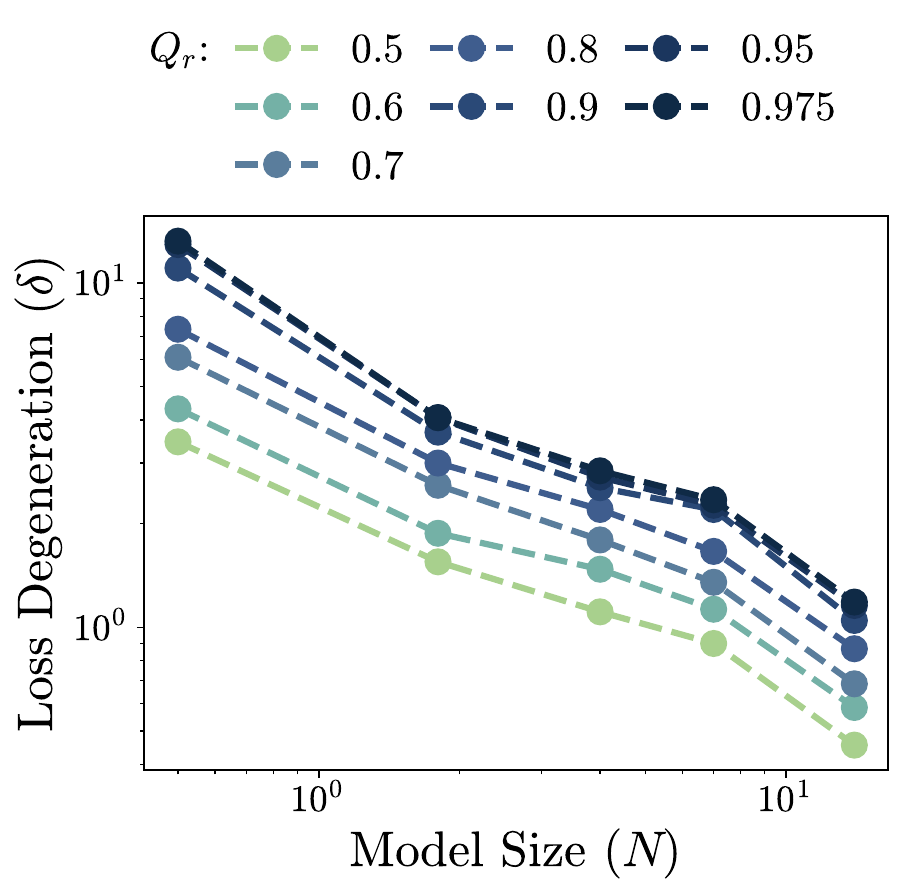}
\captionsetup{font=scriptsize}
\caption{Qwen-1.5 Actual Loss}
\end{subfigure}
\hfill
\begin{subfigure}[b]{0.3\textwidth} \centering
\includegraphics[width=\textwidth]{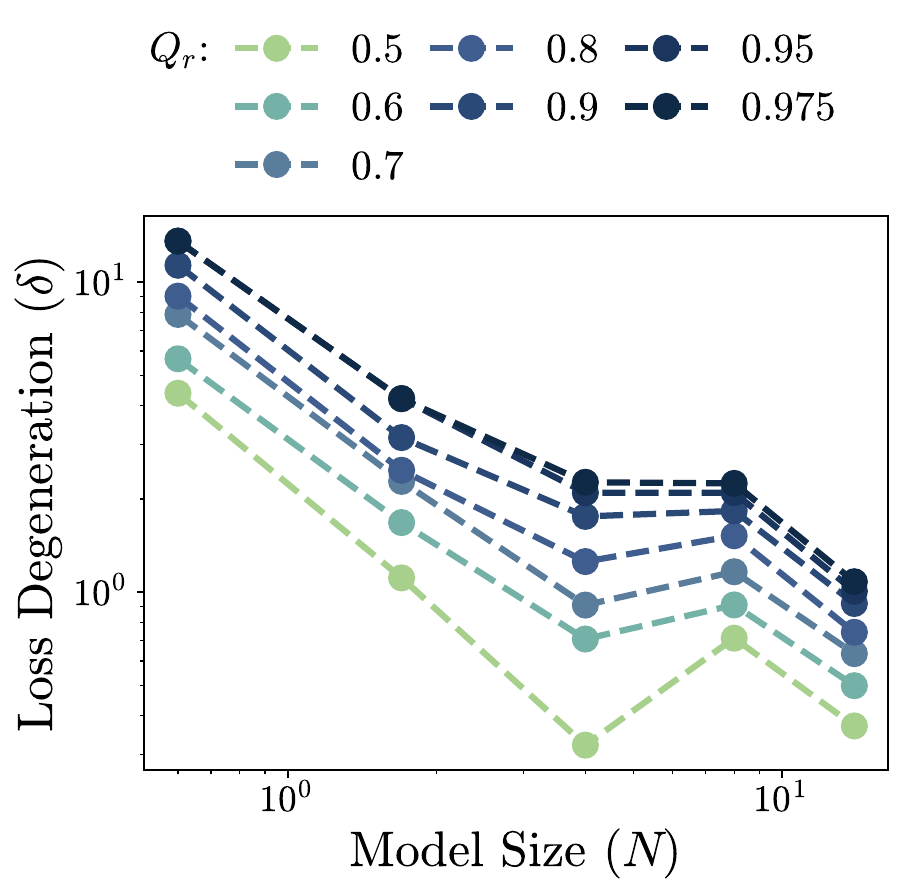}
\captionsetup{font=scriptsize}
\caption{Qwen-3 Actual Loss}
\end{subfigure}

\begin{subfigure}[b]{0.3\textwidth} \centering
\includegraphics[width=\textwidth]{figures/figures-main/llama-layer/logloss_vs_logpara-size-fitted.pdf}
\captionsetup{font=scriptsize}
\caption{CLM Predicted Loss}
\end{subfigure}
\hfill
\begin{subfigure}[b]{0.3\textwidth} \centering
\includegraphics[width=\textwidth]{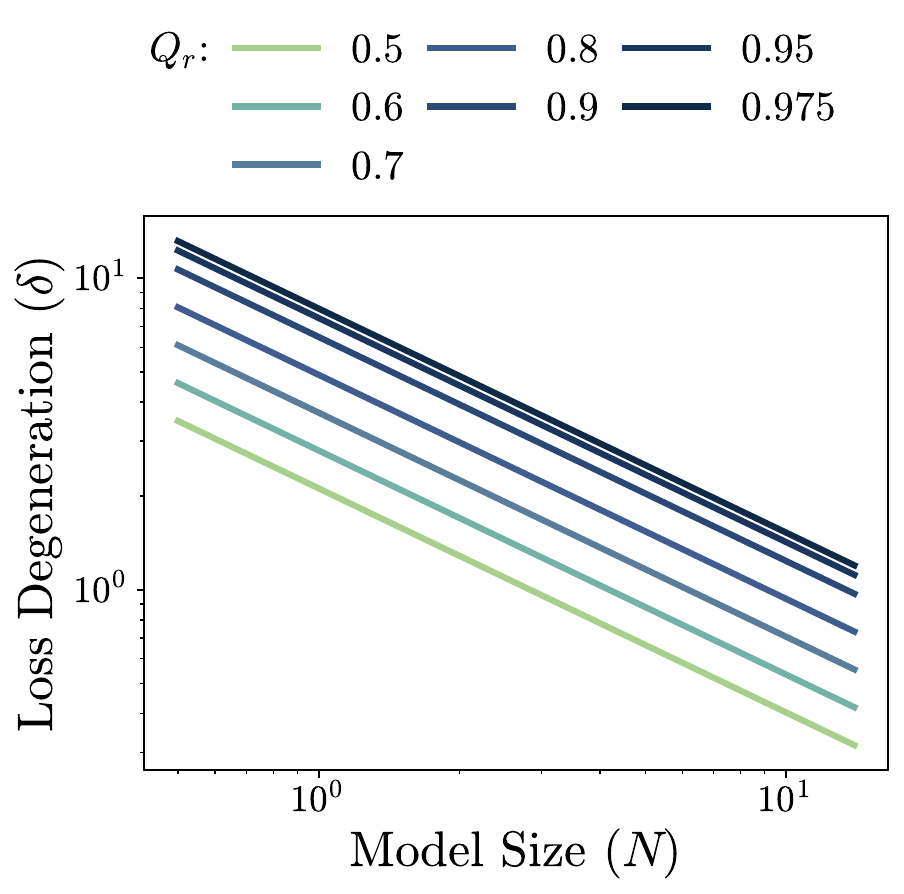}
\captionsetup{font=scriptsize}
\caption{Qwen-1.5 Predicted Loss}
\end{subfigure}
\hfill
\begin{subfigure}[b]{0.3\textwidth} \centering
\includegraphics[width=\textwidth]{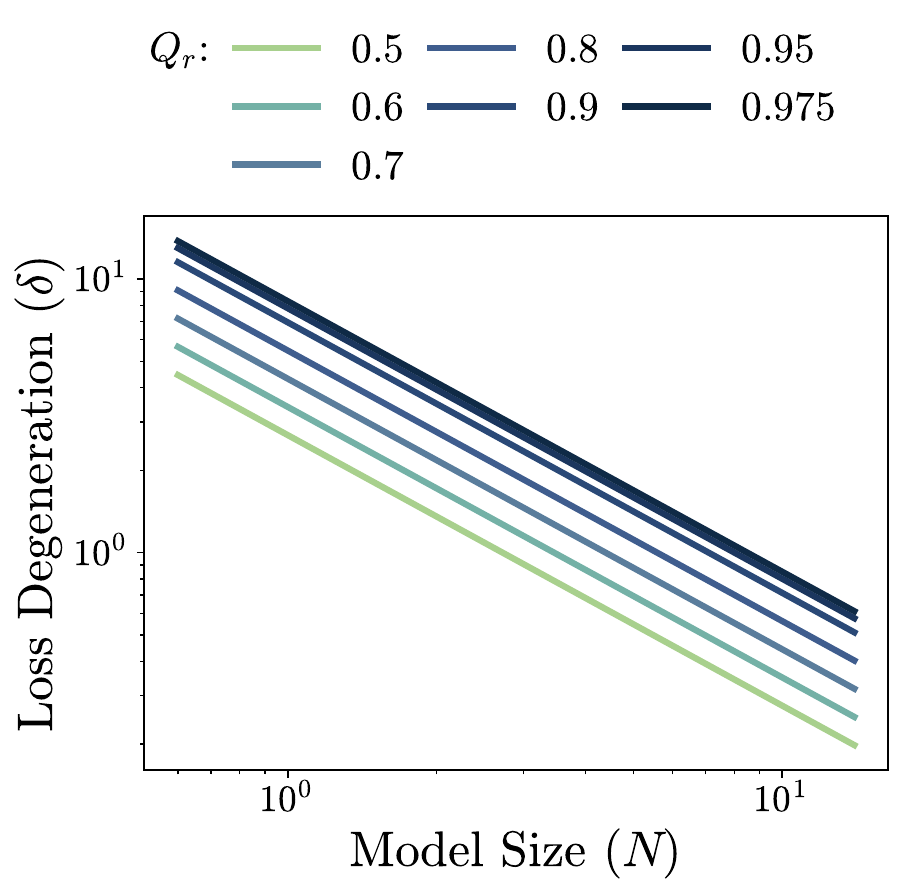}
\captionsetup{font=scriptsize}
\caption{Qwen-3 Predicted Loss}
\end{subfigure}

\begin{subfigure}[b]{0.3\textwidth} \centering
\includegraphics[width=\textwidth]{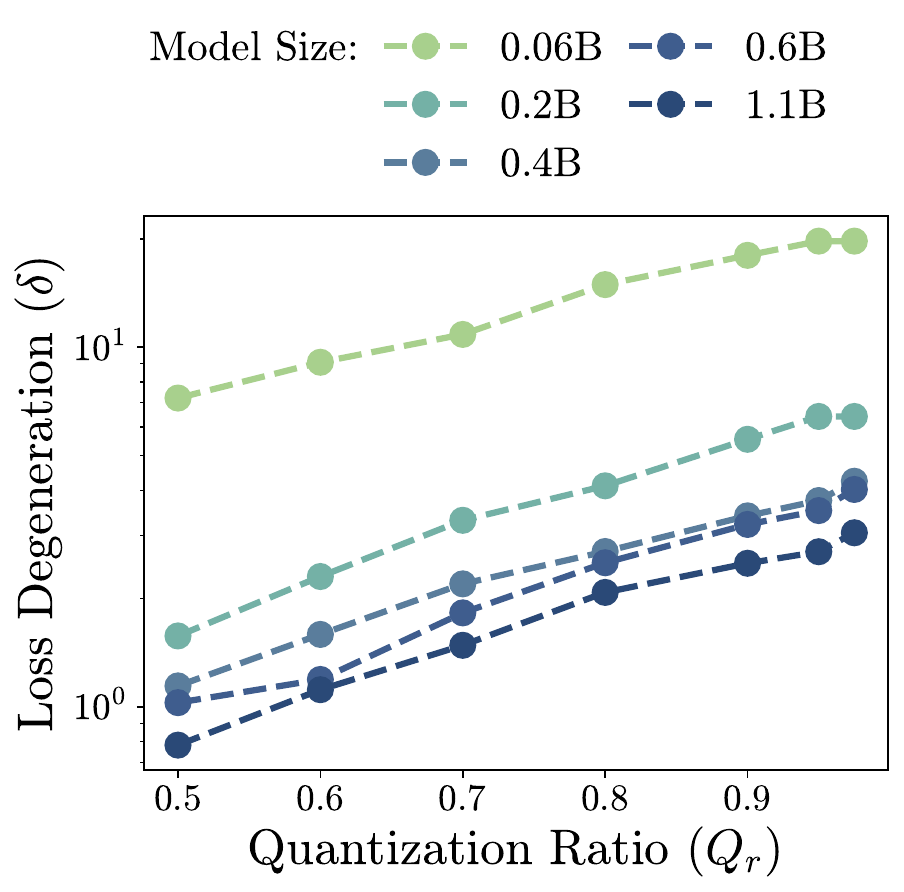}
\captionsetup{font=scriptsize}
\caption{CLM Actual Loss}
\end{subfigure}
\hfill
\begin{subfigure}[b]{0.3\textwidth} \centering
\includegraphics[width=\textwidth]{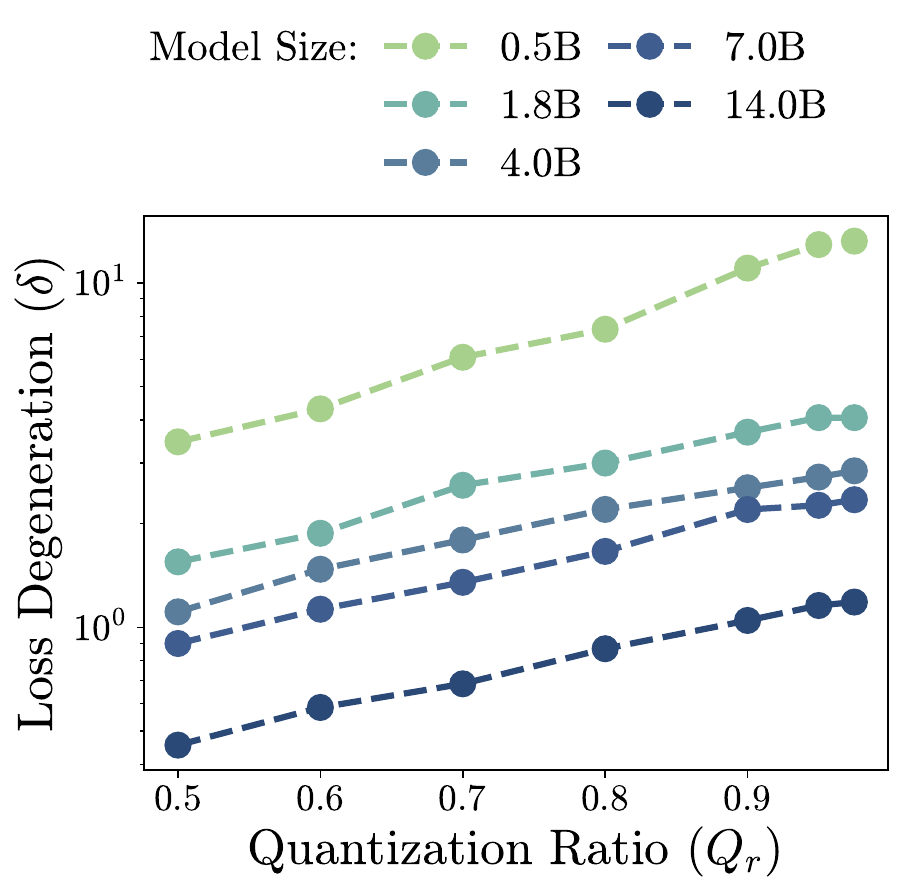}
\captionsetup{font=scriptsize}
\caption{Qwen-1.5 Actual Loss}
\end{subfigure}
\hfill
\begin{subfigure}[b]{0.3\textwidth} \centering
\includegraphics[width=\textwidth]{figures/figures-main/qwen3-layer/logloss_vs_qratio.pdf}
\captionsetup{font=scriptsize}
\caption{Qwen-3 Actual Loss}
\end{subfigure}

\begin{subfigure}[b]{0.3\textwidth} \centering
\includegraphics[width=\textwidth]{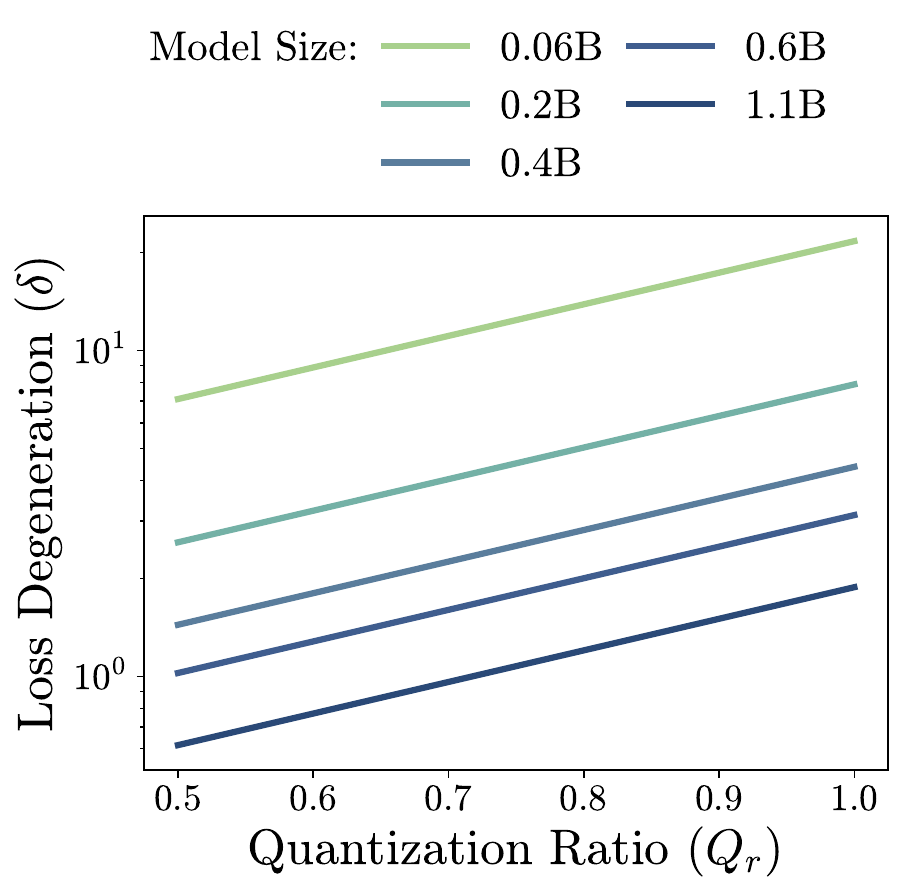}
\captionsetup{font=scriptsize}
\caption{CLM Predicted Loss}
\end{subfigure}
\hfill
\begin{subfigure}[b]{0.3\textwidth} \centering
\includegraphics[width=\textwidth]{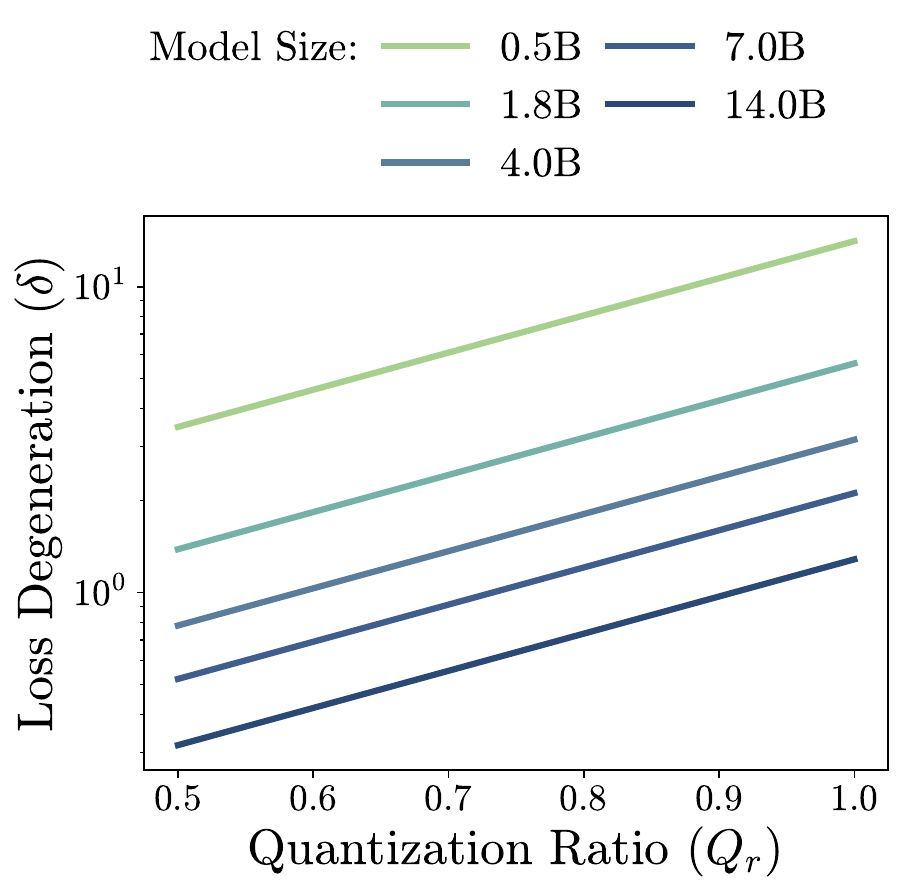}
\captionsetup{font=scriptsize}
\caption{Qwen-1.5 Predicted Loss}
\end{subfigure}
\hfill
\begin{subfigure}[b]{0.3\textwidth} \centering
\includegraphics[width=\textwidth]{figures/figures-main/qwen3-layer/logloss_vs_qratio-fitted.pdf}
\captionsetup{font=scriptsize}
\caption{Qwen-3 Predicted Loss}
\end{subfigure}

\caption{\textbf{Layer-wise ($\delta^{\text{opt}}$)} (a,d,g,h) CLM layer-wise results; (b,e,h,k) Qwen-1.5 layer-wise results; (c,f,i,l) Qwen-3 layer-wise results.}
  \label{fig:appendix-layerwise}
\end{figure}

\begin{figure}[htbp]
\centering
\begin{subfigure}[b]{0.3\textwidth} \centering
\includegraphics[width=\textwidth]{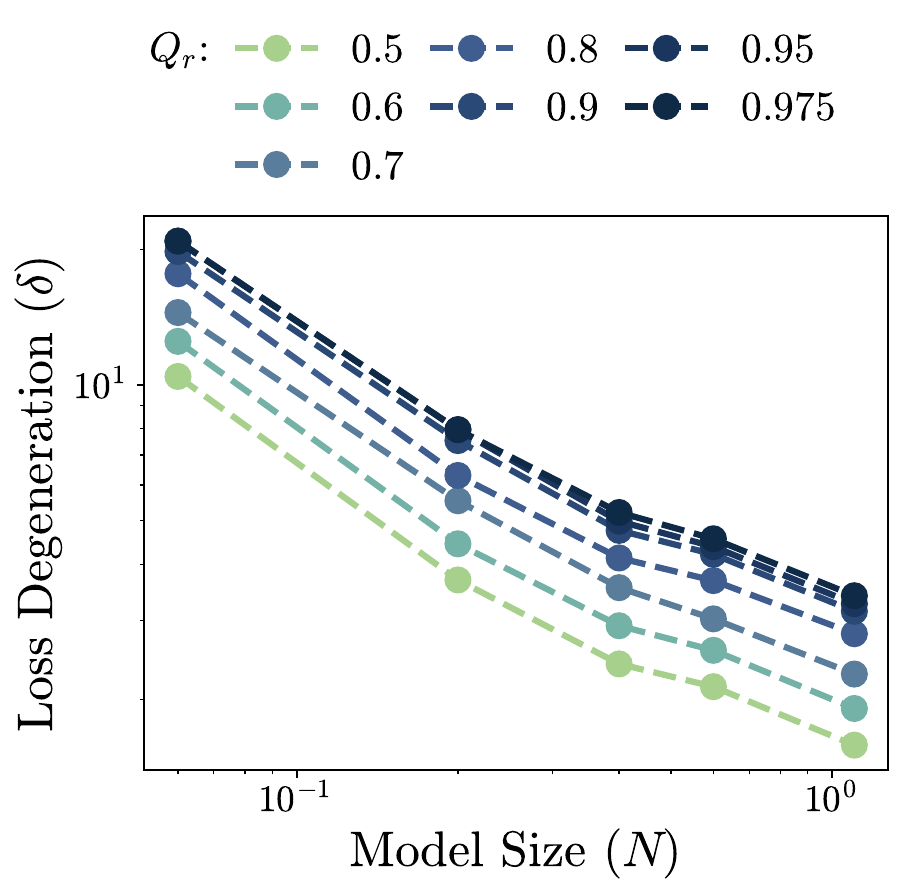}
\captionsetup{font=scriptsize}
\caption{CLM Actual Loss}
\end{subfigure}
\hfill
\begin{subfigure}[b]{0.3\textwidth} \centering
\includegraphics[width=\textwidth]{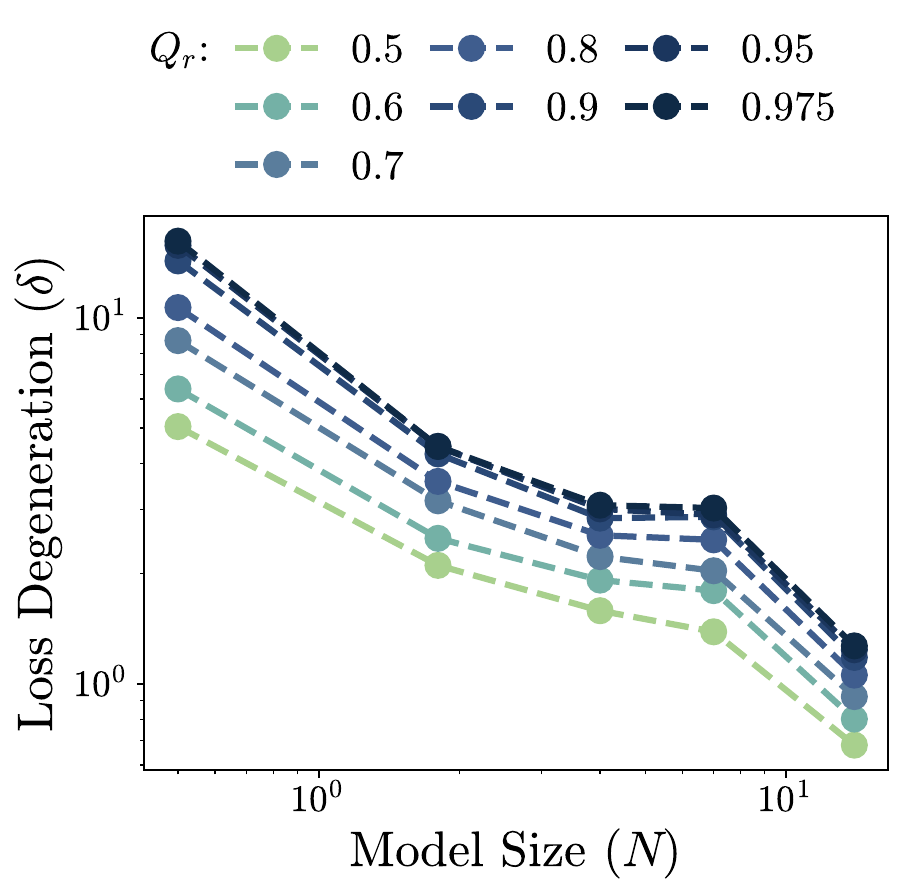}
\captionsetup{font=scriptsize}
\caption{Qwen-1.5 Actual Loss}
\end{subfigure}
\hfill
\begin{subfigure}[b]{0.3\textwidth} \centering
\includegraphics[width=\textwidth]{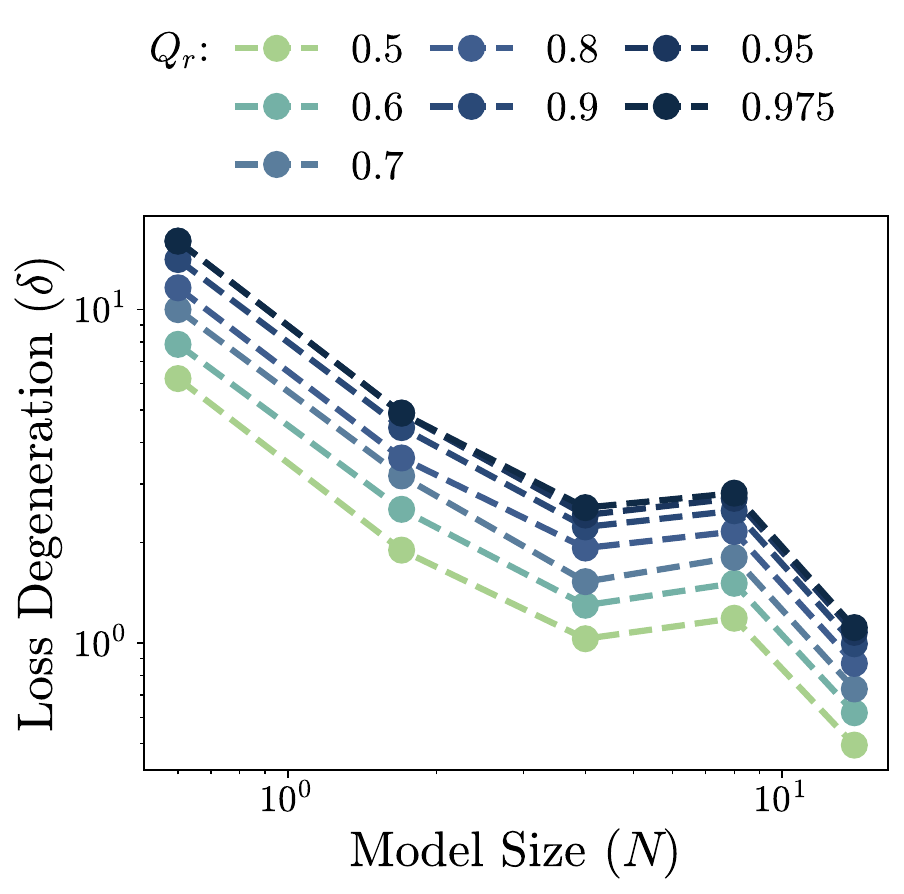}
\captionsetup{font=scriptsize}
\caption{Qwen-3 Actual Loss}
\end{subfigure}

\begin{subfigure}[b]{0.3\textwidth} \centering
\includegraphics[width=\textwidth]{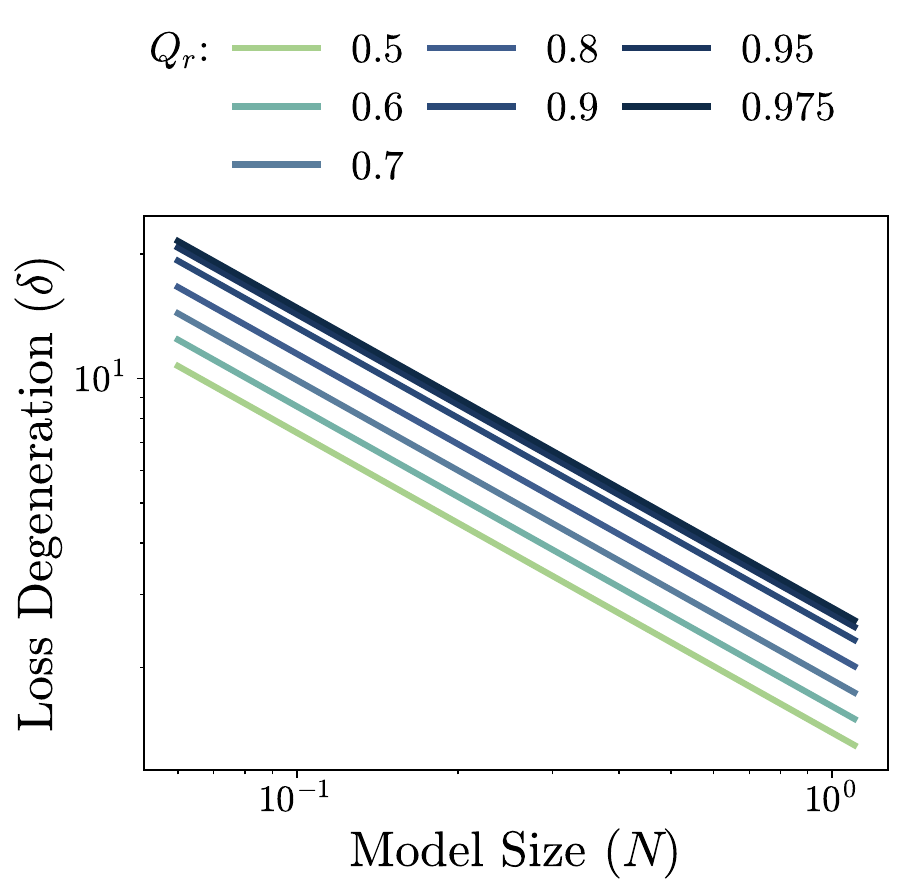}
\captionsetup{font=scriptsize}
\caption{CLM Predicted Loss}
\end{subfigure}
\hfill
\begin{subfigure}[b]{0.3\textwidth} \centering
\includegraphics[width=\textwidth]{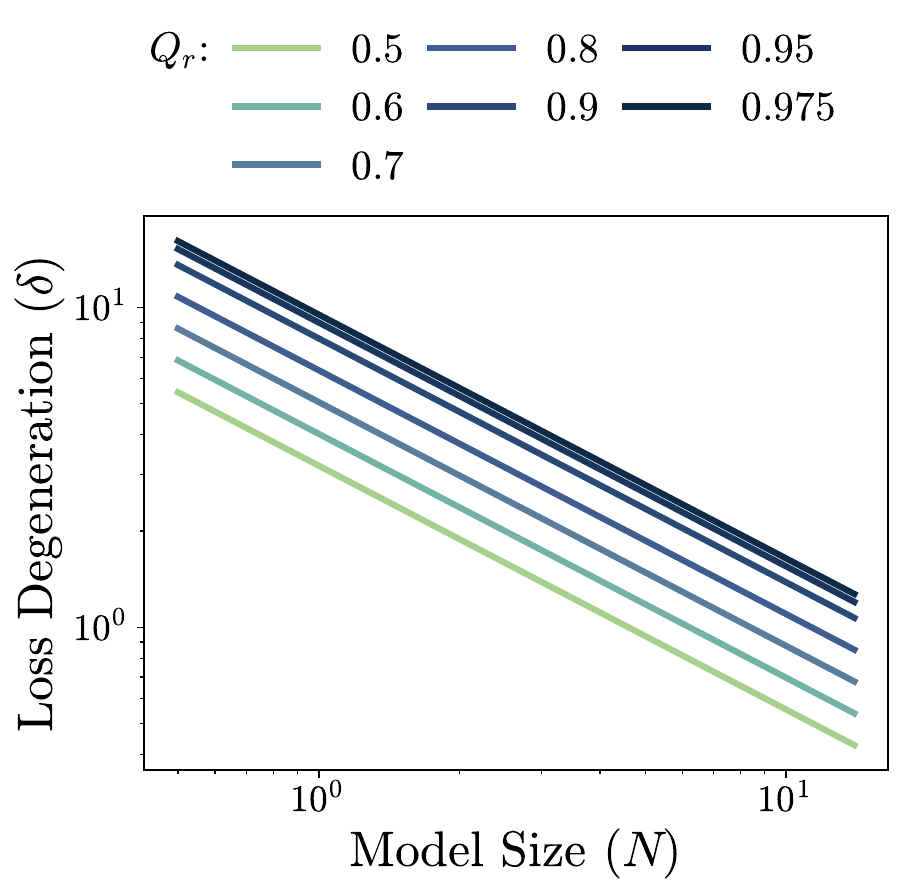}
\captionsetup{font=scriptsize}
\caption{Qwen-1.5 Predicted Loss}
\end{subfigure}
\hfill
\begin{subfigure}[b]{0.3\textwidth} \centering
\includegraphics[width=\textwidth]{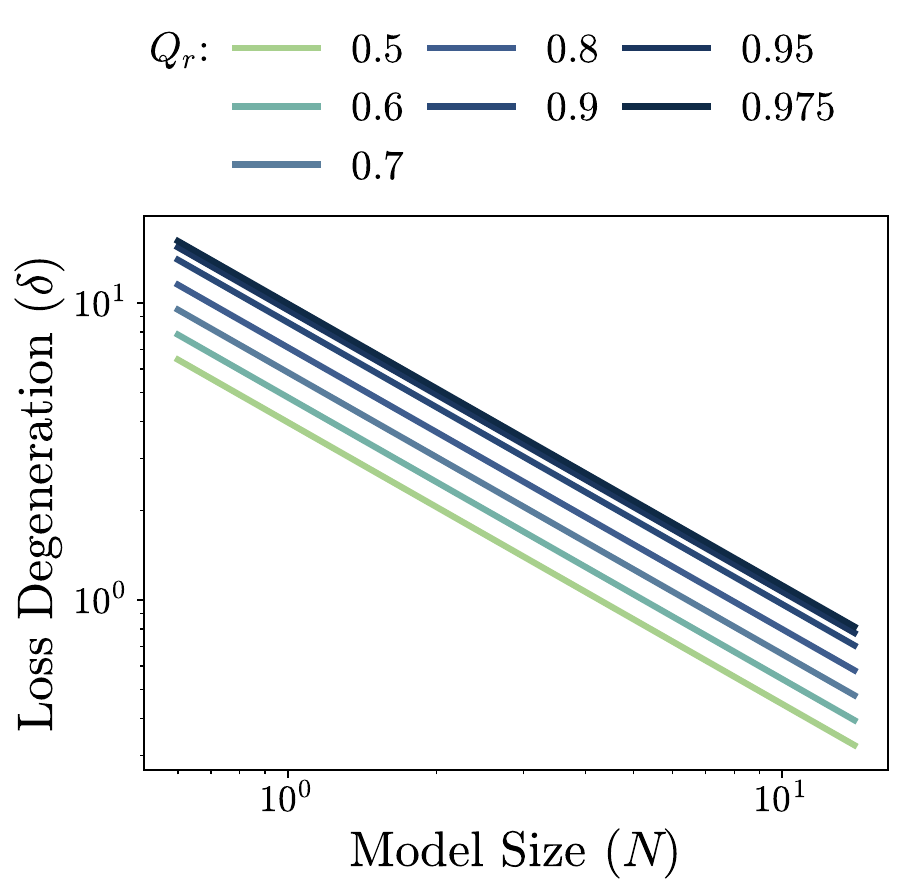}
\captionsetup{font=scriptsize}
\caption{Qwen-3 Predicted Loss}
\end{subfigure}

\begin{subfigure}[b]{0.3\textwidth} \centering
\includegraphics[width=\textwidth]{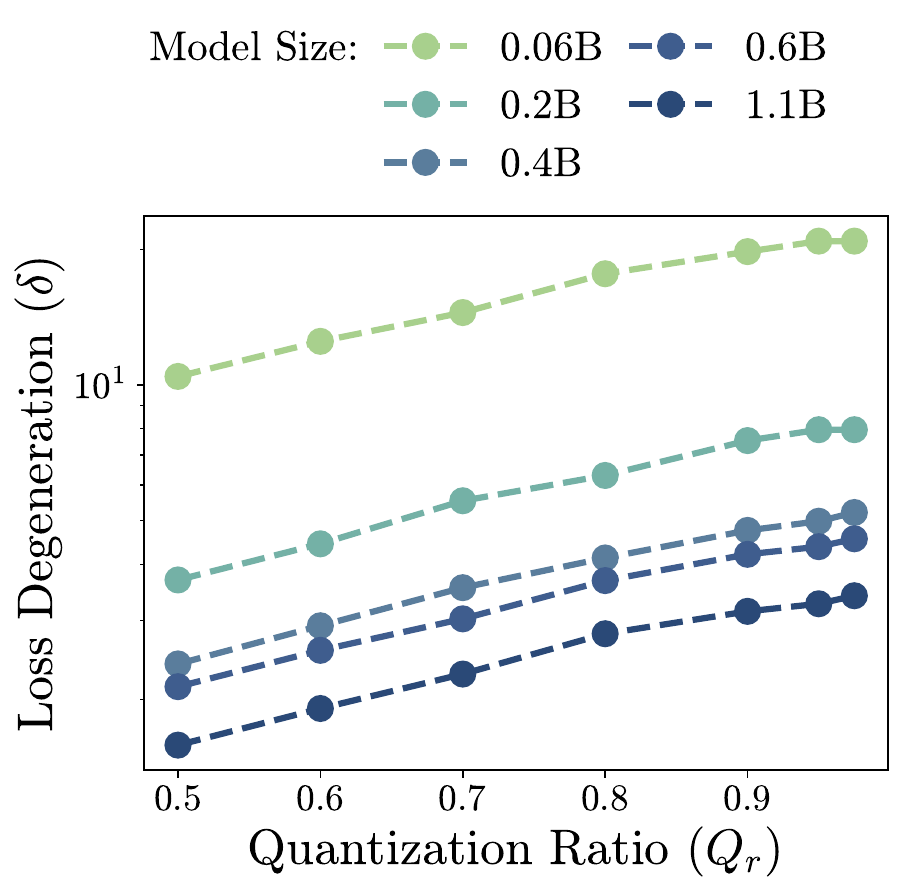}
\captionsetup{font=scriptsize}
\caption{CLM Actual Loss}
\end{subfigure}
\hfill
\begin{subfigure}[b]{0.3\textwidth} \centering
\includegraphics[width=\textwidth]{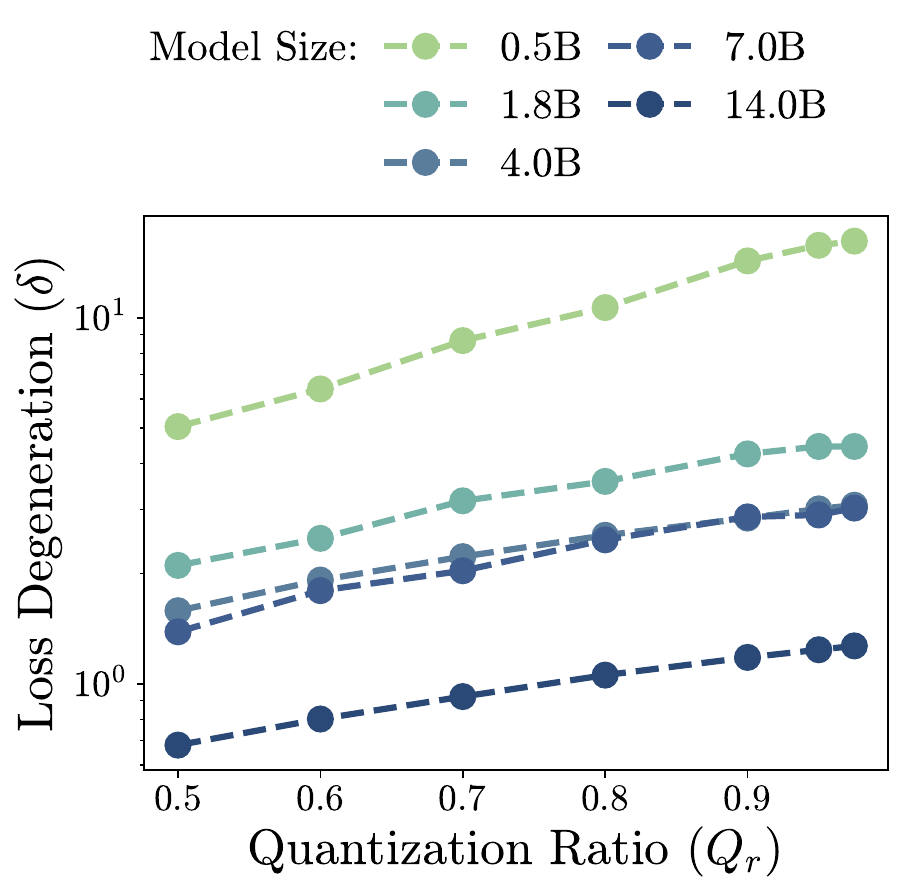}
\captionsetup{font=scriptsize}
\caption{Qwen-1.5 Actual Loss}
\end{subfigure}
\hfill
\begin{subfigure}[b]{0.3\textwidth} \centering
\includegraphics[width=\textwidth]{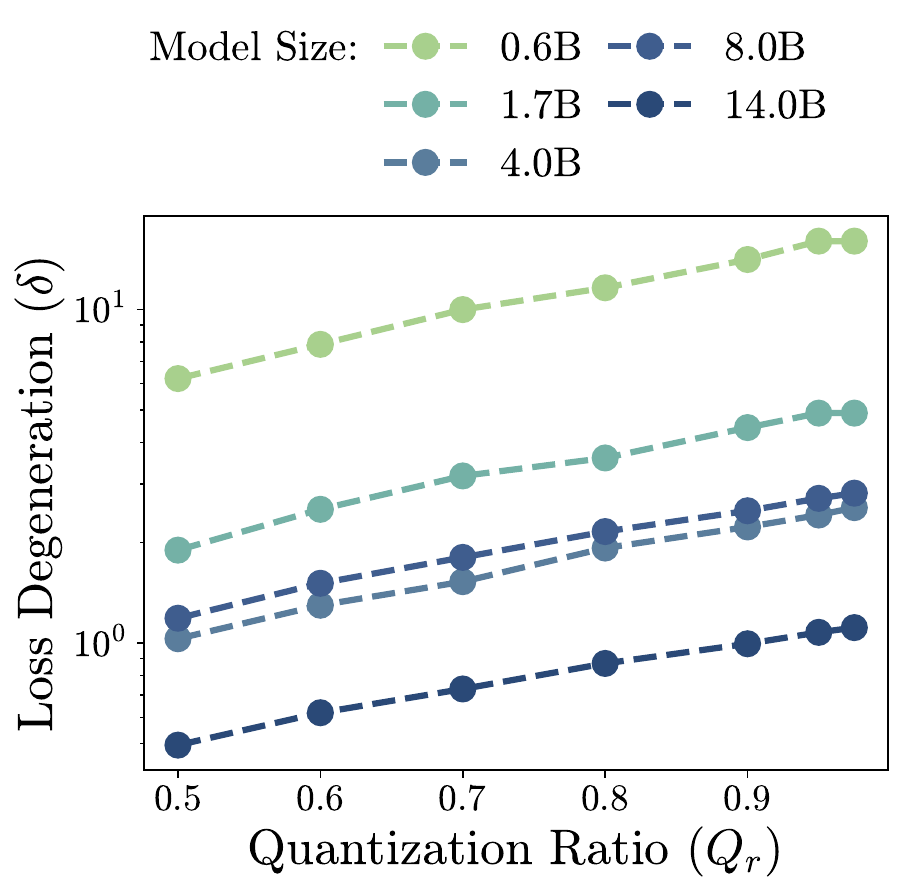}
\captionsetup{font=scriptsize}
\caption{Qwen-3 Actual Loss}
\end{subfigure}

\begin{subfigure}[b]{0.3\textwidth} \centering
\includegraphics[width=\textwidth]{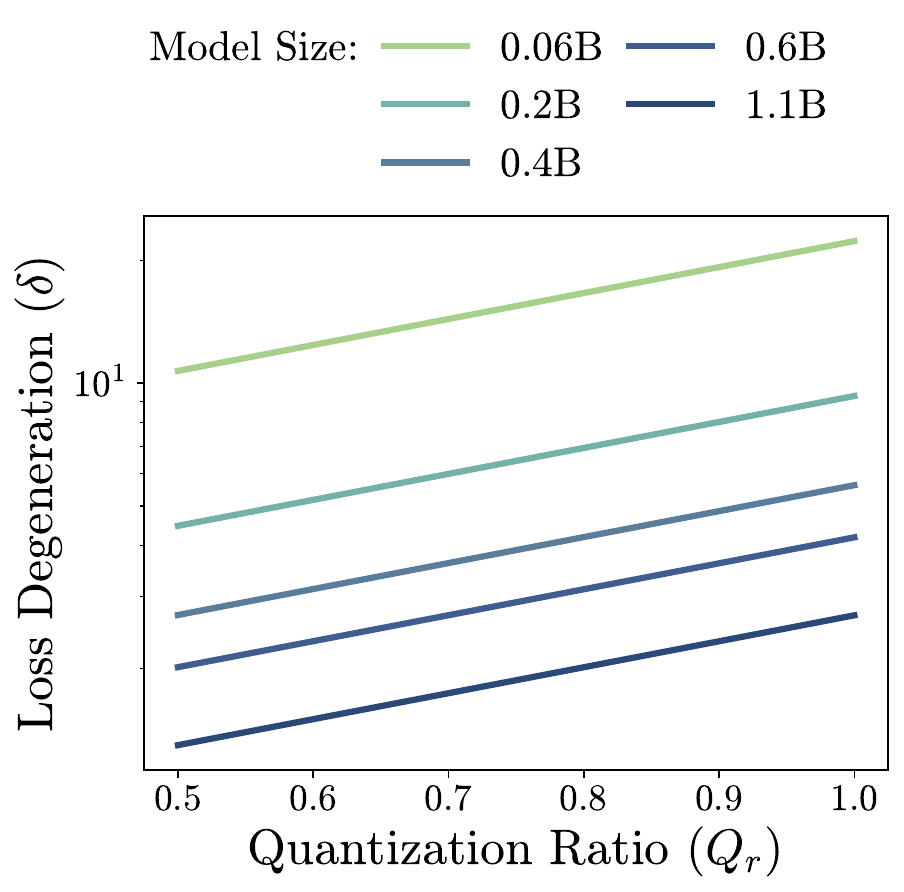}
\captionsetup{font=scriptsize}
\caption{CLM Predicted Loss}
\end{subfigure}
\hfill
\begin{subfigure}[b]{0.3\textwidth} \centering
\includegraphics[width=\textwidth]{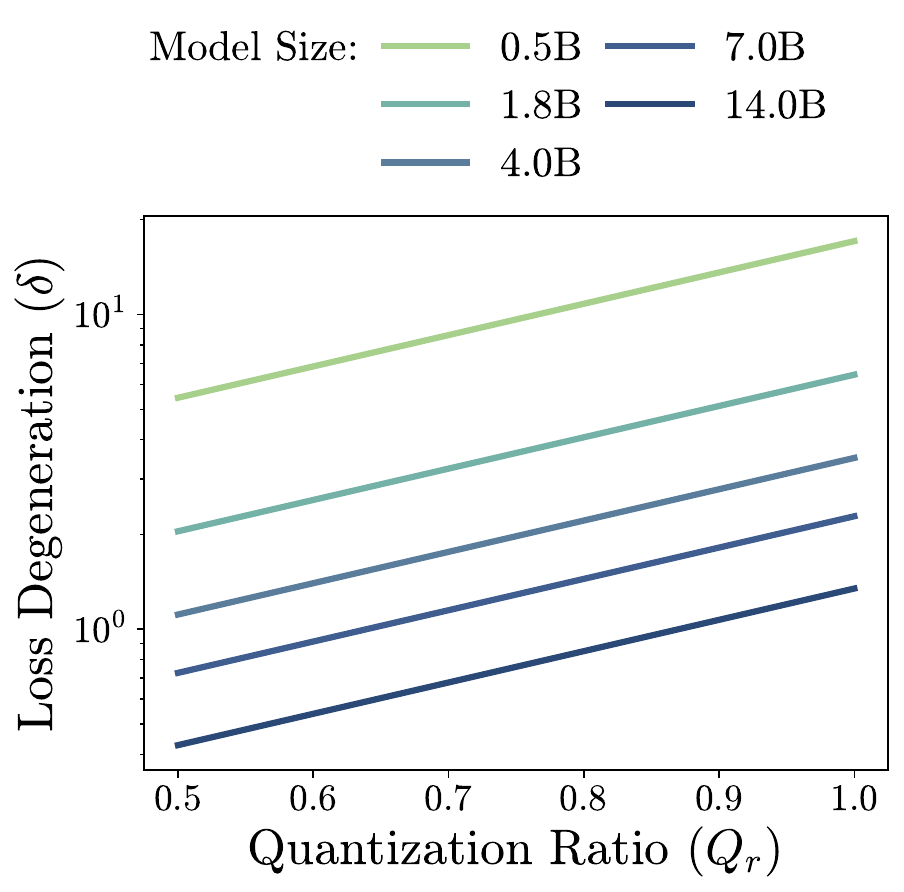}
\captionsetup{font=scriptsize}
\caption{Qwen-1.5 Predicted Loss}
\end{subfigure}
\hfill
\begin{subfigure}[b]{0.3\textwidth} \centering
\includegraphics[width=\textwidth]{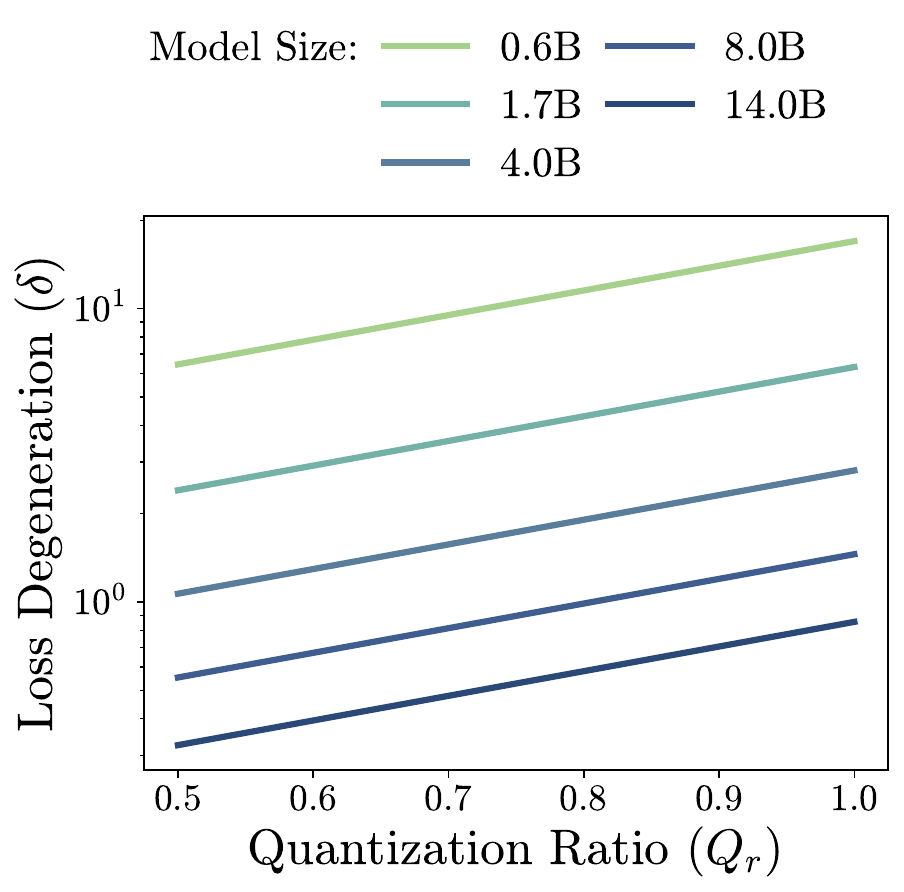}
\captionsetup{font=scriptsize}
\caption{Qwen-3 Predicted Loss}
\end{subfigure}

\caption{\textbf{Layer-wise ($\delta_\mu$)} (a,d,g,h) CLM layer-wise results; (b,e,h,k) Qwen-1.5 layer-wise results; (c,f,i,l) Qwen-3 layer-wise results.}
  \label{fig:appendix-layerwise-mean}
\end{figure}

As shown in \Cref{sec:experiments},~\Cref{fig:appendix-layerwise} and~\Cref{fig:appendix-layerwise-mean} show the actual and fitted loss contour of CLM, Qwen1.5, and Qwen3 under MXINT4 layerwise quantization. For each model family we present two contours: loss versus model size $N$ and loss versus quantization ratio $Q_r$. ~\Cref{fig:appendix-layerwise} shows the contour for the minimum value $\delta^{\text{opt}}$ and Figure~\ref{fig:appendix-layerwise-mean} shows the expectation value $\delta_\mu$. There are a few outliers that do not match our fitted contour, which could be caused by the instability of the sampling process. In addition, note that the actual losses themselves are estimations.

\section{Matrix Multiplication-wise Results}
\label{app:matmul-result}
\begin{figure}[htbp]
\centering
\begin{subfigure}[b]{0.3\textwidth} \centering
\includegraphics[width=\textwidth]{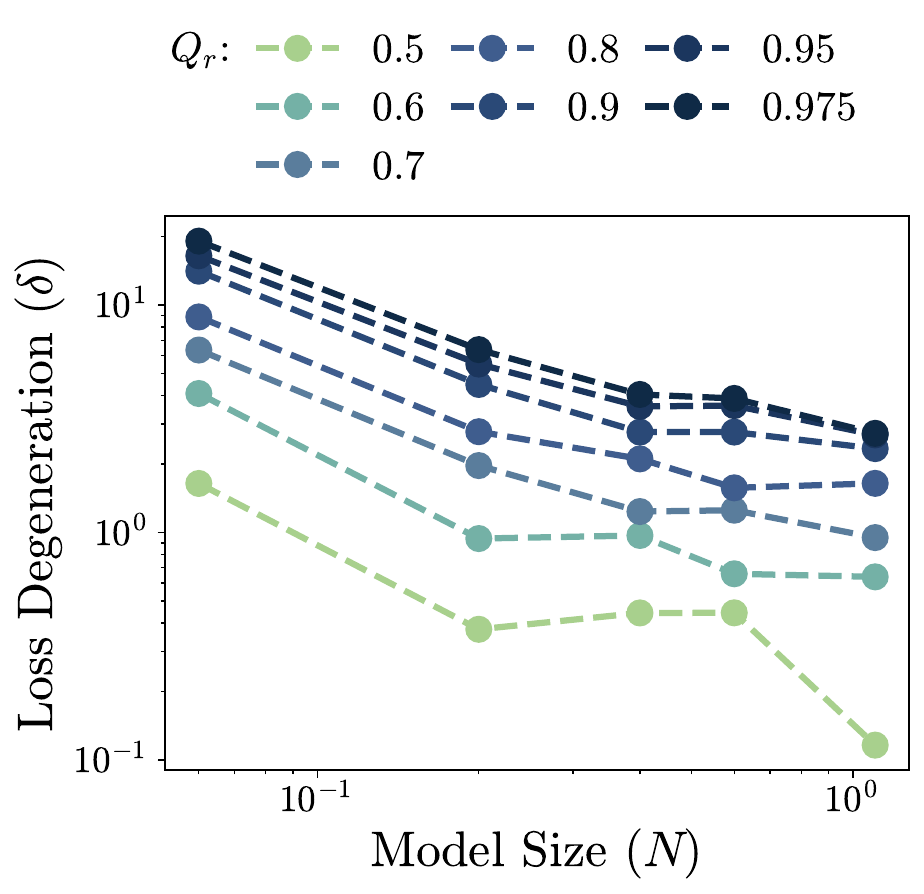}
\captionsetup{font=scriptsize}
\caption{CLM Actual Loss}
\end{subfigure}
\hfill
\begin{subfigure}[b]{0.3\textwidth} \centering
\includegraphics[width=\textwidth]{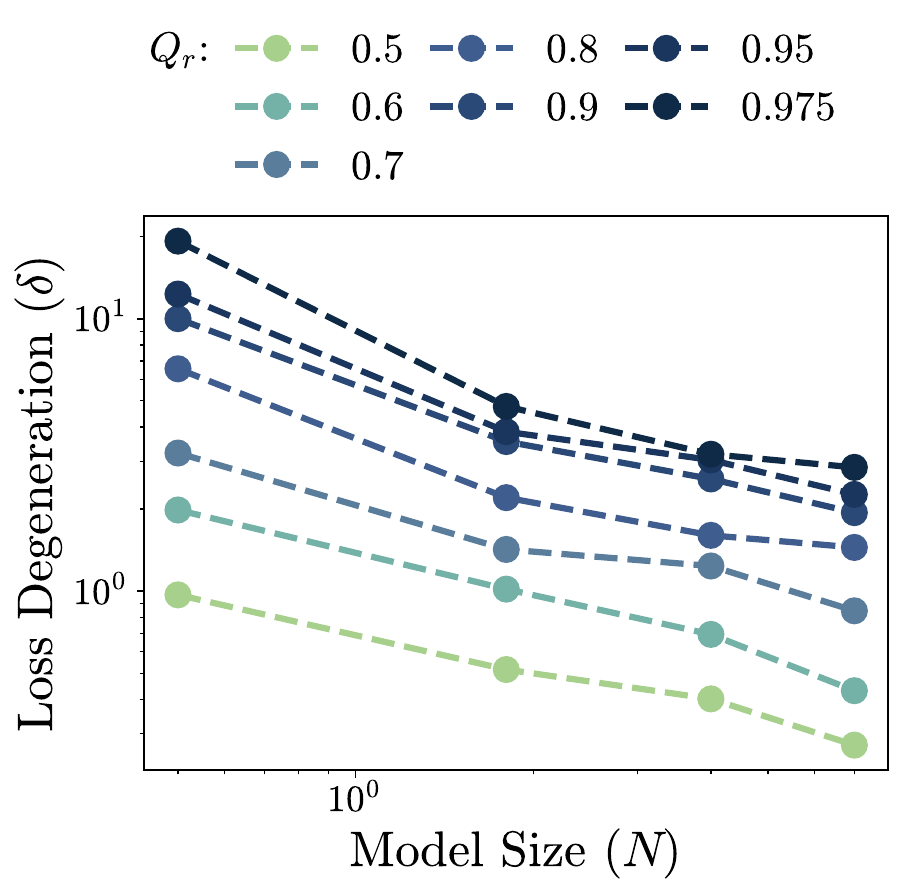}
\captionsetup{font=scriptsize}
\caption{Qwen-1.5 Actual Loss}
\end{subfigure}
\hfill
\begin{subfigure}[b]{0.3\textwidth} \centering
\includegraphics[width=\textwidth]{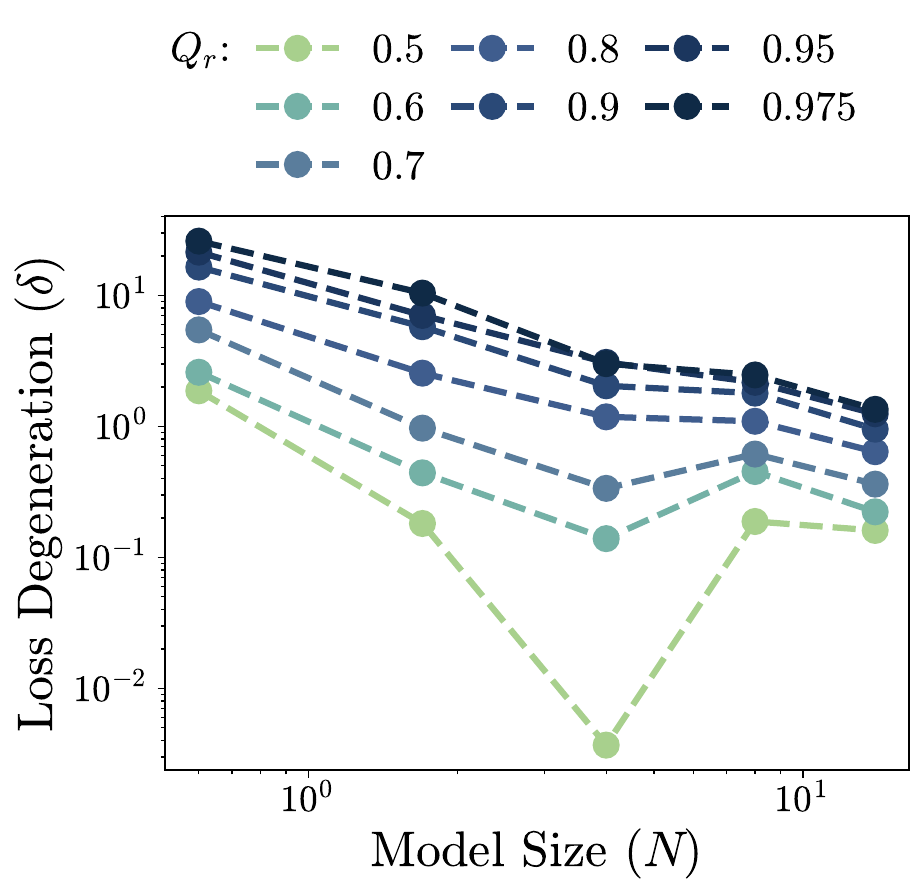}
\captionsetup{font=scriptsize}
\caption{Qwen-3 Actual Loss}
\end{subfigure}

\begin{subfigure}[b]{0.3\textwidth} \centering
\includegraphics[width=\textwidth]{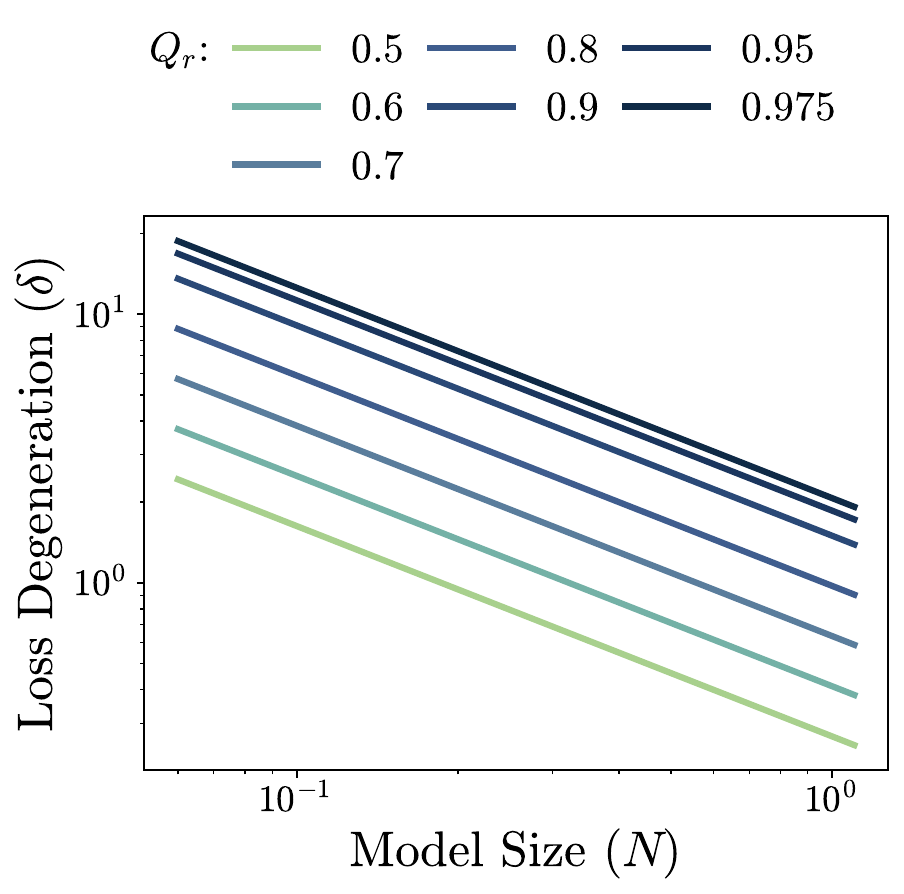}
\captionsetup{font=scriptsize}
\caption{CLM Predicted Loss}
\end{subfigure}
\hfill
\begin{subfigure}[b]{0.3\textwidth} \centering
\includegraphics[width=\textwidth]{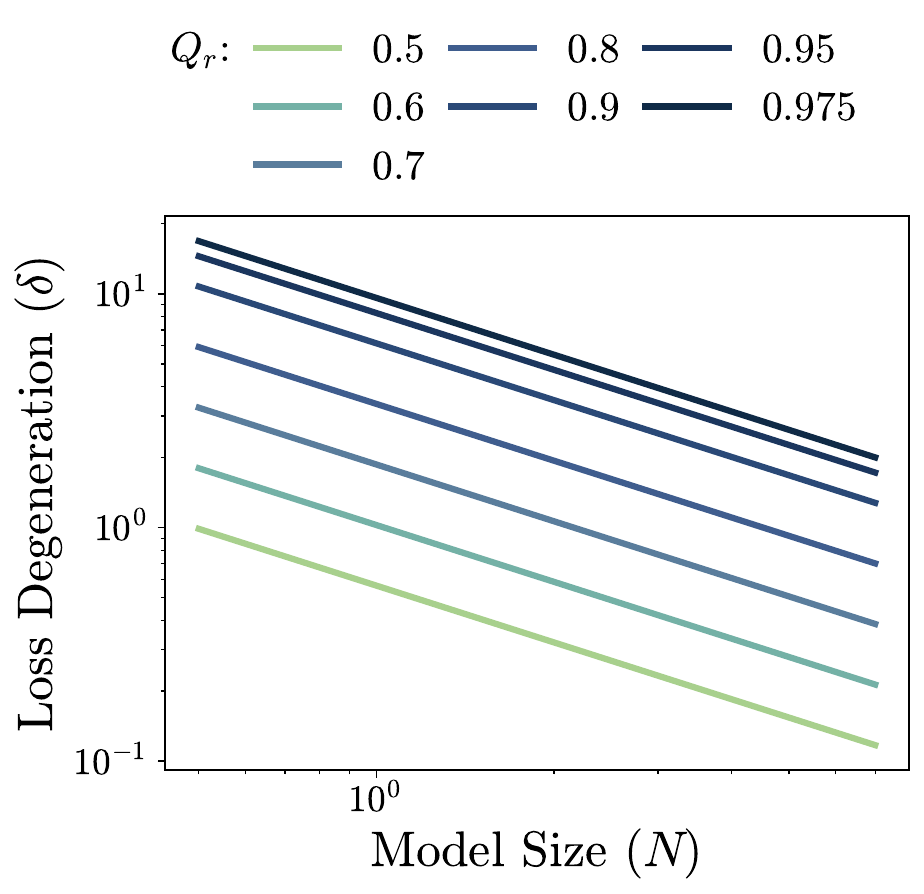}
\captionsetup{font=scriptsize}
\caption{Qwen-1.5 Predicted Loss}
\end{subfigure}
\hfill
\begin{subfigure}[b]{0.3\textwidth} \centering
\includegraphics[width=\textwidth]{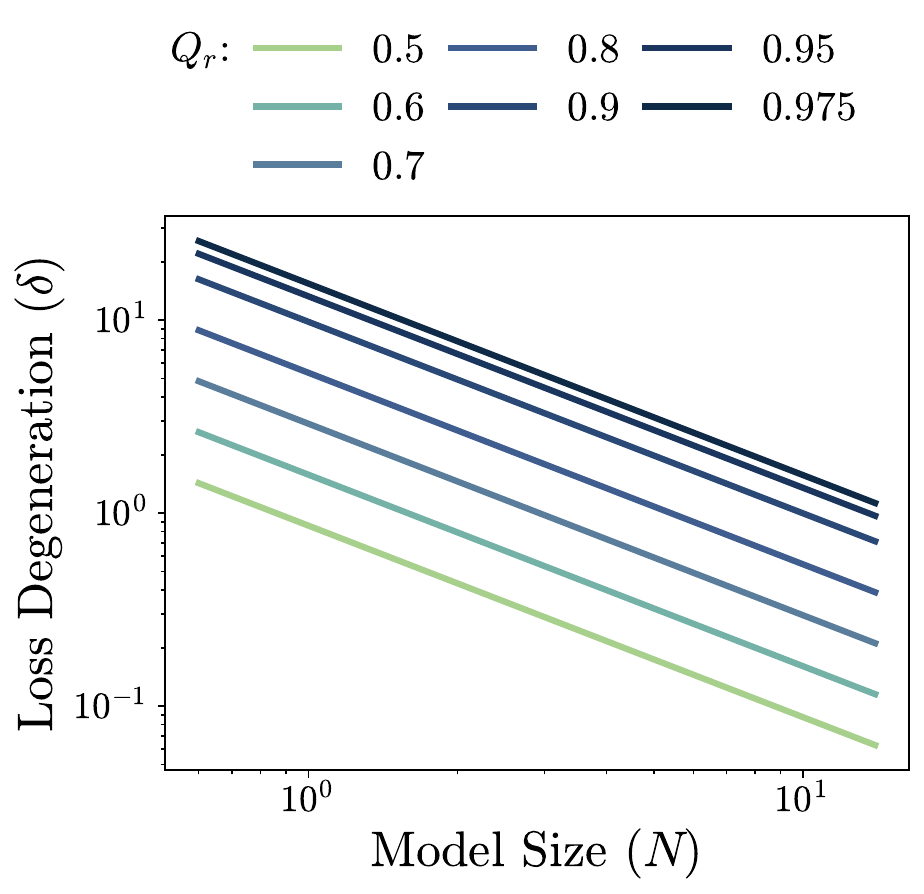}
\captionsetup{font=scriptsize}
\caption{Qwen-3 Predicted Loss}
\end{subfigure}

\begin{subfigure}[b]{0.3\textwidth} \centering
\includegraphics[width=\textwidth]{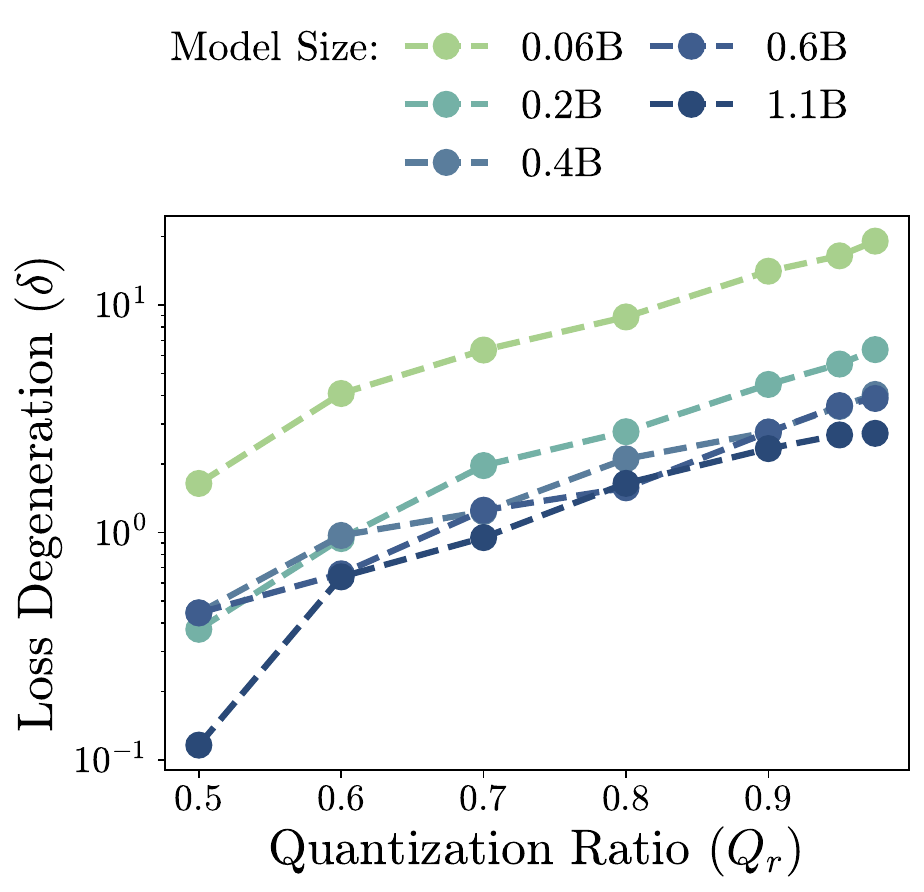}
\captionsetup{font=scriptsize}
\caption{CLM Actual Loss}
\end{subfigure}
\hfill
\begin{subfigure}[b]{0.3\textwidth} \centering
\includegraphics[width=\textwidth]{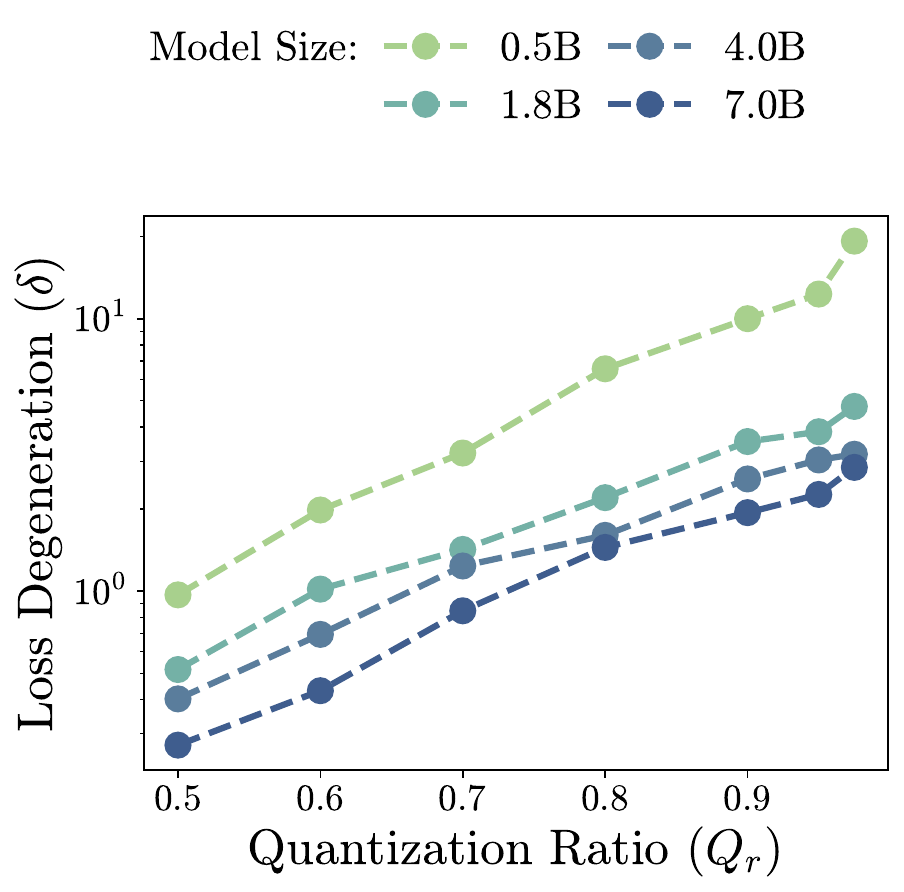}
\captionsetup{font=scriptsize}
\caption{Qwen-1.5 Actual Loss}
\end{subfigure}
\hfill
\begin{subfigure}[b]{0.3\textwidth} \centering
\includegraphics[width=\textwidth]{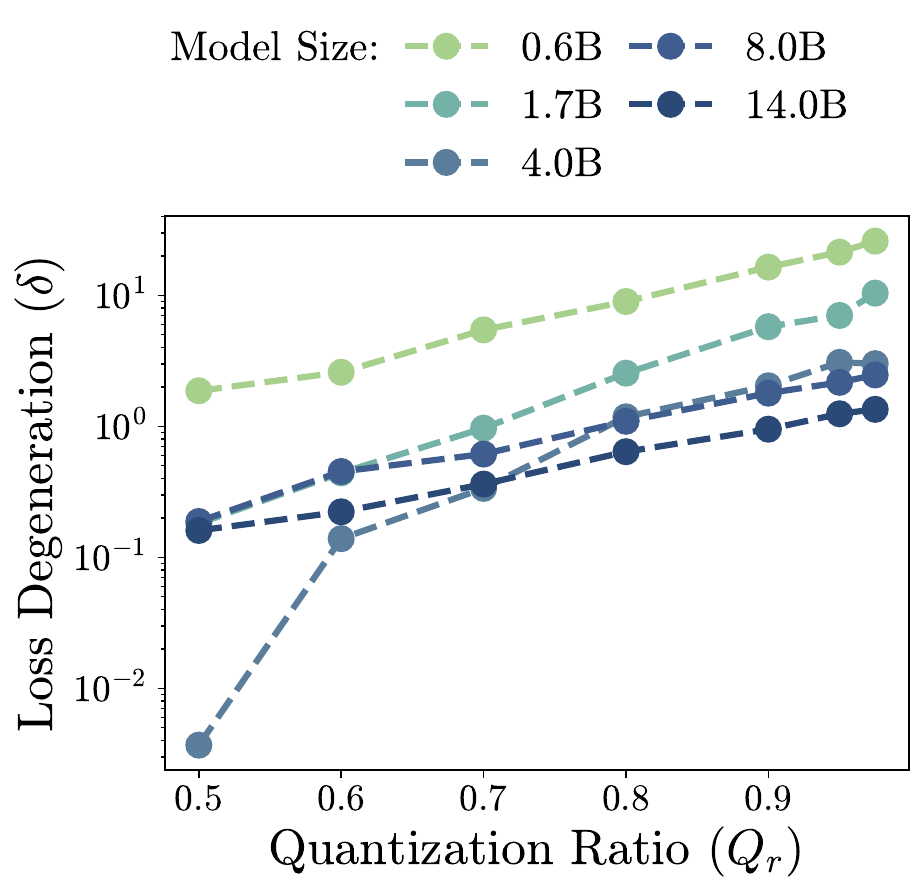}
\captionsetup{font=scriptsize}
\caption{Qwen-3 Actual Loss}
\end{subfigure}

\begin{subfigure}[b]{0.3\textwidth} \centering
\includegraphics[width=\textwidth]{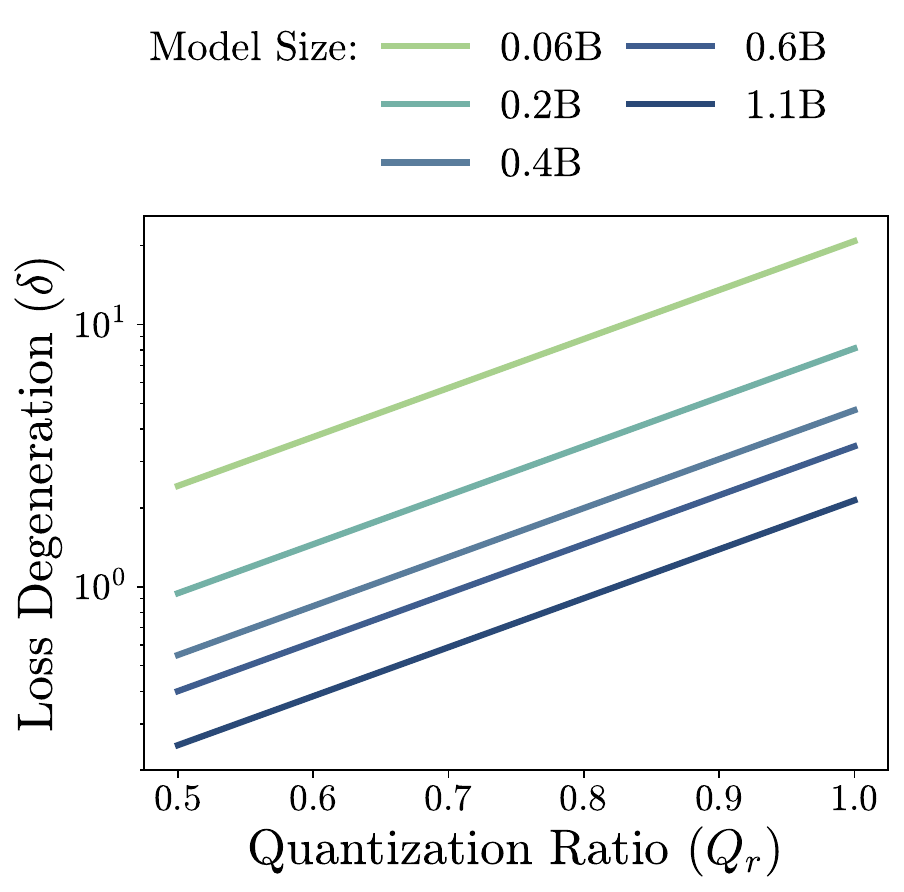}
\captionsetup{font=scriptsize}
\caption{CLM Predicted Loss}
\end{subfigure}
\hfill
\begin{subfigure}[b]{0.3\textwidth} \centering
\includegraphics[width=\textwidth]{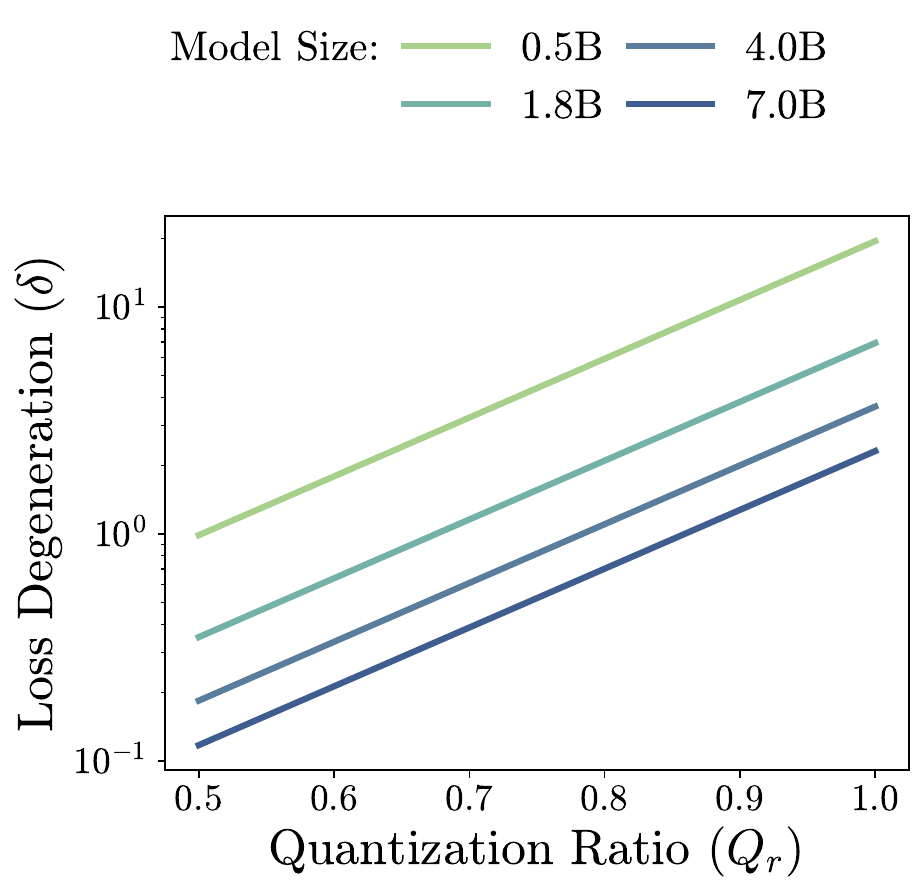}
\captionsetup{font=scriptsize}
\caption{Qwen-1.5 Predicted Loss}
\end{subfigure}
\hfill
\begin{subfigure}[b]{0.3\textwidth} \centering
\includegraphics[width=\textwidth]{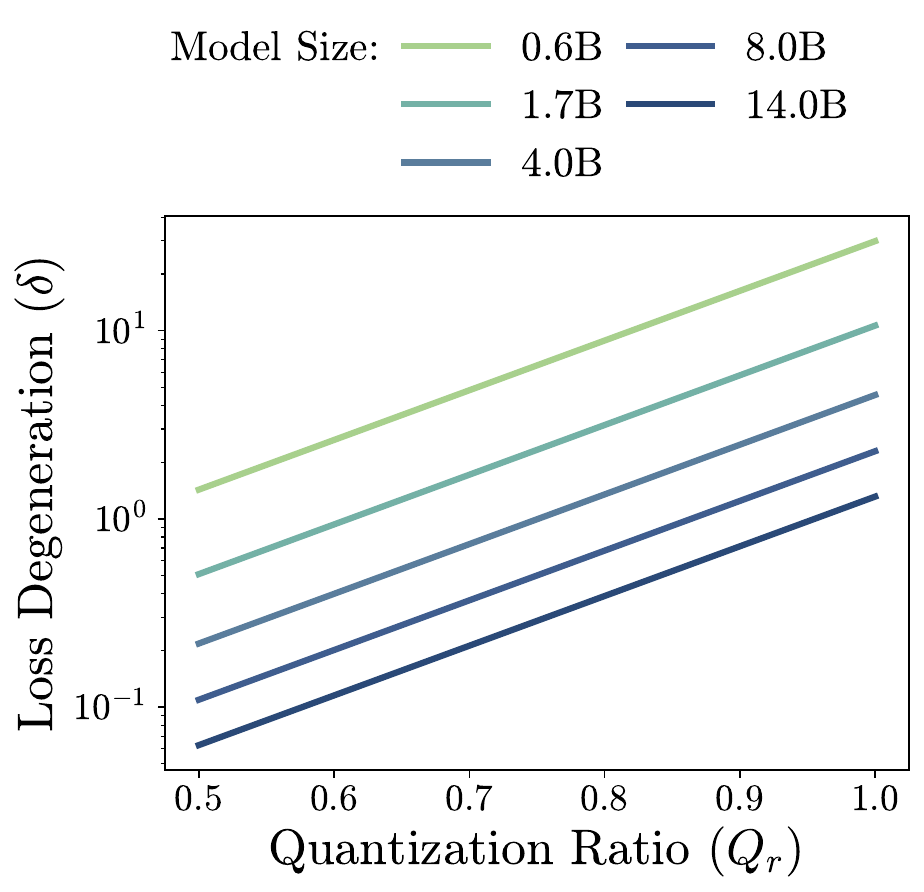}
\captionsetup{font=scriptsize}
\caption{Qwen-3 Predicted Loss}
\end{subfigure}

\caption{\textbf{Matrix Multiplication-wise ($\delta^{\text{opt}}$)} (a,d,g,h) CLM matrix multiplication-wise results; (b,e,h,k) Qwen-1.5 matrix multiplication-wise results; (c,f,i,l) Qwen-3 matrix multiplication-wise results.}
  \label{fig:appendix-matmul}
\end{figure}

\begin{figure}[htbp]
\centering
\begin{subfigure}[b]{0.3\textwidth} \centering
\includegraphics[width=\textwidth]{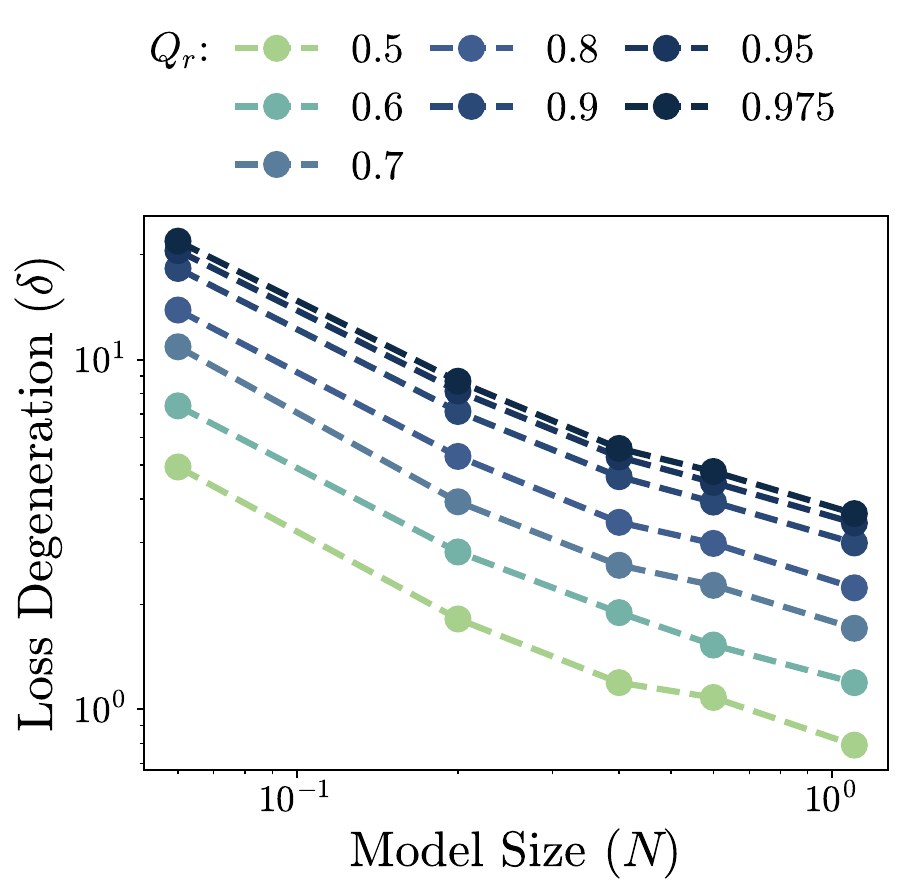}
\captionsetup{font=scriptsize}
\caption{CLM Actual Loss}
\end{subfigure}
\hfill
\begin{subfigure}[b]{0.3\textwidth} \centering
\includegraphics[width=\textwidth]{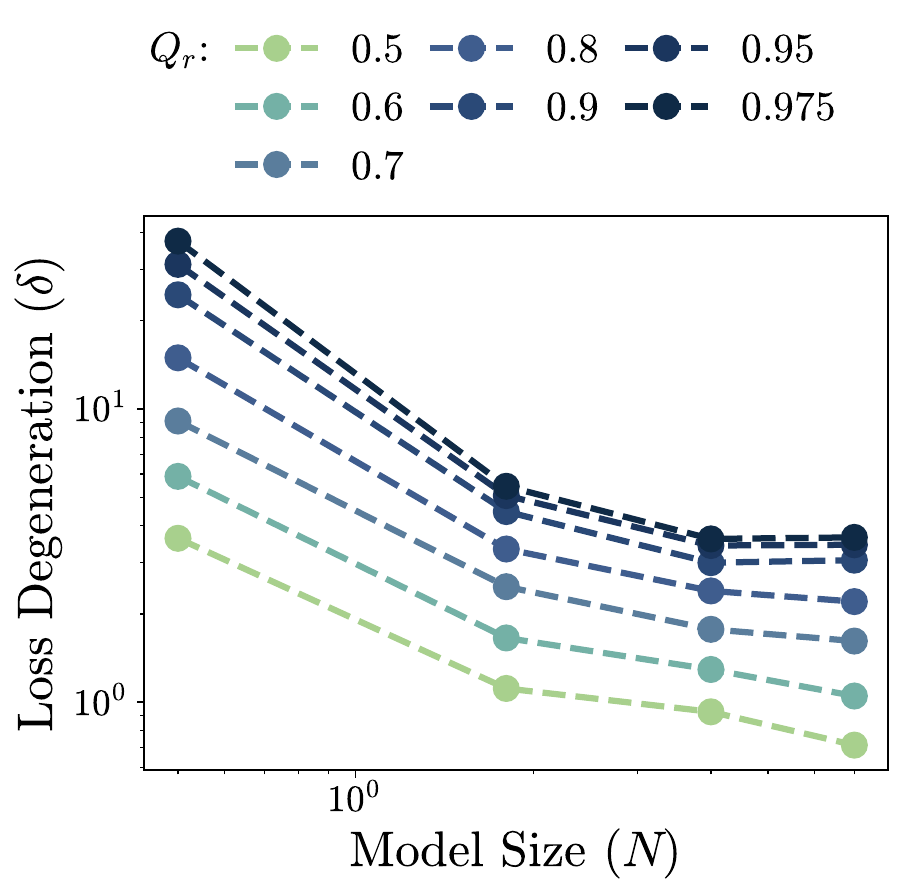}
\captionsetup{font=scriptsize}
\caption{Qwen-1.5 Actual Loss}
\end{subfigure}
\hfill
\begin{subfigure}[b]{0.3\textwidth} \centering
\includegraphics[width=\textwidth]{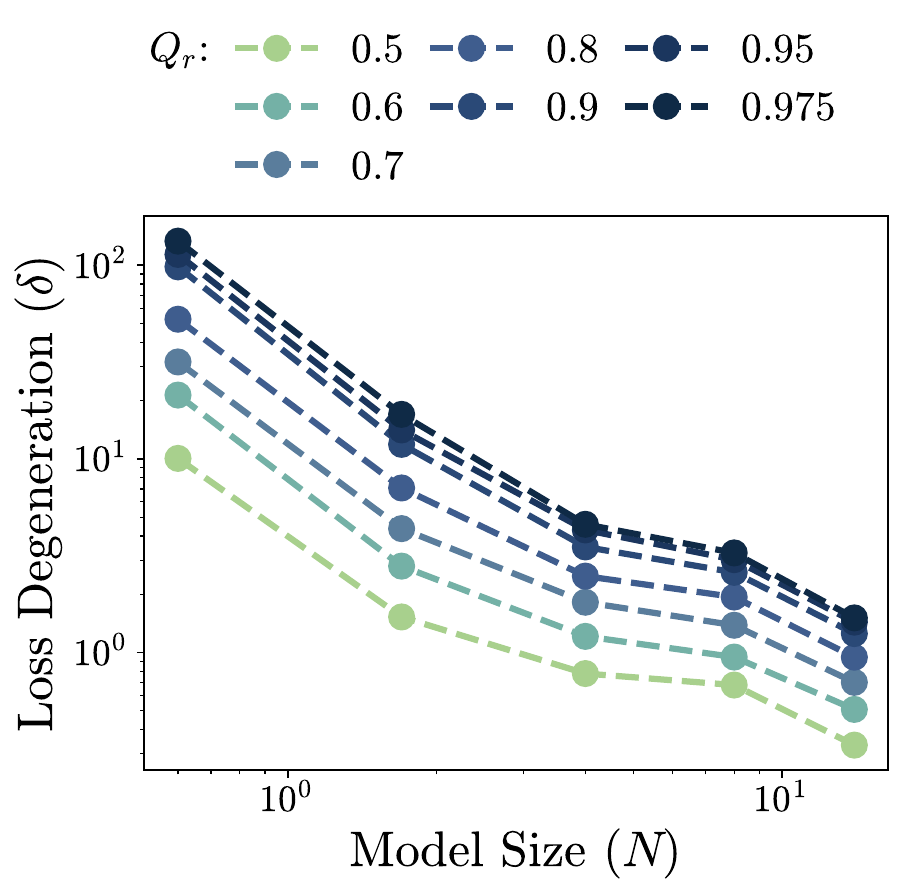}
\captionsetup{font=scriptsize}
\caption{Qwen-3 Actual Loss}
\end{subfigure}

\begin{subfigure}[b]{0.3\textwidth} \centering
\includegraphics[width=\textwidth]{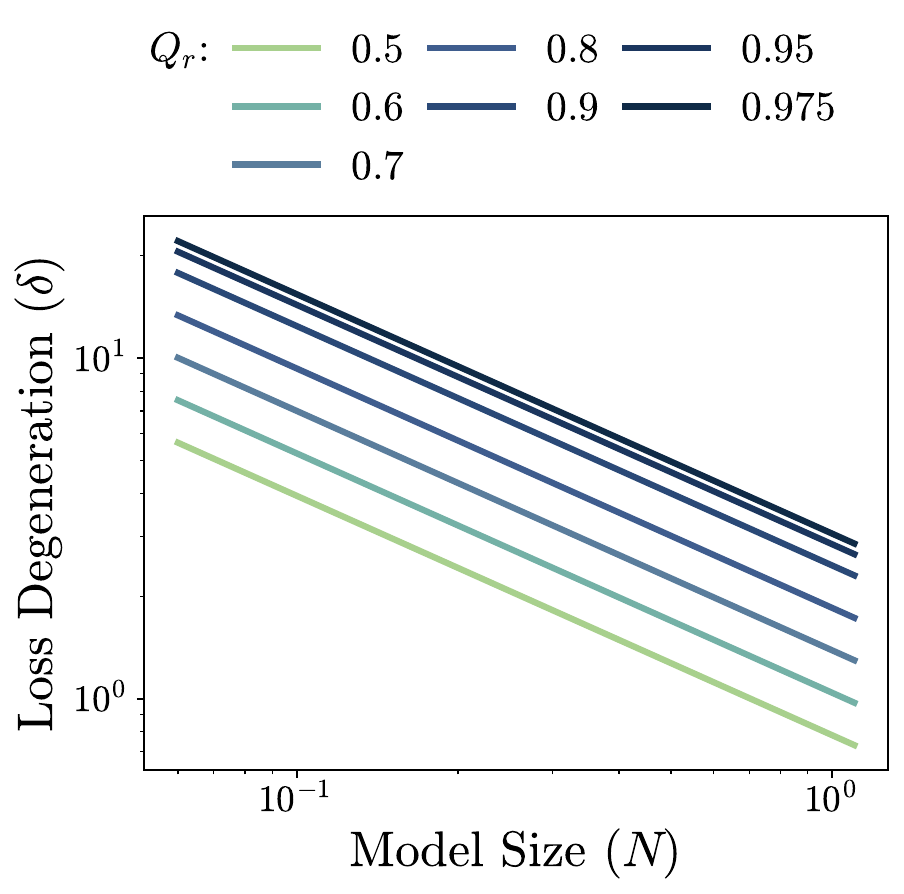}
\captionsetup{font=scriptsize}
\caption{CLM Predicted Loss}
\end{subfigure}
\hfill
\begin{subfigure}[b]{0.3\textwidth} \centering
\includegraphics[width=\textwidth]{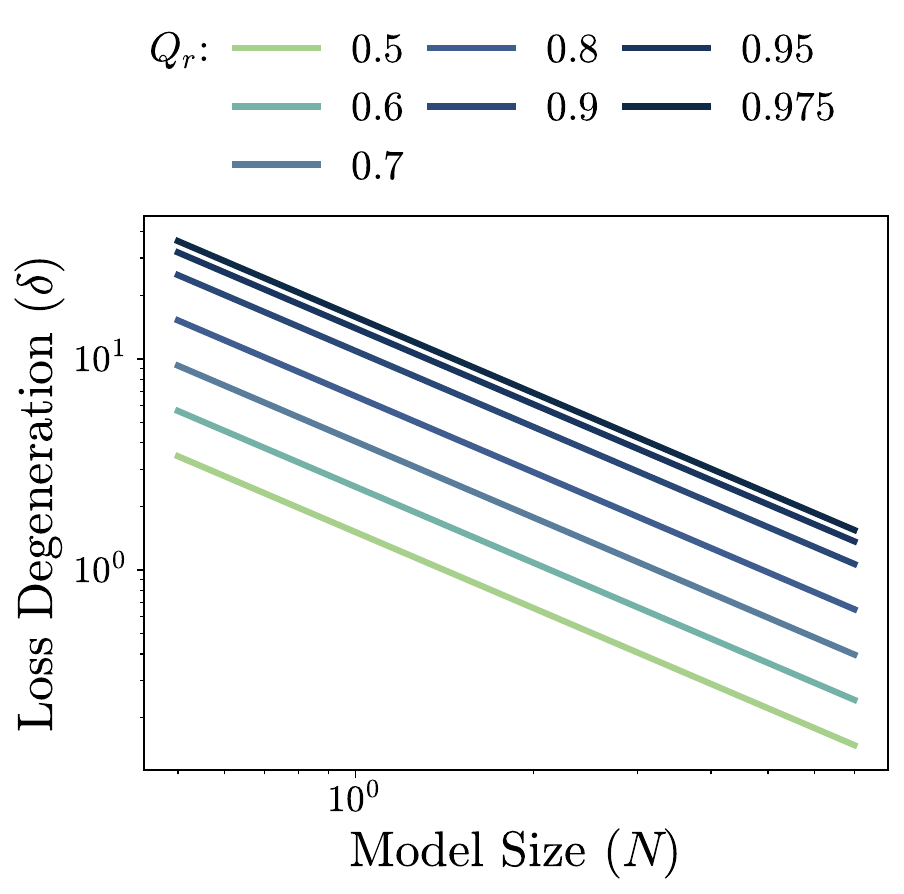}
\captionsetup{font=scriptsize}
\caption{Qwen-1.5 Predicted Loss}
\end{subfigure}
\hfill
\begin{subfigure}[b]{0.3\textwidth} \centering
\includegraphics[width=\textwidth]{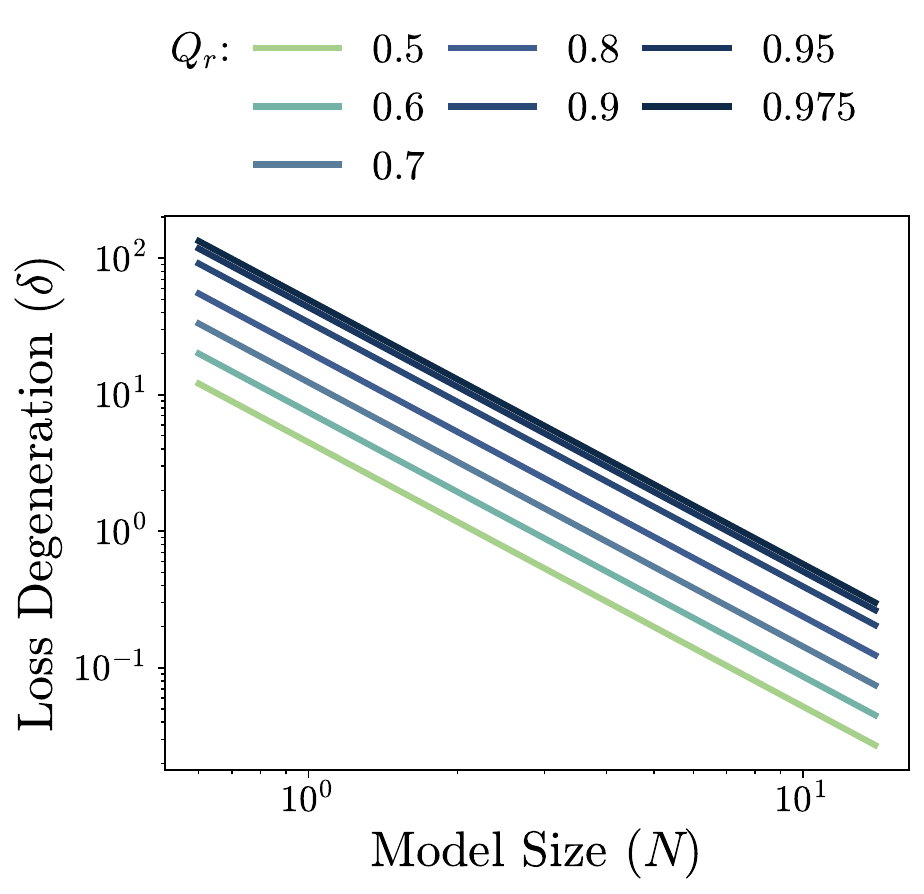}
\captionsetup{font=scriptsize}
\caption{Qwen-3 Predicted Loss}
\end{subfigure}

\begin{subfigure}[b]{0.3\textwidth} \centering
\includegraphics[width=\textwidth]{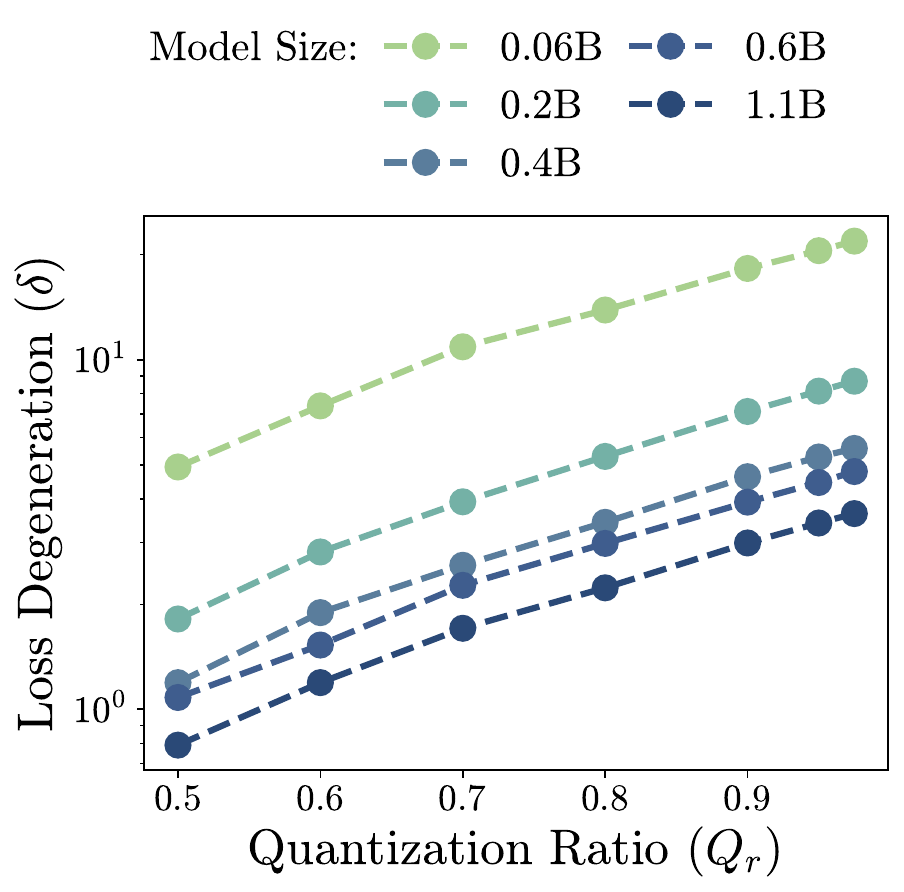}
\captionsetup{font=scriptsize}
\caption{CLM Actual Loss}
\end{subfigure}
\hfill
\begin{subfigure}[b]{0.3\textwidth} \centering
\includegraphics[width=\textwidth]{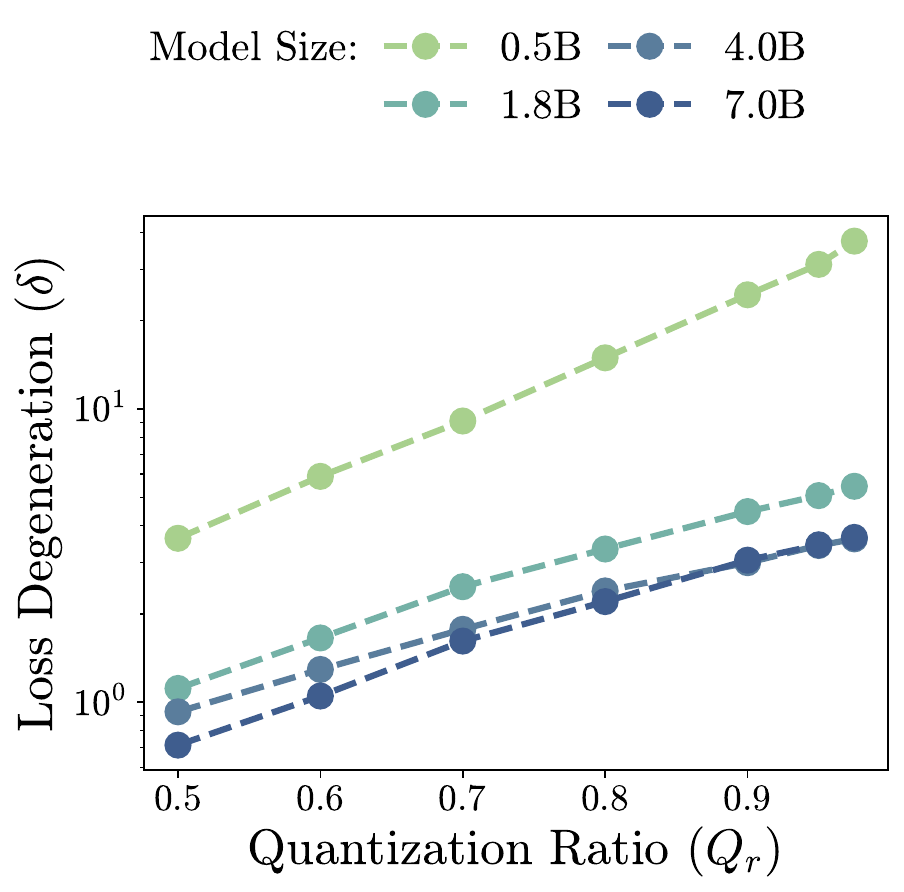}
\captionsetup{font=scriptsize}
\caption{Qwen-1.5 Actual Loss}
\end{subfigure}
\hfill
\begin{subfigure}[b]{0.3\textwidth} \centering
\includegraphics[width=\textwidth]{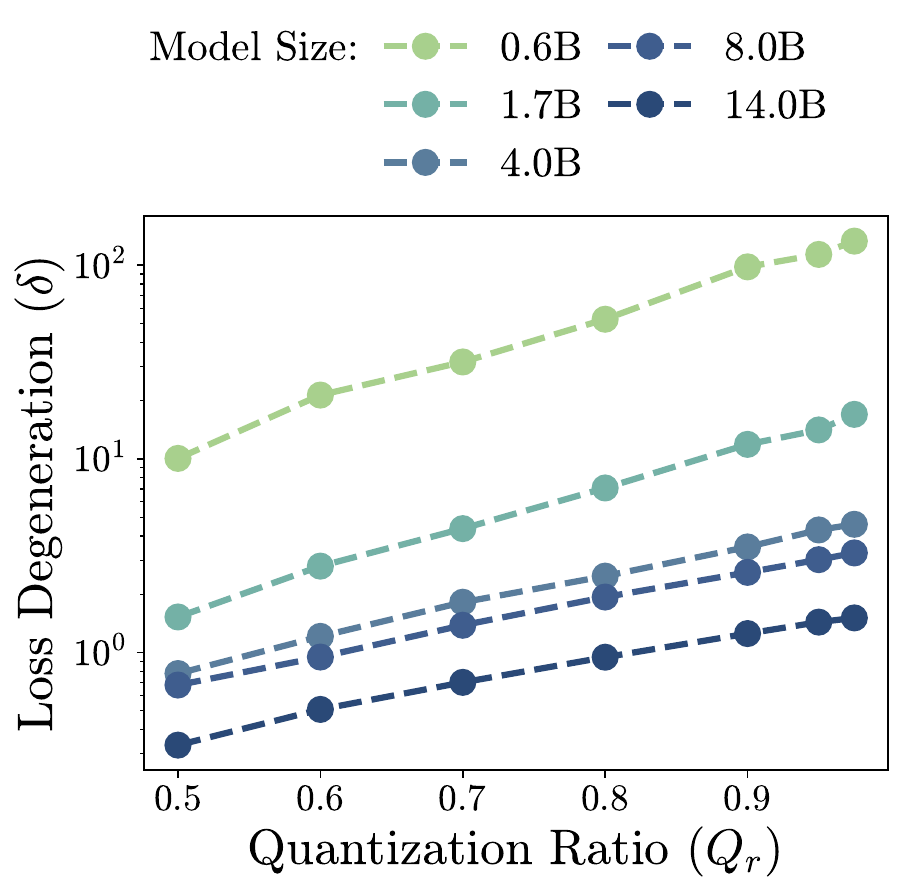}
\captionsetup{font=scriptsize}
\caption{Qwen-3 Actual Loss}
\end{subfigure}

\begin{subfigure}[b]{0.3\textwidth} \centering
\includegraphics[width=\textwidth]{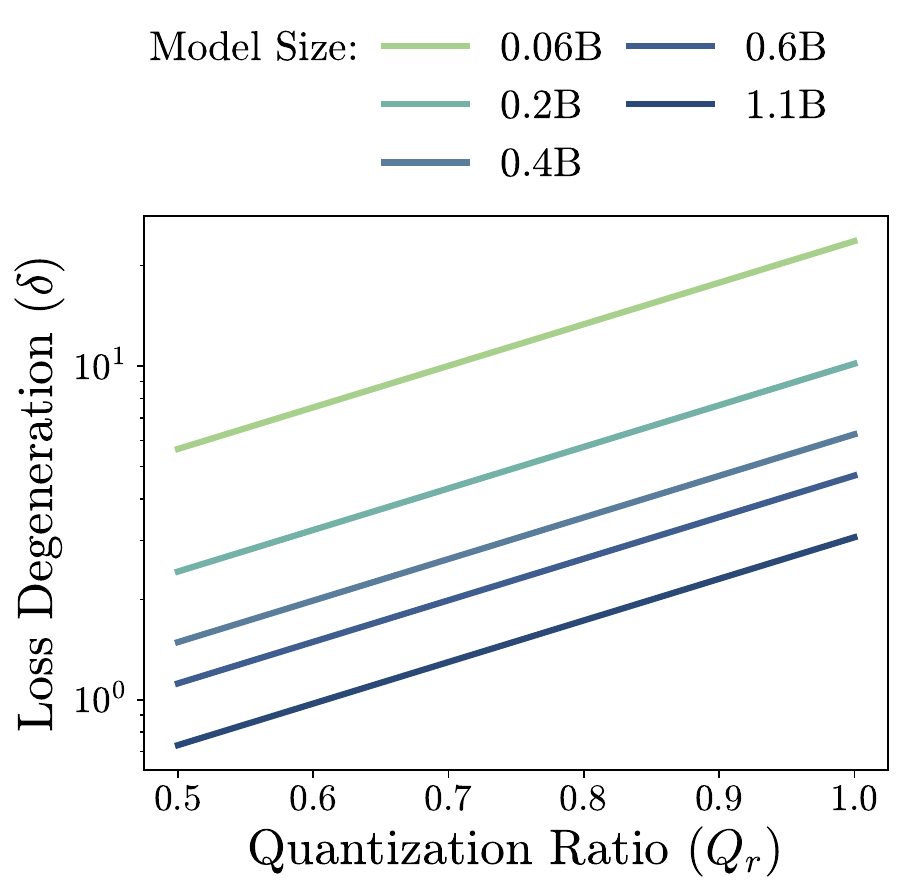}
\captionsetup{font=scriptsize}
\caption{CLM Predicted Loss}
\end{subfigure}
\hfill
\begin{subfigure}[b]{0.3\textwidth} \centering
\includegraphics[width=\textwidth]{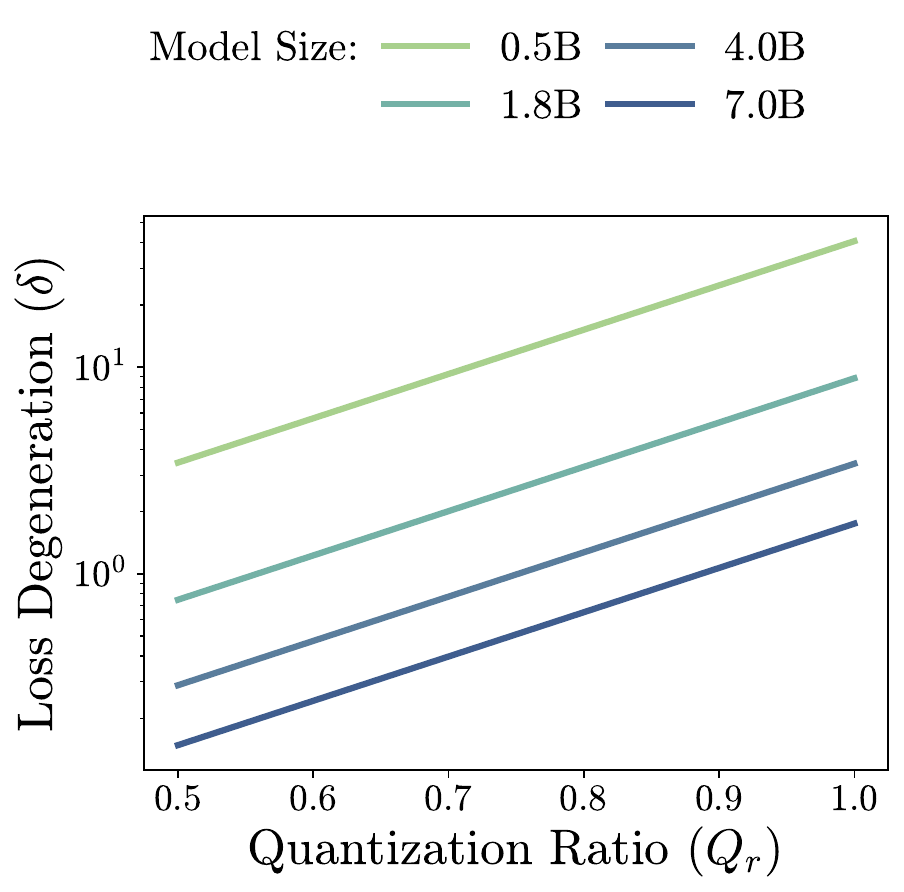}
\captionsetup{font=scriptsize}
\caption{Qwen-1.5 Predicted Loss}
\end{subfigure}
\hfill
\begin{subfigure}[b]{0.3\textwidth} \centering
\includegraphics[width=\textwidth]{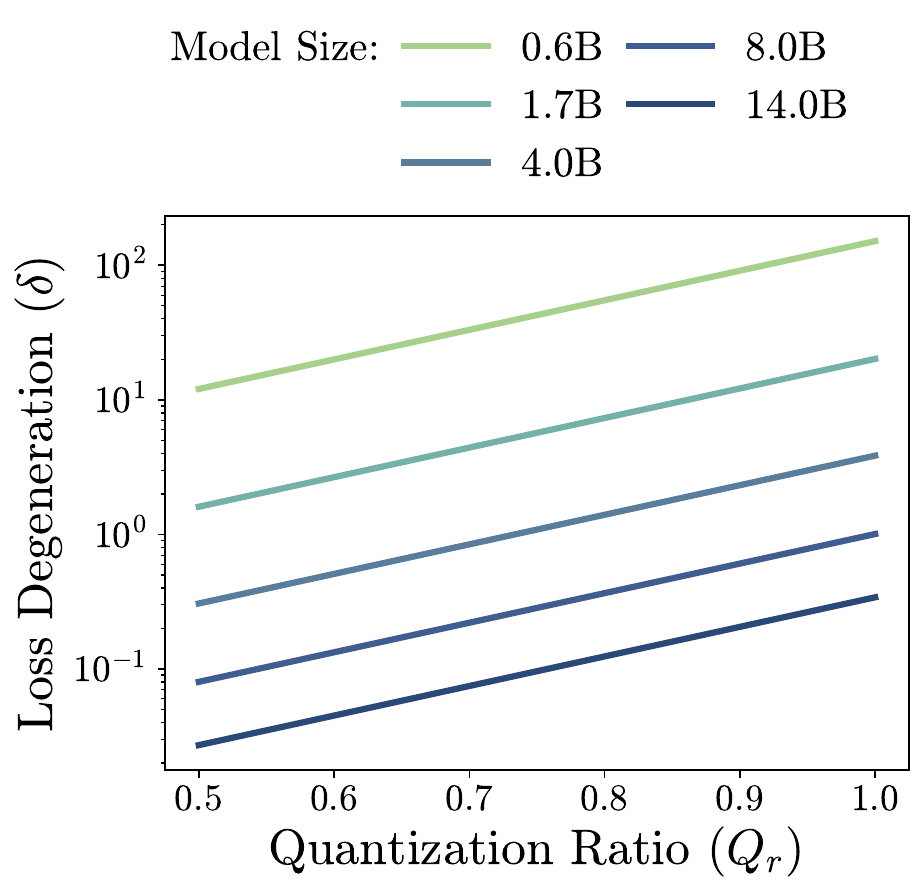}
\captionsetup{font=scriptsize}
\caption{Qwen-3 Predicted Loss}
\end{subfigure}

\caption{\textbf{Matrix Multiplication-wise ($\delta_\mu$)} (a,d,g,h) CLM matrix multiplication-wise results; (b,e,h,k) Qwen-1.5 matrix multiplication-wise results; (c,f,i,l) Qwen-3 matrix multiplication-wise results.} 
\label{fig:appendix-matmul-mean}
\end{figure}

Following the layer-wise results in~\Cref{sec:experiments}, the matrix multiplication-wise with ~\Cref{fig:appendix-matmul} and~\Cref{fig:appendix-matmul-mean} shows the actual and fitted loss contour of CLM, Qwen-1.5, and Qwen-3 under MXINT-4 matrix multiplication-wise quantization. For each model family, we present two contours: loss versus model size $N$ and loss versus quantization ratio $Q_r$. ~\Cref{fig:appendix-matmul} shows the contour for the minimum value $\delta^{\text{opt}}$ and ~\Cref{fig:appendix-matmul-mean} shows the expectation value $\delta_\mu$. As with the layer-wise quantization results, a few outliers deviate from the fitted contours due to the variability in the sampling process.

\begin{figure}[htbp]
    \centering
\begin{subfigure}[b]{0.24\textwidth}
        \centering
        \includegraphics[width=\textwidth]{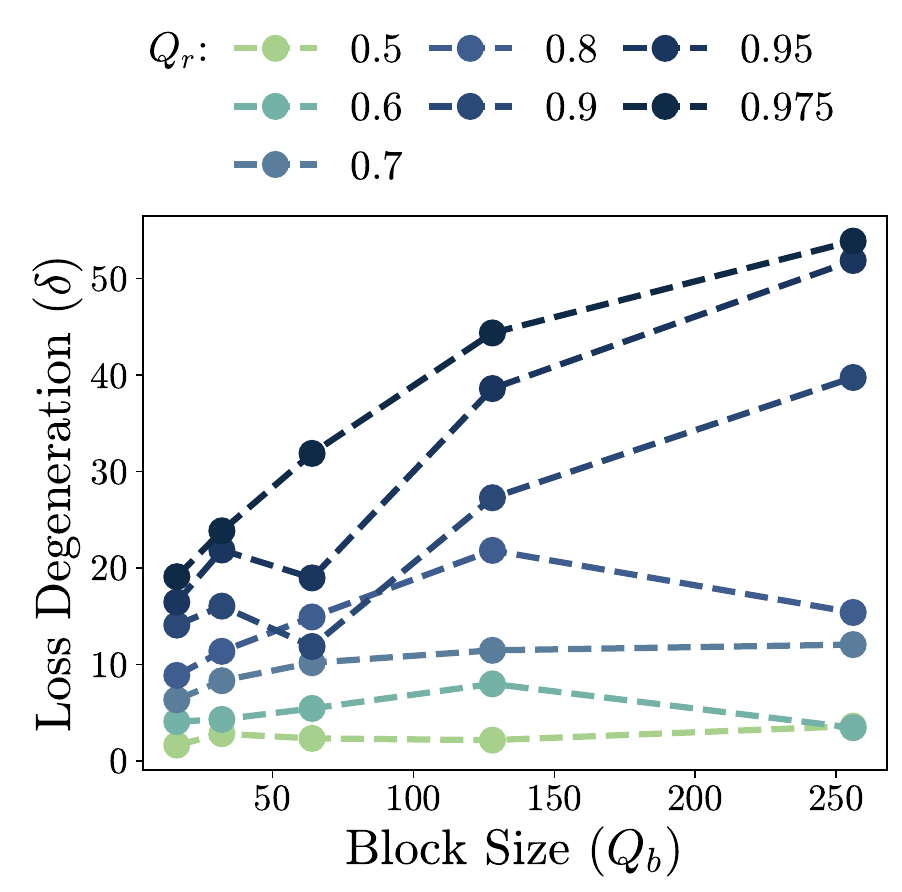}
        \captionsetup{font=scriptsize}
        \caption{60M Actual (min)}
    \end{subfigure}\hfill
    \begin{subfigure}[b]{0.24\textwidth}
        \centering
        \includegraphics[width=\textwidth]{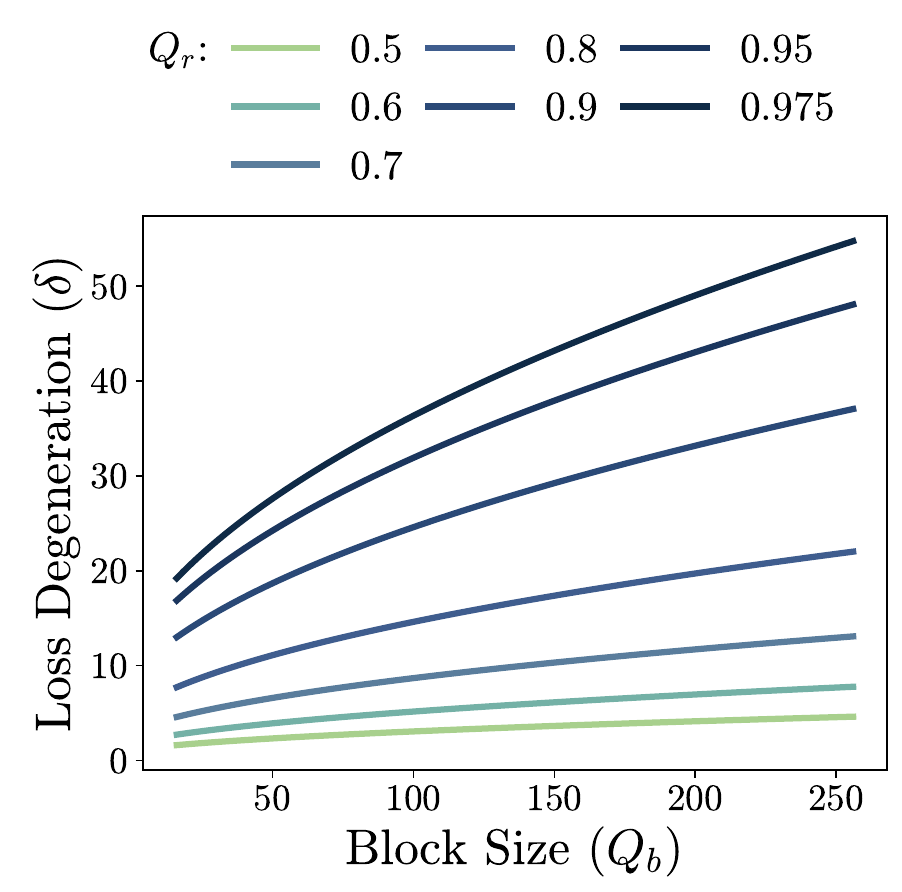}
        \captionsetup{font=scriptsize}
        \caption{60M Predicted (min)}
    \end{subfigure}\hfill
    \begin{subfigure}[b]{0.24\textwidth}
        \centering
        \includegraphics[width=\textwidth]{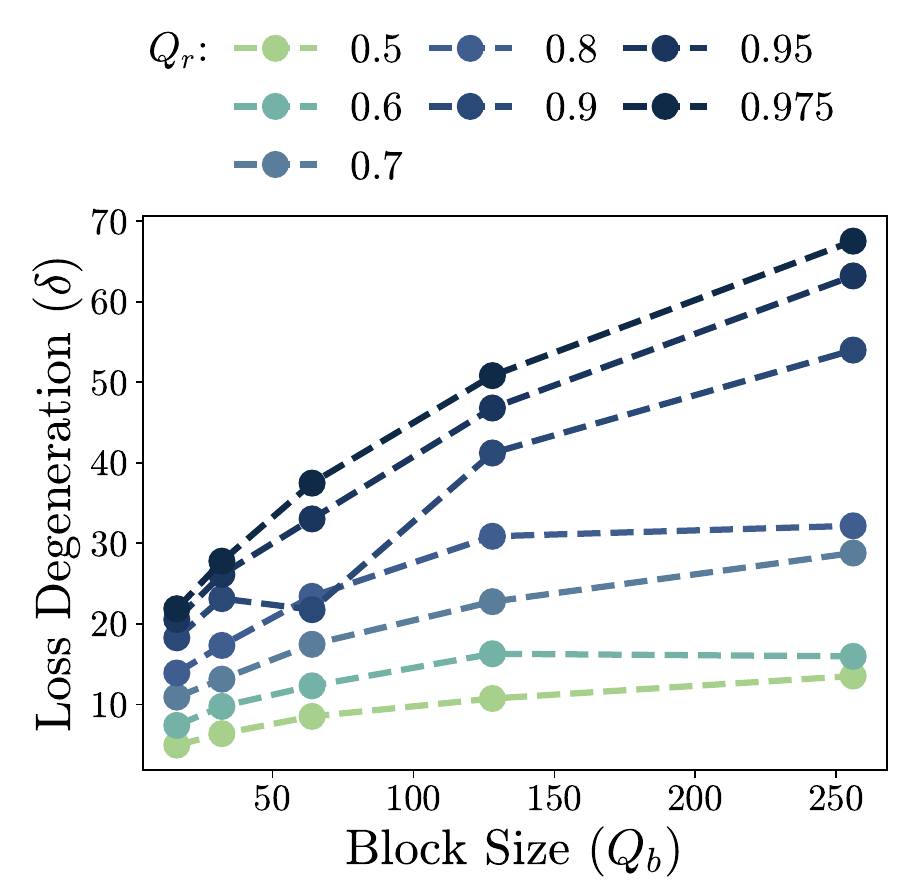}
        \captionsetup{font=scriptsize}
        \caption{60M Actual (mean)}
    \end{subfigure}\hfill
    \begin{subfigure}[b]{0.24\textwidth}
        \centering
        \includegraphics[width=\textwidth]{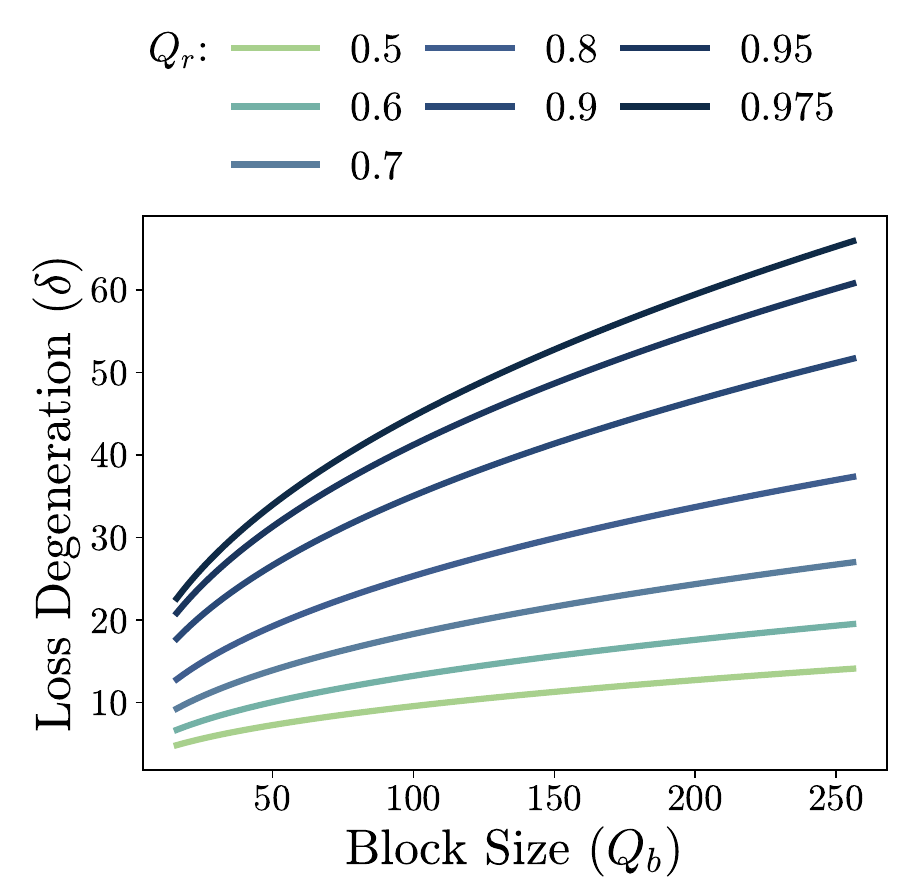}
        \captionsetup{font=scriptsize}
        \caption{60M Predicted (mean)}
    \end{subfigure}
    
\begin{subfigure}[b]{0.24\textwidth}
        \centering
        \includegraphics[width=\textwidth]{figures/figures-blocksize/clm-200m/loss_vs_blocksize.pdf}
        \captionsetup{font=scriptsize}
        \caption{200M Actual (min)}
    \end{subfigure}\hfill
    \begin{subfigure}[b]{0.24\textwidth}
        \centering
        \includegraphics[width=\textwidth]{figures/figures-blocksize/clm-200m/loss_vs_blocksize-fitted.pdf}
        \captionsetup{font=scriptsize}
        \caption{200M Predicted (min)}
    \end{subfigure}\hfill
    \begin{subfigure}[b]{0.24\textwidth}
        \centering
        \includegraphics[width=\textwidth]{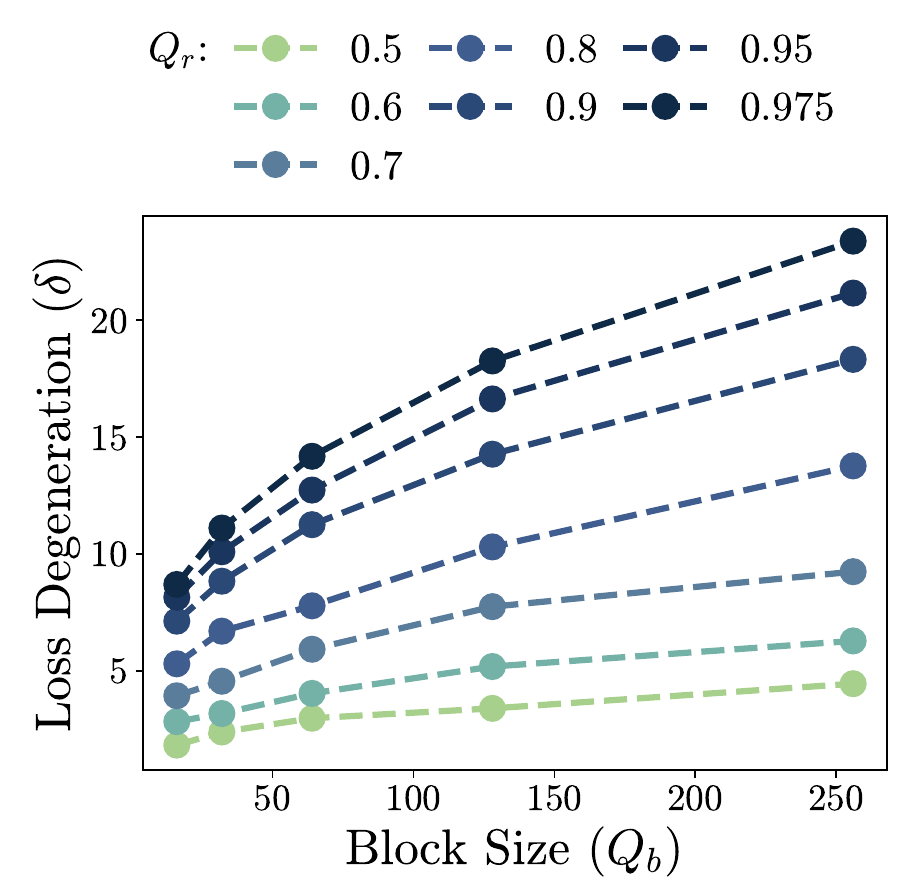}
        \captionsetup{font=scriptsize}
        \caption{200M Actual (mean)}
    \end{subfigure}\hfill
    \begin{subfigure}[b]{0.24\textwidth}
        \centering
        \includegraphics[width=\textwidth]{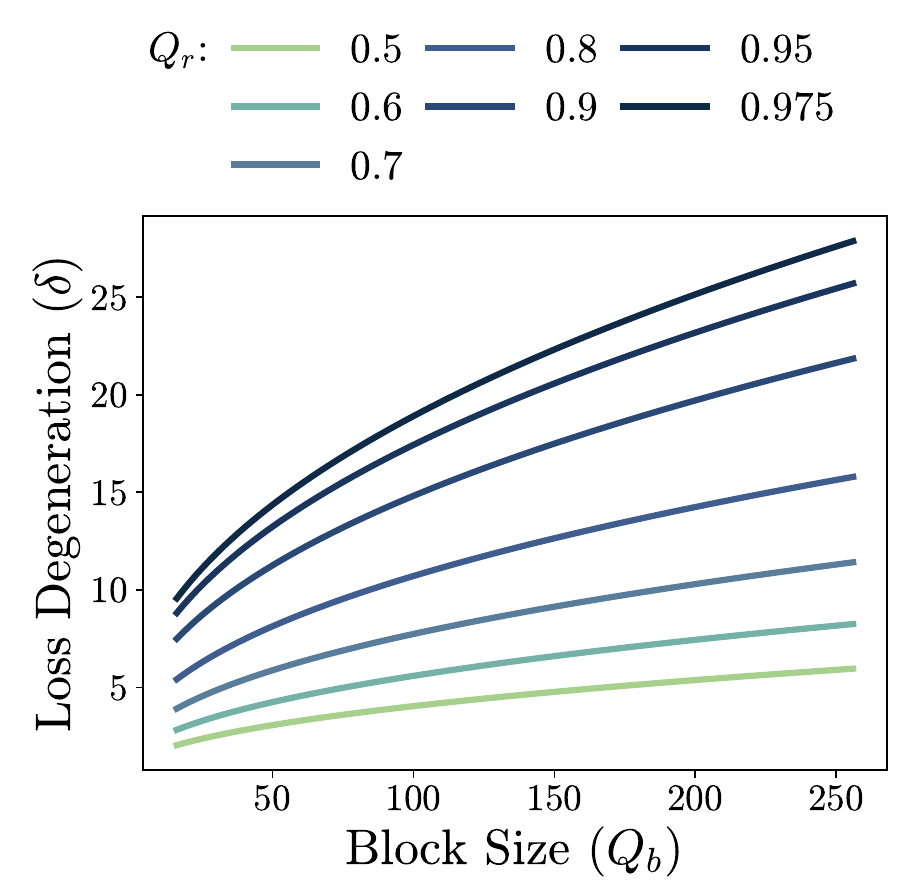}
        \captionsetup{font=scriptsize}
        \caption{200M Predicted (mean)}
    \end{subfigure}
    
\begin{subfigure}[b]{0.24\textwidth}
        \centering
        \includegraphics[width=\textwidth]{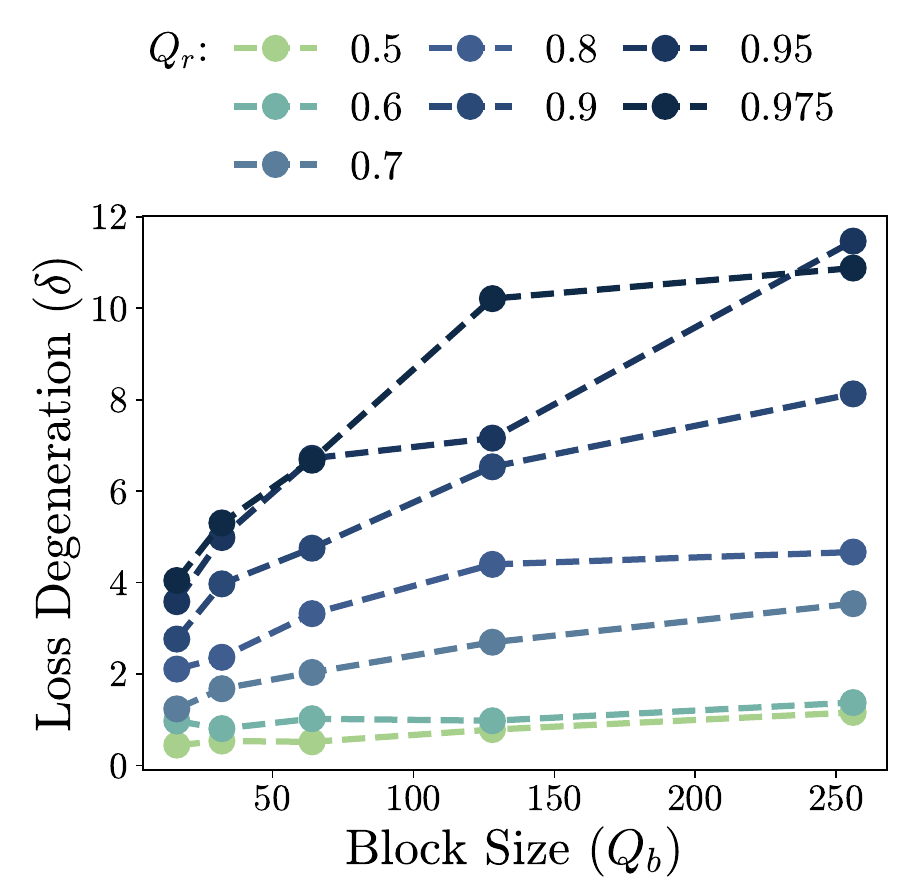}
        \captionsetup{font=scriptsize}
        \caption{400M Actual (min)}
    \end{subfigure}\hfill
    \begin{subfigure}[b]{0.24\textwidth}
        \centering
        \includegraphics[width=\textwidth]{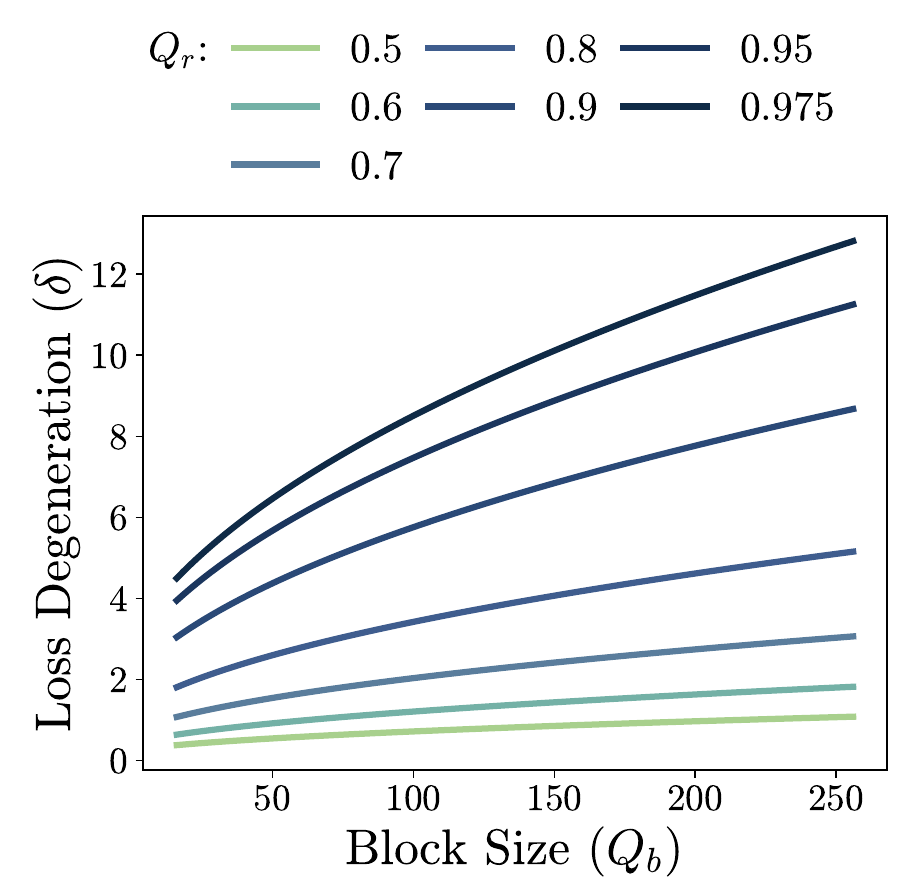}
        \captionsetup{font=scriptsize}
        \caption{400M Predicted (min)}
    \end{subfigure}\hfill
    \begin{subfigure}[b]{0.24\textwidth}
        \centering
        \includegraphics[width=\textwidth]{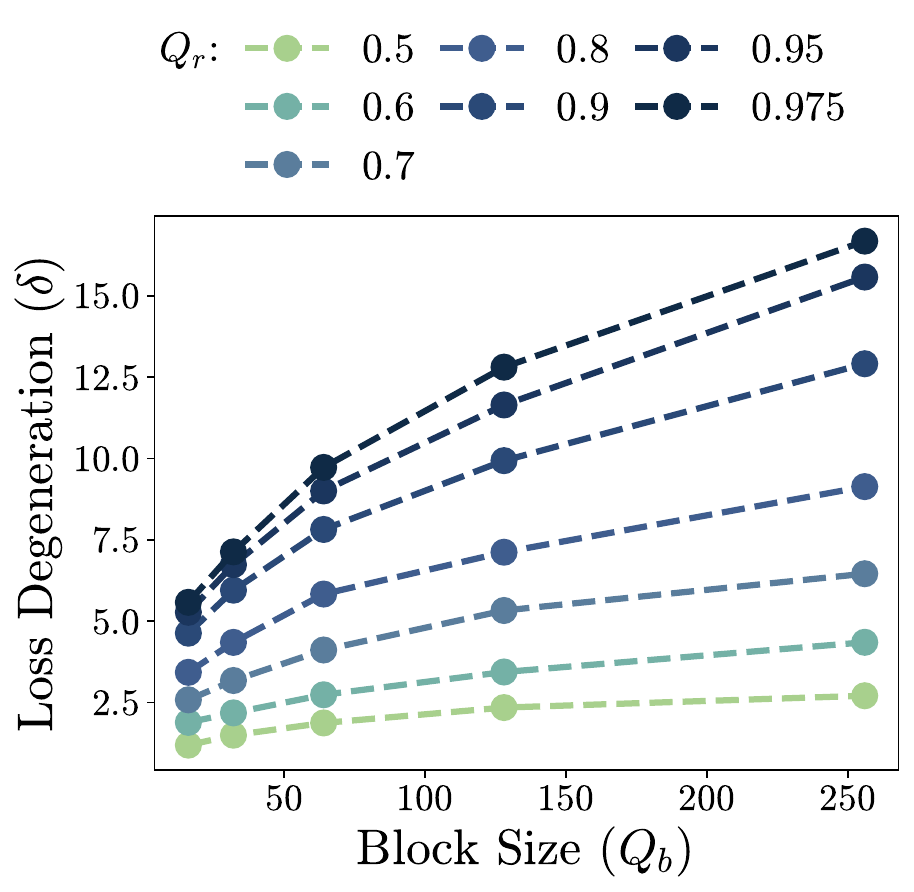}
        \captionsetup{font=scriptsize}
        \caption{400M Actual (mean)}
    \end{subfigure}\hfill
    \begin{subfigure}[b]{0.24\textwidth}
        \centering
        \includegraphics[width=\textwidth]{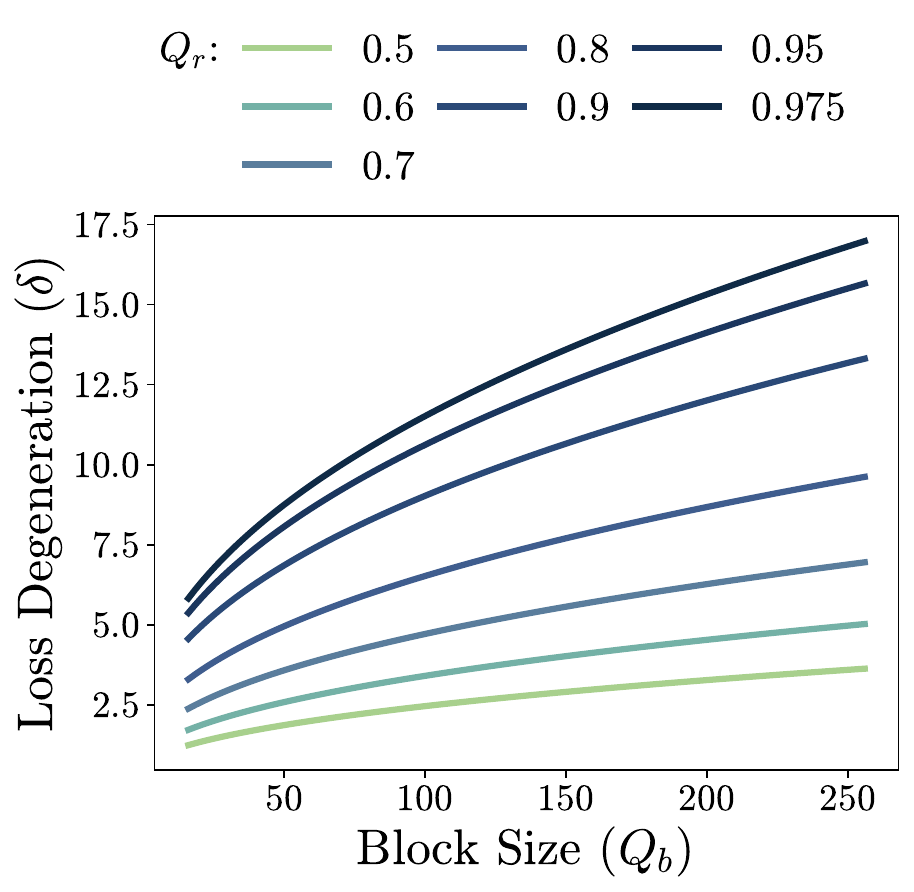}
        \captionsetup{font=scriptsize}
        \caption{400M Predicted (mean)}
    \end{subfigure}
    
\begin{subfigure}[b]{0.24\textwidth}
        \centering
        \includegraphics[width=\textwidth]{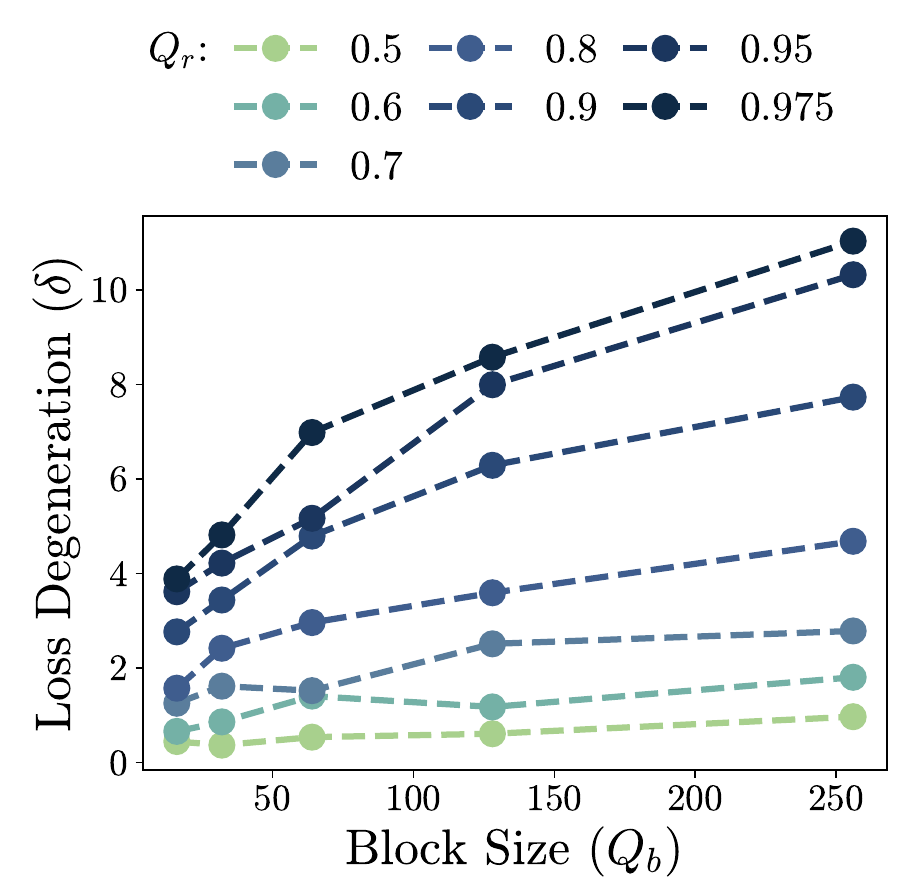}
        \captionsetup{font=scriptsize}
        \caption{600M Actual (min)}
    \end{subfigure}\hfill
    \begin{subfigure}[b]{0.24\textwidth}
        \centering
        \includegraphics[width=\textwidth]{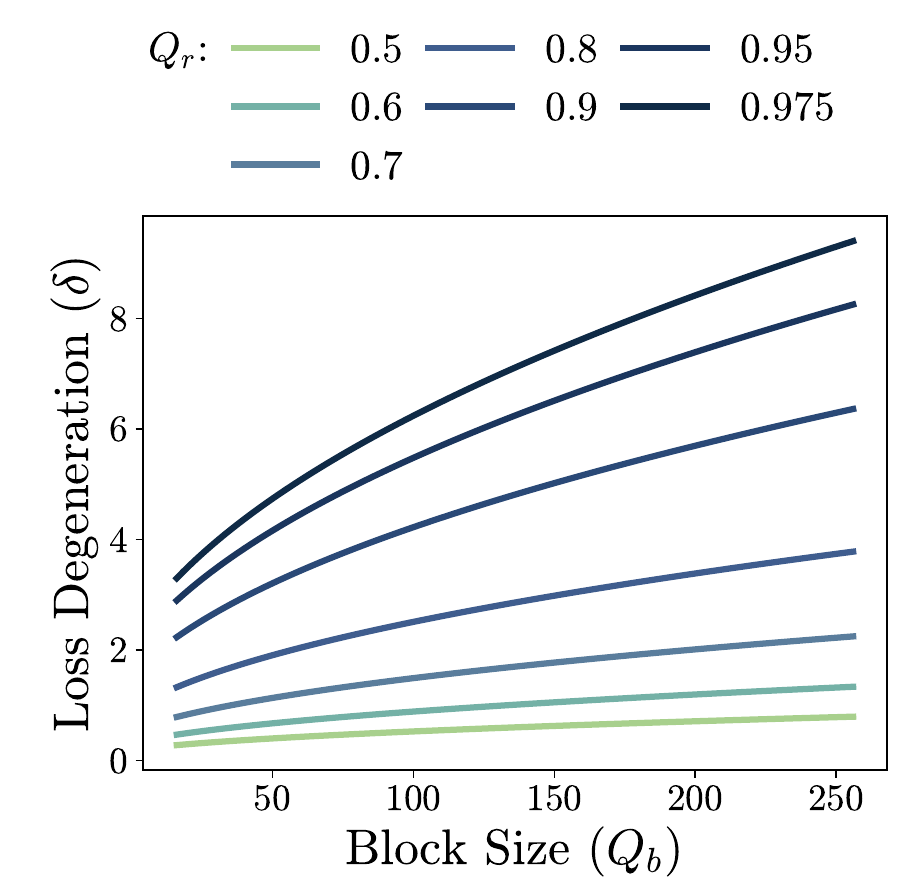}
        \captionsetup{font=scriptsize}
        \caption{600M Predicted (min)}
    \end{subfigure}\hfill
    \begin{subfigure}[b]{0.24\textwidth}
        \centering
        \includegraphics[width=\textwidth]{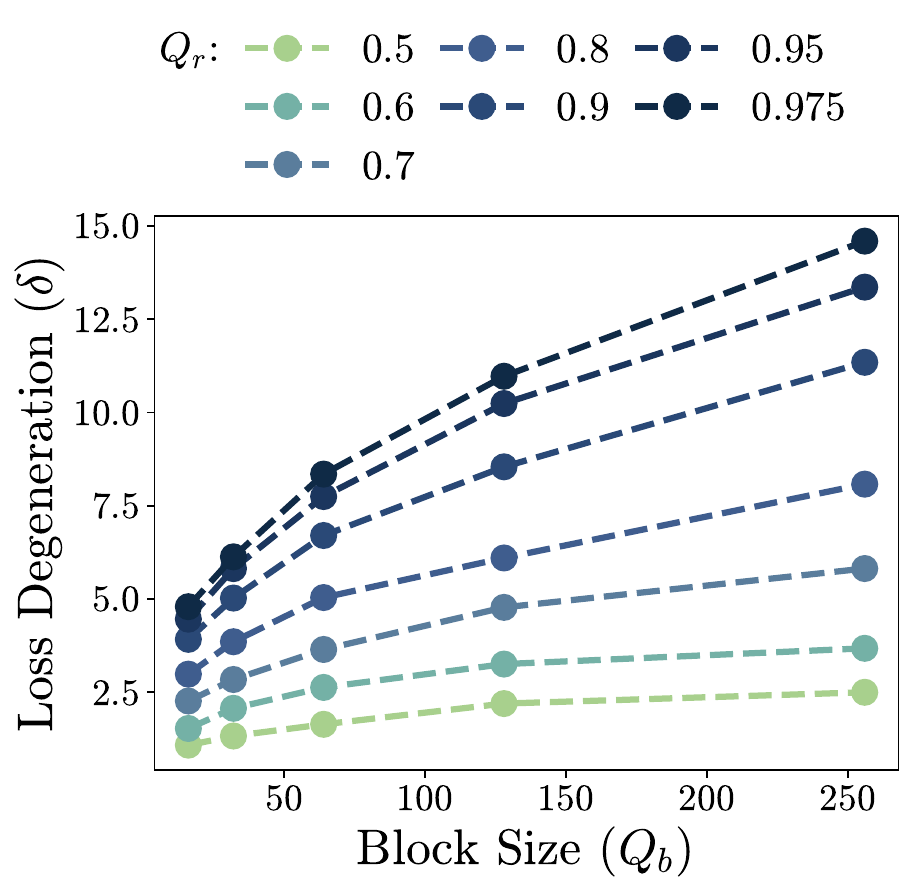}
        \captionsetup{font=scriptsize}
        \caption{600M Actual (mean)}
    \end{subfigure}\hfill
    \begin{subfigure}[b]{0.24\textwidth}
        \centering
        \includegraphics[width=\textwidth]{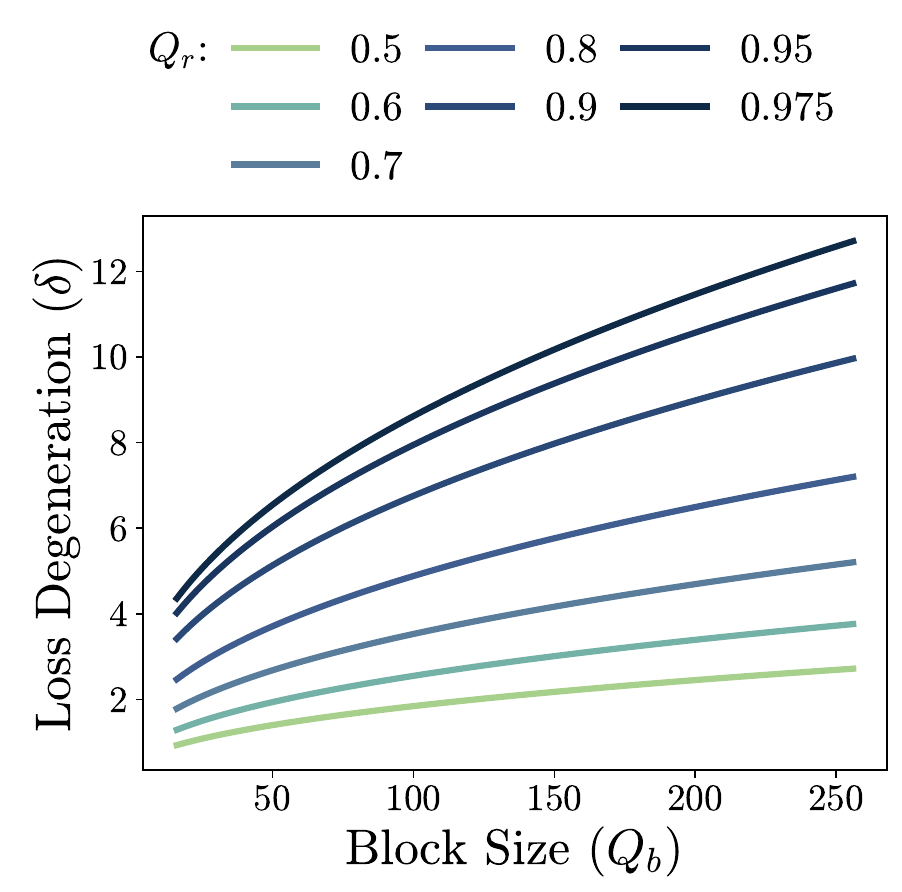}
        \captionsetup{font=scriptsize}
        \caption{600M Predicted (mean)}
    \end{subfigure}
    
\begin{subfigure}[b]{0.24\textwidth}
        \centering
        \includegraphics[width=\textwidth]{figures/figures-blocksize/clm-1.1b/loss_vs_blocksize.pdf}
        \captionsetup{font=scriptsize}
        \caption{1.1B Actual (min)}
    \end{subfigure}\hfill
    \begin{subfigure}[b]{0.24\textwidth}
        \centering
        \includegraphics[width=\textwidth]{figures/figures-blocksize/clm-1.1b/loss_vs_blocksize-fitted.pdf}
        \captionsetup{font=scriptsize}
        \caption{1.1B Predicted (min)}
    \end{subfigure}\hfill
    \begin{subfigure}[b]{0.24\textwidth}
        \centering
        \includegraphics[width=\textwidth]{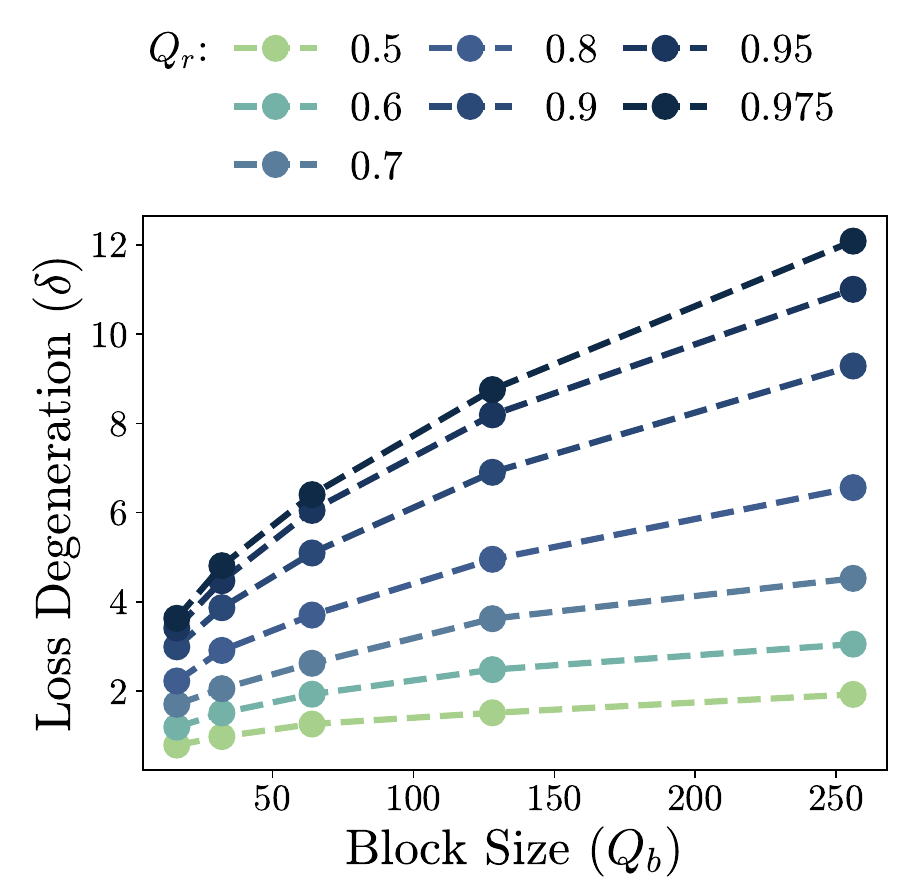}
        \captionsetup{font=scriptsize}
        \caption{1.1B Actual (mean)}
    \end{subfigure}\hfill
    \begin{subfigure}[b]{0.24\textwidth}
        \centering
        \includegraphics[width=\textwidth]{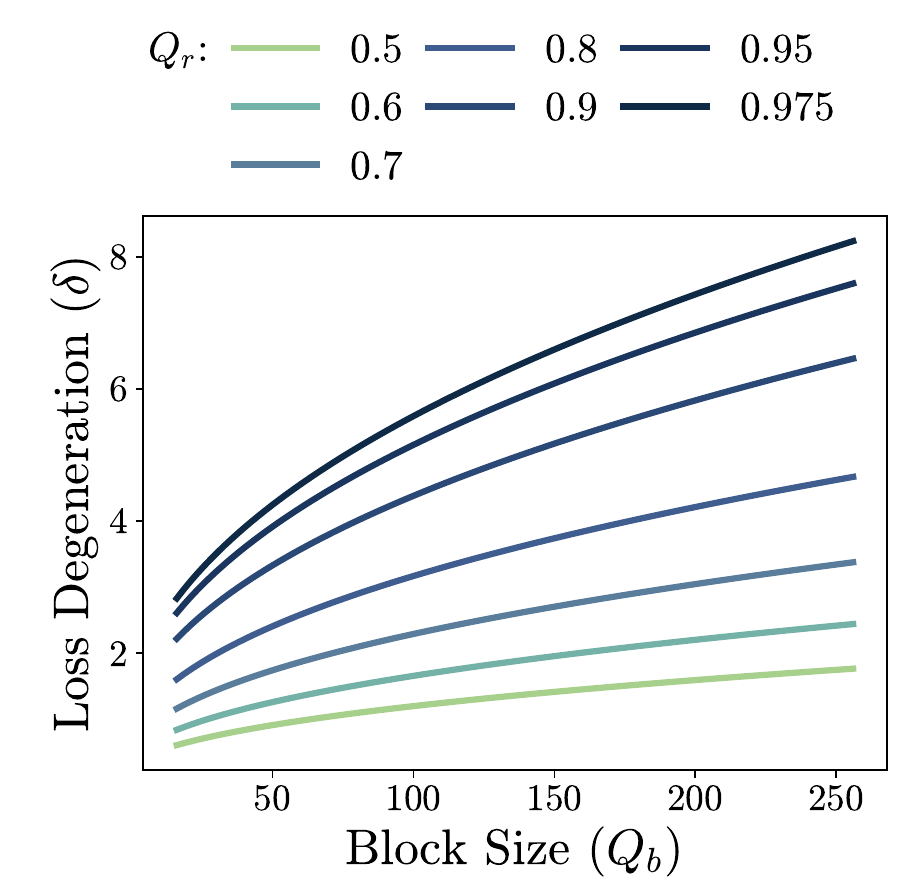}
        \captionsetup{font=scriptsize}
        \caption{1.1B Predicted (mean)}
    \end{subfigure}
    
\caption{\textbf{Strong Law CLM Matrix Multiplication-wise ($\delta^{\text{opt}}$, $\delta_\mu$)} (a,b,e,f,i,j,m,n,q,r) CLM matrix multiplication-wise $\delta^{\text{opt}}$ results; (c,d,g,h,k,l,o,p,s,t) CLM matrix multiplication-wise $\delta_\mu$ results.}
    \label{fig:appendix-blcoksize-clm}
    \end{figure}

\begin{figure}[htbp]
    \centering
\begin{subfigure}[b]{0.24\textwidth}
        \centering
        \includegraphics[width=\textwidth]{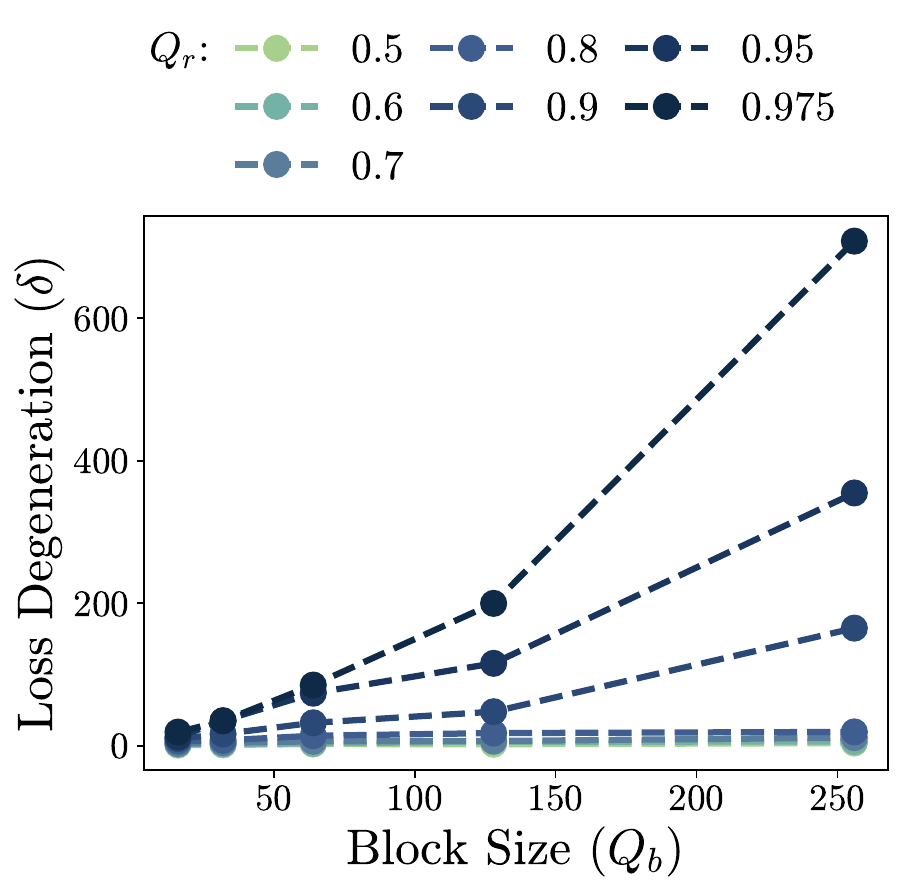}
        \captionsetup{font=scriptsize}
        \caption{0.5B Actual (min)}
    \end{subfigure}\hfill
    \begin{subfigure}[b]{0.24\textwidth}
        \centering
        \includegraphics[width=\textwidth]{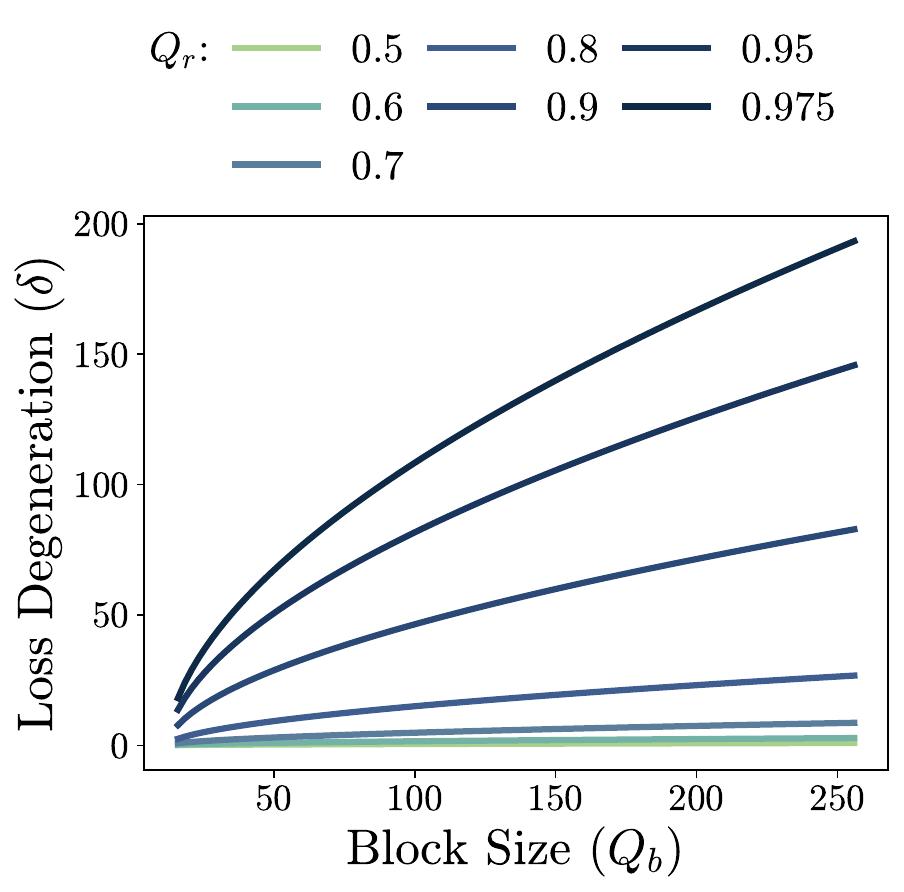}
        \captionsetup{font=scriptsize}
        \caption{0.5B Predicted (min)}
    \end{subfigure}\hfill
    \begin{subfigure}[b]{0.24\textwidth}
        \centering
        \includegraphics[width=\textwidth]{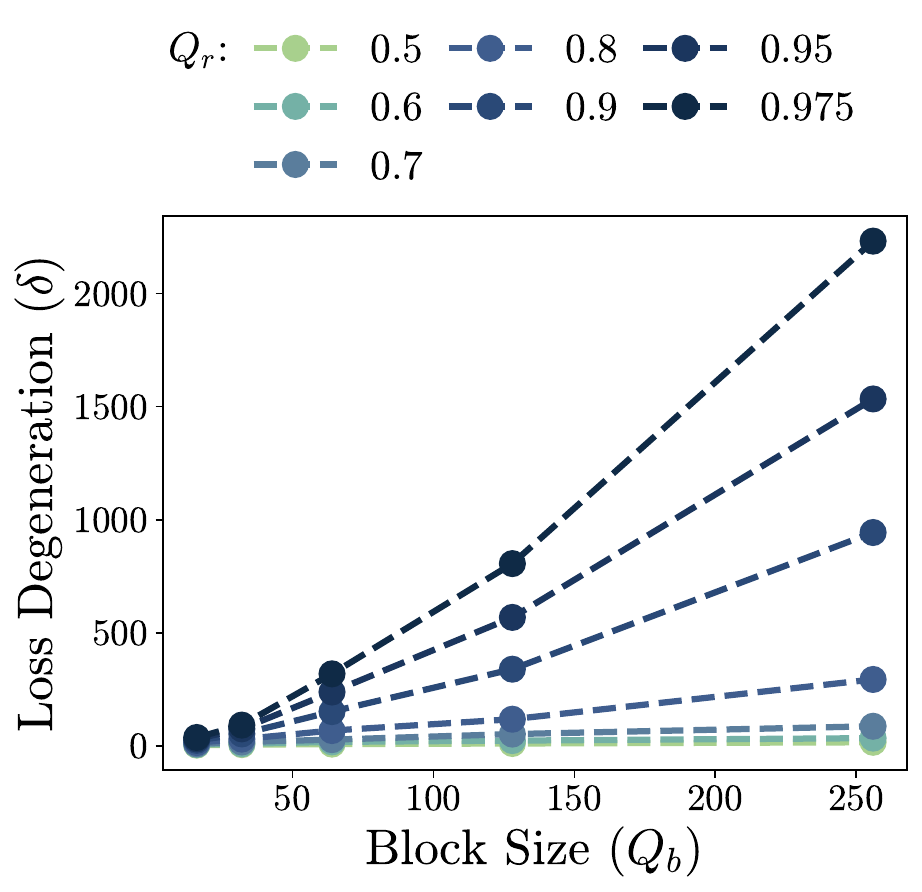}
        \captionsetup{font=scriptsize}
        \caption{0.5B Actual (mean)}
    \end{subfigure}\hfill
    \begin{subfigure}[b]{0.24\textwidth}
        \centering
        \includegraphics[width=\textwidth]{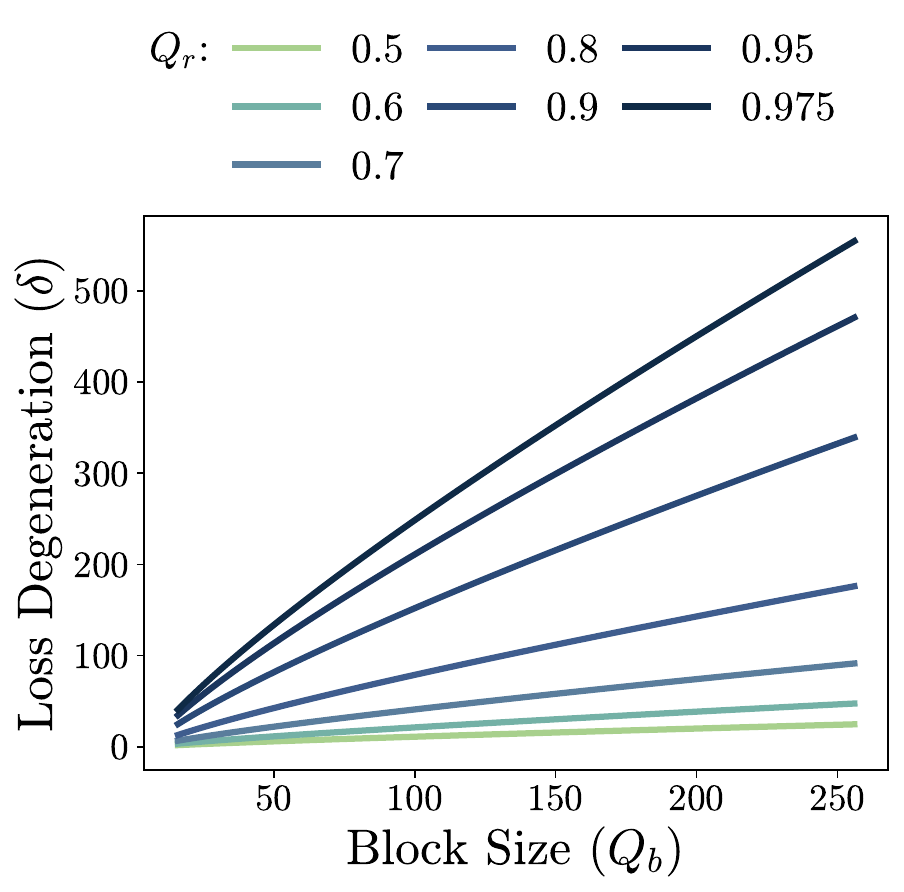}
        \captionsetup{font=scriptsize}
        \caption{0.5B Predicted (mean)}
    \end{subfigure}
    
\begin{subfigure}[b]{0.24\textwidth}
        \centering
        \includegraphics[width=\textwidth]{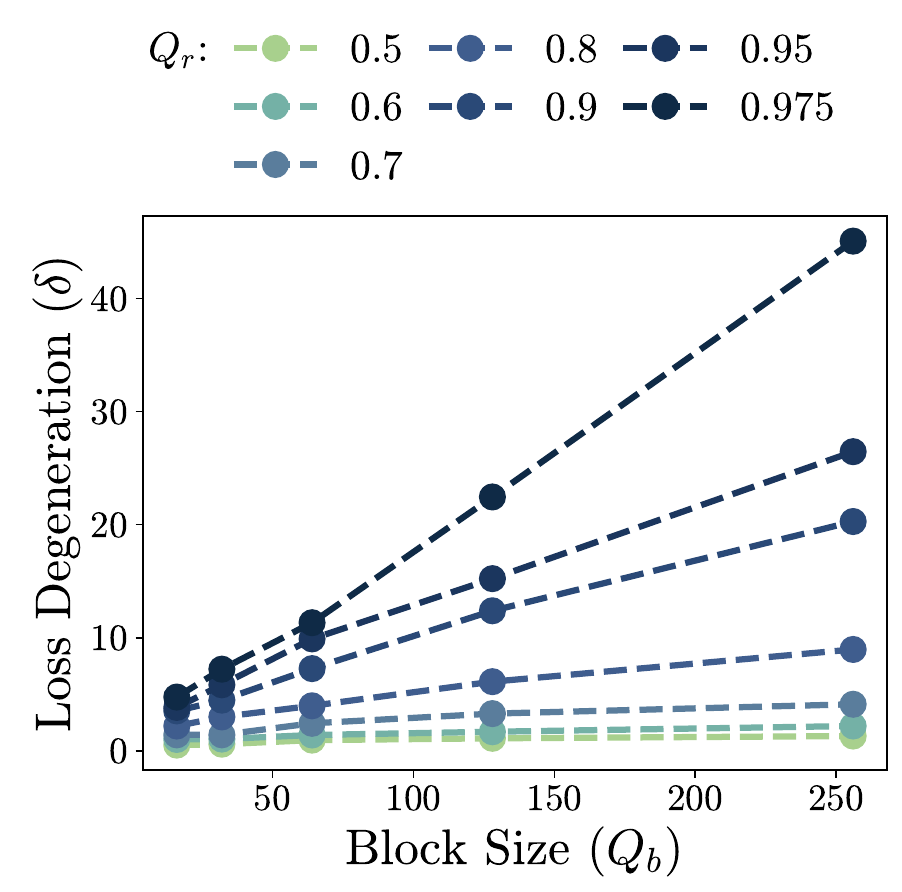}
        \captionsetup{font=scriptsize}
        \caption{1.8B Actual (min)}
    \end{subfigure}\hfill
    \begin{subfigure}[b]{0.24\textwidth}
        \centering
        \includegraphics[width=\textwidth]{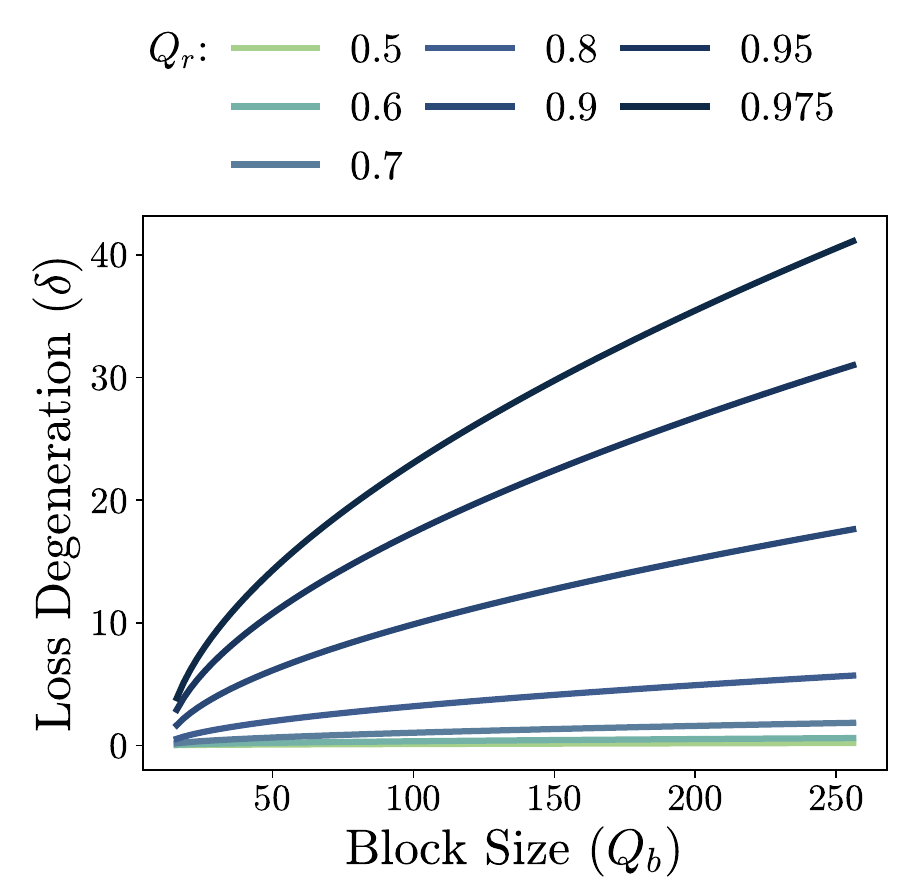}
        \captionsetup{font=scriptsize}
        \caption{1.8B Predicted (min)}
    \end{subfigure}\hfill
    \begin{subfigure}[b]{0.24\textwidth}
        \centering
        \includegraphics[width=\textwidth]{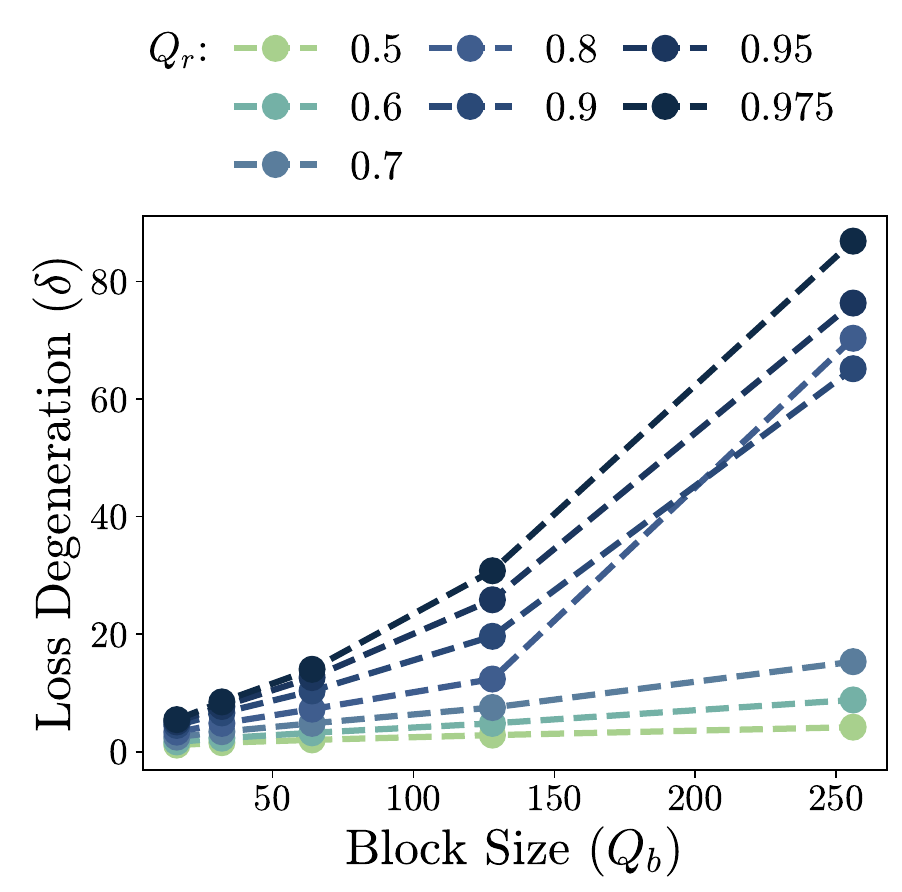}
        \captionsetup{font=scriptsize}
        \caption{1.8B Actual (mean)}
    \end{subfigure}\hfill
    \begin{subfigure}[b]{0.24\textwidth}
        \centering
        \includegraphics[width=\textwidth]{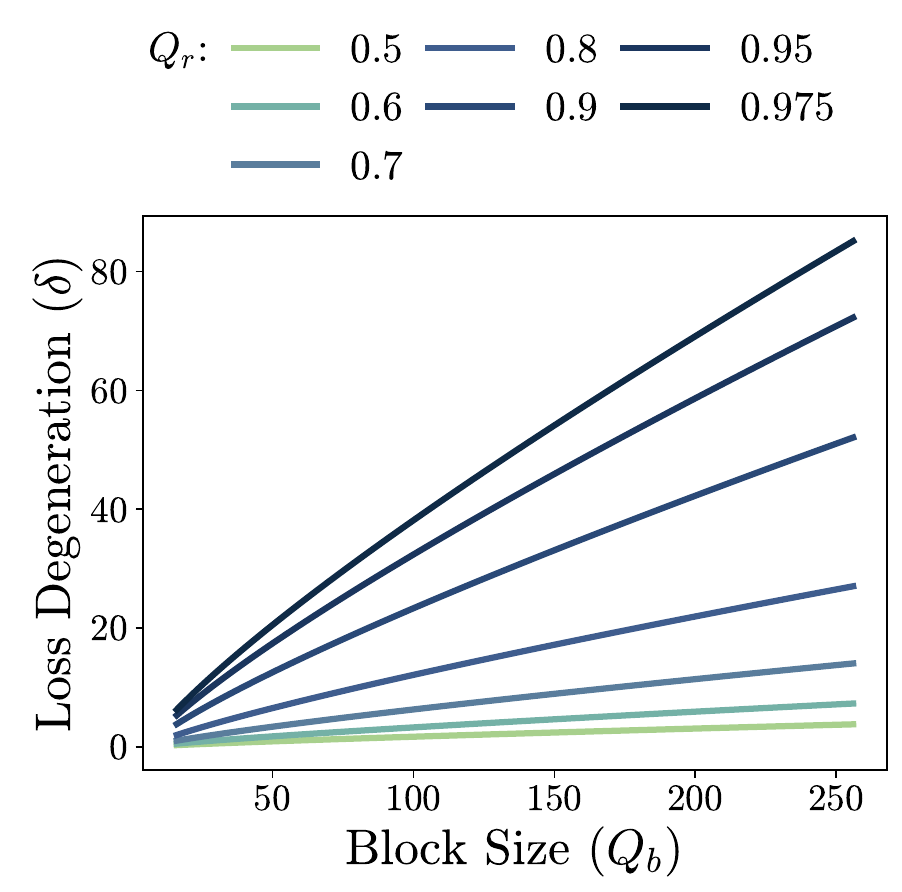}
        \captionsetup{font=scriptsize}
        \caption{1.8B Predicted (mean)}
    \end{subfigure}
    
\begin{subfigure}[b]{0.24\textwidth}
        \centering
        \includegraphics[width=\textwidth]{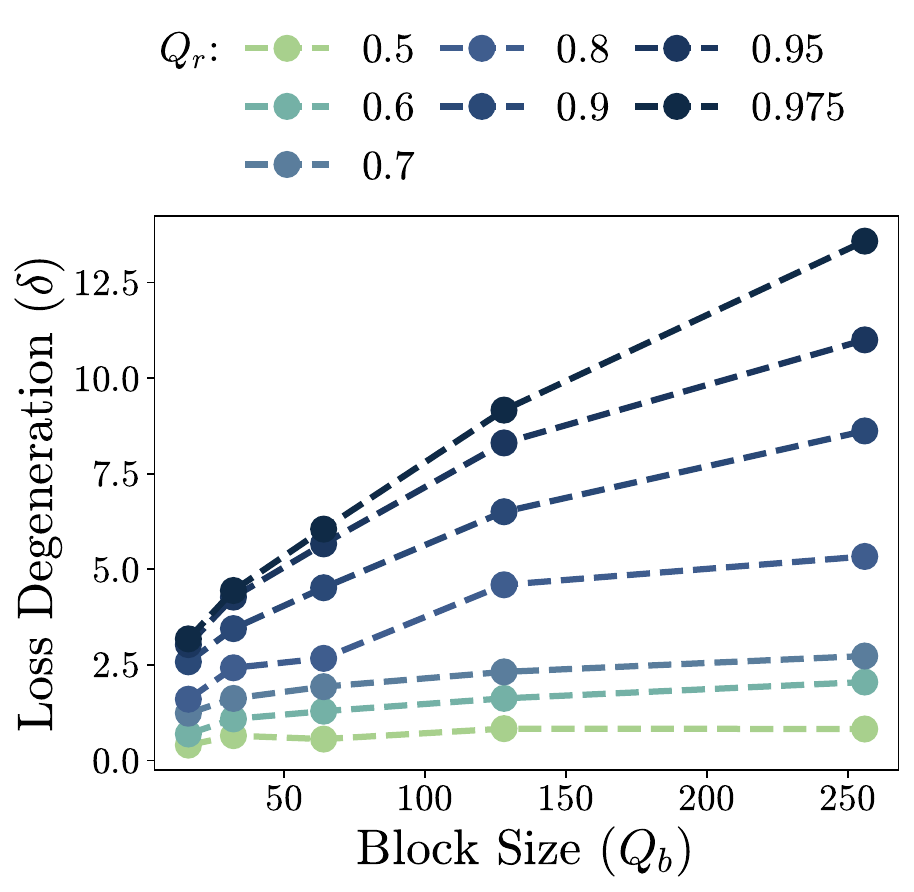}
        \captionsetup{font=scriptsize}
        \caption{4B Actual (min)}
    \end{subfigure}\hfill
    \begin{subfigure}[b]{0.24\textwidth}
        \centering
        \includegraphics[width=\textwidth]{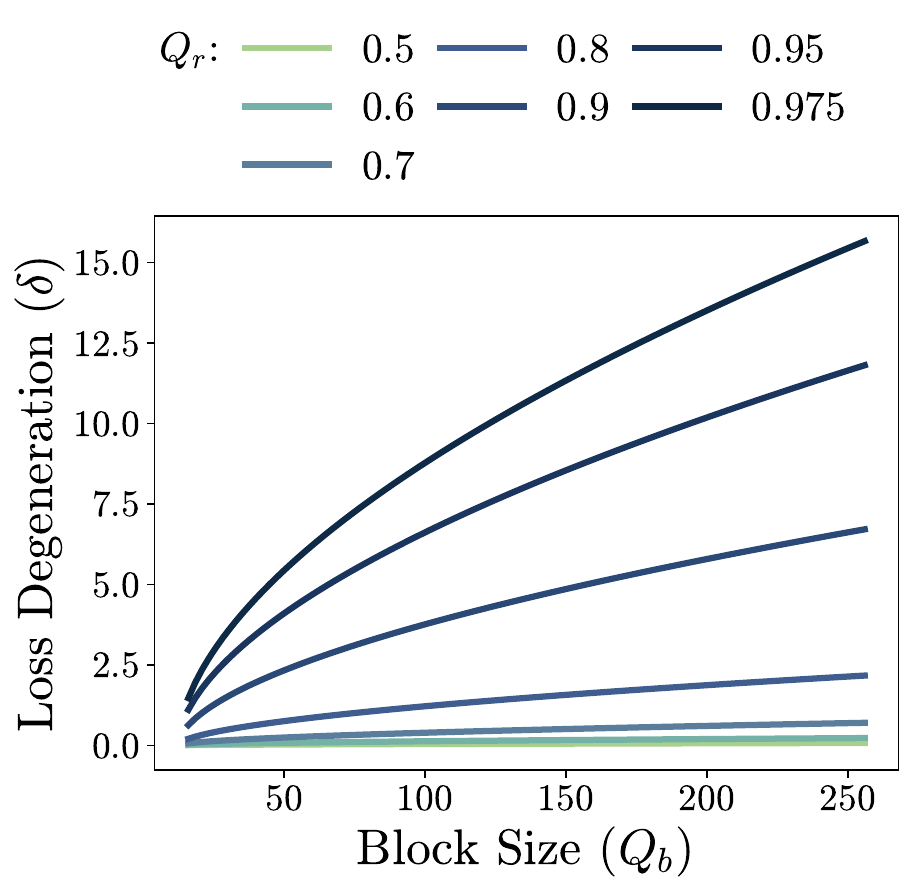}
        \captionsetup{font=scriptsize}
        \caption{4B Predicted (min)}
    \end{subfigure}\hfill
    \begin{subfigure}[b]{0.24\textwidth}
        \centering
        \includegraphics[width=\textwidth]{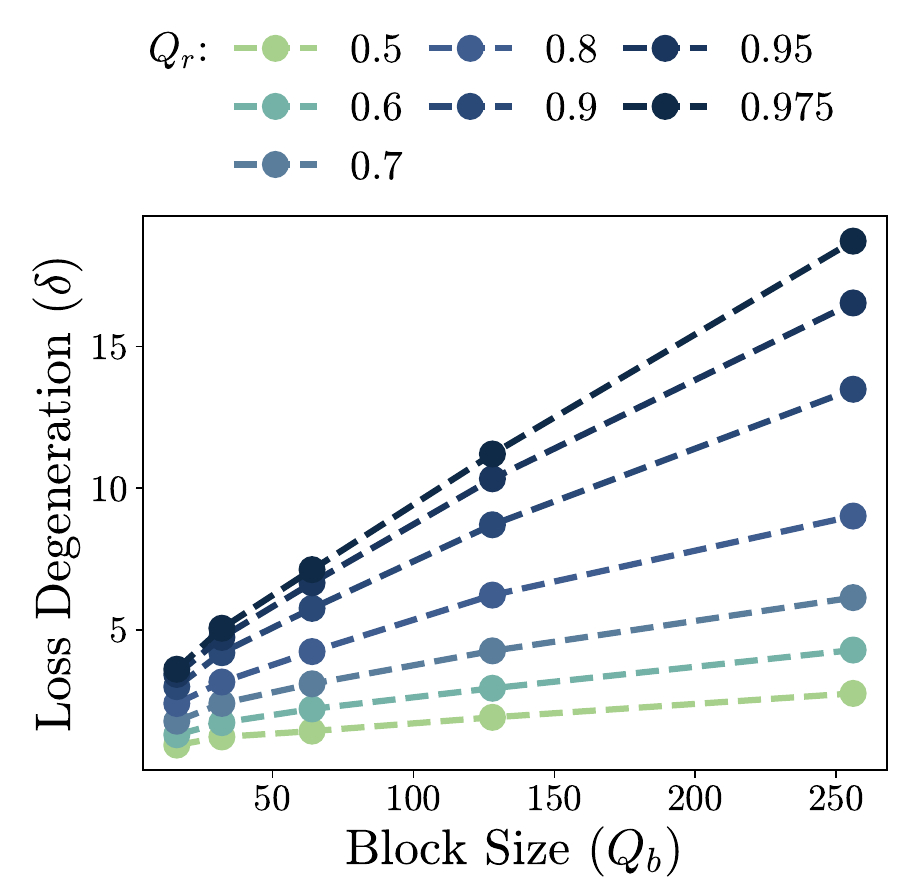}
        \captionsetup{font=scriptsize}
        \caption{4B Actual (mean)}
    \end{subfigure}\hfill
    \begin{subfigure}[b]{0.24\textwidth}
        \centering
        \includegraphics[width=\textwidth]{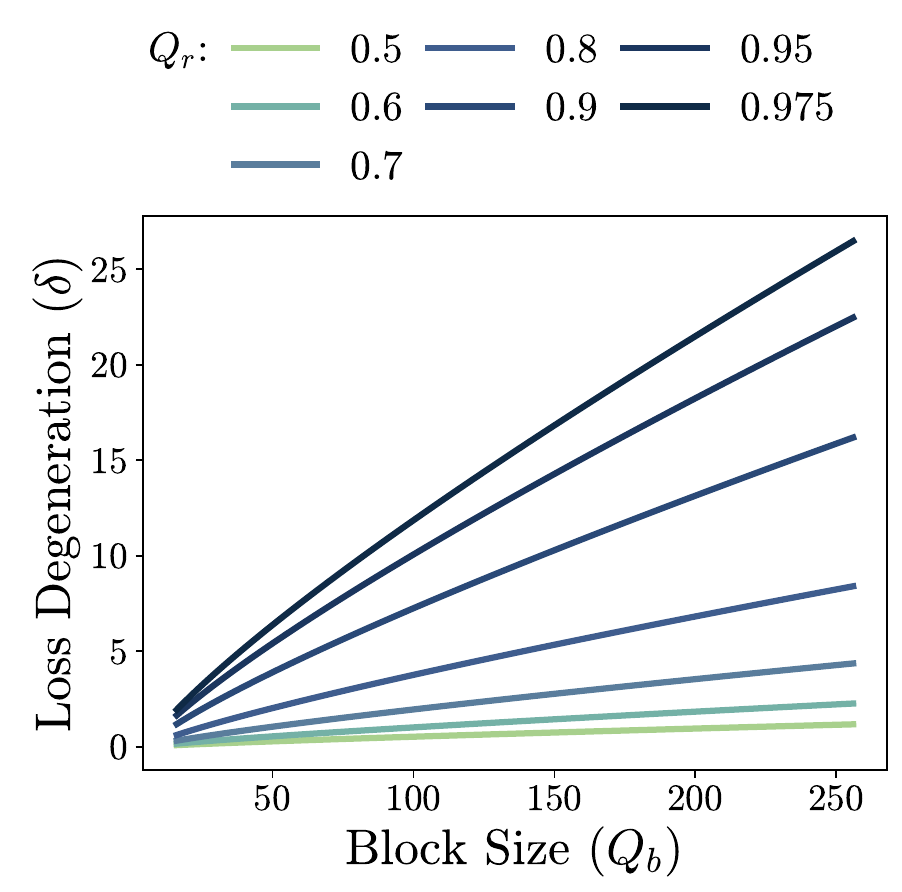}
        \captionsetup{font=scriptsize}
        \caption{4B Predicted (mean)}
    \end{subfigure}
    
\begin{subfigure}[b]{0.24\textwidth}
        \centering
        \includegraphics[width=\textwidth]{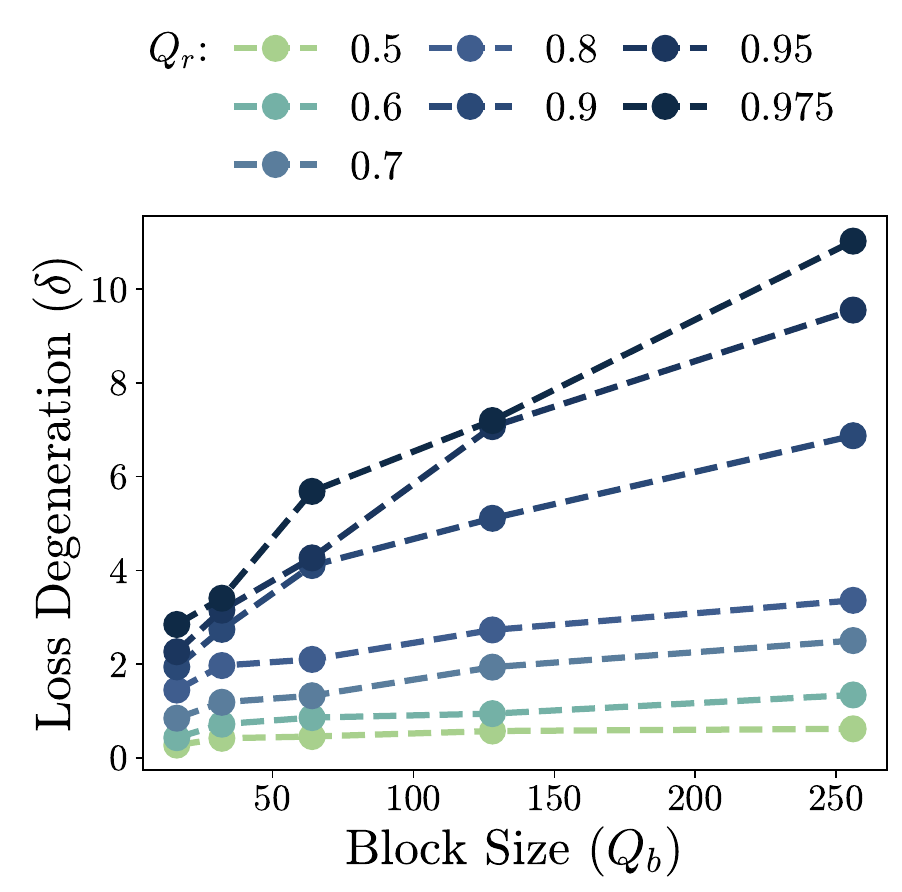}
        \captionsetup{font=scriptsize}
        \caption{7B Actual (min)}
    \end{subfigure}\hfill
    \begin{subfigure}[b]{0.24\textwidth}
        \centering
        \includegraphics[width=\textwidth]{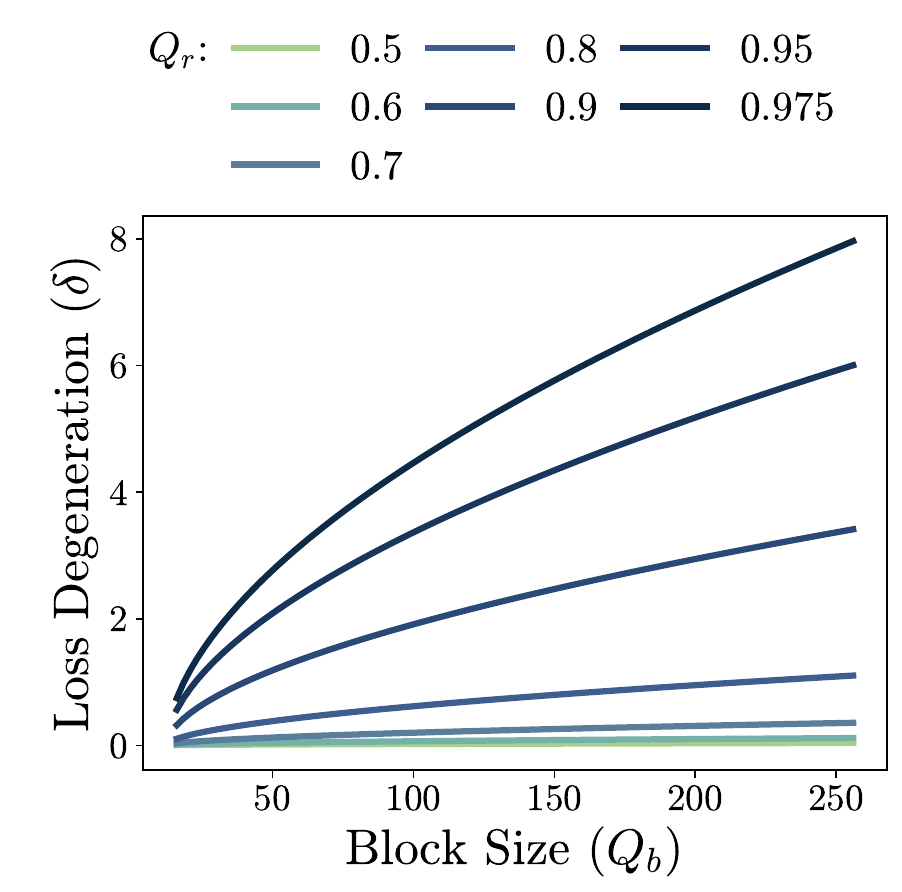}
        \captionsetup{font=scriptsize}
        \caption{7B Predicted (min)}
    \end{subfigure}\hfill
    \begin{subfigure}[b]{0.24\textwidth}
        \centering
        \includegraphics[width=\textwidth]{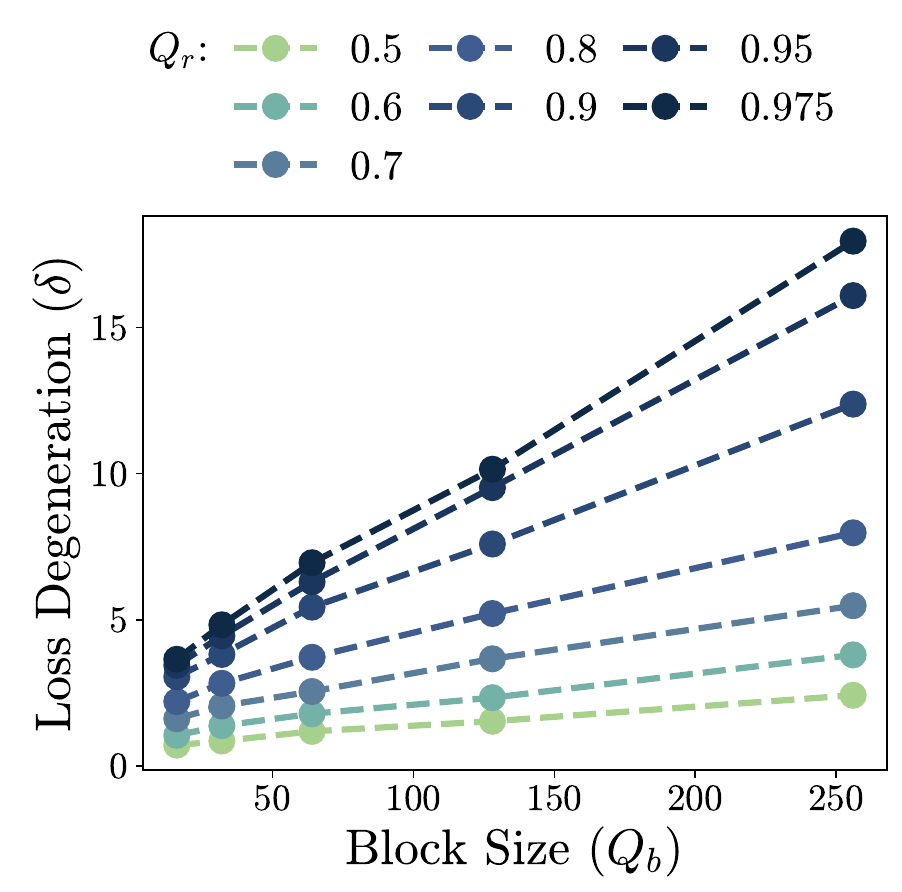}
        \captionsetup{font=scriptsize}
        \caption{7B Actual (mean)}
    \end{subfigure}\hfill
    \begin{subfigure}[b]{0.24\textwidth}
        \centering
        \includegraphics[width=\textwidth]{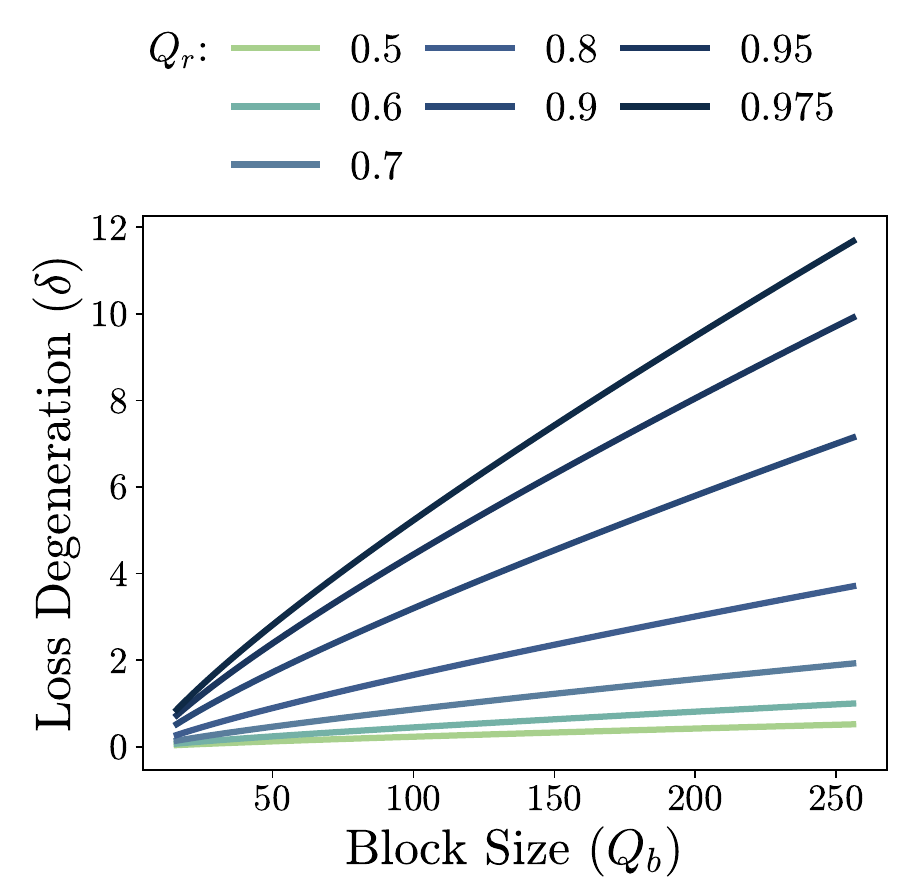}
        \captionsetup{font=scriptsize}
        \caption{7B Predicted (mean)}
    \end{subfigure}
    
\caption{\textbf{Strong Law Qwen-1.5 Matrix Multiplication-wise ($\delta^{\text{opt}}$, $\delta_\mu$)} (a,b,e,f,i,j,m,n) Qwen1.5 matrix multiplication-wise $\delta^{\text{opt}}$ results; (c,d,g,h,k,l,o,p) Qwen1.5 matrix multiplication-wise $\delta_\mu$ results.}
    \label{fig:appendix-blcoksize-qwen1.5}
    \end{figure}

\begin{figure}[htbp]
    \centering
\begin{subfigure}[b]{0.24\textwidth}
        \centering
        \includegraphics[width=\textwidth]{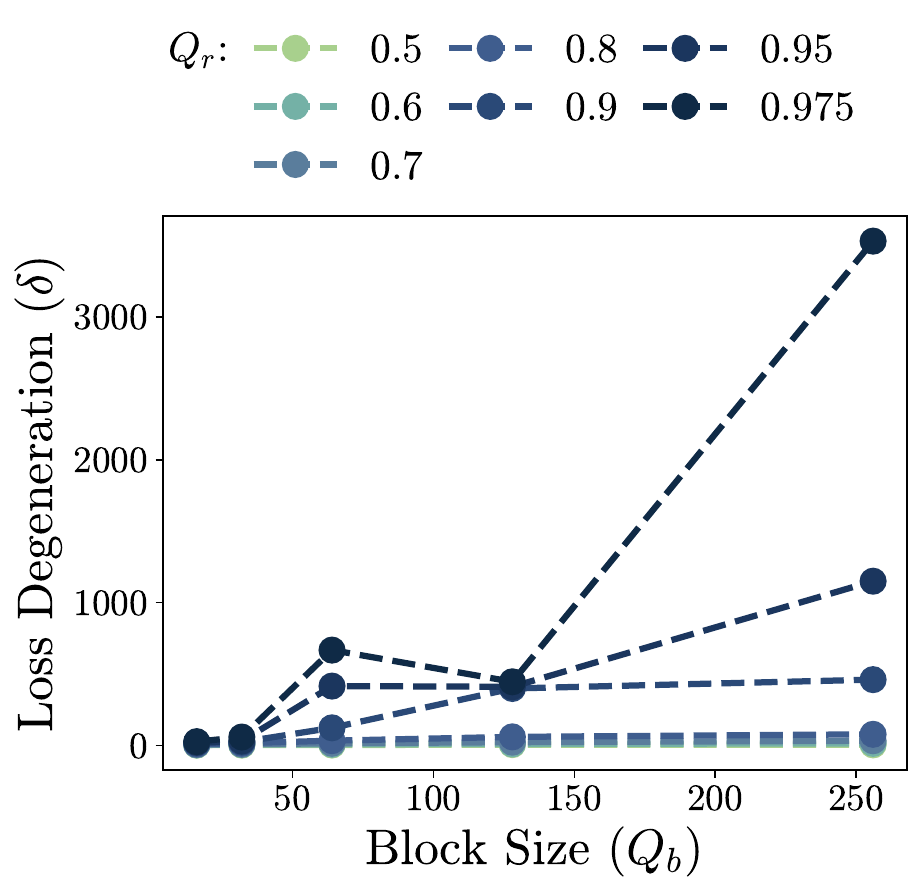}
        \captionsetup{font=scriptsize}
        \caption{0.6B Actual (min)}
    \end{subfigure}\hfill
    \begin{subfigure}[b]{0.24\textwidth}
        \centering
        \includegraphics[width=\textwidth]{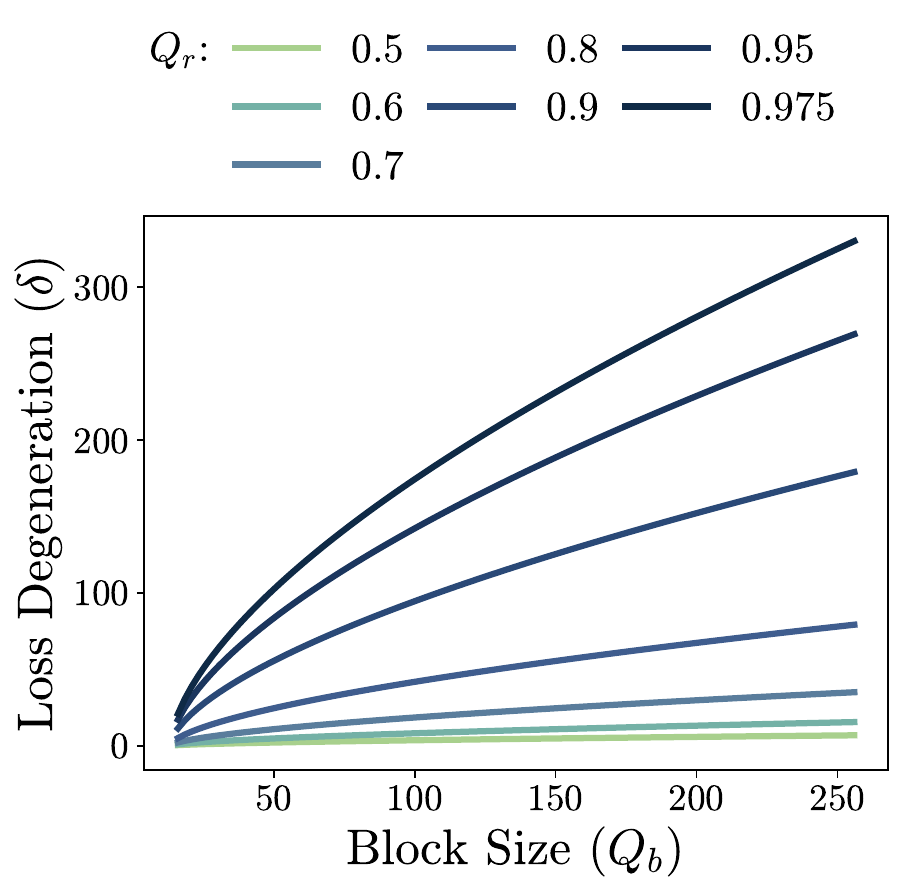}
        \captionsetup{font=scriptsize}
        \caption{0.6B Predicted (min)}
    \end{subfigure}\hfill
    \begin{subfigure}[b]{0.24\textwidth}
        \centering
        \includegraphics[width=\textwidth]{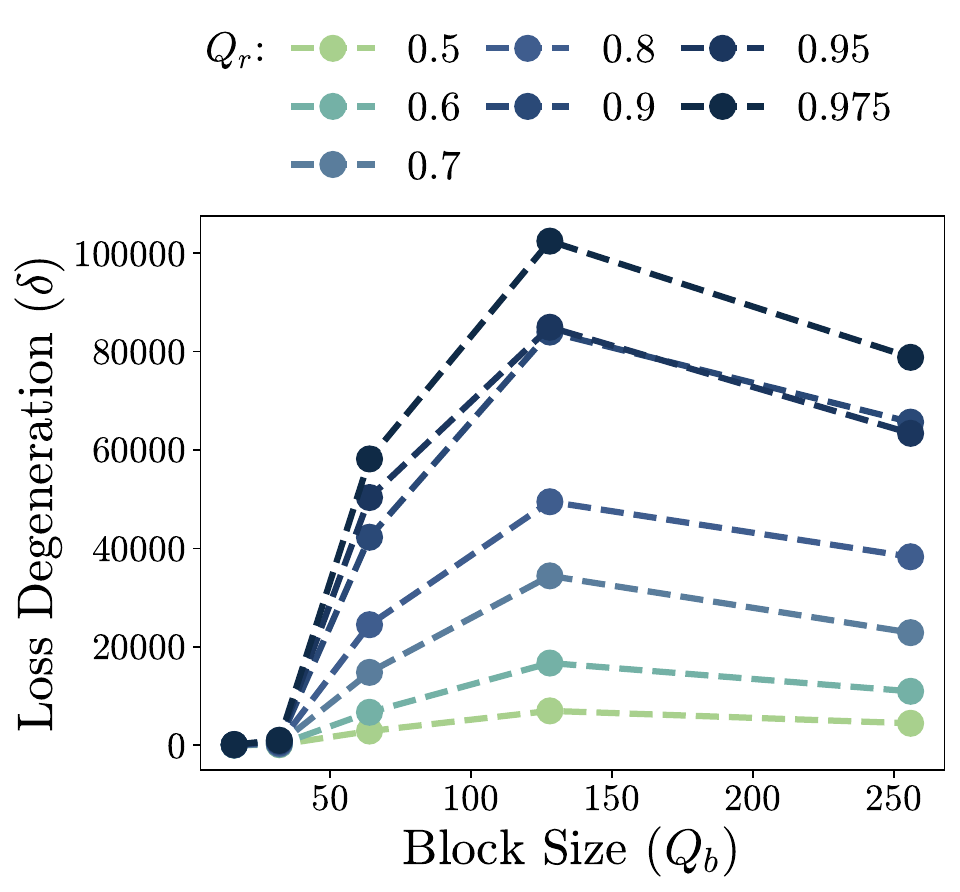}
        \captionsetup{font=scriptsize}
        \caption{0.6B Actual (mean)}
    \end{subfigure}\hfill
    \begin{subfigure}[b]{0.24\textwidth}
        \centering
        \includegraphics[width=\textwidth]{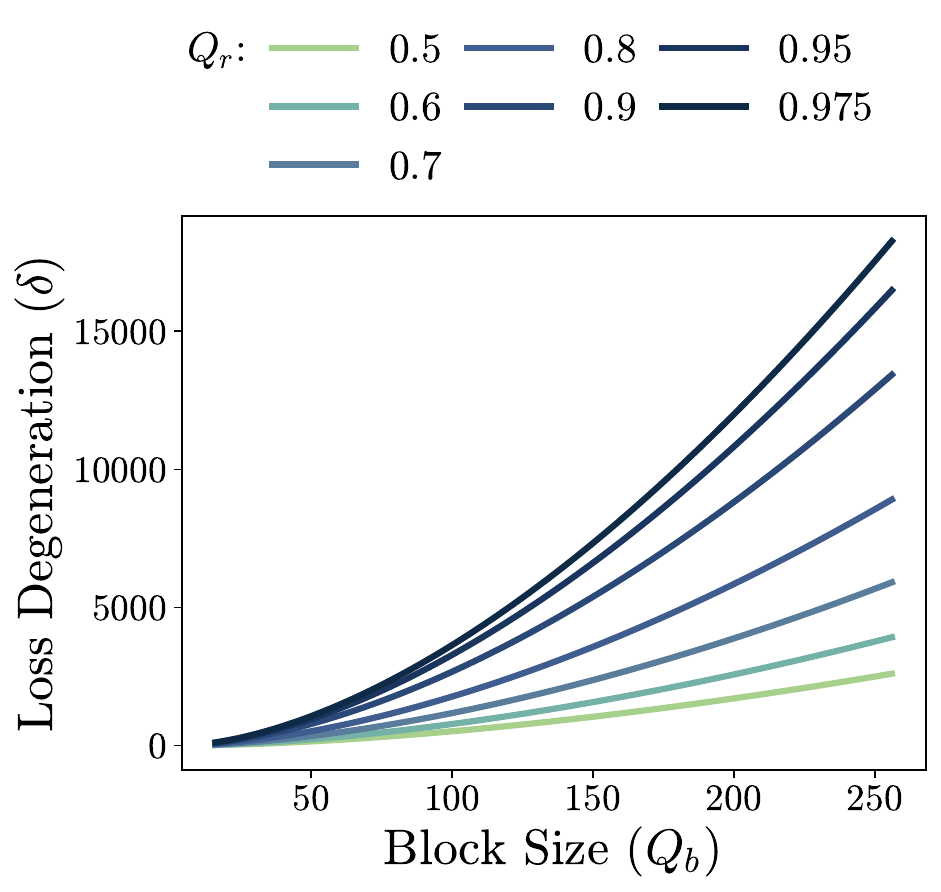}
        \captionsetup{font=scriptsize}
        \caption{0.6B Predicted (mean)}
    \end{subfigure}
    
\begin{subfigure}[b]{0.24\textwidth}
        \centering
        \includegraphics[width=\textwidth]{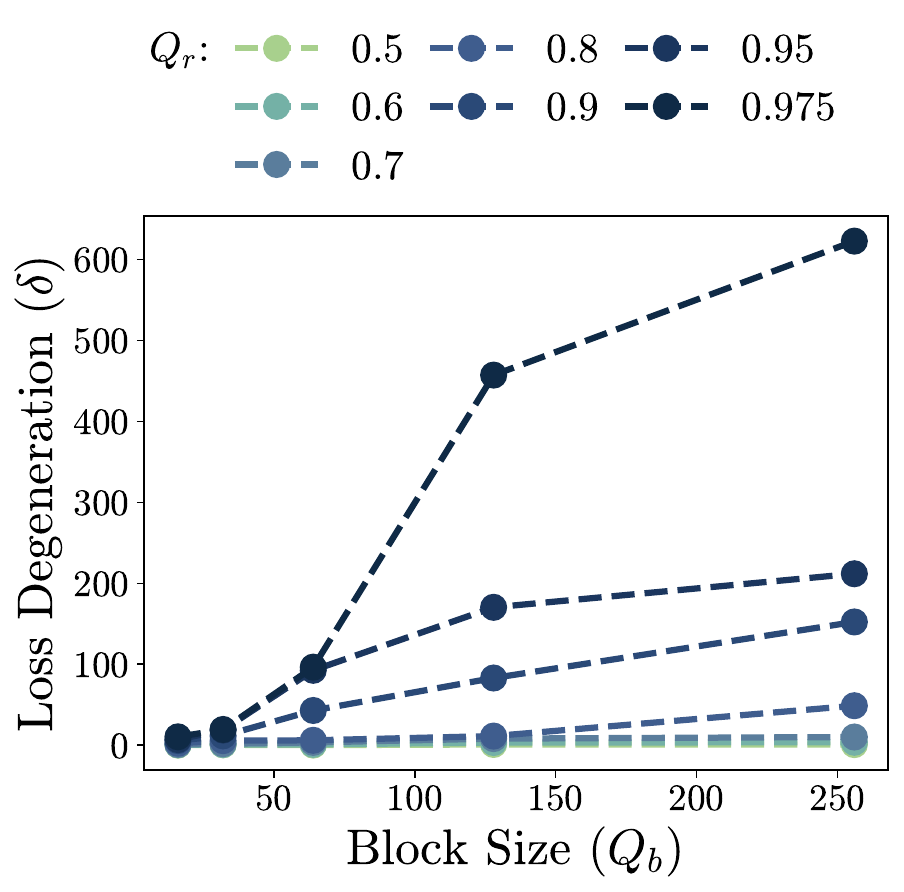}
        \captionsetup{font=scriptsize}
        \caption{1.7B Actual (min)}
    \end{subfigure}\hfill
    \begin{subfigure}[b]{0.24\textwidth}
        \centering
        \includegraphics[width=\textwidth]{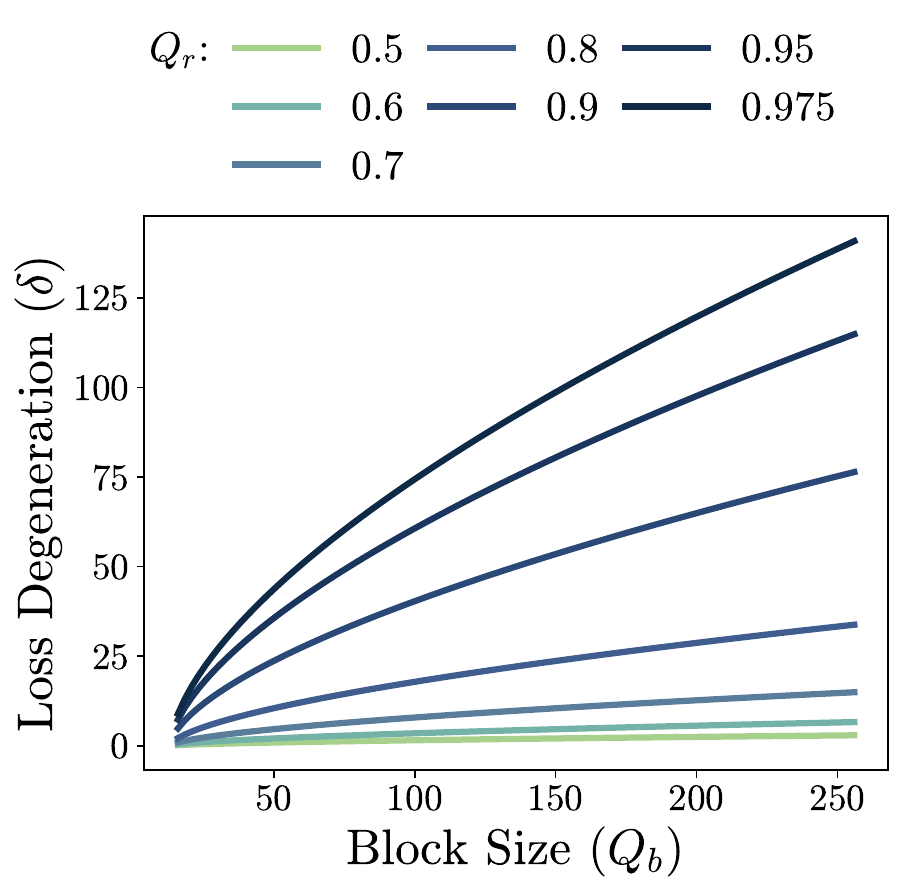}
        \captionsetup{font=scriptsize}
        \caption{1.7B Predicted (min)}
    \end{subfigure}\hfill
    \begin{subfigure}[b]{0.24\textwidth}
        \centering
        \includegraphics[width=\textwidth]{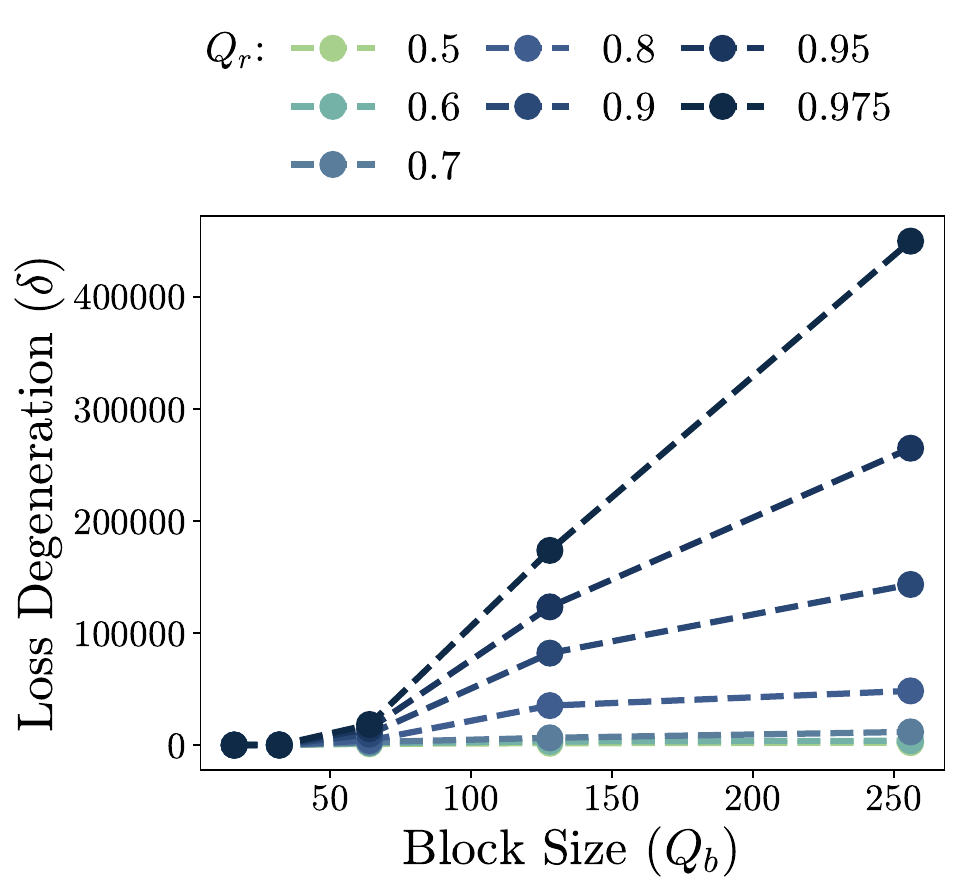}
        \captionsetup{font=scriptsize}
        \caption{1.7B Actual (mean)}
    \end{subfigure}\hfill
    \begin{subfigure}[b]{0.24\textwidth}
        \centering
        \includegraphics[width=\textwidth]{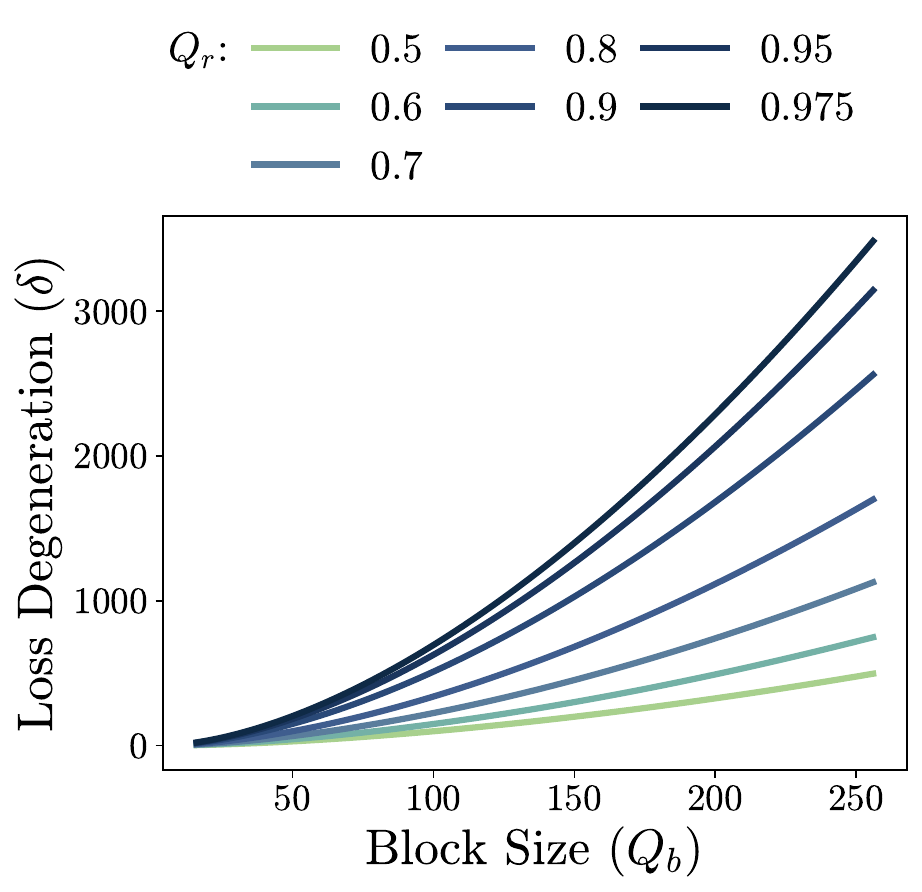}
        \captionsetup{font=scriptsize}
        \caption{1.7B Predicted (mean)}
    \end{subfigure}
    
\begin{subfigure}[b]{0.24\textwidth}
        \centering
        \includegraphics[width=\textwidth]{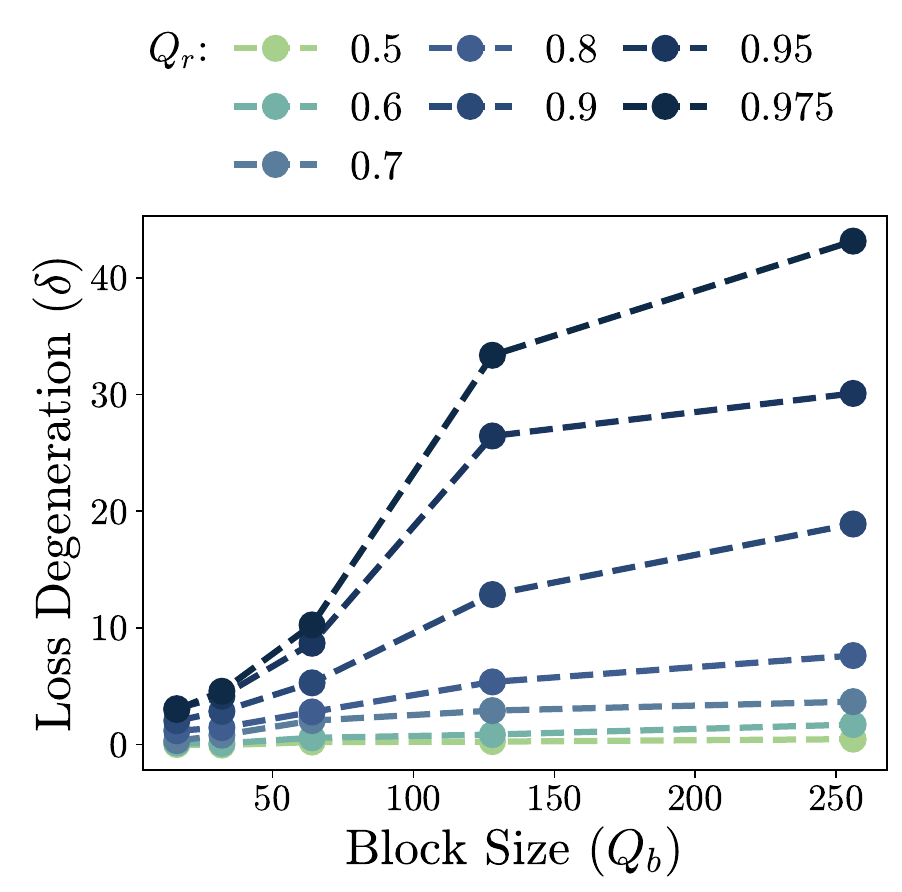}
        \captionsetup{font=scriptsize}
        \caption{4B Actual (min)}
    \end{subfigure}\hfill
    \begin{subfigure}[b]{0.24\textwidth}
        \centering
        \includegraphics[width=\textwidth]{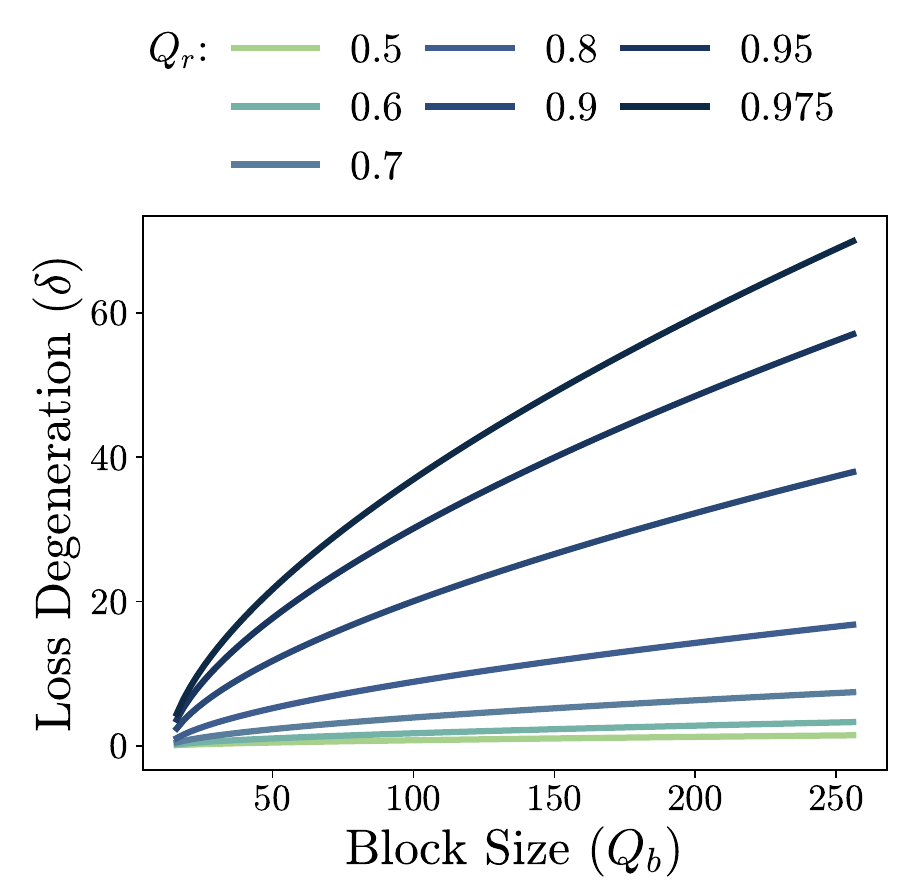}
        \captionsetup{font=scriptsize}
        \caption{4B Predicted (min)}
    \end{subfigure}\hfill
    \begin{subfigure}[b]{0.24\textwidth}
        \centering
        \includegraphics[width=\textwidth]{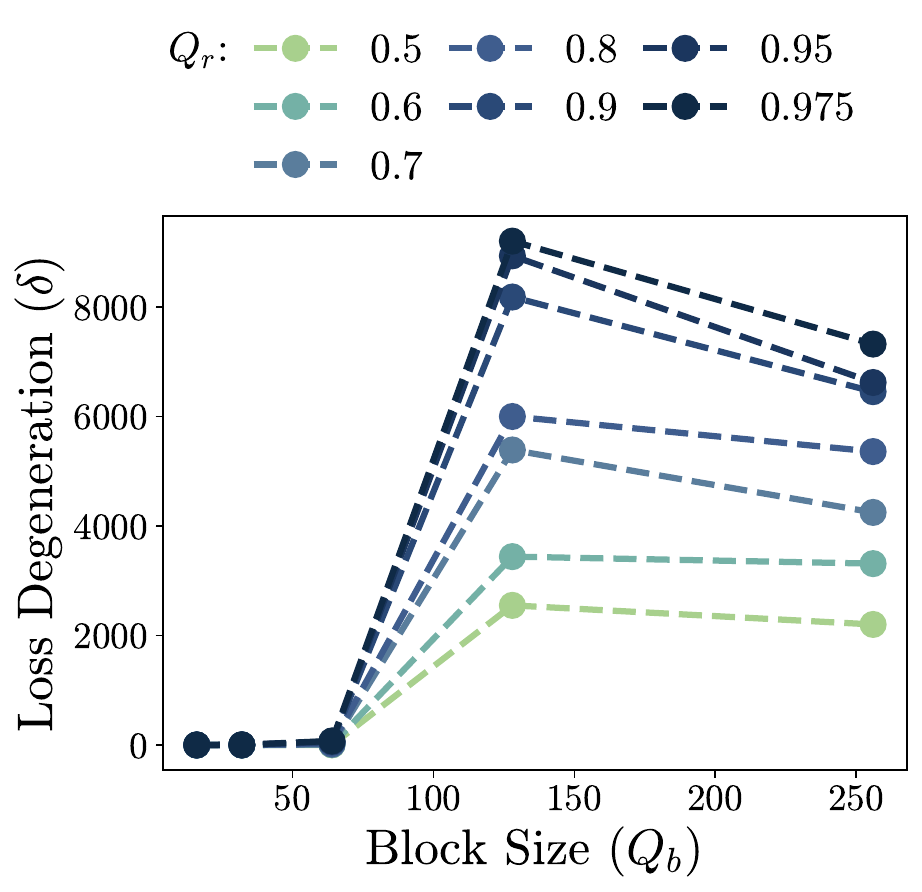}
        \captionsetup{font=scriptsize}
        \caption{4B Actual (mean)}
    \end{subfigure}\hfill
    \begin{subfigure}[b]{0.24\textwidth}
        \centering
        \includegraphics[width=\textwidth]{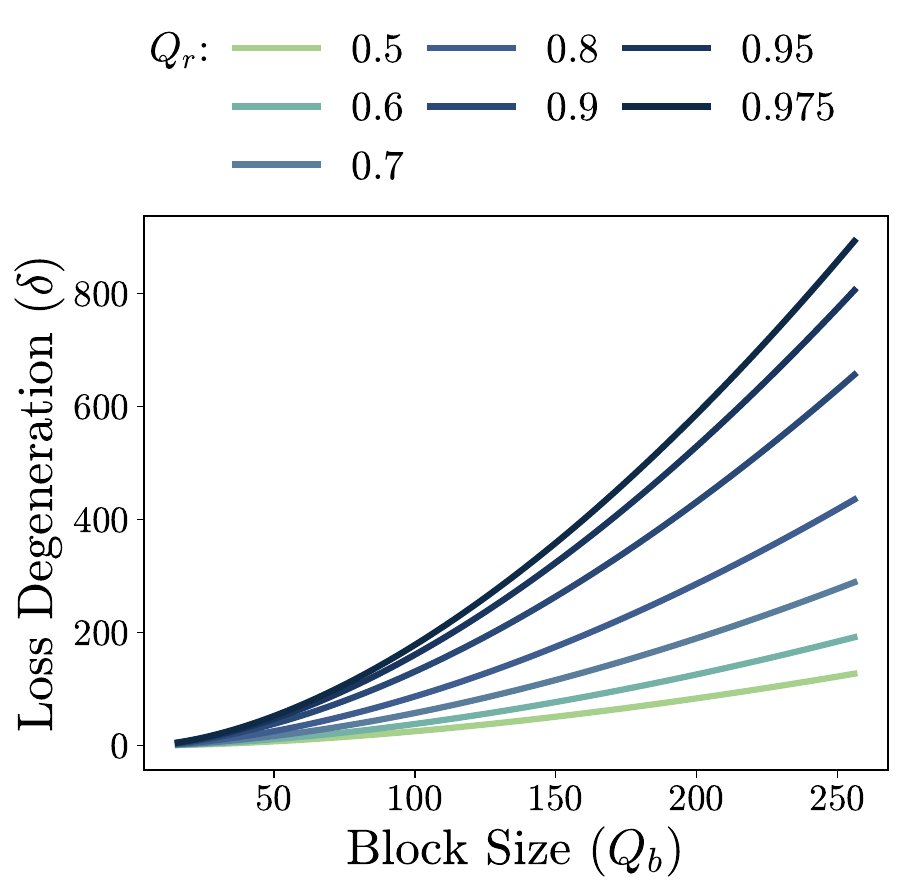}
        \captionsetup{font=scriptsize}
        \caption{4B Predicted (mean)}
    \end{subfigure}
    
\begin{subfigure}[b]{0.24\textwidth}
        \centering
        \includegraphics[width=\textwidth]{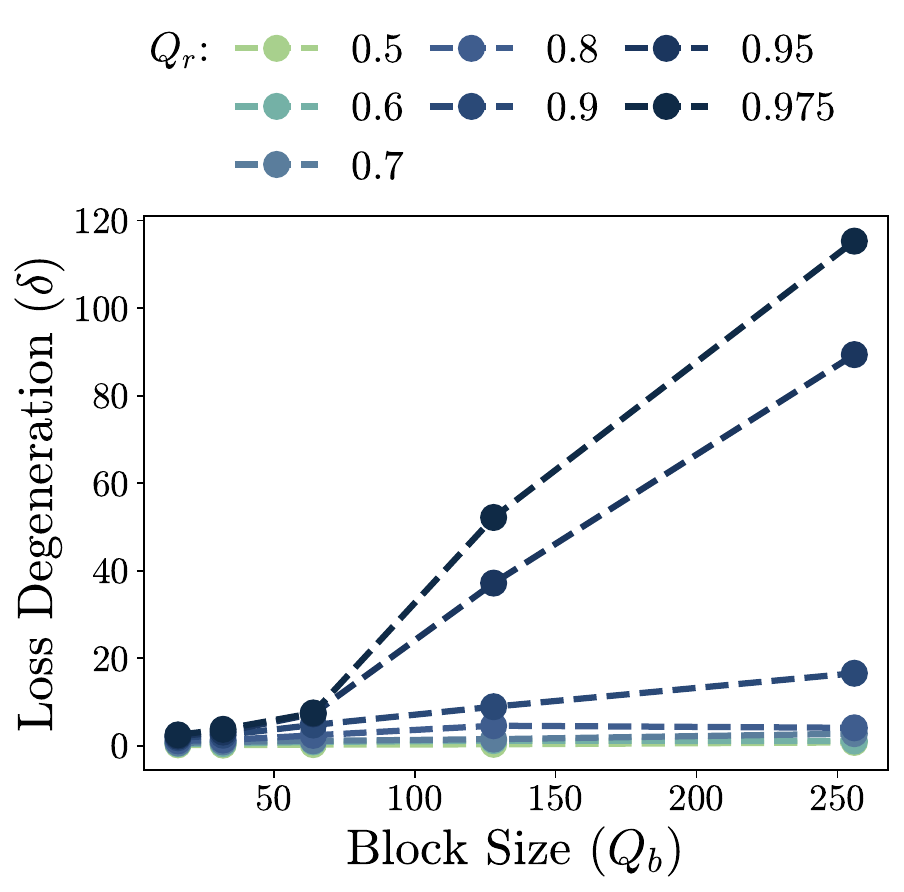}
        \captionsetup{font=scriptsize}
        \caption{8B Actual (min)}
    \end{subfigure}\hfill
    \begin{subfigure}[b]{0.24\textwidth}
        \centering
        \includegraphics[width=\textwidth]{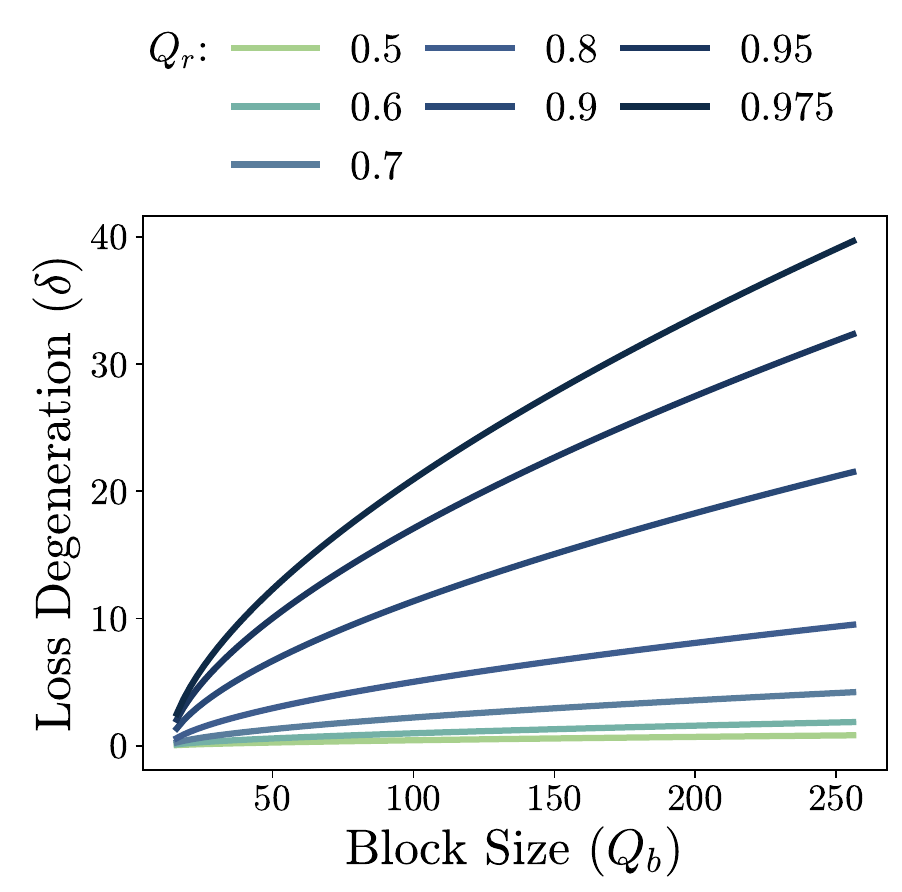}
        \captionsetup{font=scriptsize}
        \caption{8B Predicted (min)}
    \end{subfigure}\hfill
    \begin{subfigure}[b]{0.24\textwidth}
        \centering
        \includegraphics[width=\textwidth]{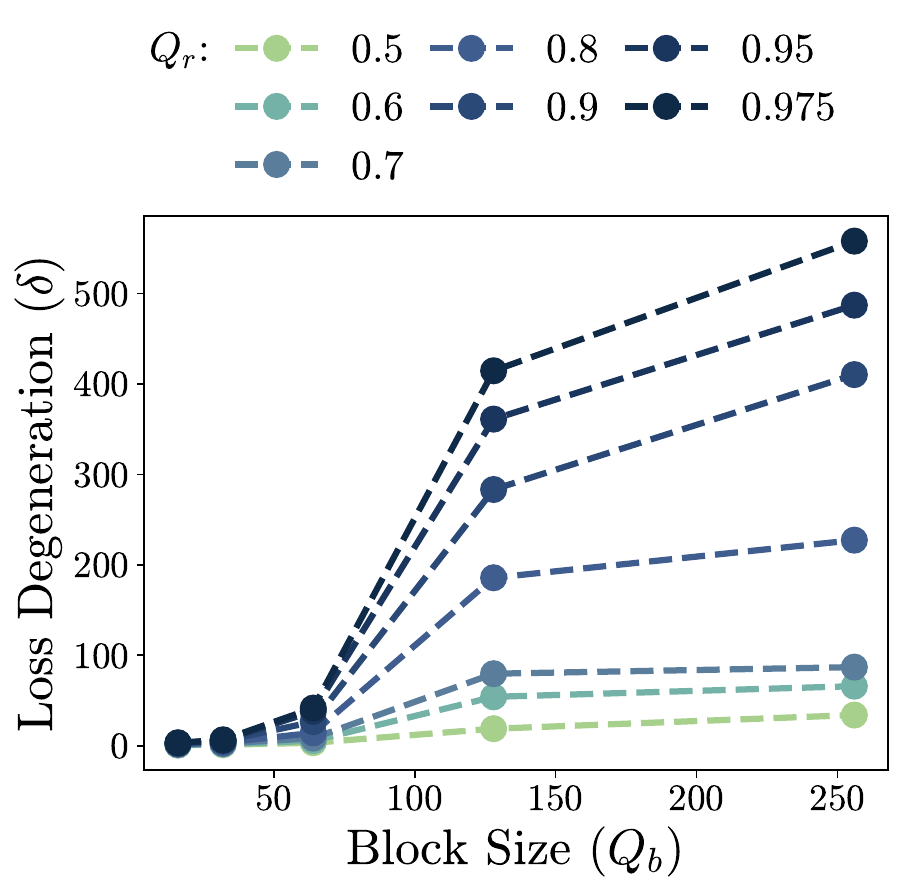}
        \captionsetup{font=scriptsize}
        \caption{8B Actual (mean)}
    \end{subfigure}\hfill
    \begin{subfigure}[b]{0.24\textwidth}
        \centering
        \includegraphics[width=\textwidth]{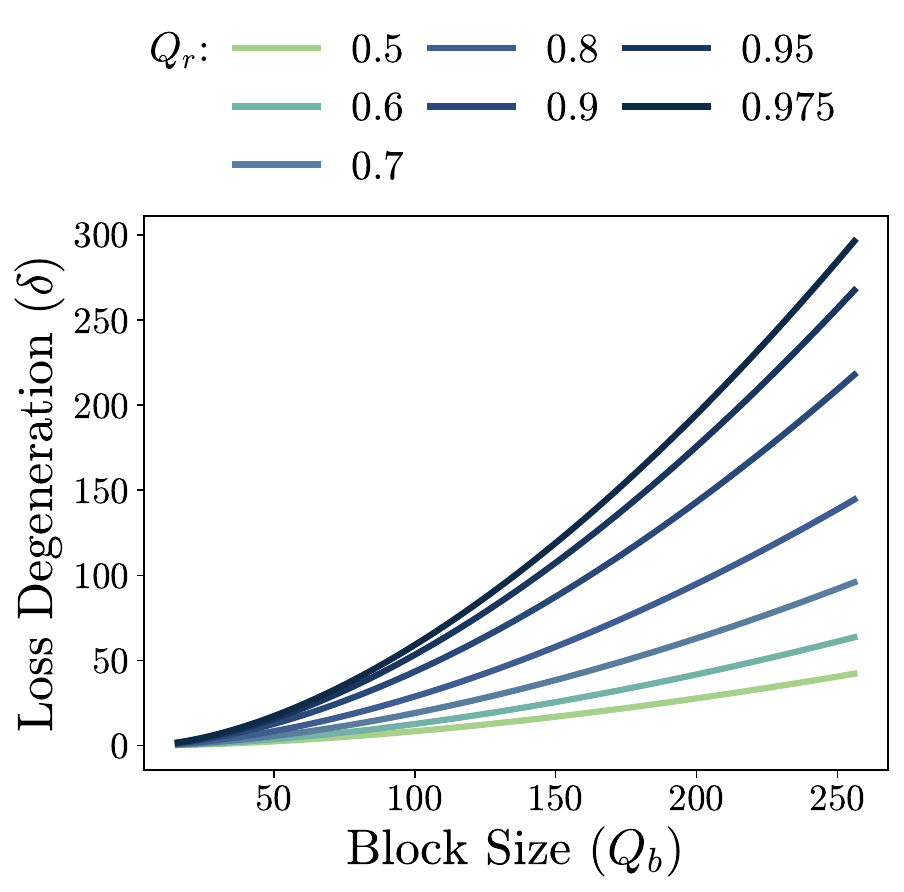}
        \captionsetup{font=scriptsize}
        \caption{8B Predicted (mean)}
    \end{subfigure}
    
\caption{\textbf{Strong Law Qwen-3 Matrix Multiplication-wise ($\delta^{\text{opt}}$, $\delta_\mu$)} (a,b,e,f,i,j,m,n) Qwen3 matrix multiplication-wise $\delta^{\text{opt}}$ results; (c,d,g,h,k,l,o,p) Qwen3 matrix multiplication-wise $\delta_\mu$ results.}
    \label{fig:appendix-blcoksize-qwen3}
    \end{figure}

While~\Cref{fig:appendix-matmul} and ~\Cref{fig:appendix-matmul-mean} show the effectiveness of the weak law, ~\Cref{fig:appendix-blcoksize-clm}, ~\Cref{fig:appendix-blcoksize-qwen1.5}, and ~\Cref{fig:appendix-blcoksize-qwen3} demonstrate the utility for the strong law. Those three figures show the actual and fitted loss contours (both minimum and mean) with respect to block size $Q_b$ with different quantization ratios $Q_r$ for CLM, Qwen-1.5, and Qwen-3. For those figures, our strong law only takes into consideration losses that are less than 100, since larger losses in perplexity are meaningless under our setting (in practical term, these models would not generate anything distinguishable). As a result, the fitted contours are not accurate when the actual loss is large, which is acceptable since we do not care about those excessive losses.

\section{Other Numerical Format Results}
\label{app:extending-other-format}
\begin{figure}[htbp]
\centering
\begin{subfigure}[b]{0.3\textwidth} \centering
    \includegraphics[width=\textwidth]{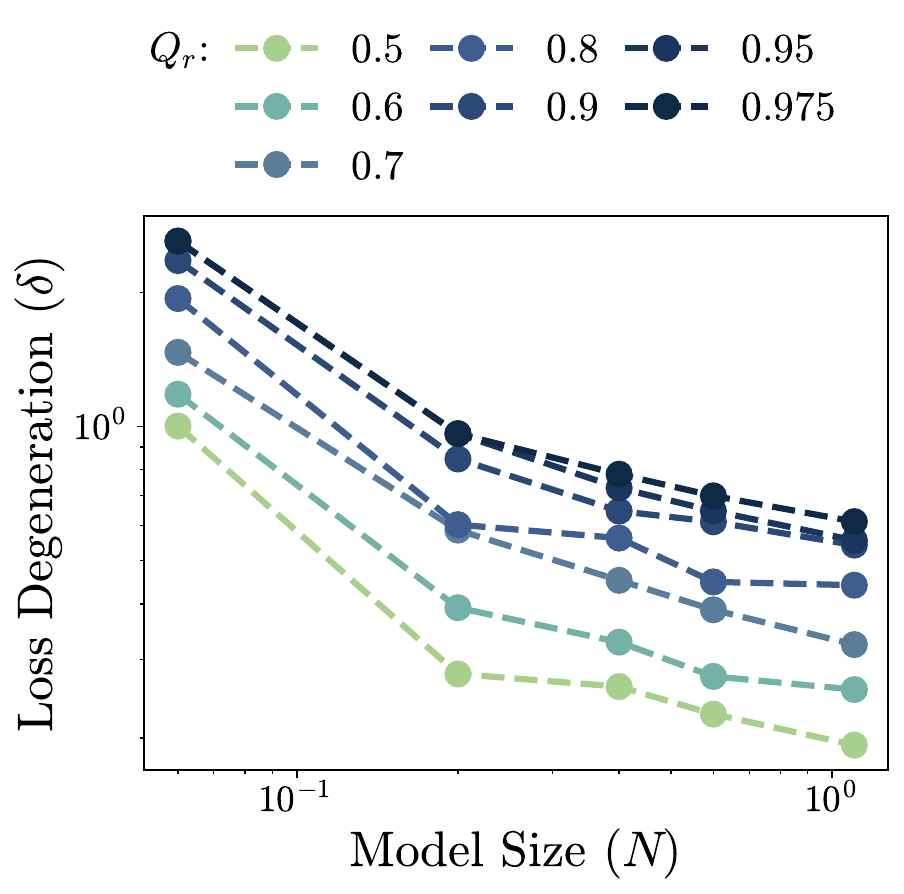}
    \captionsetup{font=scriptsize}
    \caption{CLM Actual Loss}
\end{subfigure}
\hfill
\begin{subfigure}[b]{0.3\textwidth} \centering
    \includegraphics[width=\textwidth]{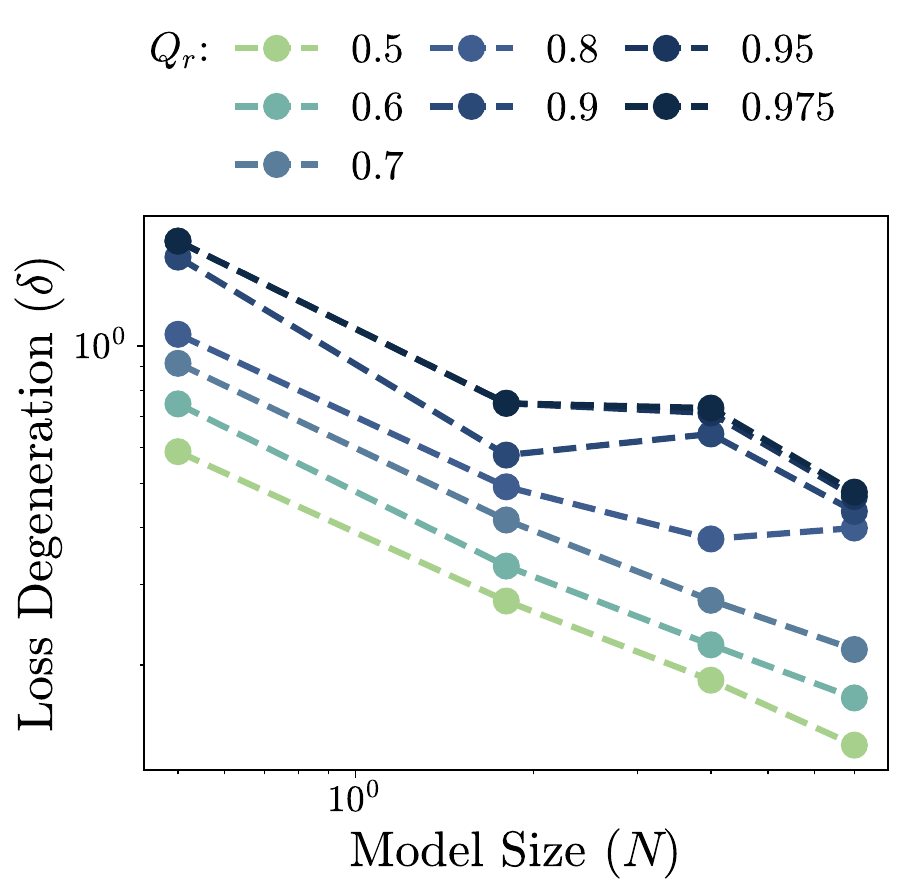}
    \captionsetup{font=scriptsize}
    \caption{Qwen-1.5 Actual Loss}
\end{subfigure}
\hfill
\begin{subfigure}[b]{0.3\textwidth} \centering
    \includegraphics[width=\textwidth]{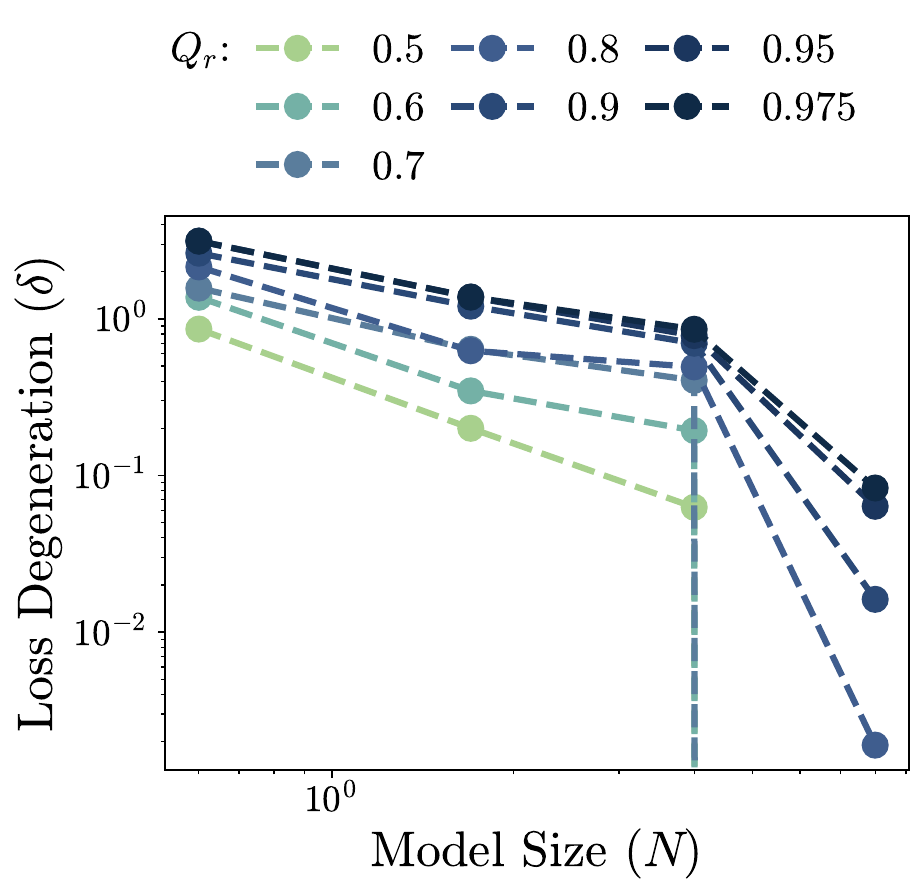}
    \captionsetup{font=scriptsize}
    \caption{Qwen-3 Actual Loss}
\end{subfigure}

\begin{subfigure}[b]{0.3\textwidth} \centering
    \includegraphics[width=\textwidth]{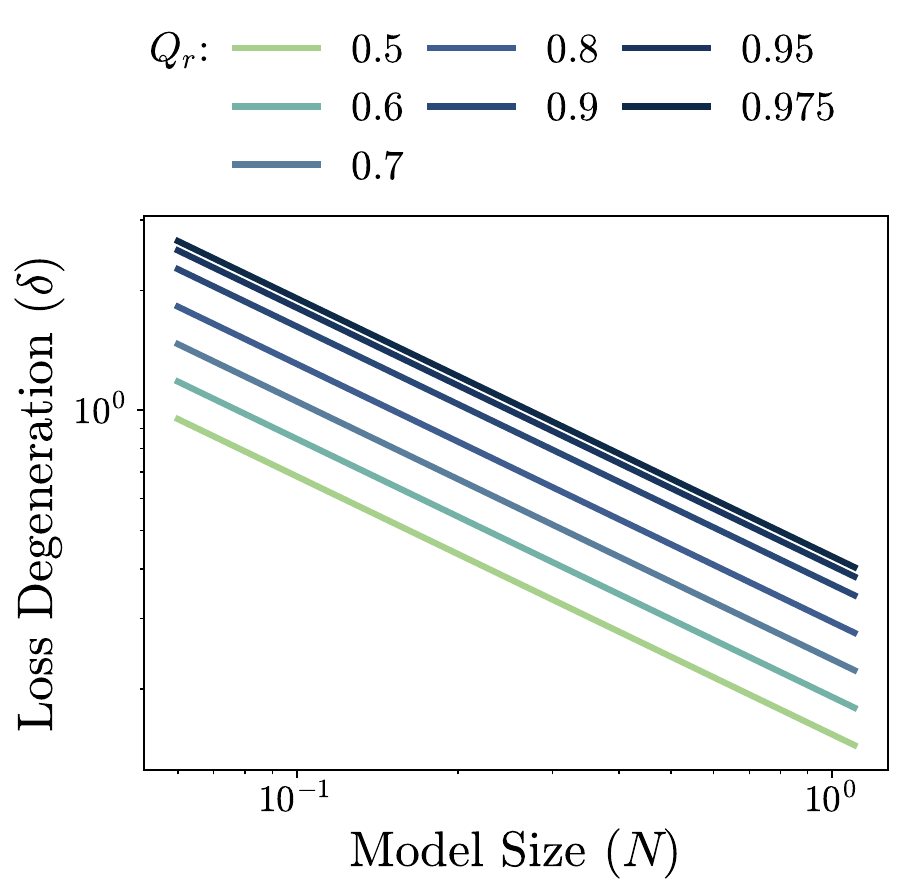}
    \captionsetup{font=scriptsize}
    \caption{CLM Predicted Loss}
\end{subfigure}
\hfill
\begin{subfigure}[b]{0.3\textwidth} \centering
    \includegraphics[width=\textwidth]{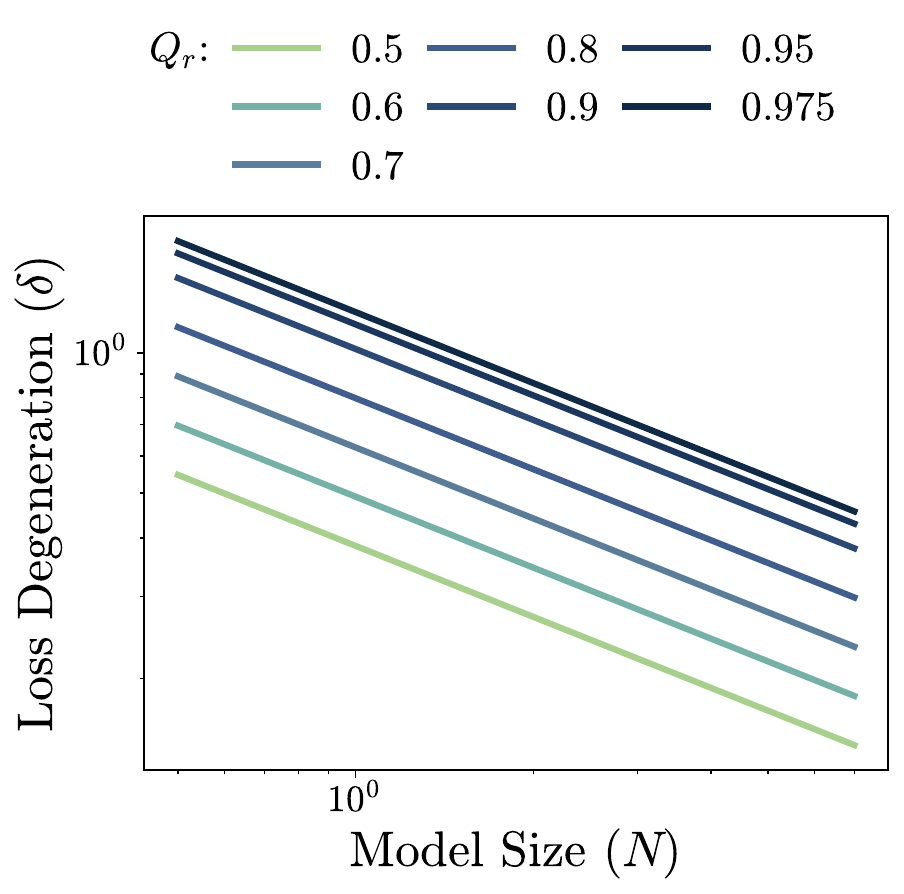}
    \captionsetup{font=scriptsize}
    \caption{Qwen-1.5 Predicted Loss}
\end{subfigure}
\hfill
\begin{subfigure}[b]{0.3\textwidth} \centering
    \includegraphics[width=\textwidth]{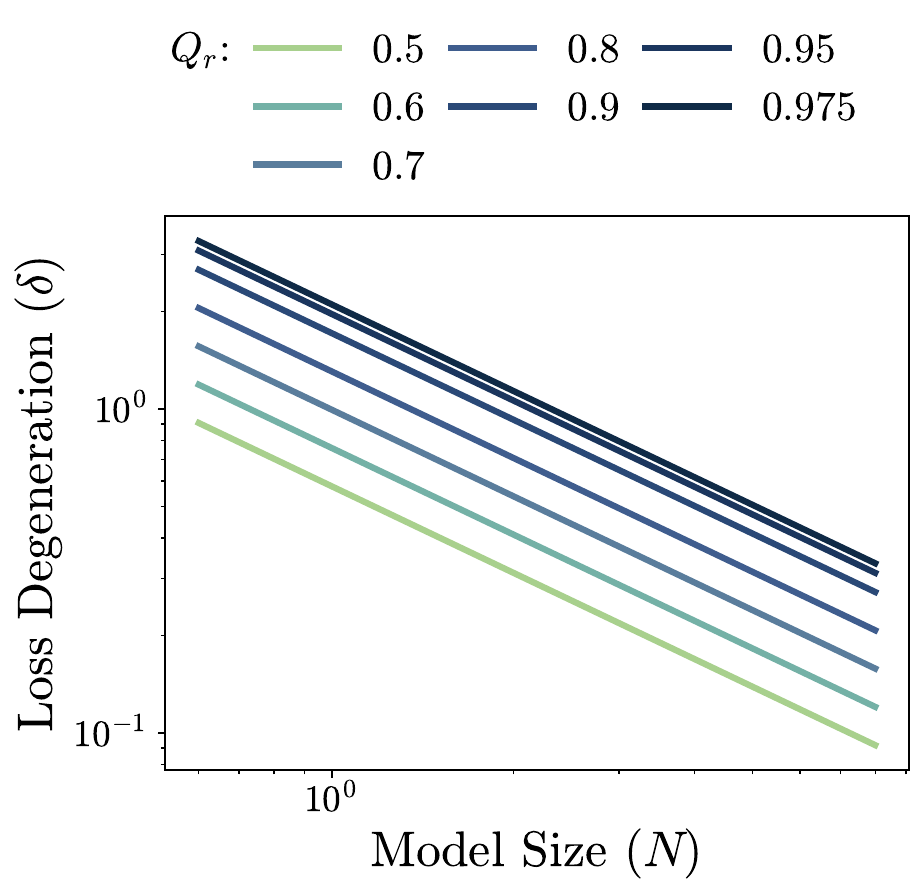}
    \captionsetup{font=scriptsize}
    \caption{Qwen-3 Predicted Loss}
\end{subfigure}

\begin{subfigure}[b]{0.3\textwidth} \centering
    \includegraphics[width=\textwidth]{figures/figures-hqq/llama-hqq/logloss_vs_qratio.pdf}
    \captionsetup{font=scriptsize}
    \caption{CLM Actual Loss}
\end{subfigure}
\hfill
\begin{subfigure}[b]{0.3\textwidth} \centering
    \includegraphics[width=\textwidth]{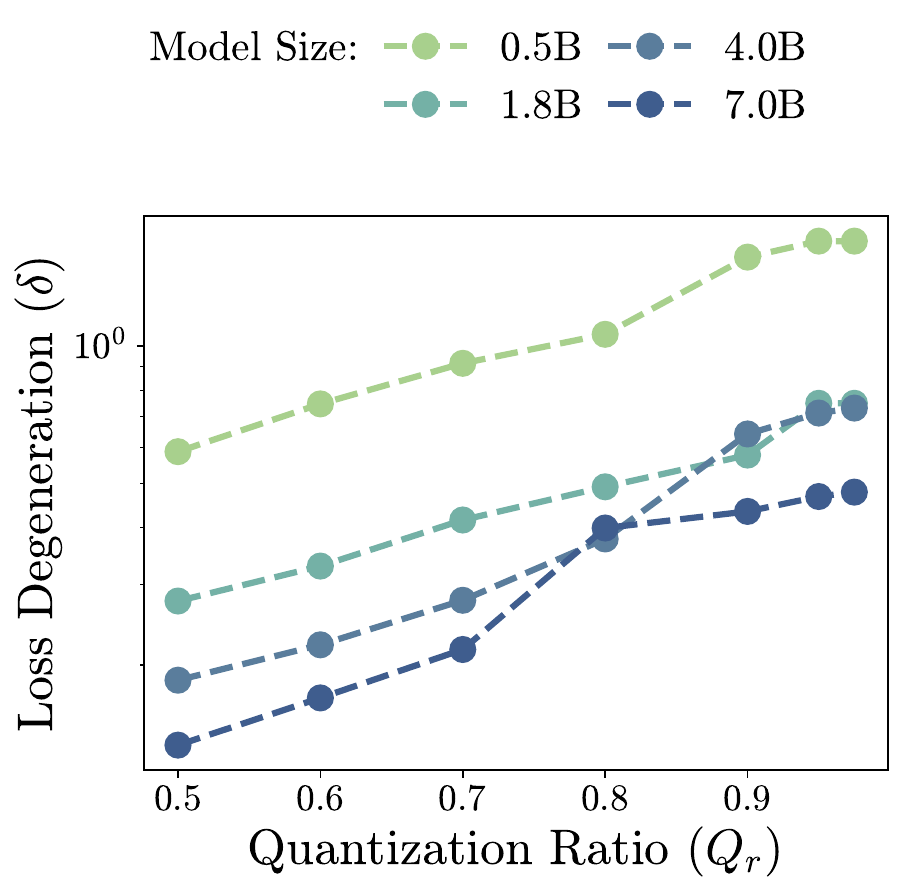}
    \captionsetup{font=scriptsize}
    \caption{Qwen-1.5 Actual Loss}
\end{subfigure}
\hfill
\begin{subfigure}[b]{0.3\textwidth} \centering
    \includegraphics[width=\textwidth]{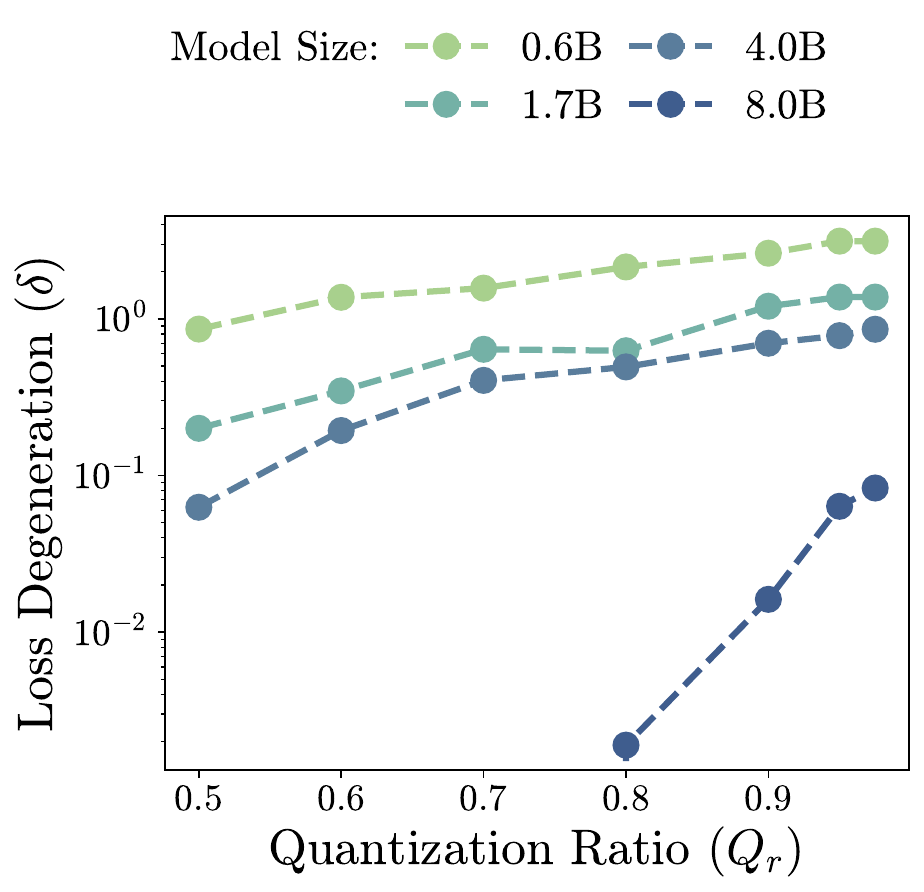}
    \captionsetup{font=scriptsize}
    \caption{Qwen-3 Actual Loss}
\end{subfigure}

\begin{subfigure}[b]{0.3\textwidth} \centering
\includegraphics[width=\textwidth]{figures/figures-hqq/llama-hqq/logloss_vs_qratio-fitted.pdf}
    \captionsetup{font=scriptsize}
    \caption{CLM Predicted Loss}
    \end{subfigure}
\hfill
\begin{subfigure}[b]{0.3\textwidth} \centering
    \includegraphics[width=\textwidth]{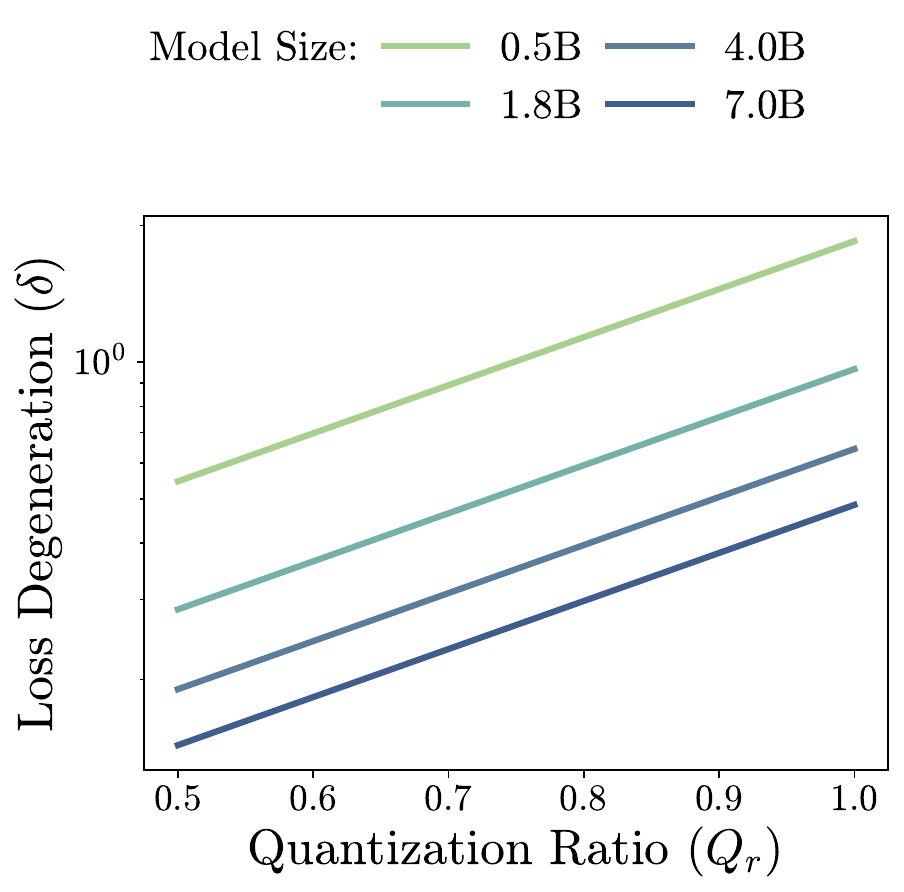}
    \captionsetup{font=scriptsize}
    \caption{Qwen-1.5 Predicted Loss}
\end{subfigure}
\hfill
\begin{subfigure}[b]{0.3\textwidth} \centering
    \includegraphics[width=\textwidth]{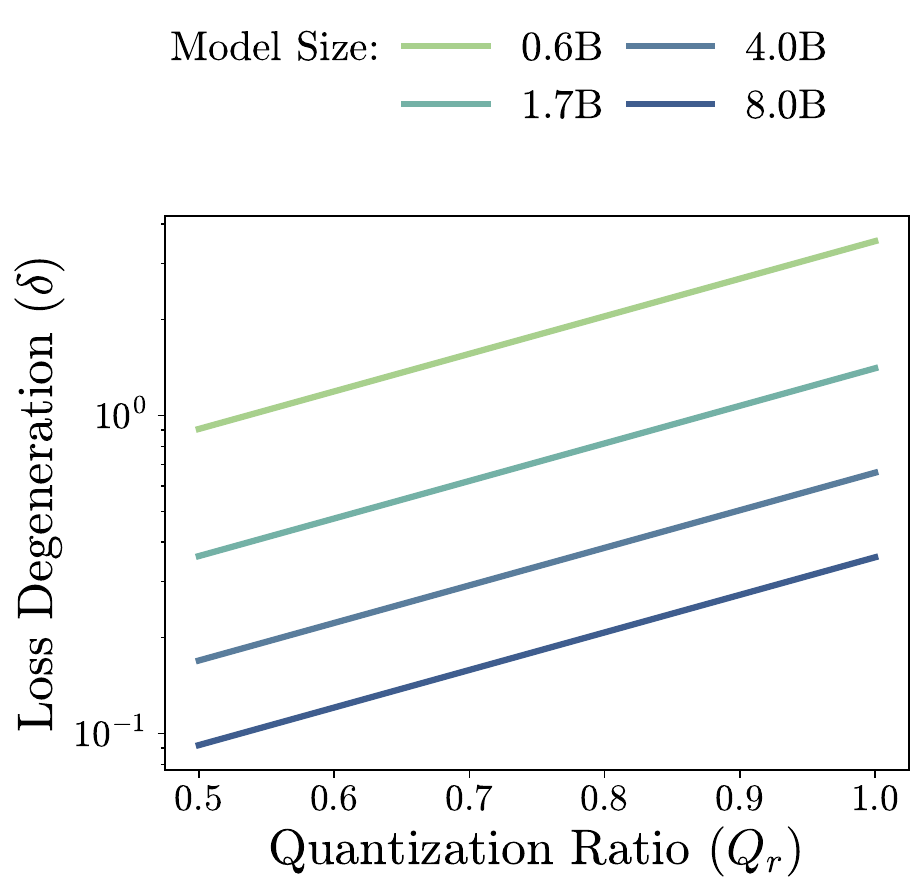}
    \captionsetup{font=scriptsize}
    \caption{Qwen-3 Predicted Loss}
\end{subfigure}

\caption{\textbf{HQQ ($\delta^{\text{opt}}$)} (a,d,g,h) CLM HQQ results; (b,e,h,k) Qwen-1.5 HQQ results; (c,f,i,l) Qwen-3 HQQ results.}
  \label{fig:appendix-hqq}
\end{figure}

\begin{figure}[htbp]
\centering
\begin{subfigure}[b]{0.3\textwidth} \centering
    \includegraphics[width=\textwidth]{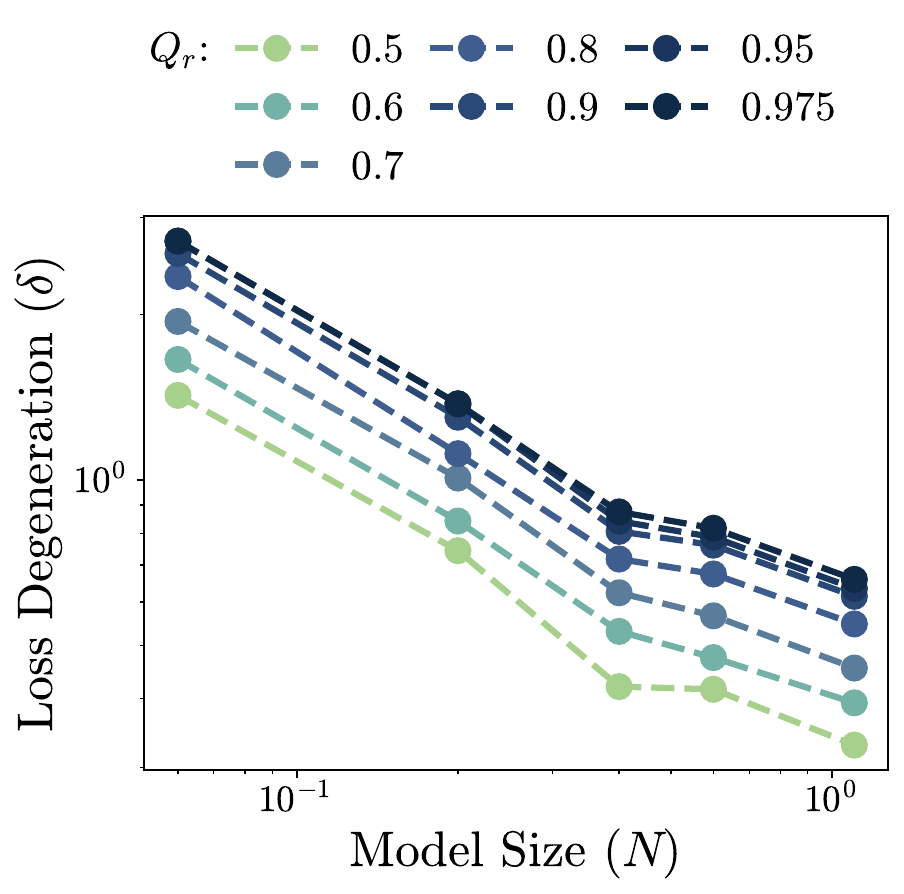}
    \captionsetup{font=scriptsize}
    \caption{CLM Actual Loss}
\end{subfigure}
\hfill
\begin{subfigure}[b]{0.3\textwidth} \centering
    \includegraphics[width=\textwidth]{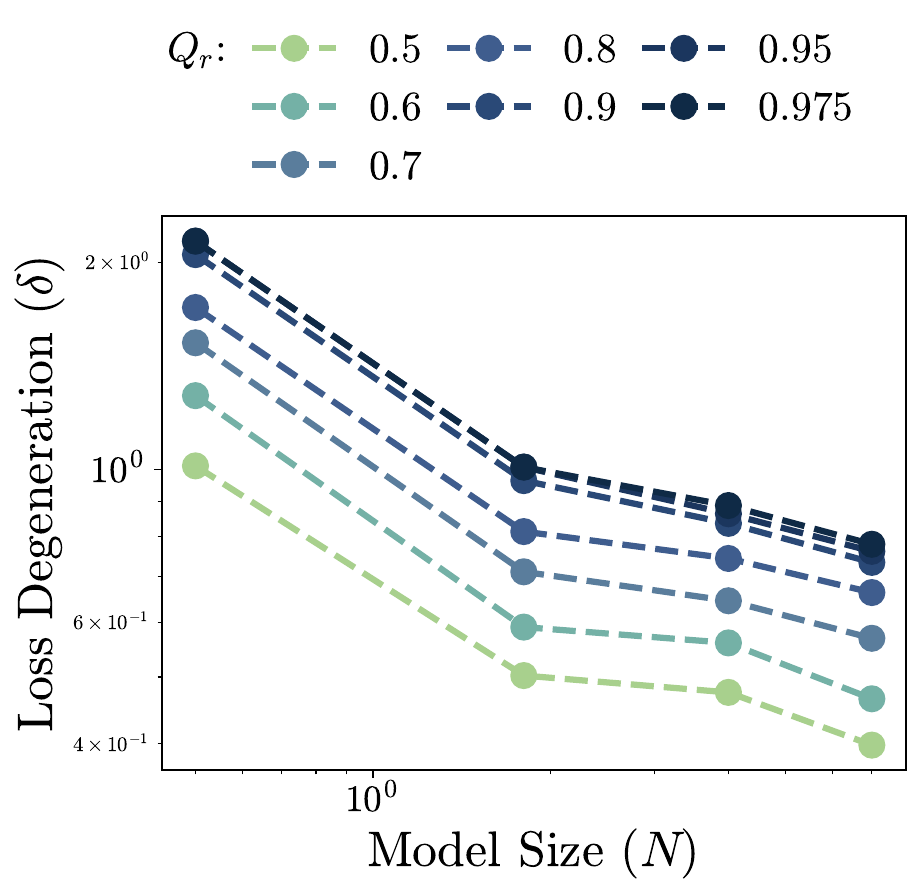}
    \captionsetup{font=scriptsize}
    \caption{Qwen1.5 Actual Loss}
\end{subfigure}
\hfill
\begin{subfigure}[b]{0.3\textwidth} \centering
    \includegraphics[width=\textwidth]{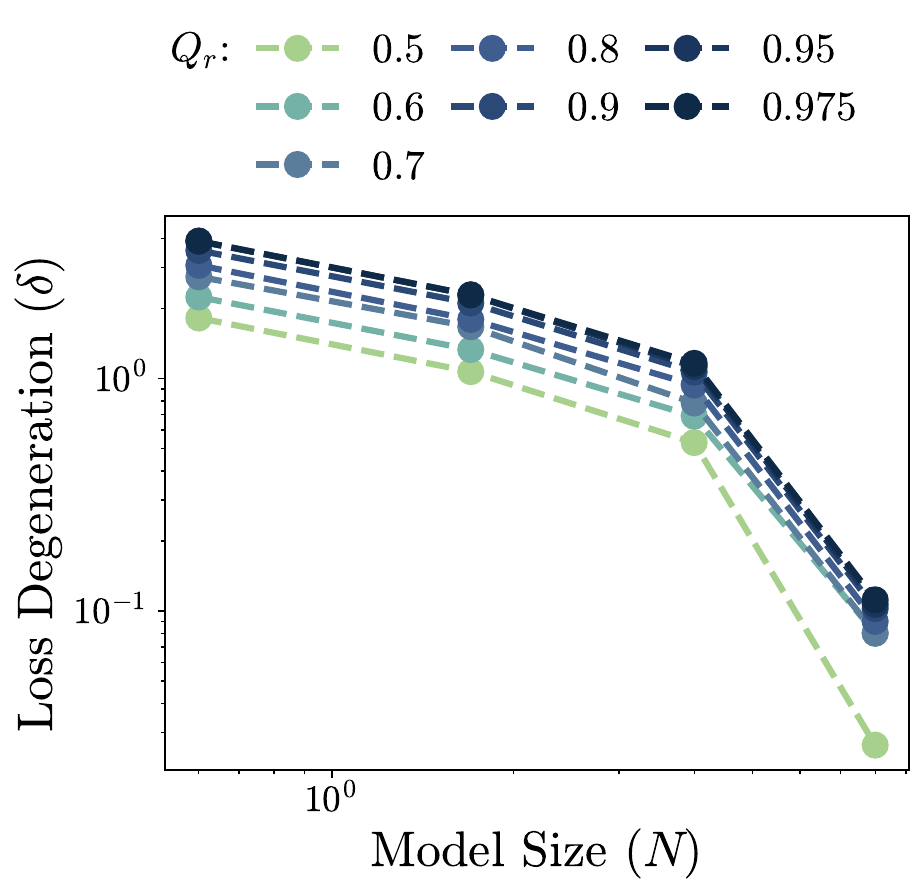}
    \captionsetup{font=scriptsize}
    \caption{Qwen3 Actual Loss}
\end{subfigure}

\begin{subfigure}[b]{0.3\textwidth} \centering
    \includegraphics[width=\textwidth]{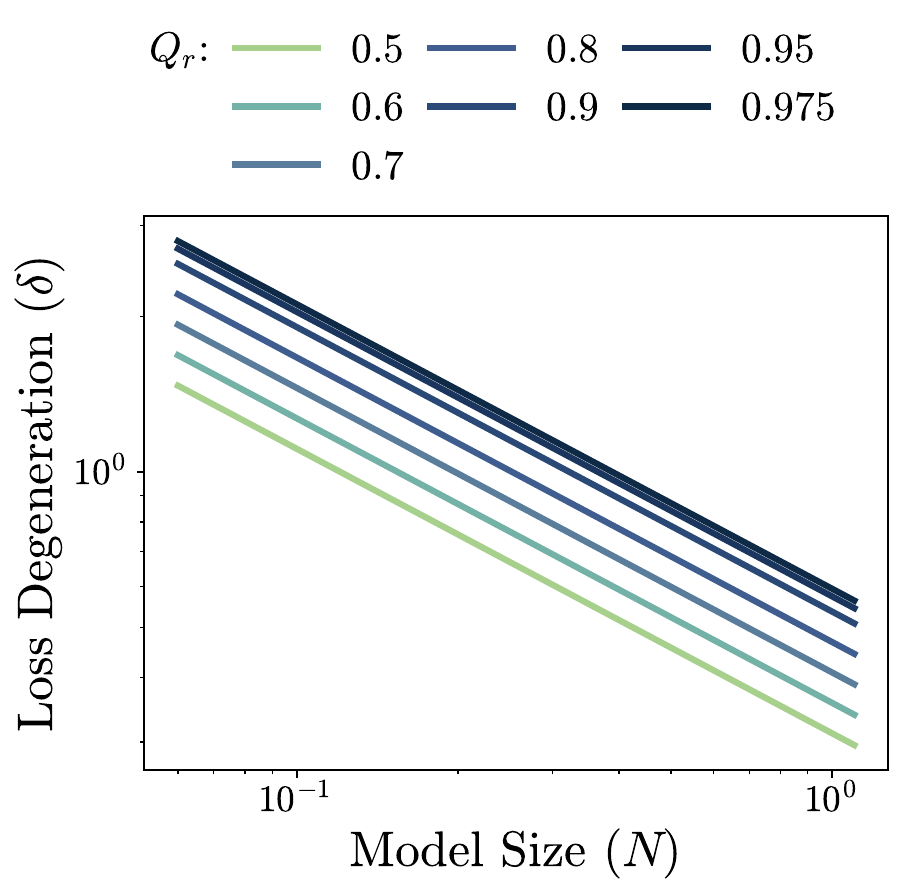}
    \captionsetup{font=scriptsize}
    \caption{CLM Predicted Loss}
\end{subfigure}
\hfill
\begin{subfigure}[b]{0.3\textwidth} \centering
    \includegraphics[width=\textwidth]{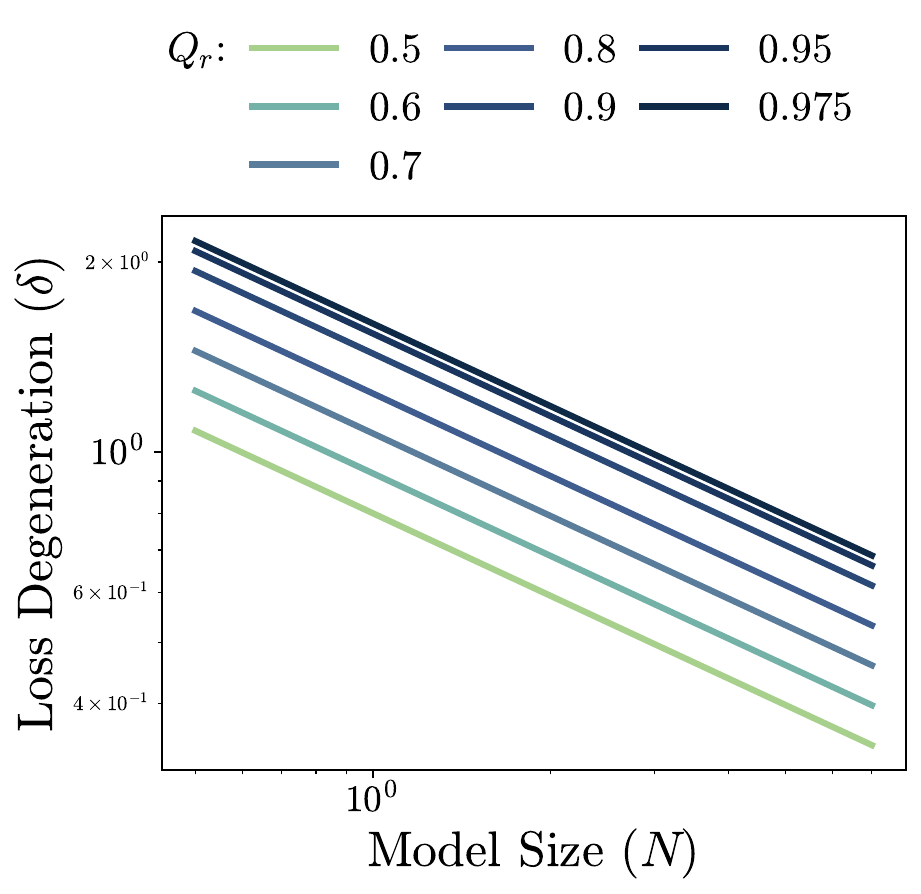}
    \captionsetup{font=scriptsize}
    \caption{Qwen1.5 Predicted Loss}
\end{subfigure}
\hfill
\begin{subfigure}[b]{0.3\textwidth} \centering
    \includegraphics[width=\textwidth]{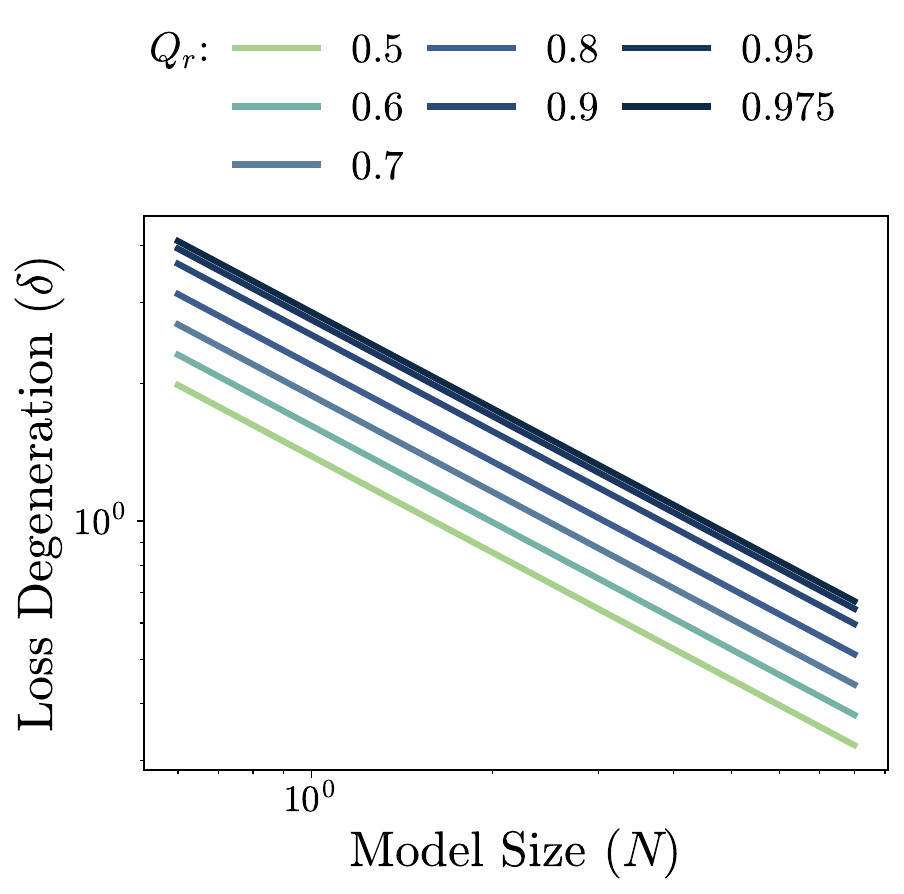}
    \captionsetup{font=scriptsize}
    \caption{Qwen3 Predicted Loss}
\end{subfigure}

\begin{subfigure}[b]{0.3\textwidth} \centering
    \includegraphics[width=\textwidth]{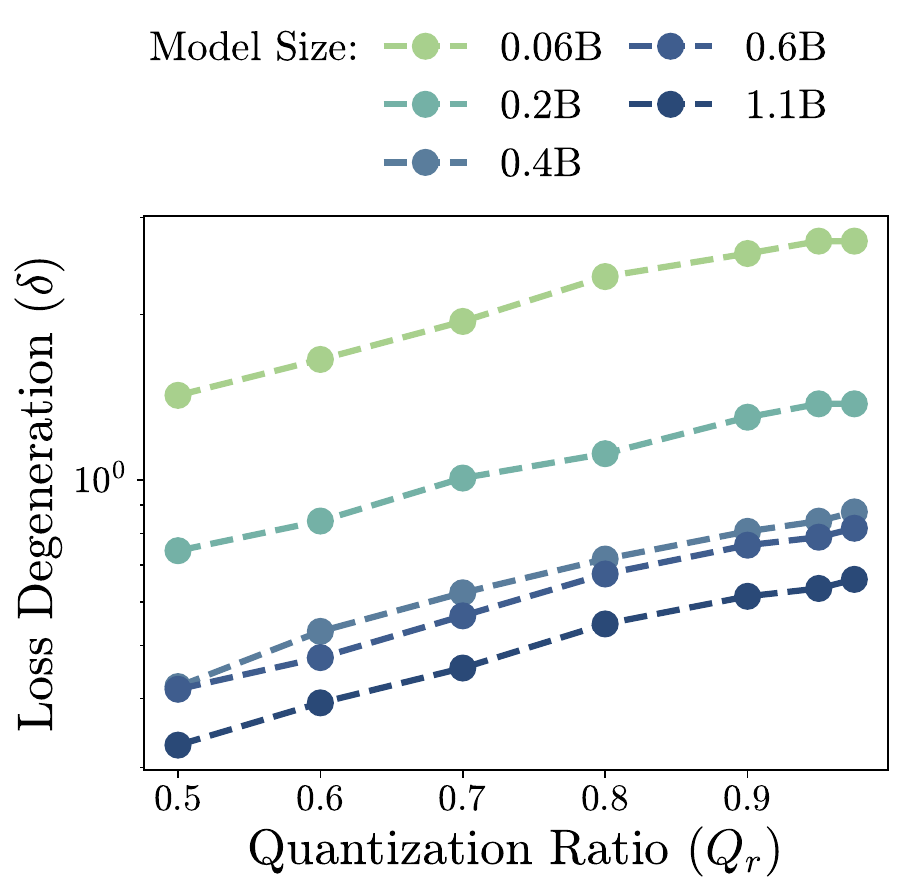}
    \captionsetup{font=scriptsize}
    \caption{CLM Actual Loss}
\end{subfigure}
\hfill
\begin{subfigure}[b]{0.3\textwidth} \centering
    \includegraphics[width=\textwidth]{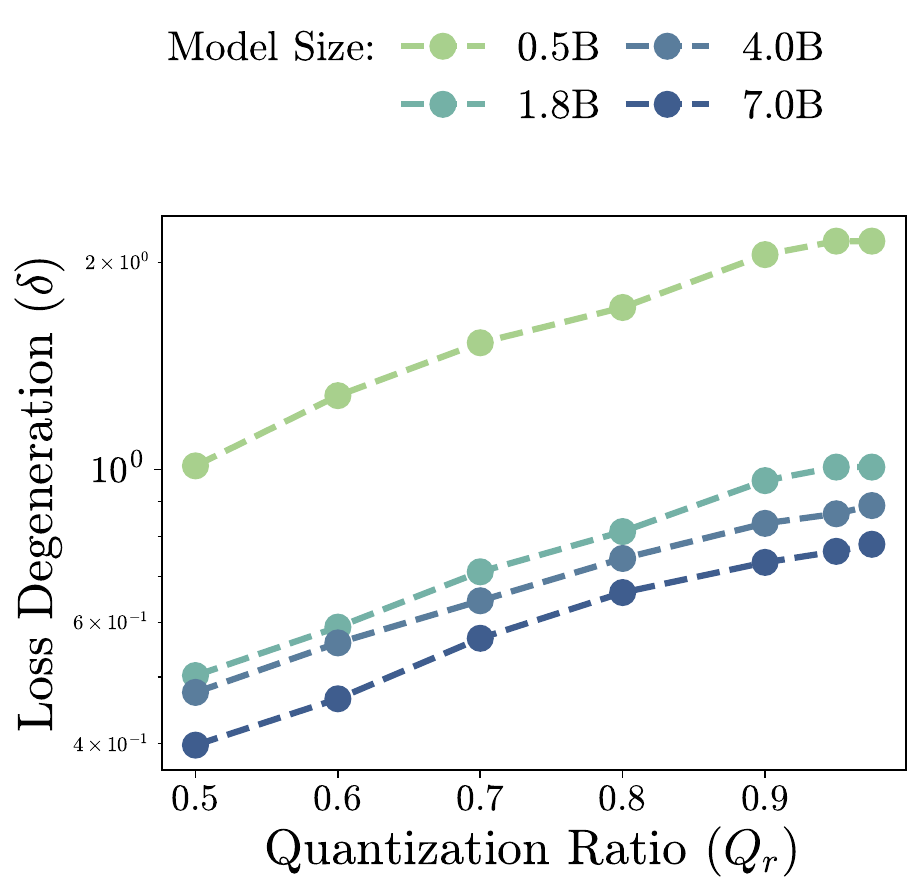}
    \captionsetup{font=scriptsize}
    \caption{Qwen1.5 Actual Loss}
\end{subfigure}
\hfill
\begin{subfigure}[b]{0.3\textwidth} \centering
    \includegraphics[width=\textwidth]{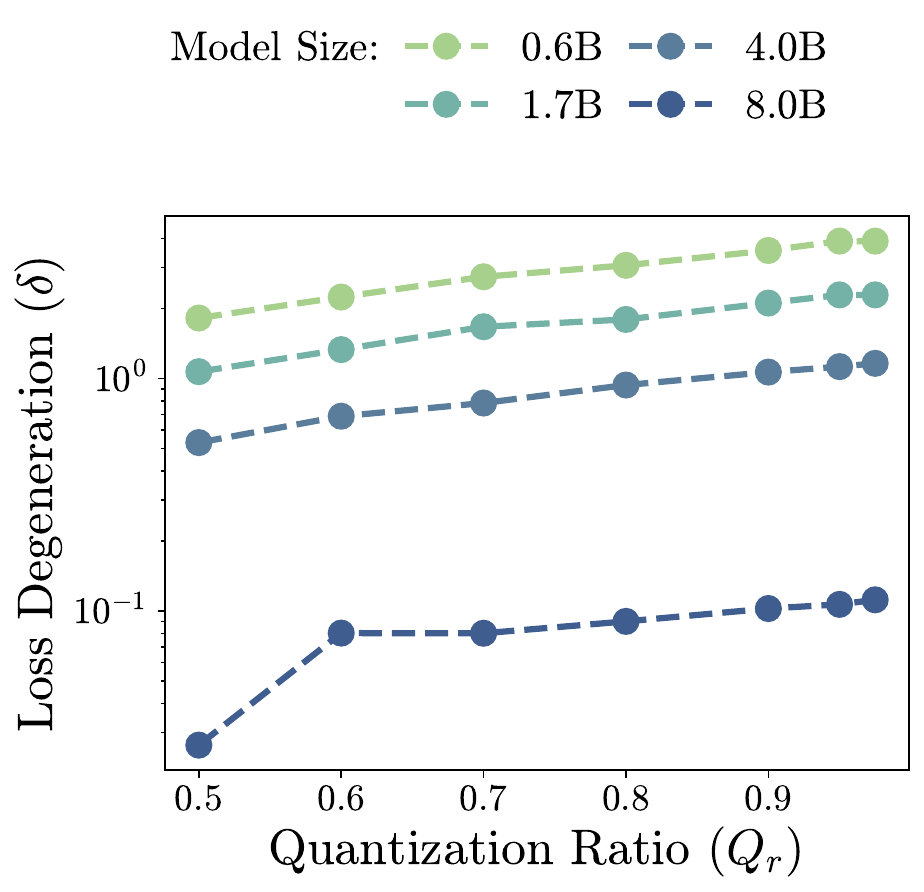}
    \captionsetup{font=scriptsize}
    \caption{Qwen3 Actual Loss}
\end{subfigure}

\begin{subfigure}[b]{0.3\textwidth} \centering
\includegraphics[width=\textwidth]{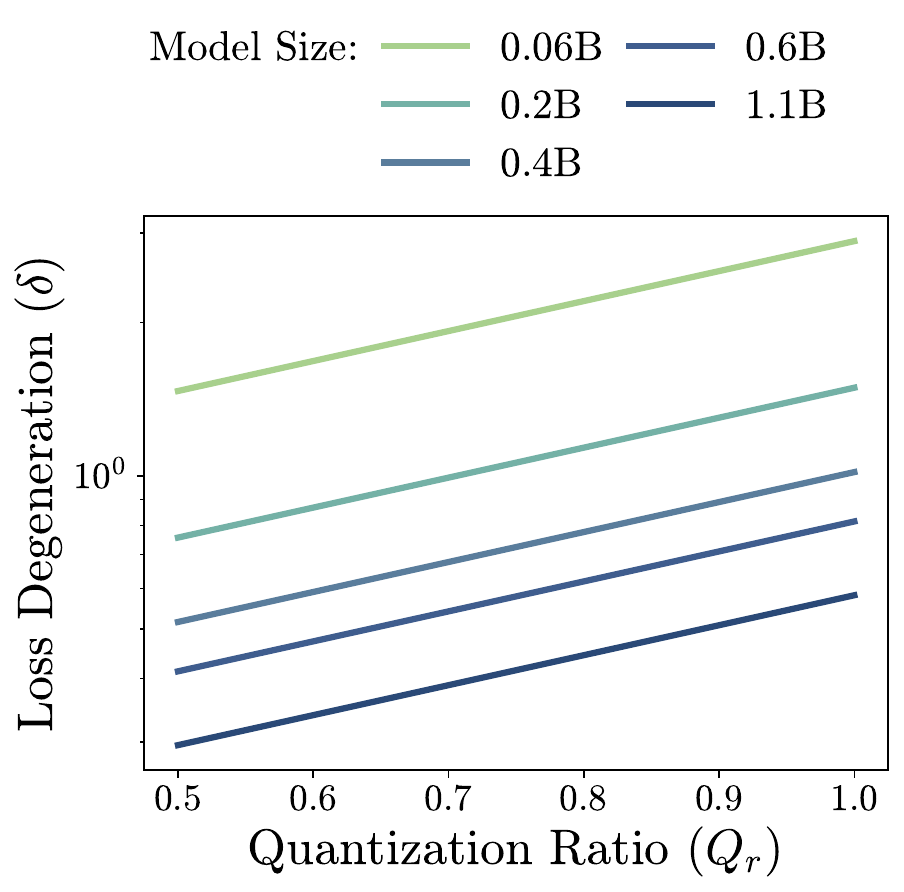}
    \captionsetup{font=scriptsize}
    \caption{CLM Predicted Loss}
    \end{subfigure}
\hfill
\begin{subfigure}[b]{0.3\textwidth} \centering
    \includegraphics[width=\textwidth]{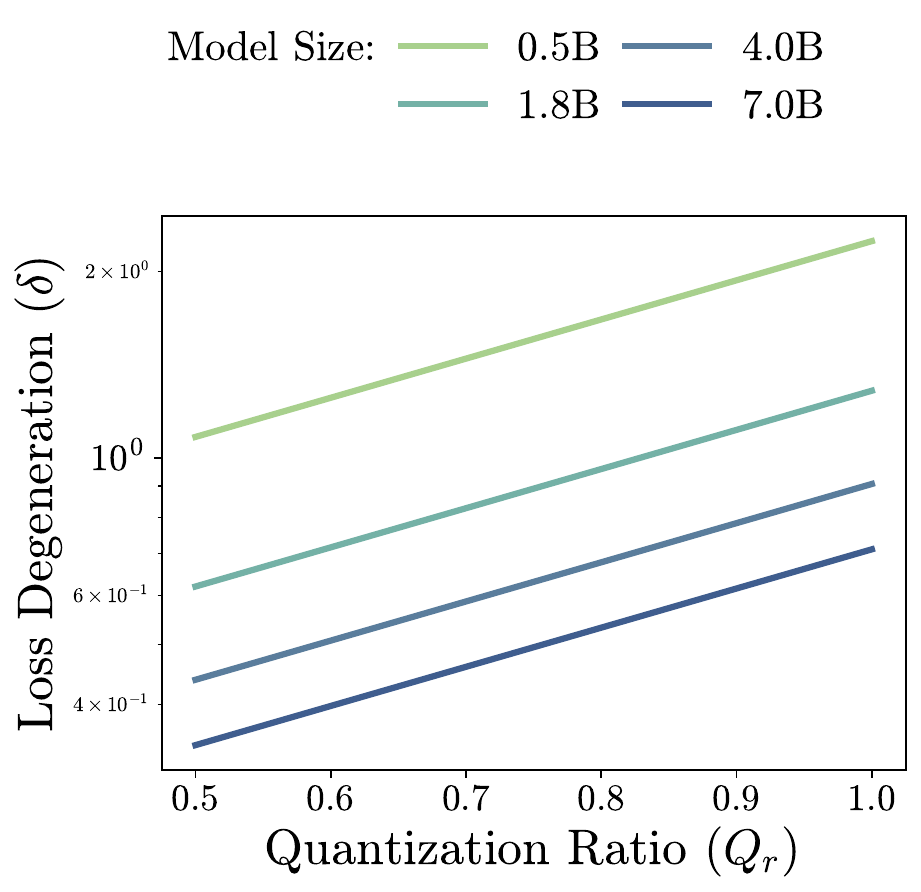}
    \captionsetup{font=scriptsize}
    \caption{Qwen1.5 Predicted Loss}
\end{subfigure}
\hfill
\begin{subfigure}[b]{0.3\textwidth} \centering
    \includegraphics[width=\textwidth]{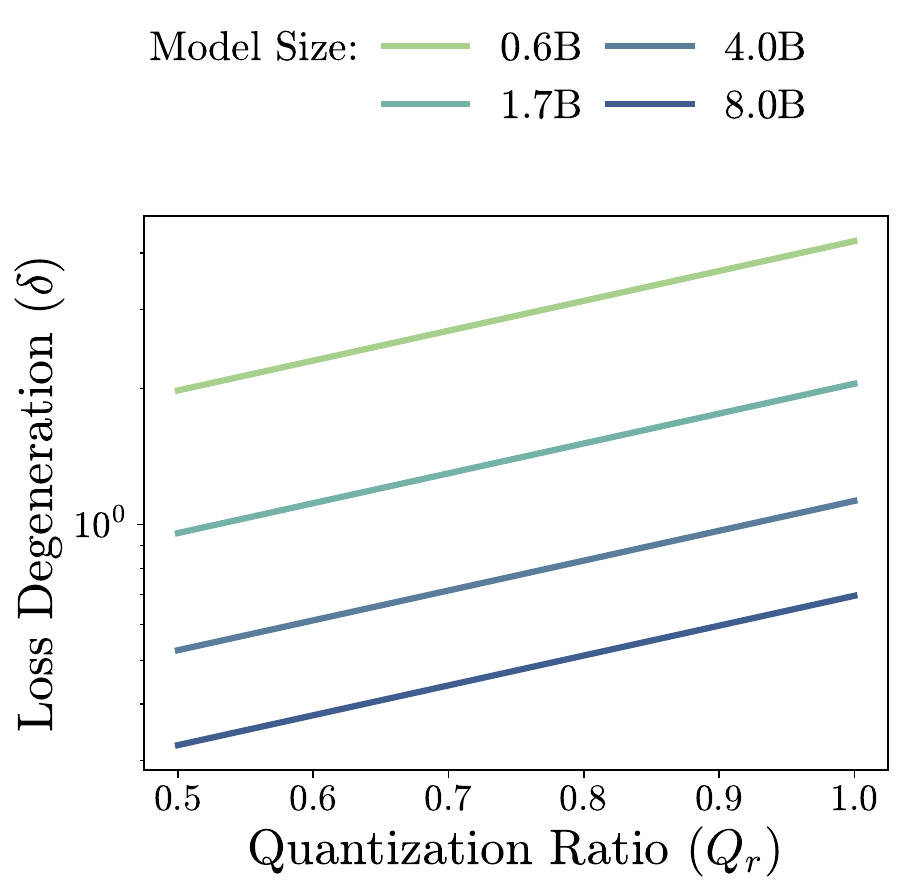}
    \captionsetup{font=scriptsize}
    \caption{Qwen3 Predicted Loss}
\end{subfigure}

\caption{\textbf{HQQ ($\delta_\mu$)} (a,d,g,h) CLM HQQ results; (b,e,h,k) Qwen-1.5 HQQ results; (c,f,i,l) Qwen-3 HQQ results.}
  \label{fig:appendix-hqq-mean}
\end{figure}

Figures~\ref{fig:appendix-hqq} and~\ref{fig:appendix-hqq-mean} present the minimum and mean loss contours with respect to model size $N$ and quantization ratio $Q_r$ for the CLM, Qwen-1.5, and Qwen-3 models under HQQ quantization. These results demonstrate that HQQ is a highly effective quantization method, successfully preserving the capabilities of the pre-trained models and maintaining low post-training quantization (PTQ) loss. Notably, the fitted contours closely match the empirical data across all model families, indicating that our scaling law formulation generalizes well, even under high-performance quantization methods. 

\clearpage
Note that, under these settings, there are some significant outliers for Qwen-3 results, with negativate PTQ losses, such negative PTQ losses might be caused by the fact that \textit{all Qwen-3 models involved in our experiments are partially distilled from larger models} as indicated in~\citep{yang2025qwen3}, invalidating our assumptions in Section~\ref{sec:laws}.

\begin{figure}[htbp]
\centering
\begin{subfigure}[b]{0.3\textwidth} \centering
    \includegraphics[width=\textwidth]{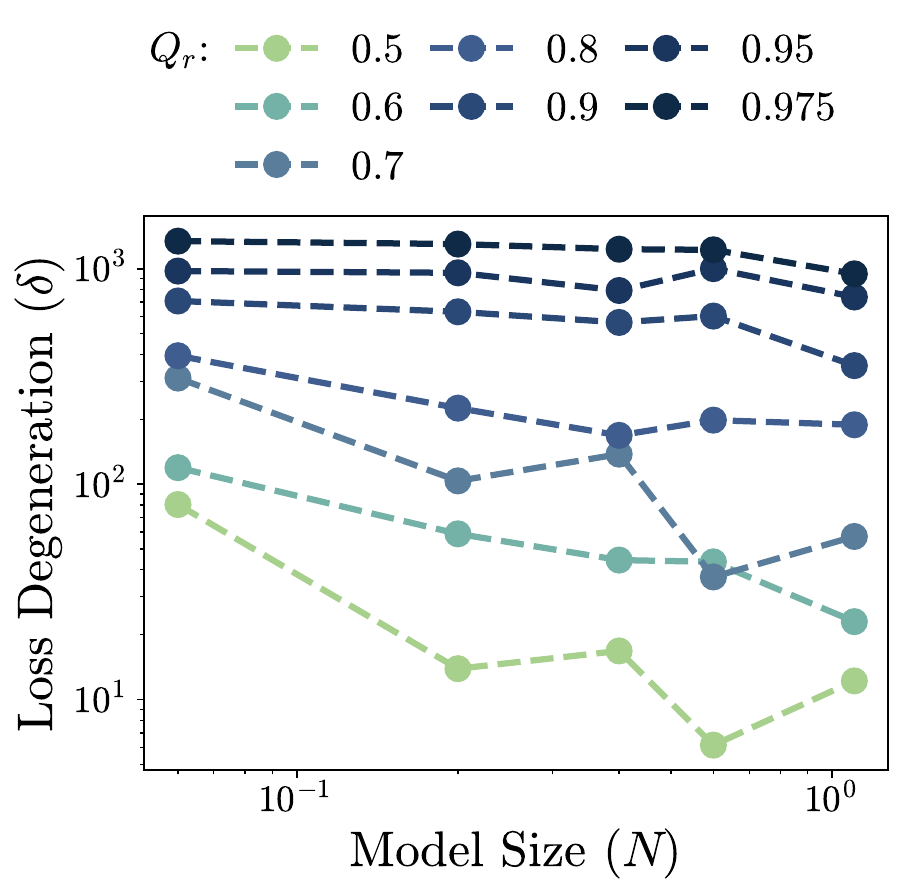}
    \captionsetup{font=scriptsize}
    \caption{CLM Actual Loss}
\end{subfigure}
\hfill
\begin{subfigure}[b]{0.3\textwidth} \centering
    \includegraphics[width=\textwidth]{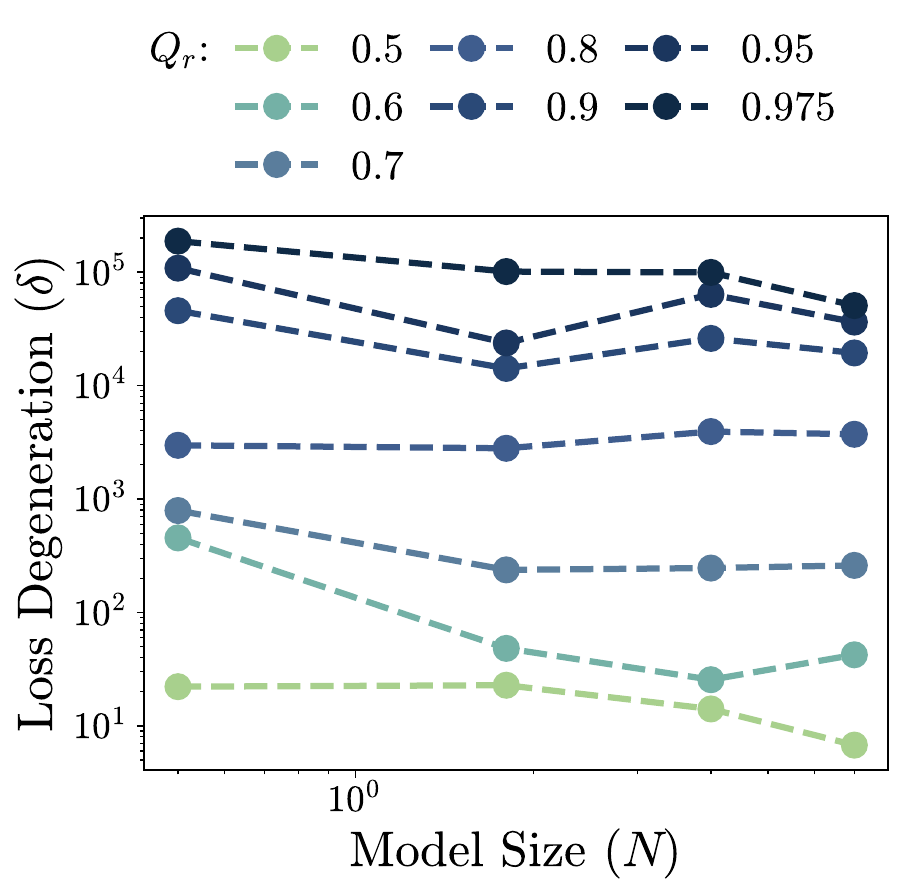}
    \captionsetup{font=scriptsize}
    \caption{Qwen-1.5 Actual Loss}
\end{subfigure}
\hfill
\begin{subfigure}[b]{0.3\textwidth} \centering
    \includegraphics[width=\textwidth]{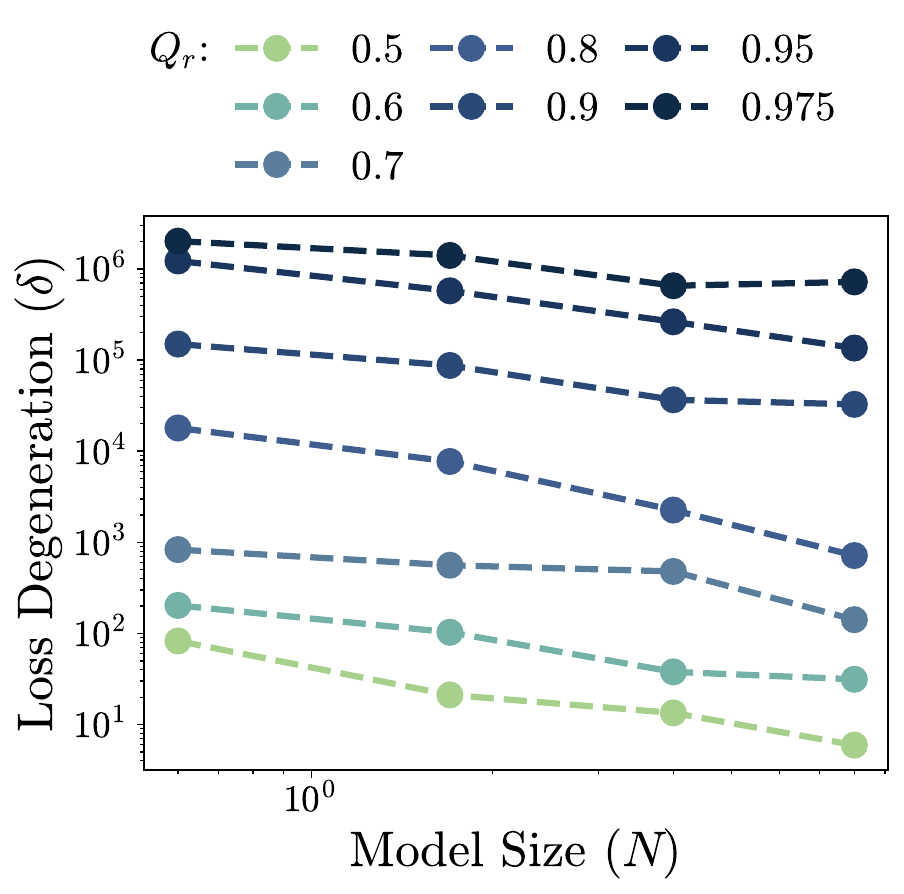}
    \captionsetup{font=scriptsize}
    \caption{Qwen-3 Actual Loss}
\end{subfigure}

\begin{subfigure}[b]{0.3\textwidth} \centering
    \includegraphics[width=\textwidth]{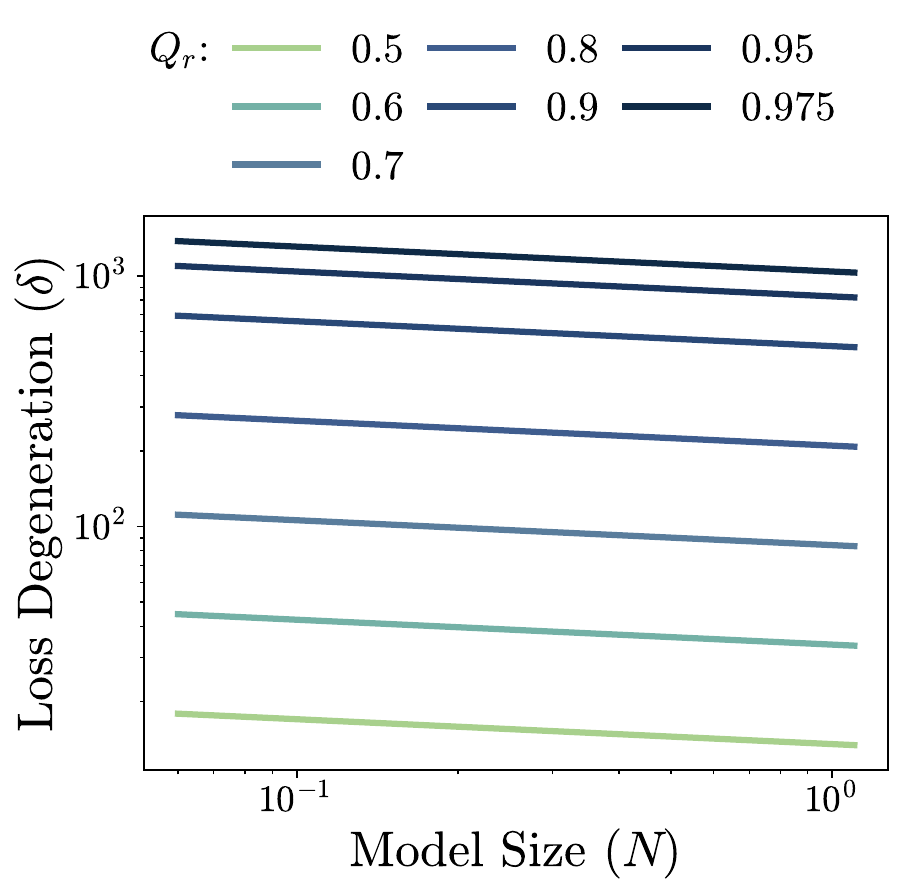}
    \captionsetup{font=scriptsize}
    \caption{CLM Predicted Loss}
\end{subfigure}
\hfill
\begin{subfigure}[b]{0.3\textwidth} \centering
    \includegraphics[width=\textwidth]{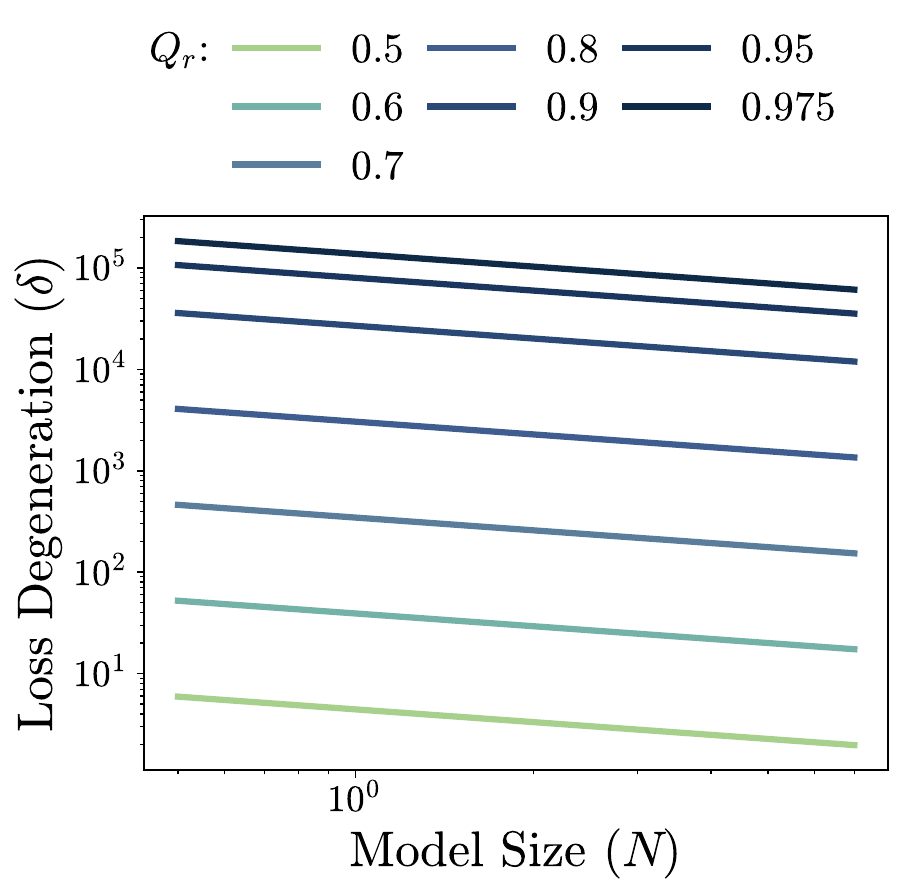}
    \captionsetup{font=scriptsize}
    \caption{Qwen-1.5 Predicted Loss}
\end{subfigure}
\hfill
\begin{subfigure}[b]{0.3\textwidth} \centering
    \includegraphics[width=\textwidth]{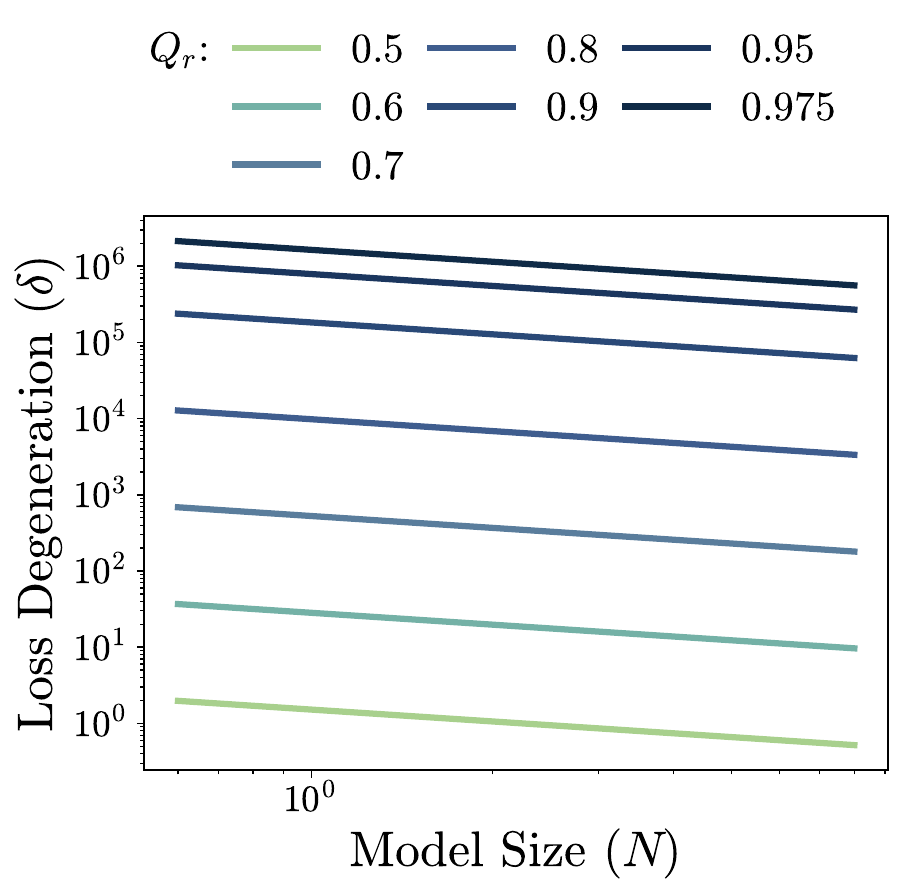}
    \captionsetup{font=scriptsize}
    \caption{Qwen-3 Predicted Loss}
\end{subfigure}

\begin{subfigure}[b]{0.3\textwidth} \centering
    \includegraphics[width=\textwidth]{figures/figures-mxint2/llama-matmul2/logloss_vs_qratio.pdf}
    \captionsetup{font=scriptsize}
    \caption{CLM Actual Loss}
\end{subfigure}
\hfill
\begin{subfigure}[b]{0.3\textwidth} \centering
    \includegraphics[width=\textwidth]{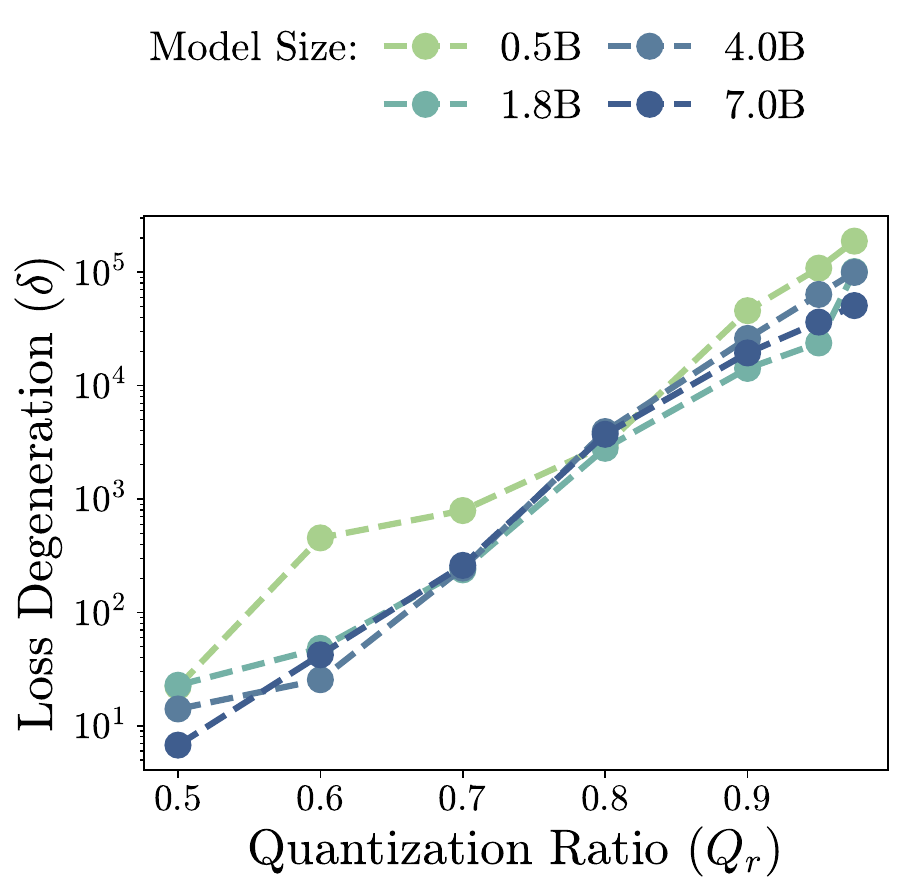}
    \captionsetup{font=scriptsize}
    \caption{Qwen-1.5 Actual Loss}
\end{subfigure}
\hfill
\begin{subfigure}[b]{0.3\textwidth} \centering
    \includegraphics[width=\textwidth]{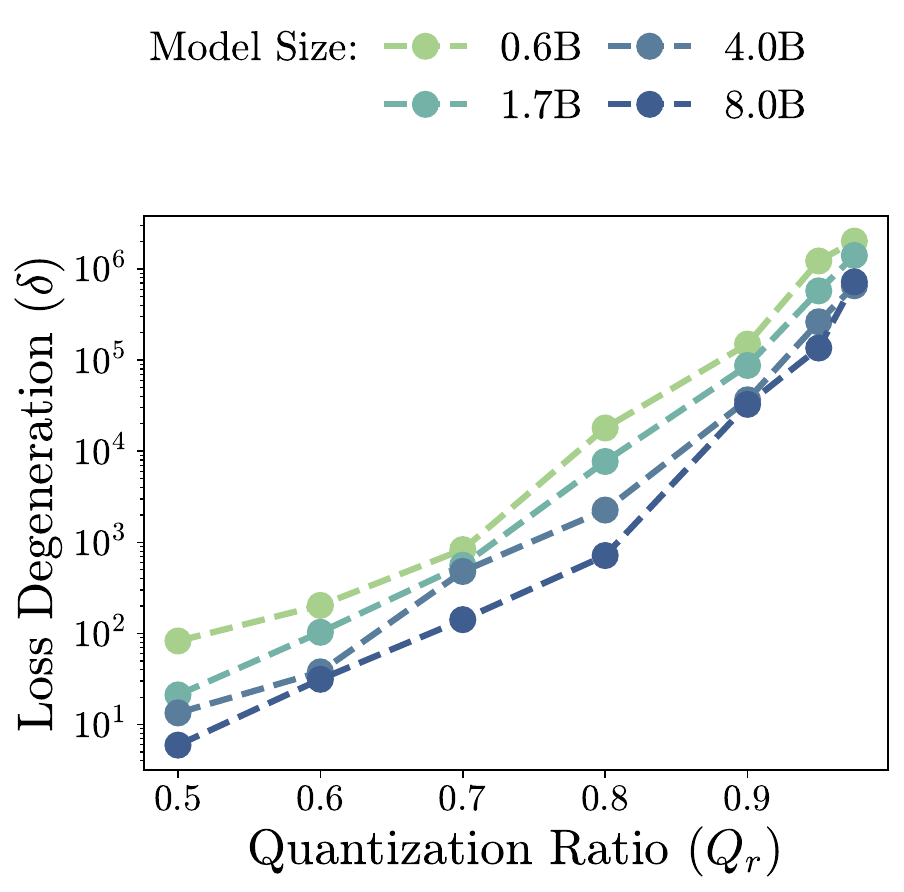}
    \captionsetup{font=scriptsize}
    \caption{Qwen-3 Actual Loss}
\end{subfigure}

\begin{subfigure}[b]{0.3\textwidth} \centering
\includegraphics[width=\textwidth]{figures/figures-mxint2/llama-matmul2/logloss_vs_qratio-fitted.pdf}
    \captionsetup{font=scriptsize}
    \caption{CLM Predicted Loss}
    \end{subfigure}
\hfill
\begin{subfigure}[b]{0.3\textwidth} \centering
    \includegraphics[width=\textwidth]{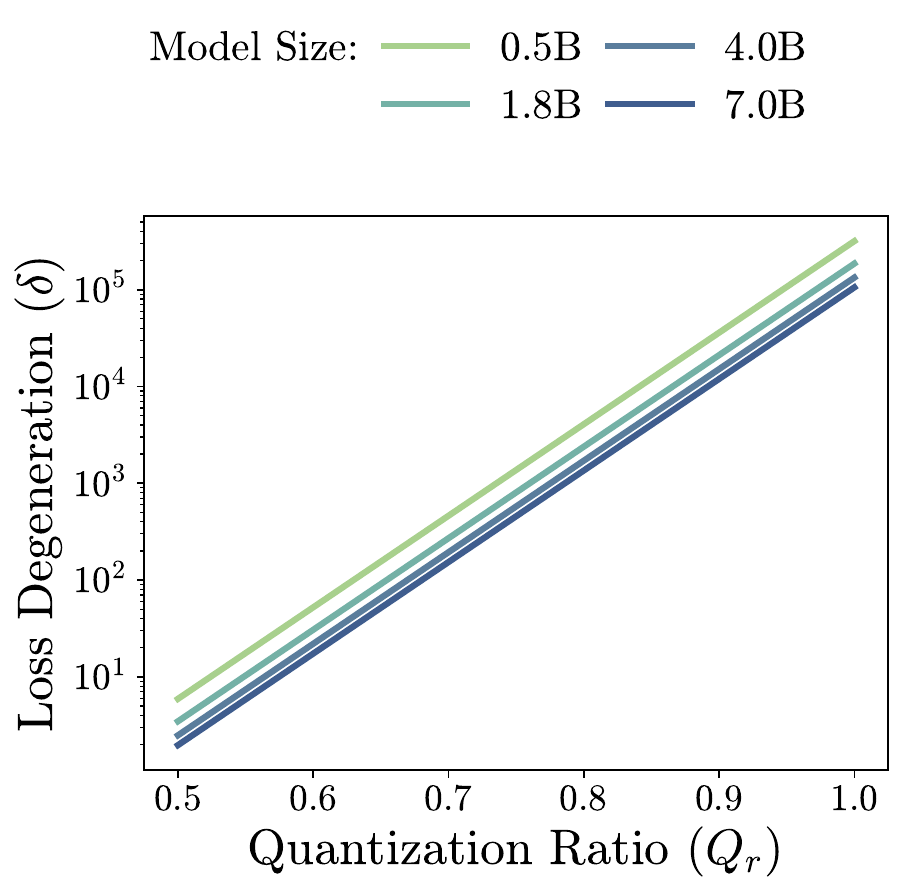}
    \captionsetup{font=scriptsize}
    \caption{Qwen-1.5 Predicted Loss}
\end{subfigure}
\hfill
\begin{subfigure}[b]{0.3\textwidth} \centering
    \includegraphics[width=\textwidth]{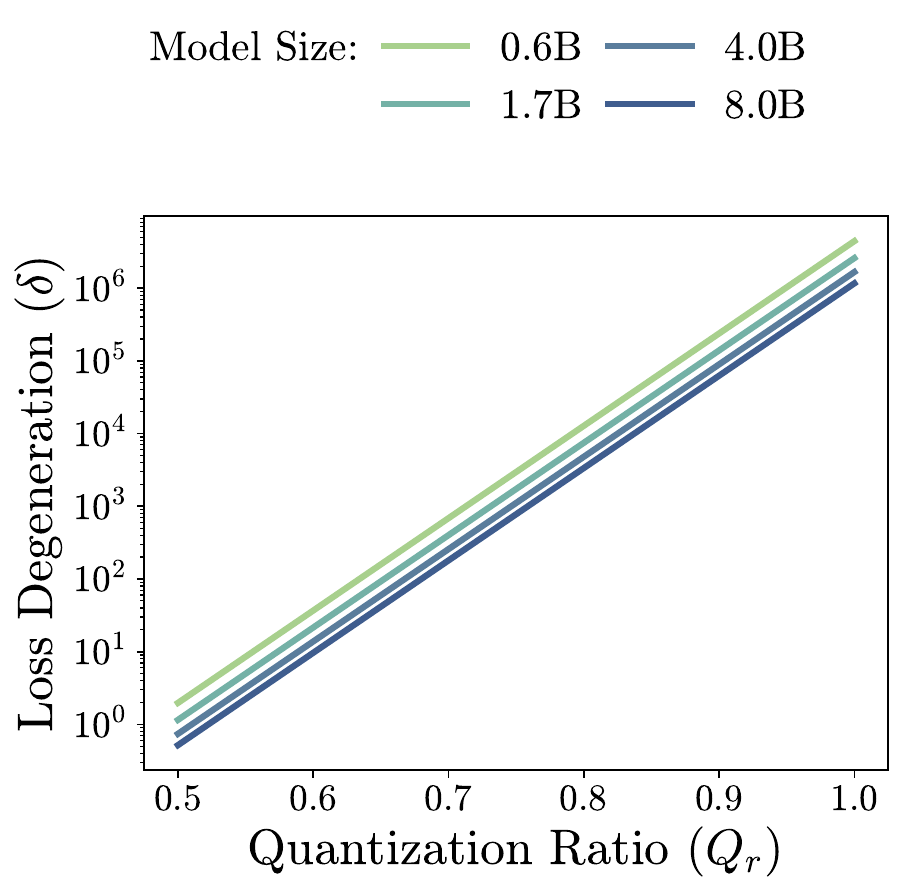}
    \captionsetup{font=scriptsize}
    \caption{Qwen-3 Predicted Loss}
\end{subfigure}

\caption{\textbf{MXINT-2 ($\delta^{\text{opt}}$)} (a,d,g,h) CLM MXINT-2 results; (b,e,h,k) Qwen-1.5 MXINT-2 results; (c,f,i,l) Qwen-3 MXINT-2 results.}
  \label{fig:appendix-mxint2}
\end{figure}

\begin{figure}[htbp]
\centering
\begin{subfigure}[b]{0.3\textwidth} \centering
    \includegraphics[width=\textwidth]{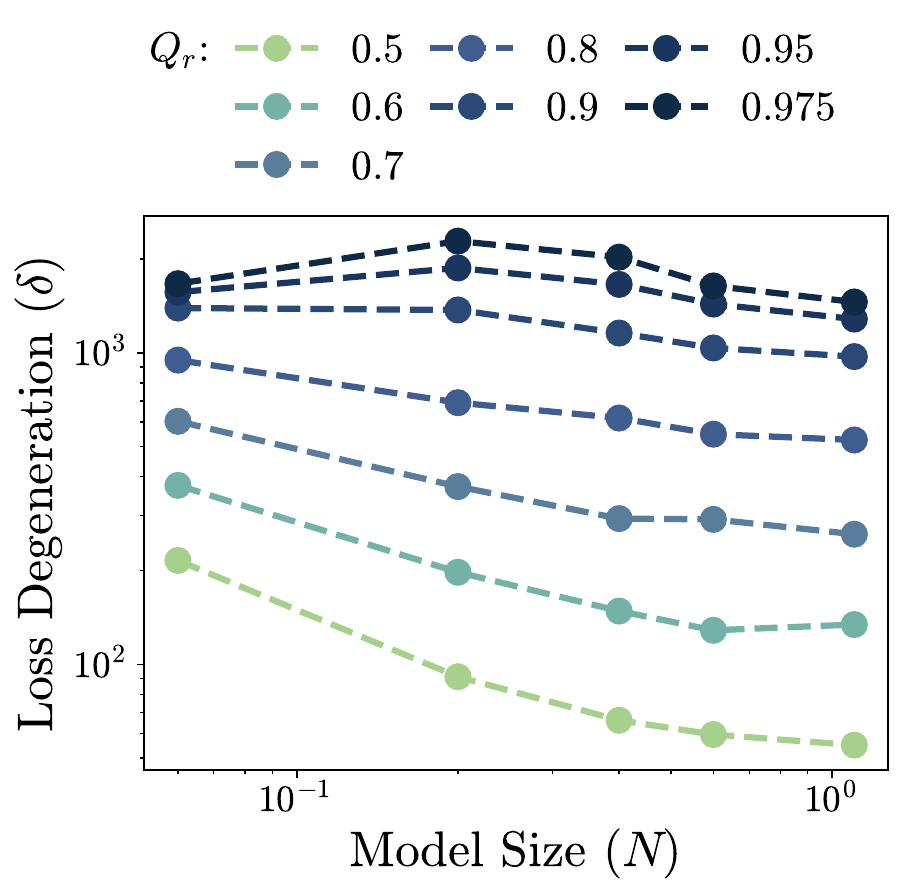}
    \captionsetup{font=scriptsize}
    \caption{CLM Actual Loss}
\end{subfigure}
\hfill
\begin{subfigure}[b]{0.3\textwidth} \centering
    \includegraphics[width=\textwidth]{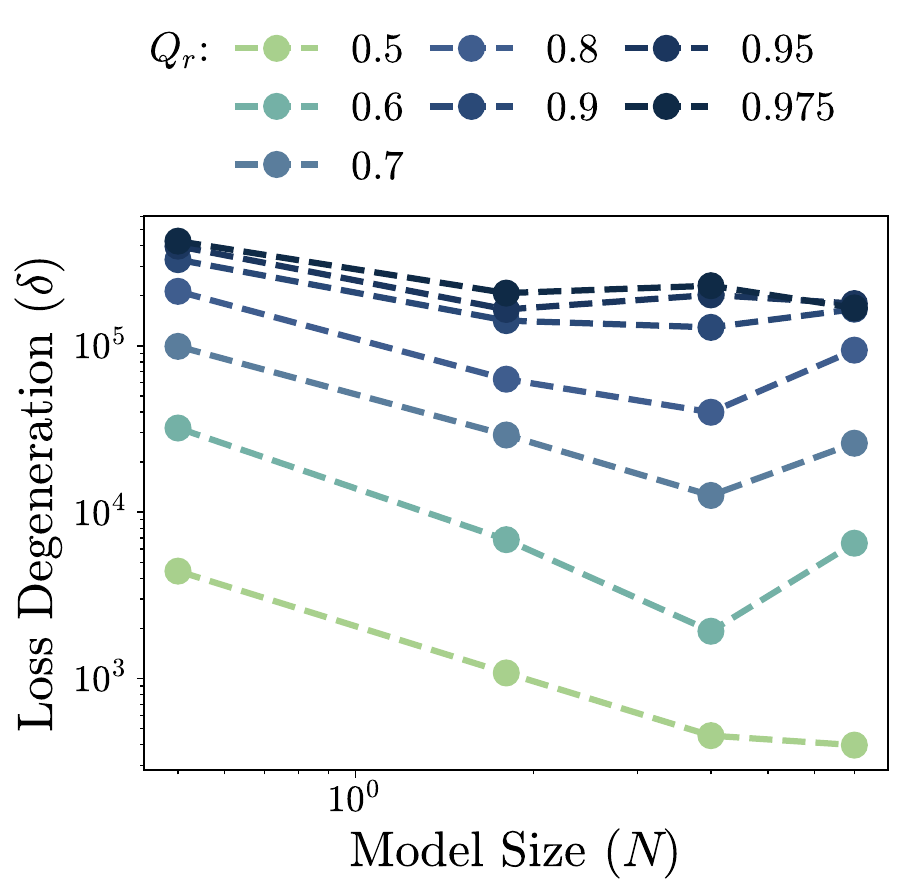}
    \captionsetup{font=scriptsize}
    \caption{Qwen-1.5 Actual Loss}
\end{subfigure}
\hfill
\begin{subfigure}[b]{0.3\textwidth} \centering
    \includegraphics[width=\textwidth]{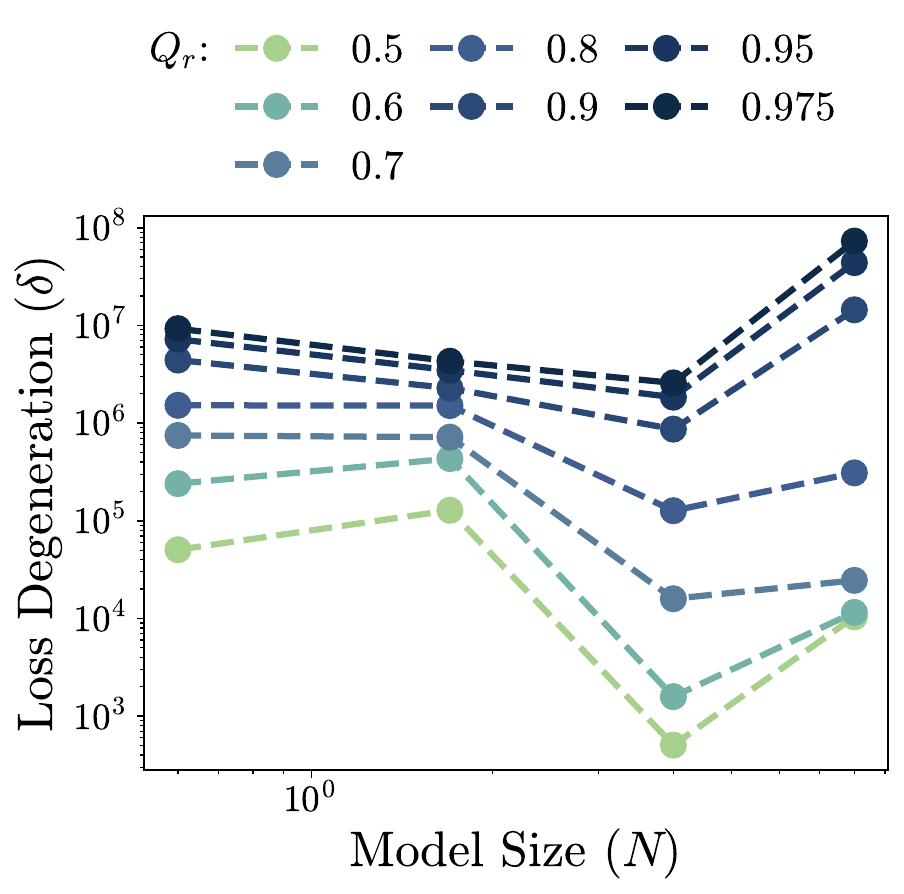}
    \captionsetup{font=scriptsize}
    \caption{Qwen-3 Actual Loss}
\end{subfigure}

\begin{subfigure}[b]{0.3\textwidth} \centering
    \includegraphics[width=\textwidth]{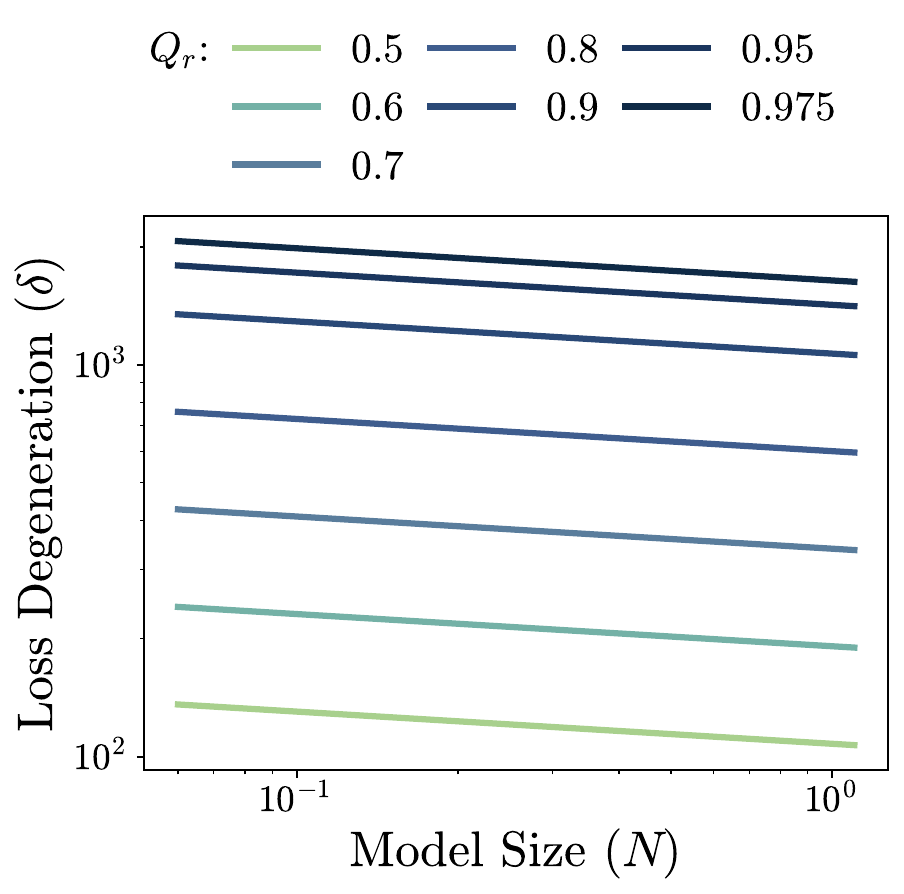}
    \captionsetup{font=scriptsize}
    \caption{CLM Predicted Loss}
\end{subfigure}
\hfill
\begin{subfigure}[b]{0.3\textwidth} \centering
    \includegraphics[width=\textwidth]{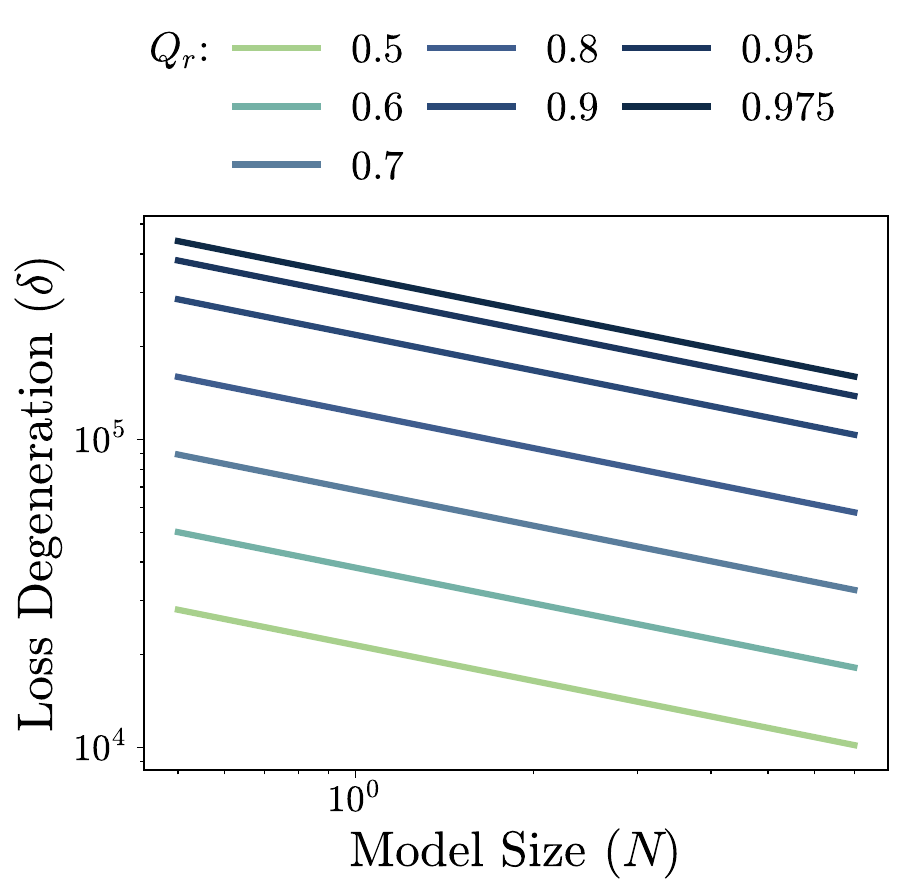}
    \captionsetup{font=scriptsize}
    \caption{Qwen-1.5 Predicted Loss}
\end{subfigure}
\hfill
\begin{subfigure}[b]{0.3\textwidth} \centering
    \includegraphics[width=\textwidth]{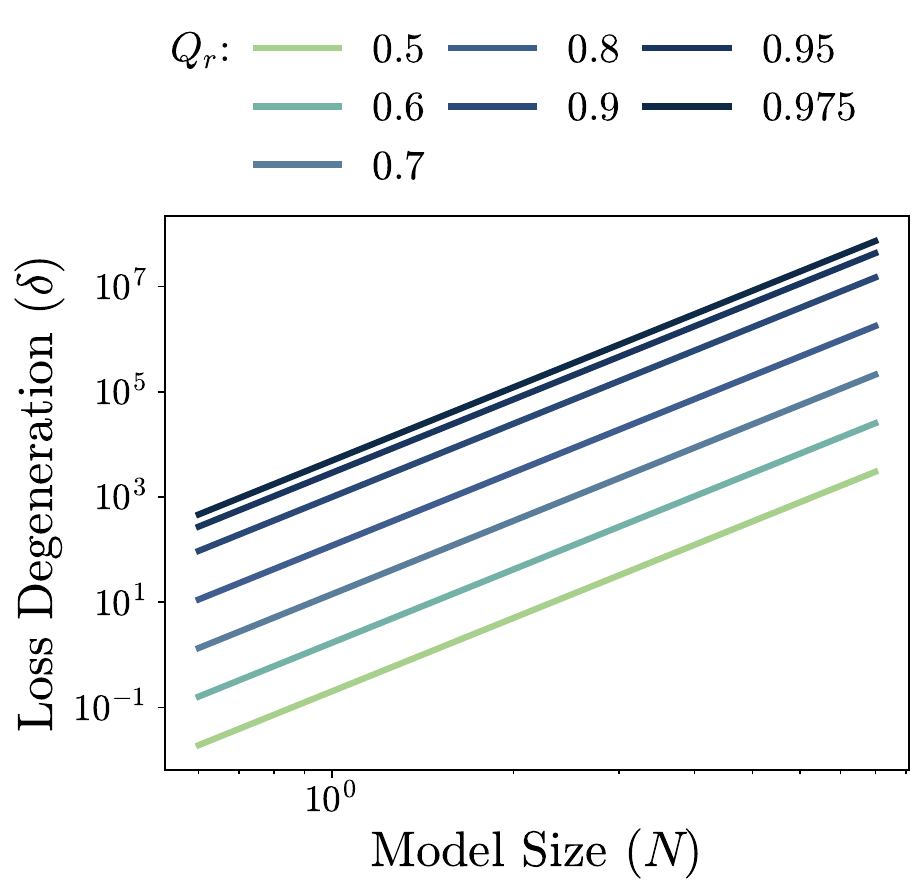}
    \captionsetup{font=scriptsize}
    \caption{Qwen-3 Predicted Loss}
\end{subfigure}

\begin{subfigure}[b]{0.3\textwidth} \centering
    \includegraphics[width=\textwidth]{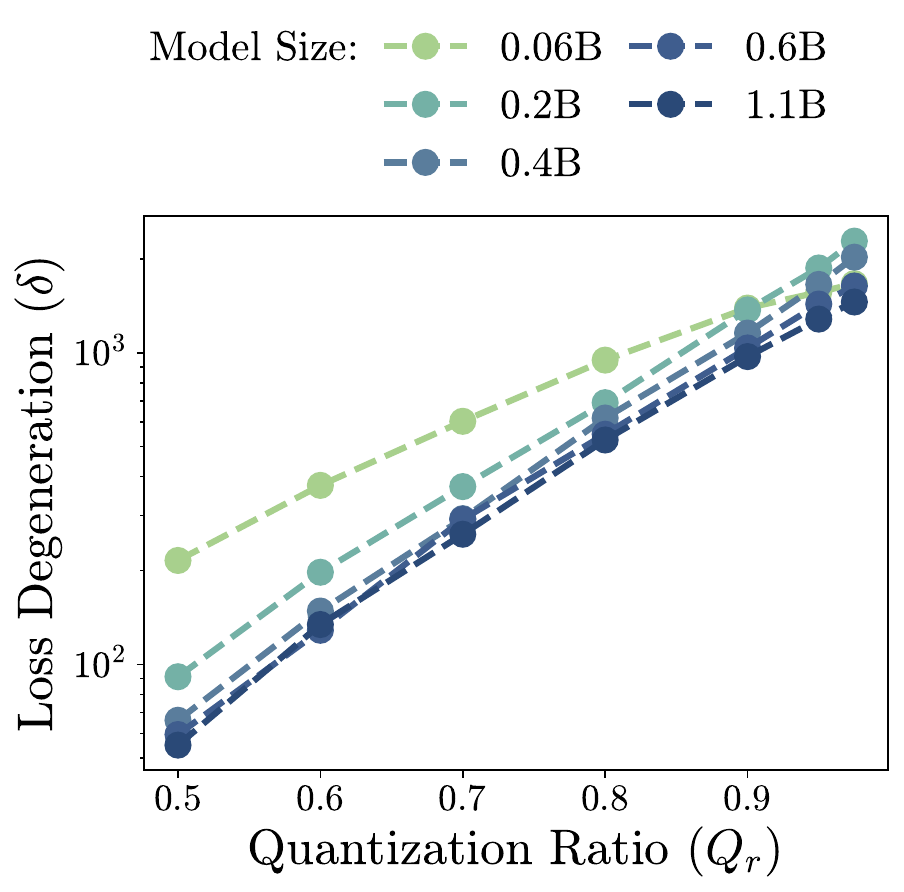}
    \captionsetup{font=scriptsize}
    \caption{CLM Actual Loss}
\end{subfigure}
\hfill
\begin{subfigure}[b]{0.3\textwidth} \centering
    \includegraphics[width=\textwidth]{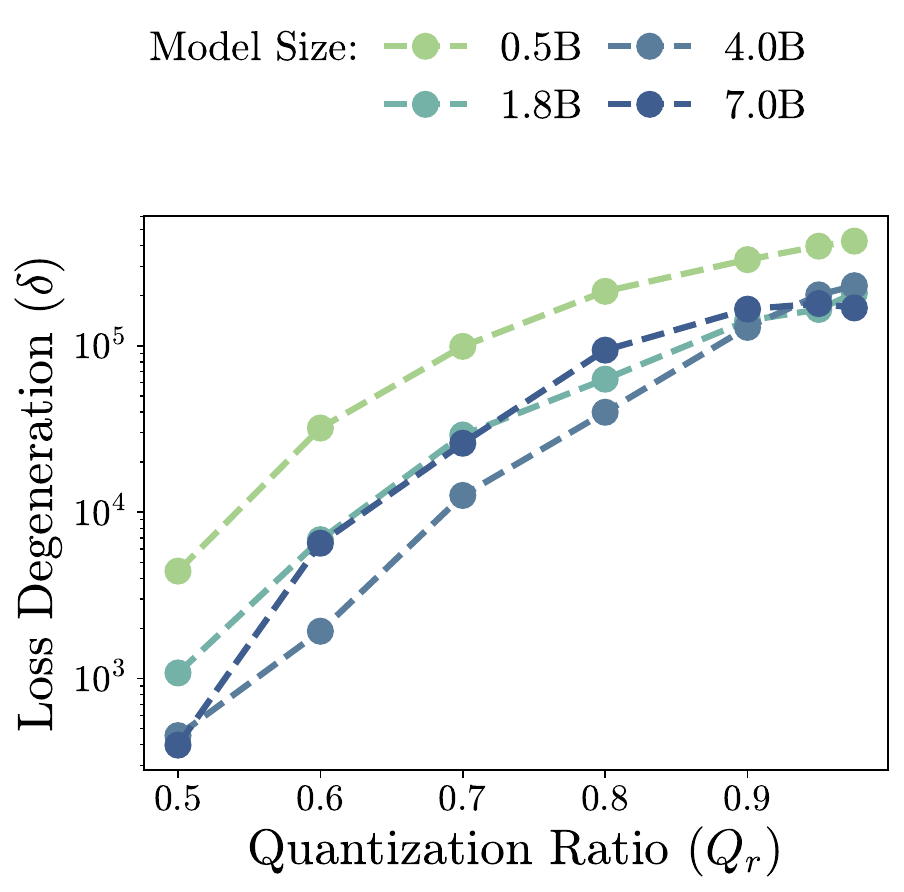}
    \captionsetup{font=scriptsize}
    \caption{Qwen-1.5 Actual Loss}
\end{subfigure}
\hfill
\begin{subfigure}[b]{0.3\textwidth} \centering
    \includegraphics[width=\textwidth]{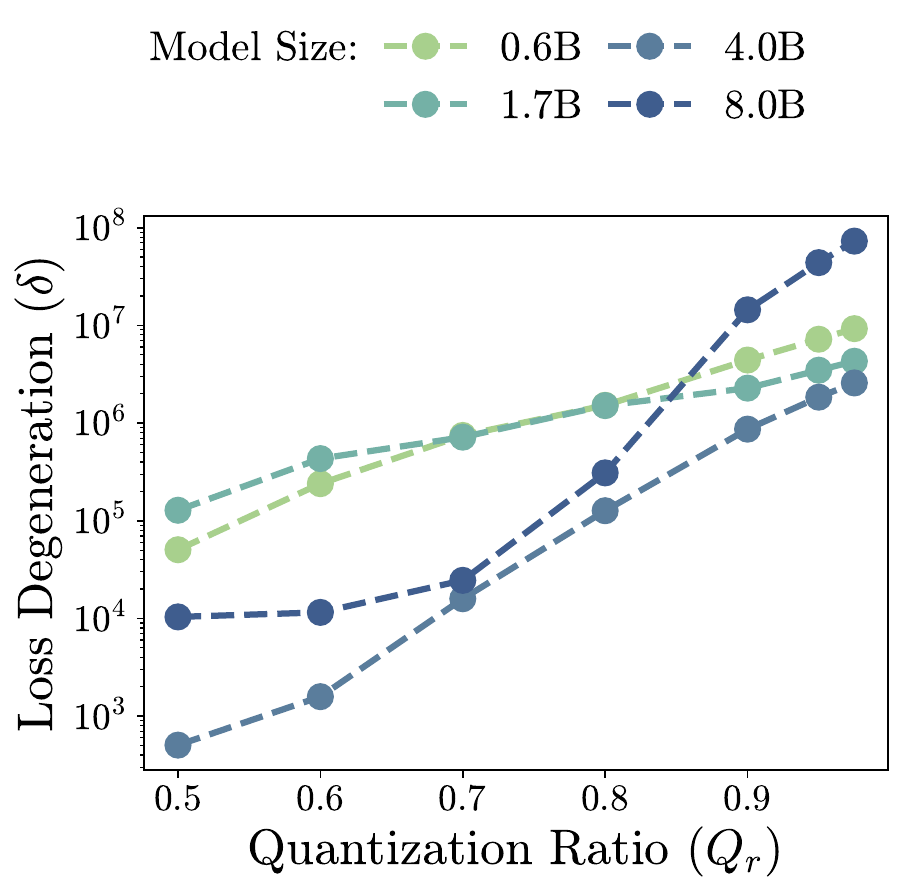}
    \captionsetup{font=scriptsize}
    \caption{Qwen-3 Actual Loss}
\end{subfigure}

\begin{subfigure}[b]{0.3\textwidth} \centering
\includegraphics[width=\textwidth]{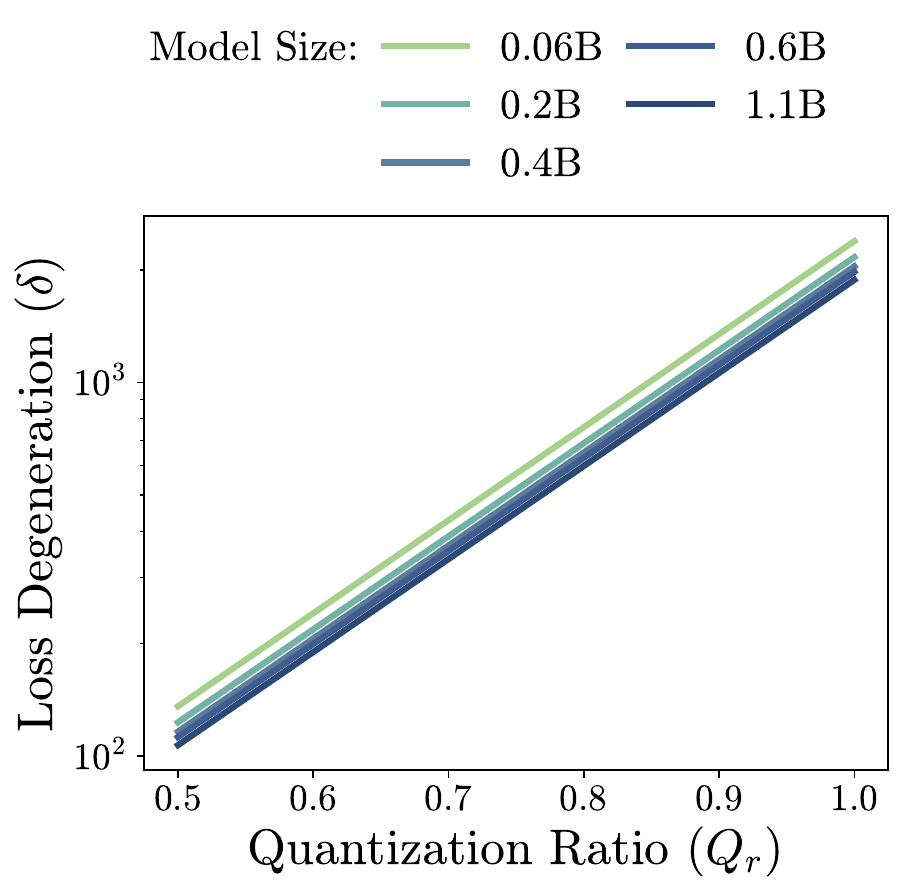}
    \captionsetup{font=scriptsize}
    \caption{CLM Predicted Loss}
    \end{subfigure}
\hfill
\begin{subfigure}[b]{0.3\textwidth} \centering
    \includegraphics[width=\textwidth]{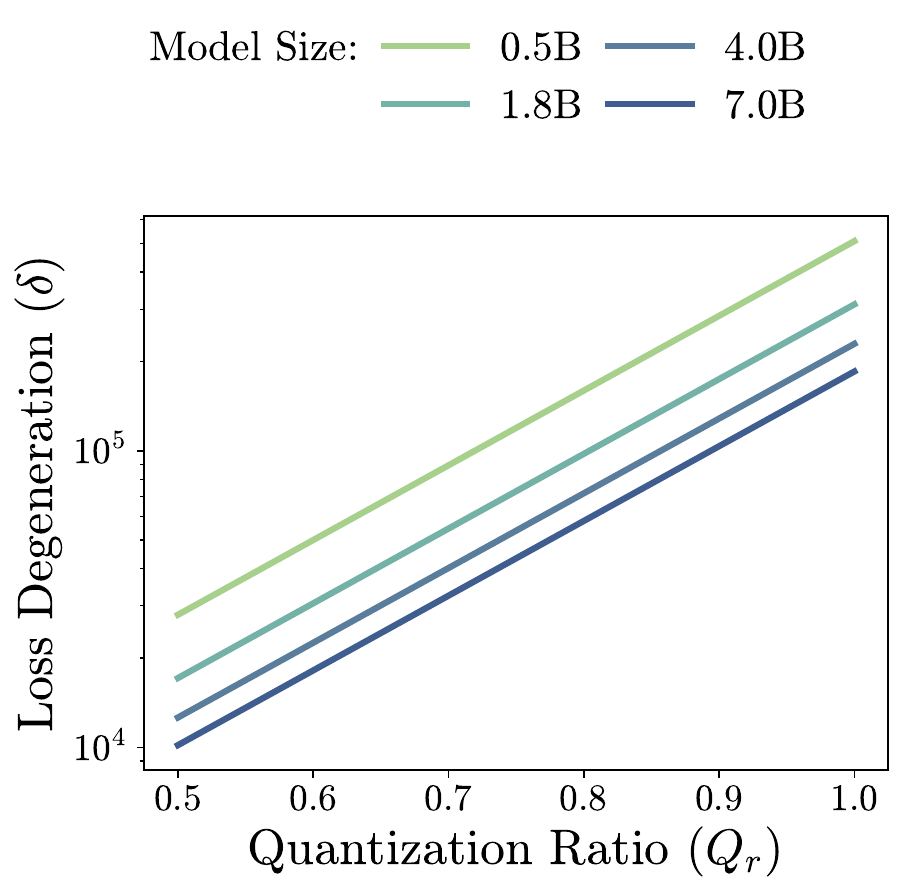}
    \captionsetup{font=scriptsize}
    \caption{Qwen-1.5 Predicted Loss}
\end{subfigure}
\hfill
\begin{subfigure}[b]{0.3\textwidth} \centering
    \includegraphics[width=\textwidth]{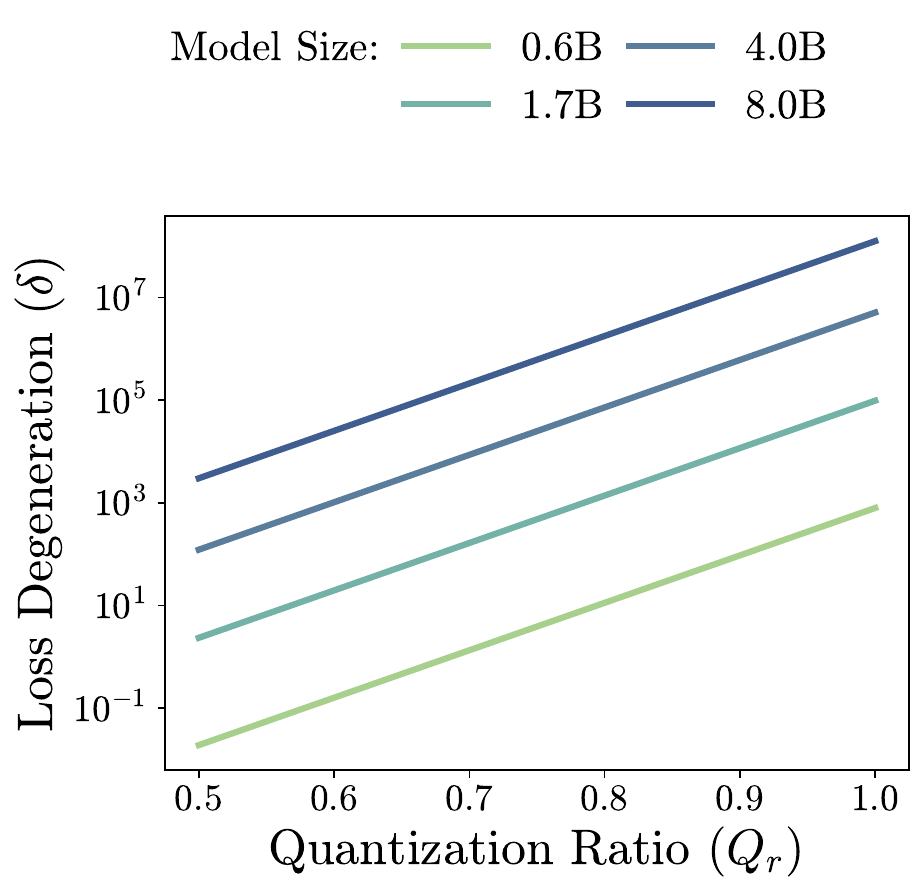}
    \captionsetup{font=scriptsize}
    \caption{Qwen-3 Predicted Loss}
\end{subfigure}

\caption{\textbf{MXINT-2 ($\delta_\mu$)} (a,d,g,h) CLM MXINT2 results; (b,e,h,k) Qwen-1.5 MXINT2 results; (c,f,i,l) Qwen-3 MXINT2 results.}
  \label{fig:appendix-mxint2-mean}
\end{figure}

\clearpage
In contrast, Figures~\ref{fig:appendix-mxint2} and~\ref{fig:appendix-mxint2-mean} show the results for MXINT-2 quantization, which performs poorly across the board. The loss increases significantly under this setting, suggesting that aggressive quantization in both weights and activations severely degrades model quality. Despite the high loss, our fitted contours remain aligned with the empirical observations, capturing the underlying loss surface accurately.

\begin{figure}[htbp]
\centering
\begin{subfigure}[b]{0.3\textwidth} \centering
    \includegraphics[width=\textwidth]{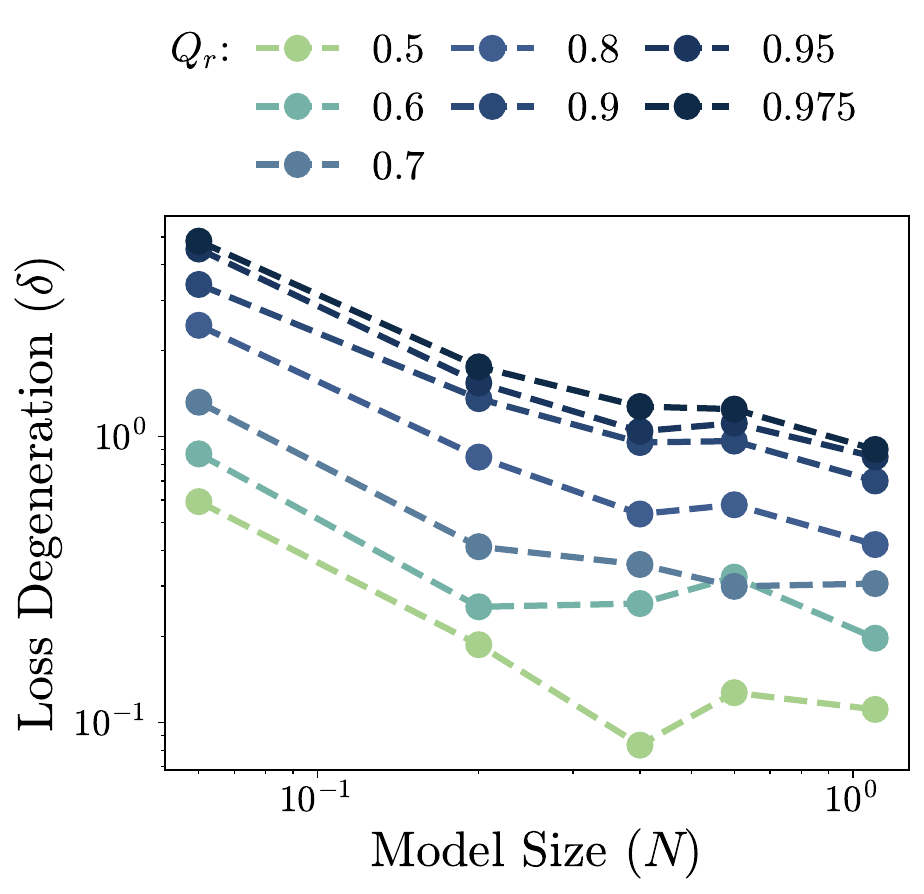}
    \captionsetup{font=scriptsize}
    \caption{CLM Actual Loss}
\end{subfigure}
\hfill
\begin{subfigure}[b]{0.3\textwidth} \centering
    \includegraphics[width=\textwidth]{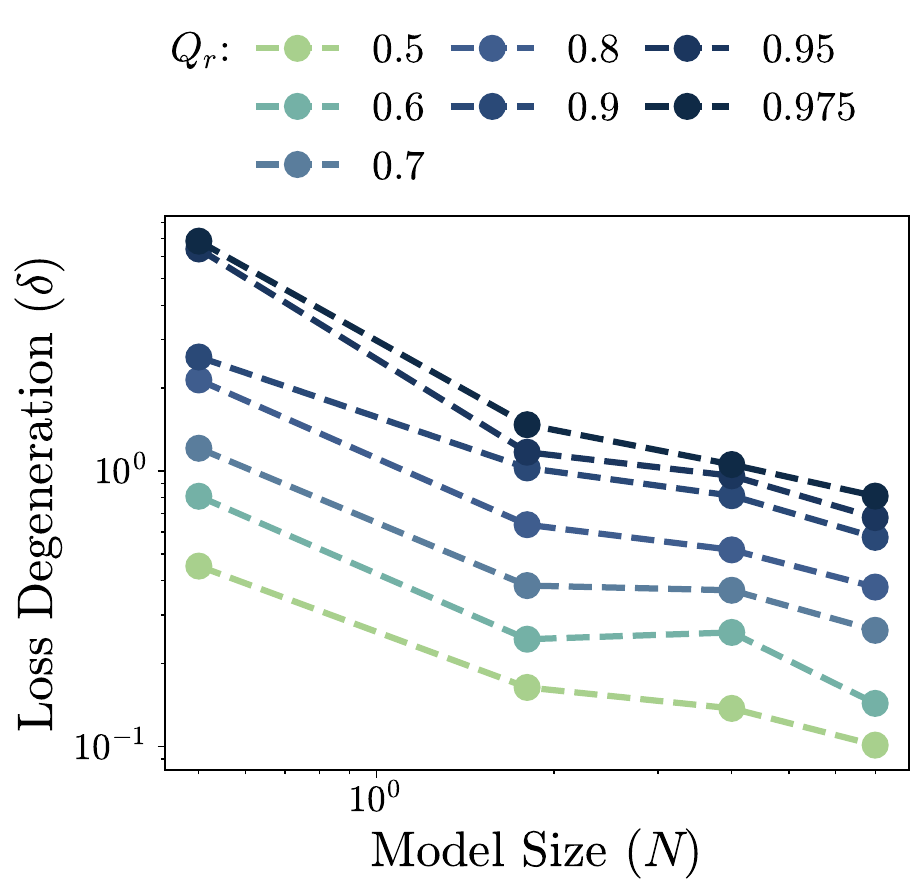}
    \captionsetup{font=scriptsize}
    \caption{Qwen-1.5 Actual Loss}
\end{subfigure}
\hfill
\begin{subfigure}[b]{0.3\textwidth} \centering
    \includegraphics[width=\textwidth]{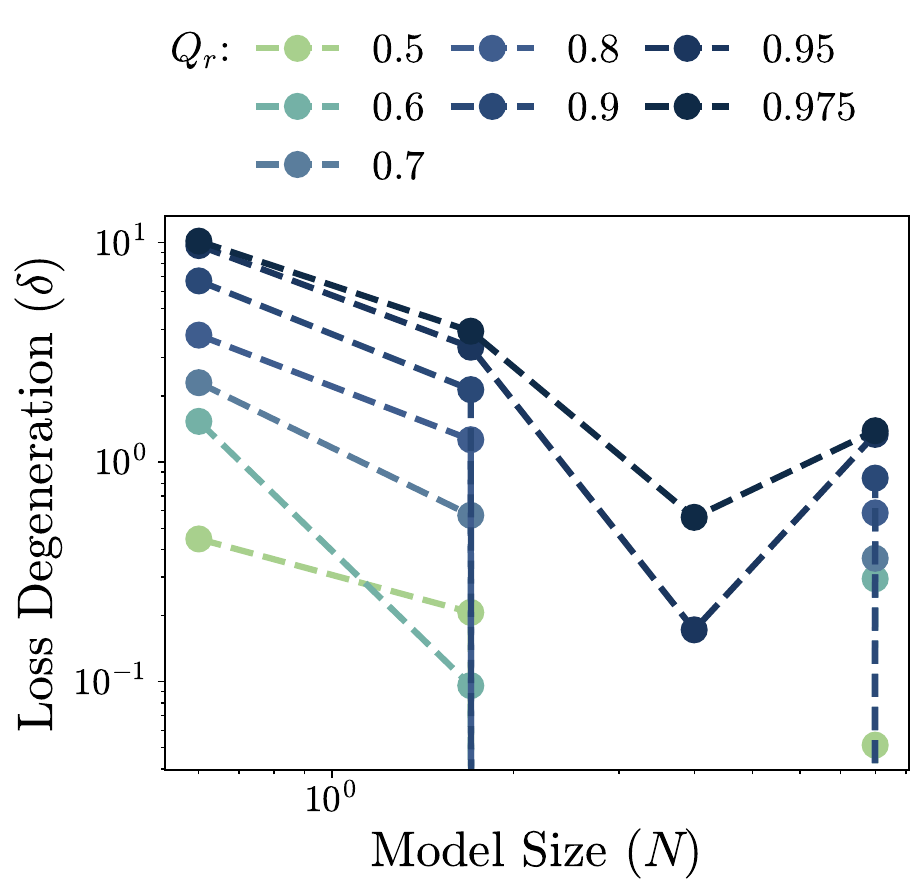}
    \captionsetup{font=scriptsize}
    \caption{Qwen-3 Actual Loss}
\end{subfigure}

\begin{subfigure}[b]{0.3\textwidth} \centering
    \includegraphics[width=\textwidth]{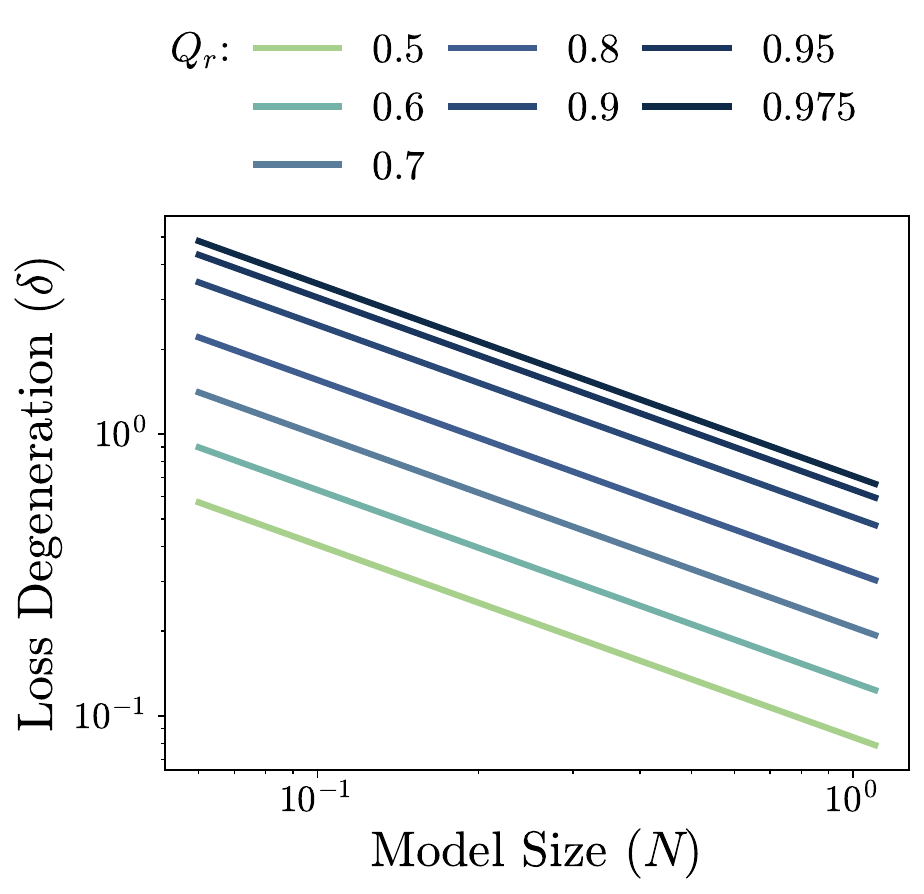}
    \captionsetup{font=scriptsize}
    \caption{CLM Predicted Loss}
\end{subfigure}
\hfill
\begin{subfigure}[b]{0.3\textwidth} \centering
    \includegraphics[width=\textwidth]{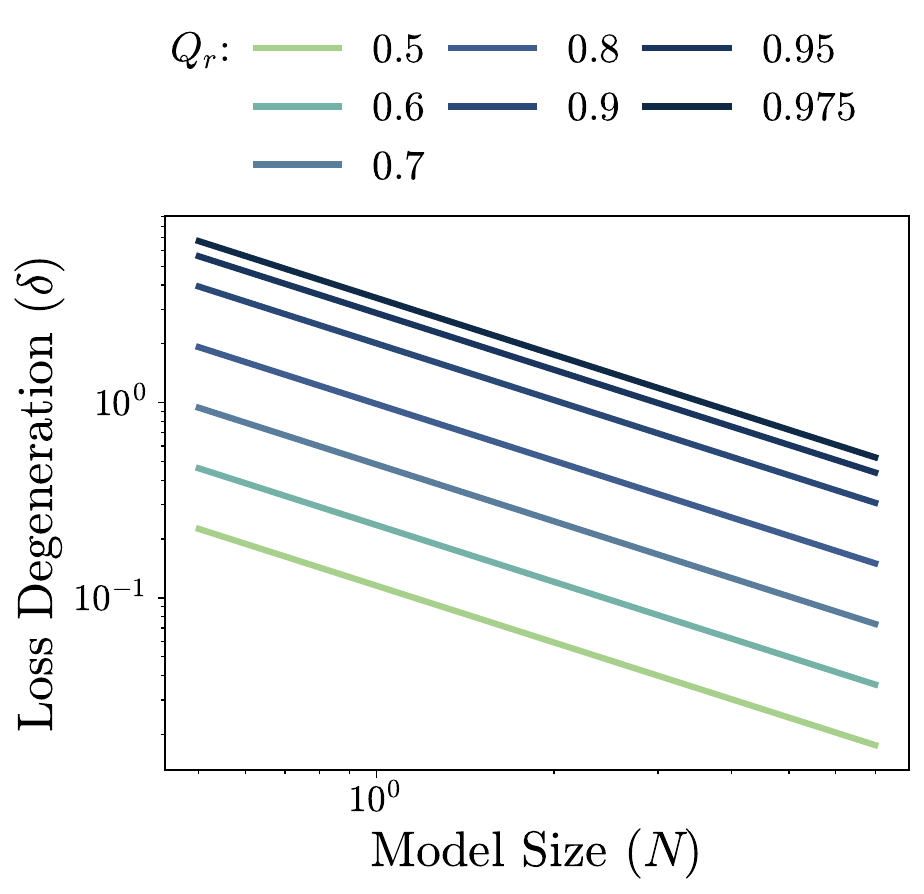}
    \captionsetup{font=scriptsize}
    \caption{Qwen-1.5 Predicted Loss}
\end{subfigure}
\hfill
\begin{subfigure}[b]{0.3\textwidth} \centering
    \includegraphics[width=\textwidth]{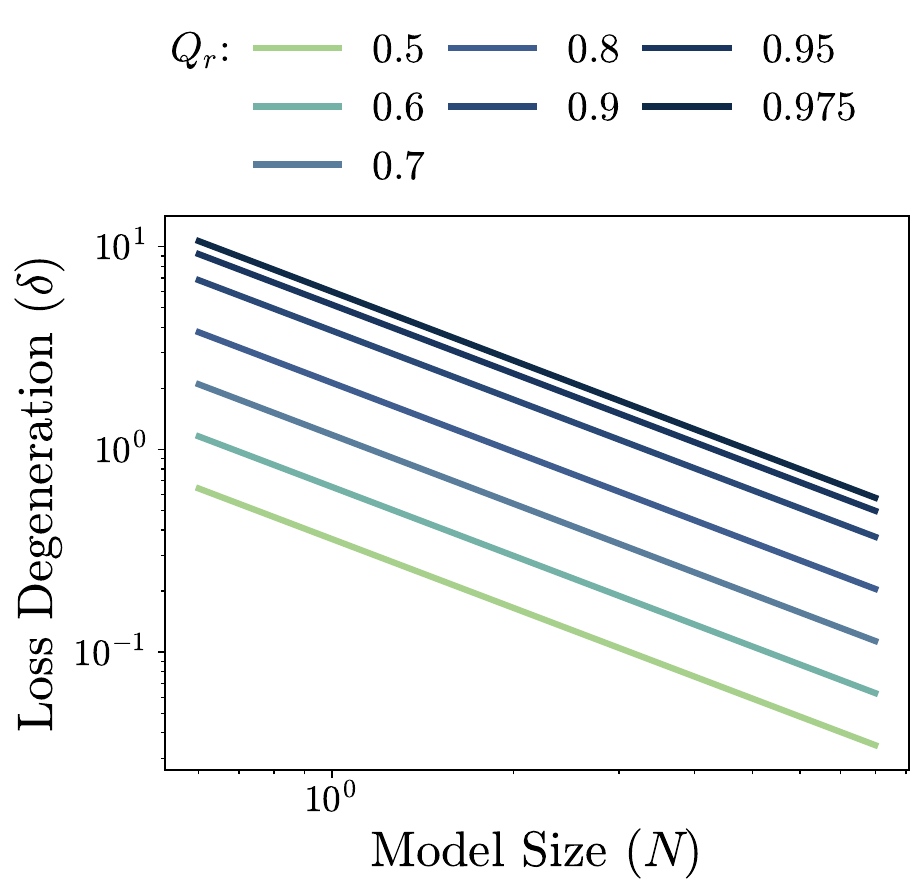}
    \captionsetup{font=scriptsize}
    \caption{Qwen-3 Predicted Loss}
\end{subfigure}

\begin{subfigure}[b]{0.3\textwidth} \centering
    \includegraphics[width=\textwidth]{figures/figures-w-only/llama-matmul/logloss_vs_qratio.pdf}
    \captionsetup{font=scriptsize}
    \caption{CLM Actual Loss}
\end{subfigure}
\hfill
\begin{subfigure}[b]{0.3\textwidth} \centering
    \includegraphics[width=\textwidth]{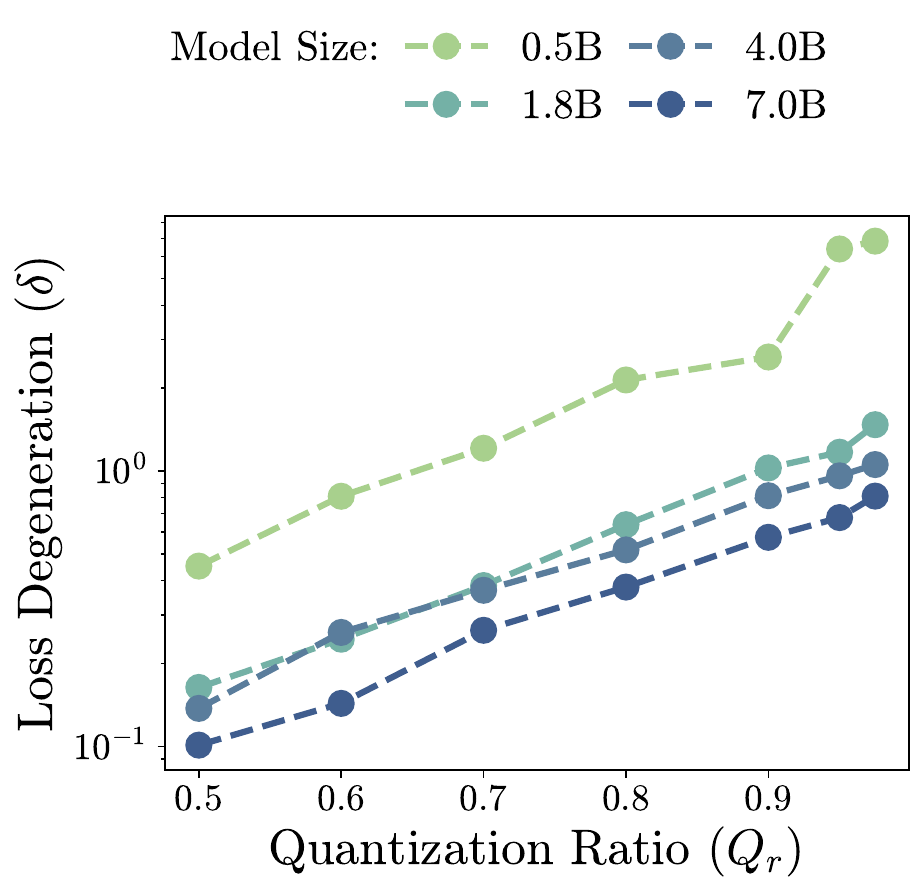}
    \captionsetup{font=scriptsize}
    \caption{Qwen-1.5 Actual Loss}
\end{subfigure}
\hfill
\begin{subfigure}[b]{0.3\textwidth} \centering
    \includegraphics[width=\textwidth]{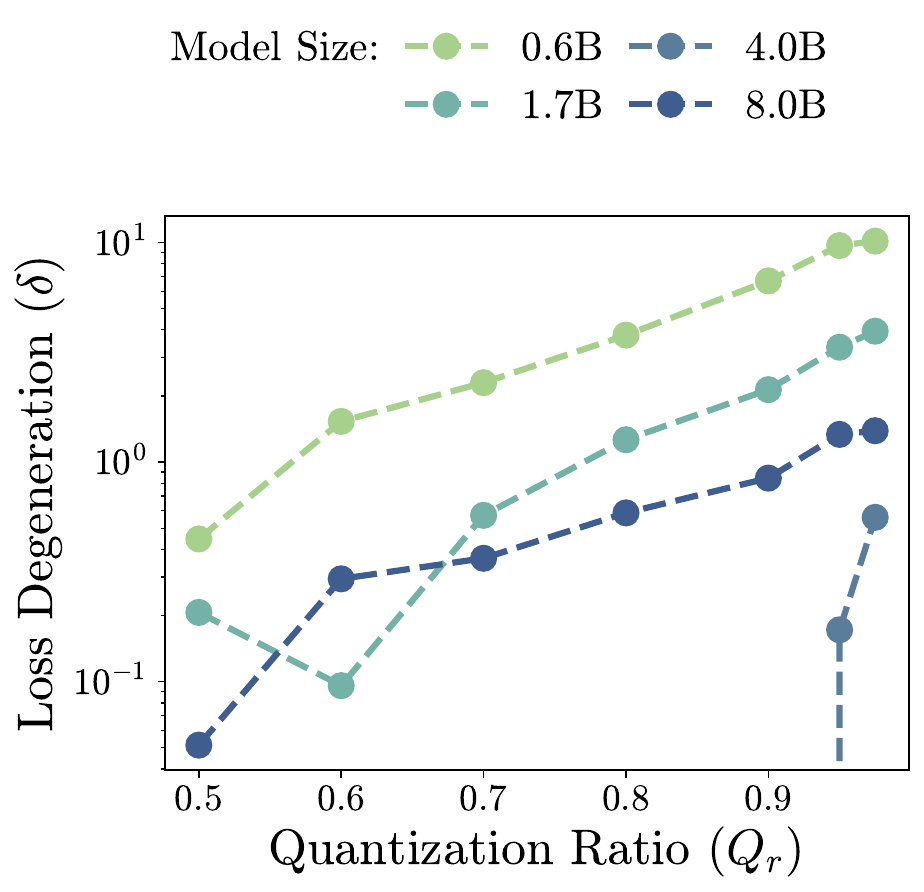}
    \captionsetup{font=scriptsize}
    \caption{Qwen-3 Actual Loss}
\end{subfigure}

\begin{subfigure}[b]{0.3\textwidth} \centering
\includegraphics[width=\textwidth]{figures/figures-w-only/llama-matmul/logloss_vs_qratio-fitted.pdf}
    \captionsetup{font=scriptsize}
    \caption{CLM Predicted Loss}
    \end{subfigure}
\hfill
\begin{subfigure}[b]{0.3\textwidth} \centering
    \includegraphics[width=\textwidth]{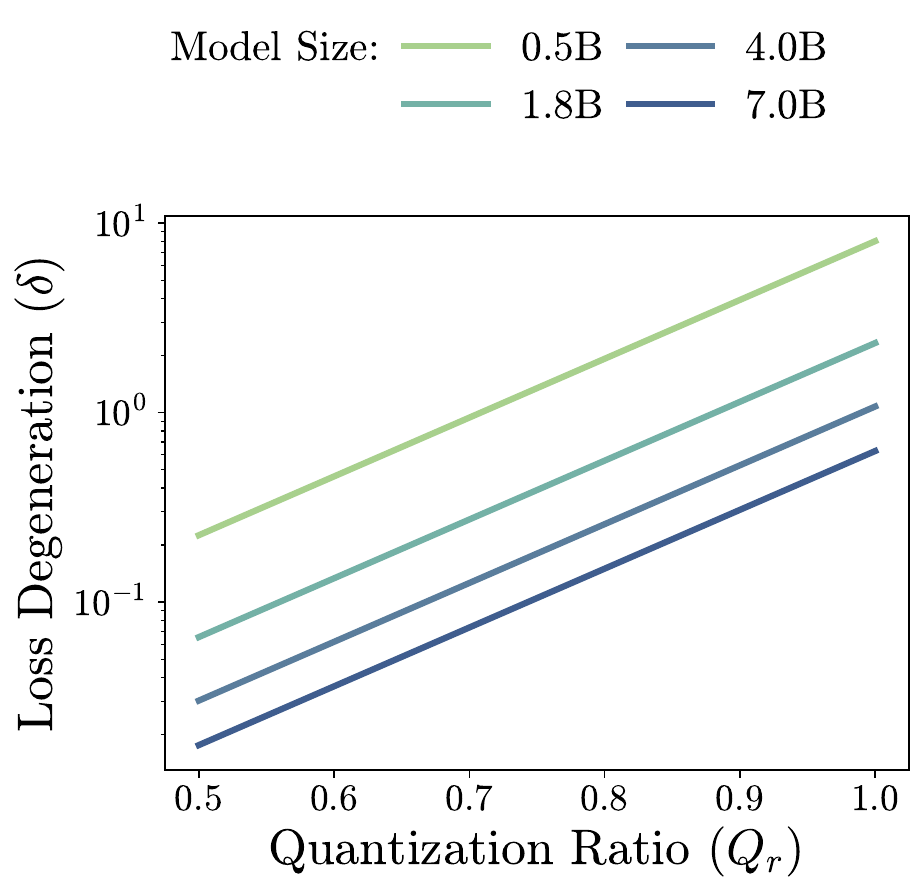}
    \captionsetup{font=scriptsize}
    \caption{Qwen-1.5 Predicted Loss}
\end{subfigure}
\hfill
\begin{subfigure}[b]{0.3\textwidth} \centering
    \includegraphics[width=\textwidth]{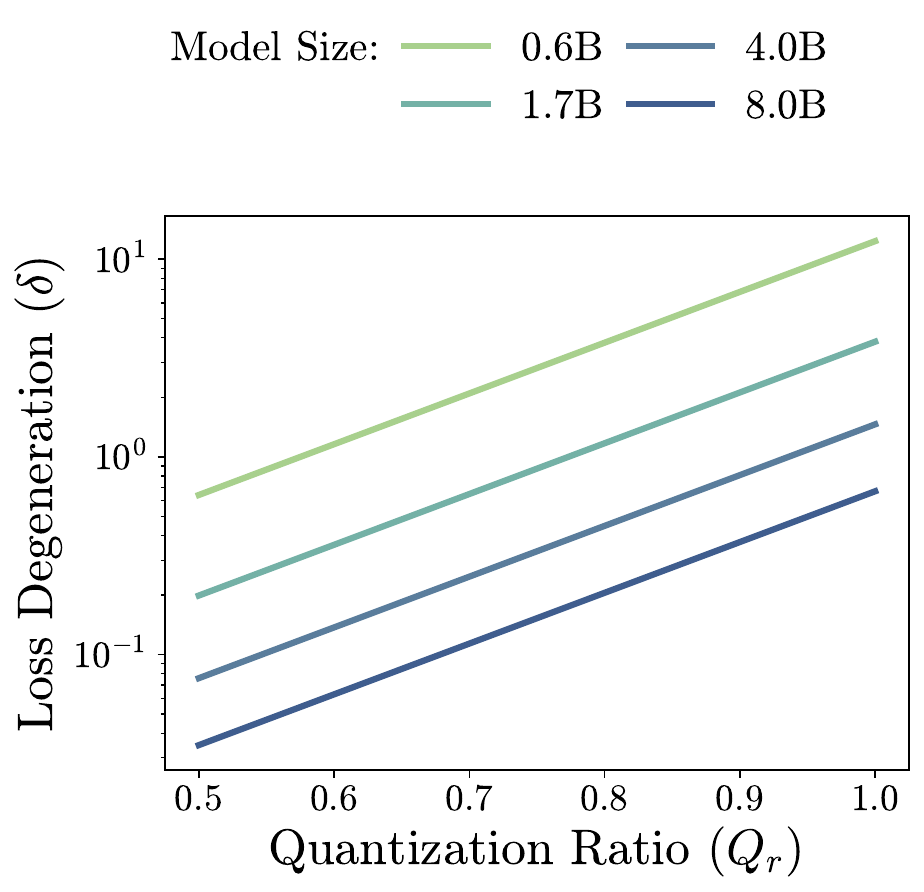}
    \captionsetup{font=scriptsize}
    \caption{Qwen-3 Predicted Loss}
\end{subfigure}

\caption{\textbf{MXINT-4 Weight-only ($\delta^{\text{opt}}$)} (a,d,g,h) CLM MXINT-4 Weight-only results; (b,e,h,k) Qwen-1.5 MXINT4 Weight-only results; (c,f,i,l) Qwen-3 MXINT-4 Weight-only results.}
  \label{fig:appendix-w-only}
\end{figure}

\begin{figure}[htbp]
\centering
\begin{subfigure}[b]{0.3\textwidth} \centering
    \includegraphics[width=\textwidth]{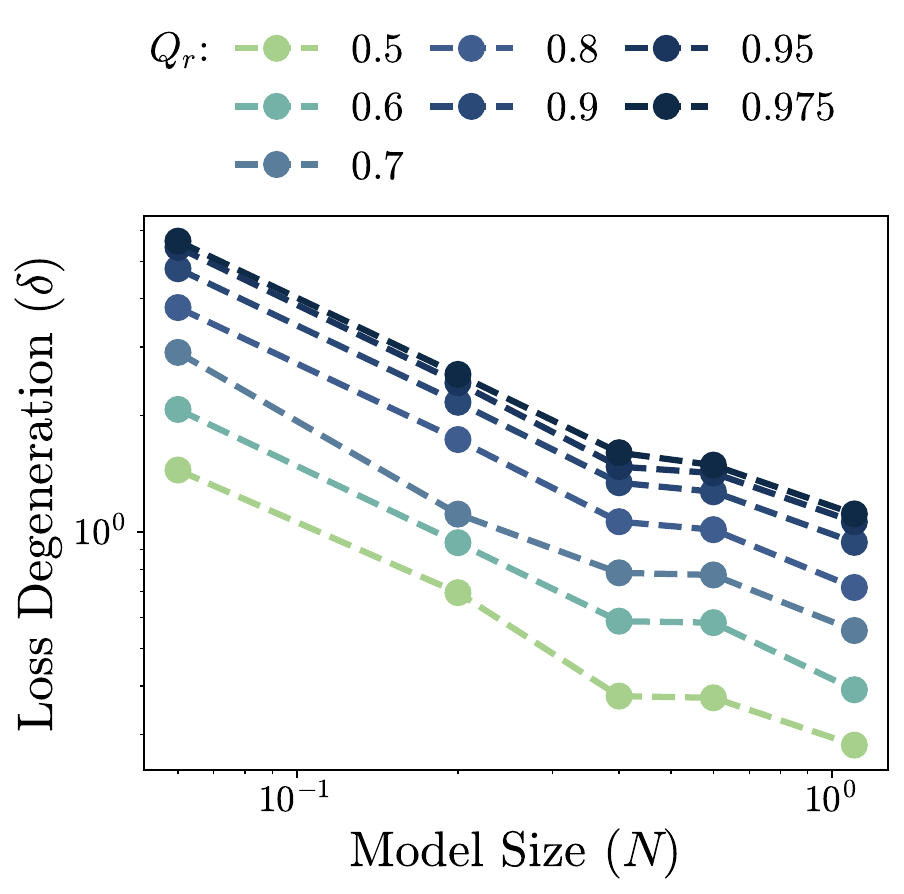}
    \captionsetup{font=scriptsize}
    \caption{CLM Actual Loss}
\end{subfigure}
\hfill
\begin{subfigure}[b]{0.3\textwidth} \centering
    \includegraphics[width=\textwidth]{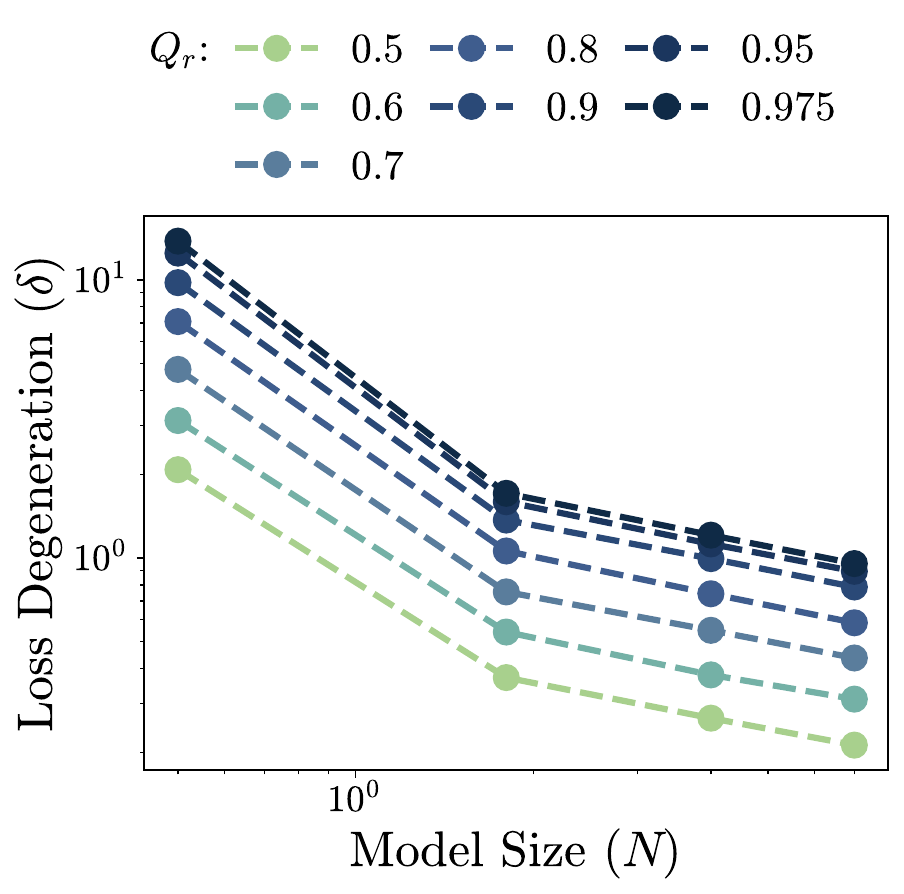}
    \captionsetup{font=scriptsize}
    \caption{Qwen1.5 Actual Loss}
\end{subfigure}
\hfill
\begin{subfigure}[b]{0.3\textwidth} \centering
    \includegraphics[width=\textwidth]{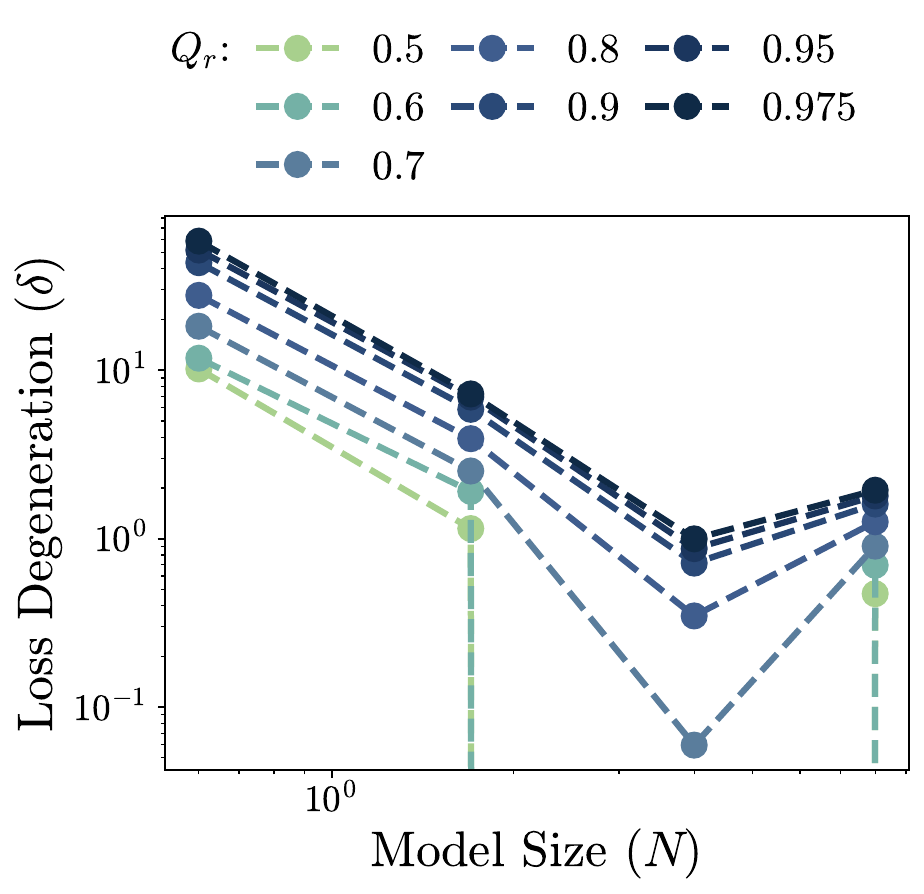}
    \captionsetup{font=scriptsize}
    \caption{Qwen3 Actual Loss}
\end{subfigure}

\begin{subfigure}[b]{0.3\textwidth} \centering
    \includegraphics[width=\textwidth]{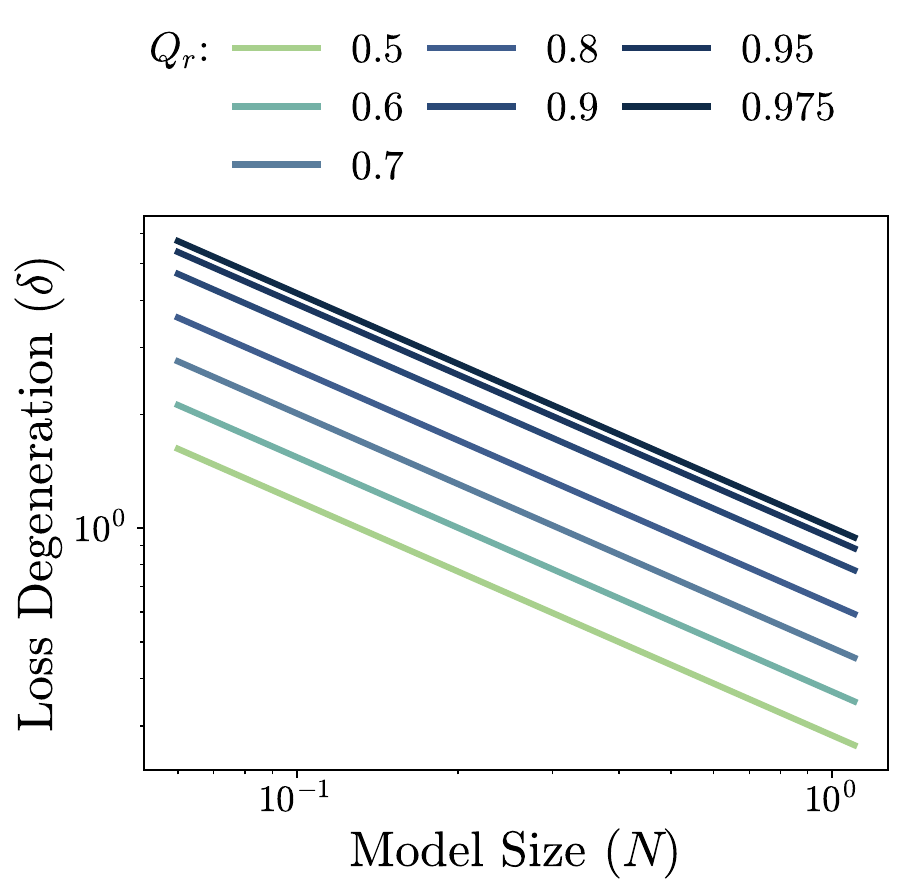}
    \captionsetup{font=scriptsize}
    \caption{CLM Predicted Loss}
\end{subfigure}
\hfill
\begin{subfigure}[b]{0.3\textwidth} \centering
    \includegraphics[width=\textwidth]{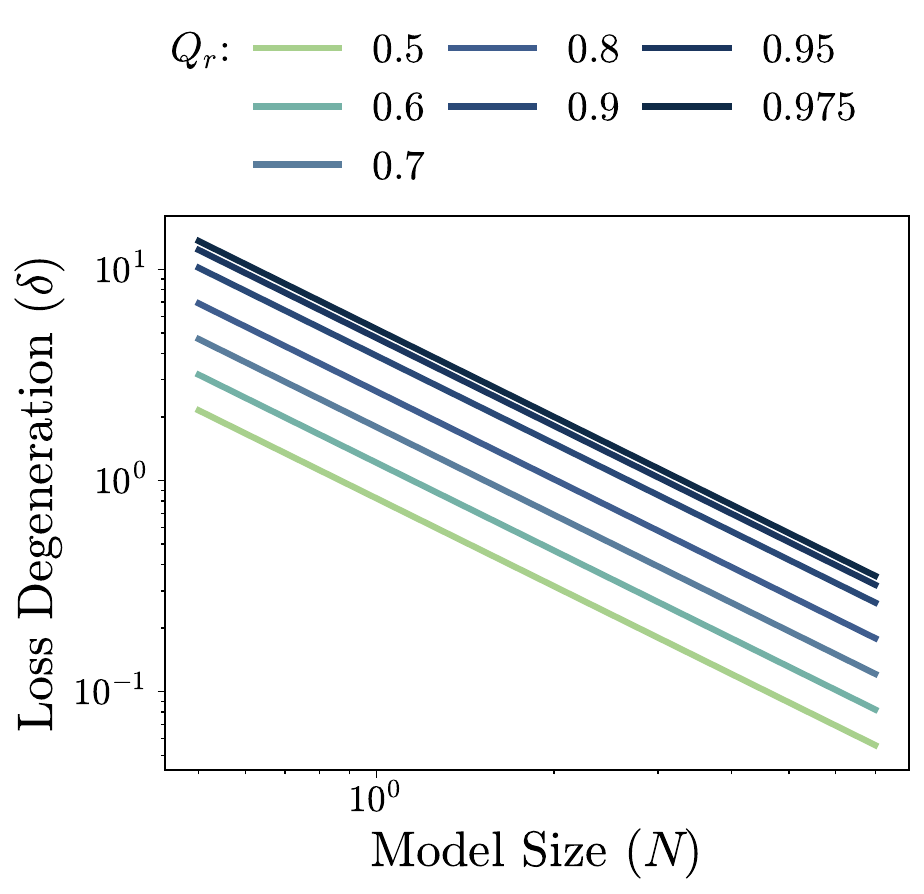}
    \captionsetup{font=scriptsize}
    \caption{Qwen1.5 Predicted Loss}
\end{subfigure}
\hfill
\begin{subfigure}[b]{0.3\textwidth} \centering
    \includegraphics[width=\textwidth]{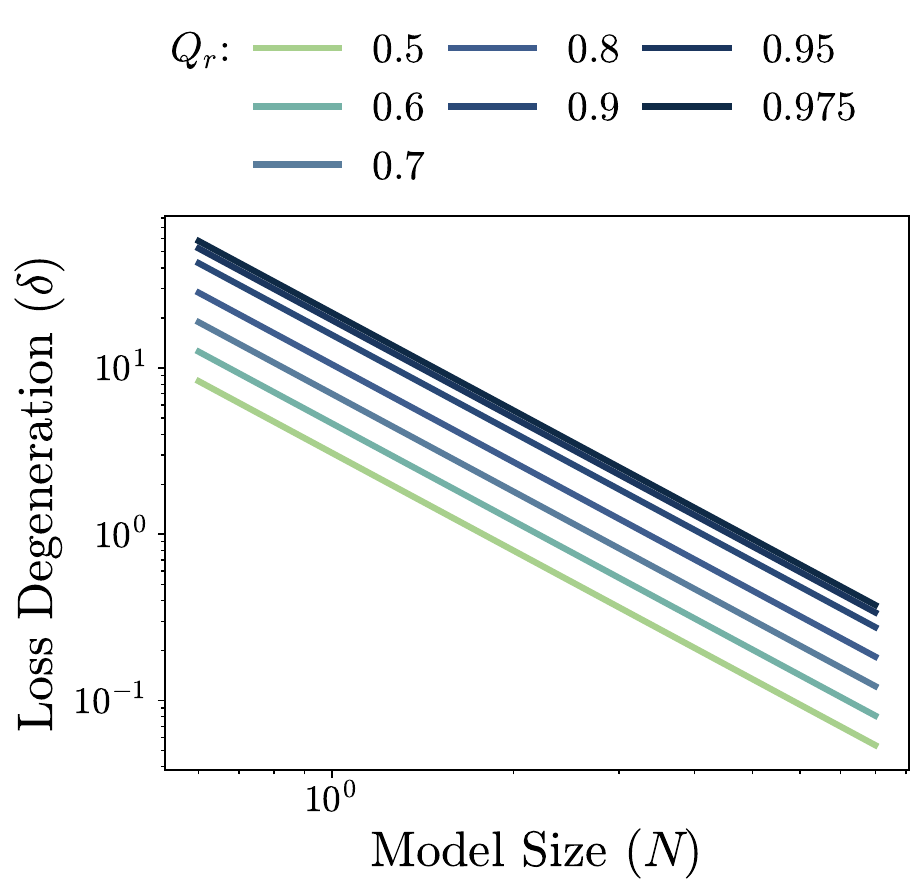}
    \captionsetup{font=scriptsize}
    \caption{Qwen3 Predicted Loss}
\end{subfigure}

\begin{subfigure}[b]{0.3\textwidth} \centering
    \includegraphics[width=\textwidth]{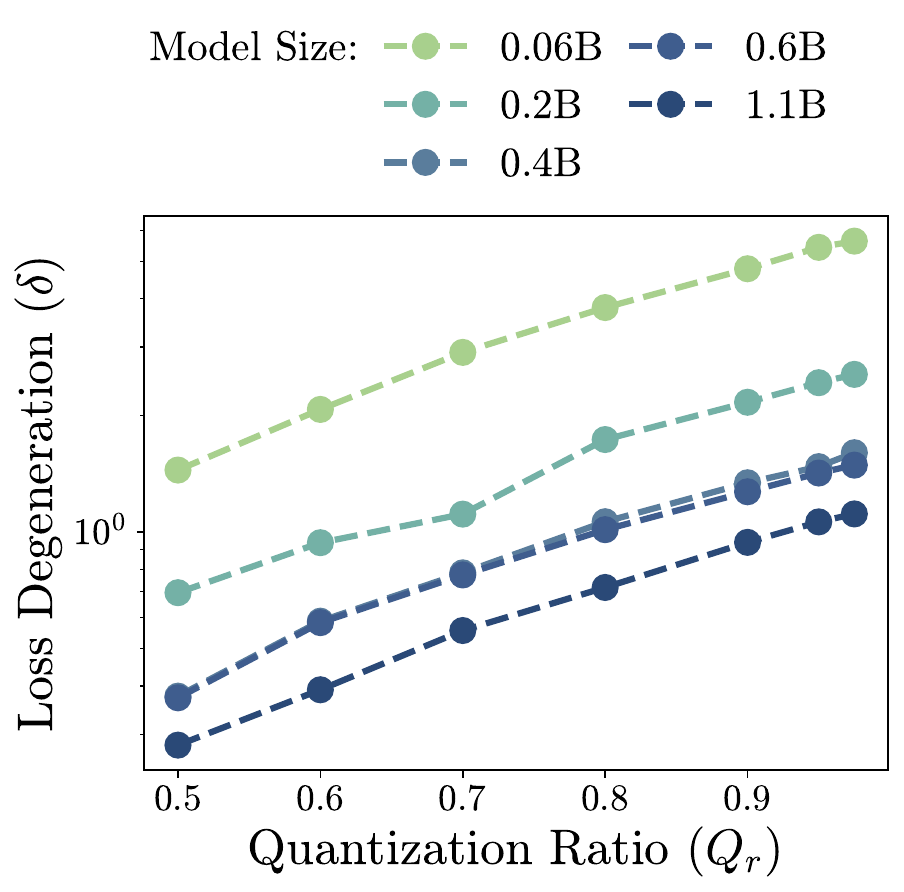}
    \captionsetup{font=scriptsize}
    \caption{CLM Actual Loss}
\end{subfigure}
\hfill
\begin{subfigure}[b]{0.3\textwidth} \centering
    \includegraphics[width=\textwidth]{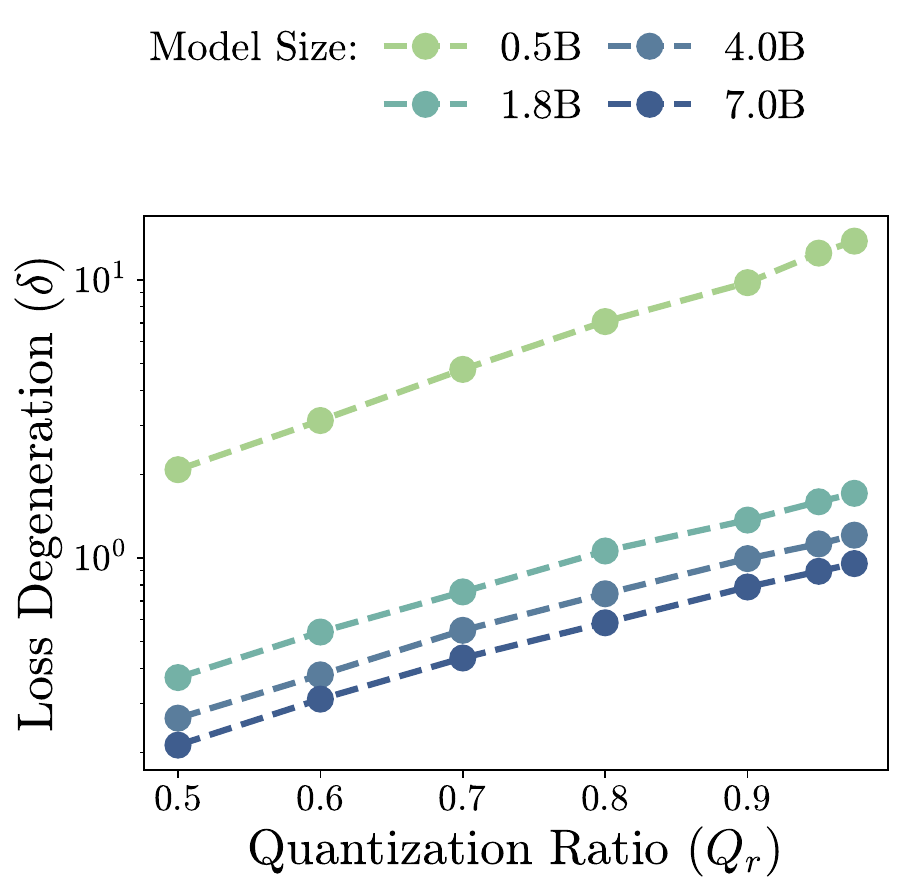}
    \captionsetup{font=scriptsize}
    \caption{Qwen1.5 Actual Loss}
\end{subfigure}
\hfill
\begin{subfigure}[b]{0.3\textwidth} \centering
    \includegraphics[width=\textwidth]{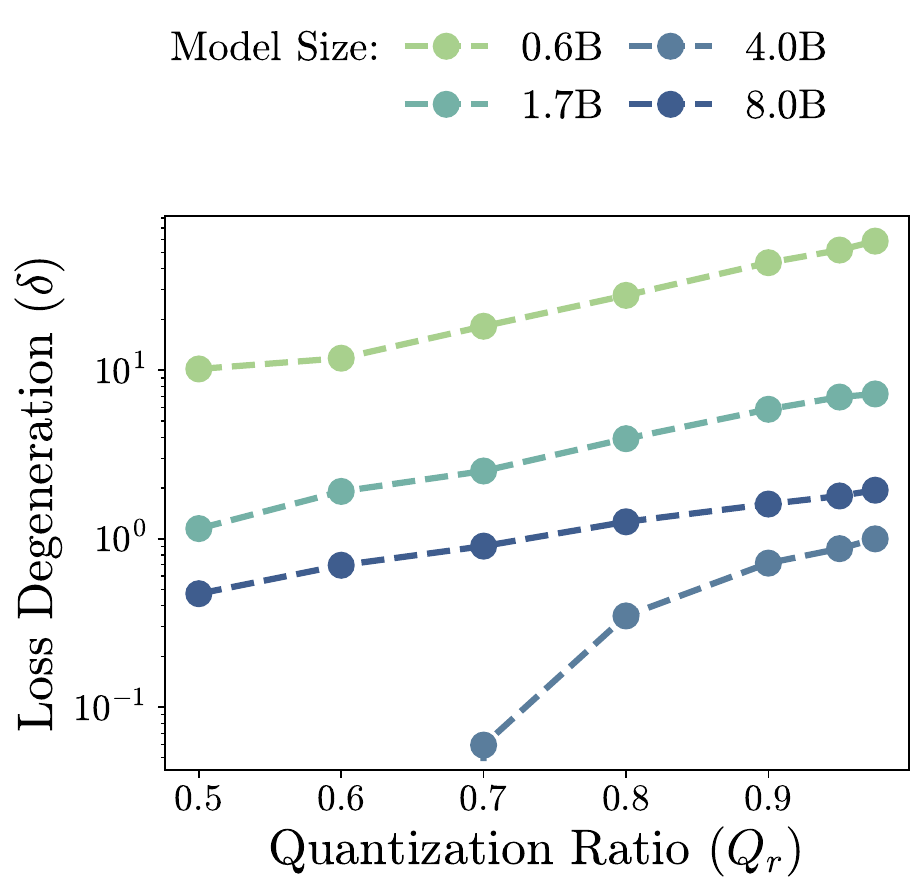}
    \captionsetup{font=scriptsize}
    \caption{Qwen3 Actual Loss}
\end{subfigure}

\begin{subfigure}[b]{0.3\textwidth} \centering
\includegraphics[width=\textwidth]{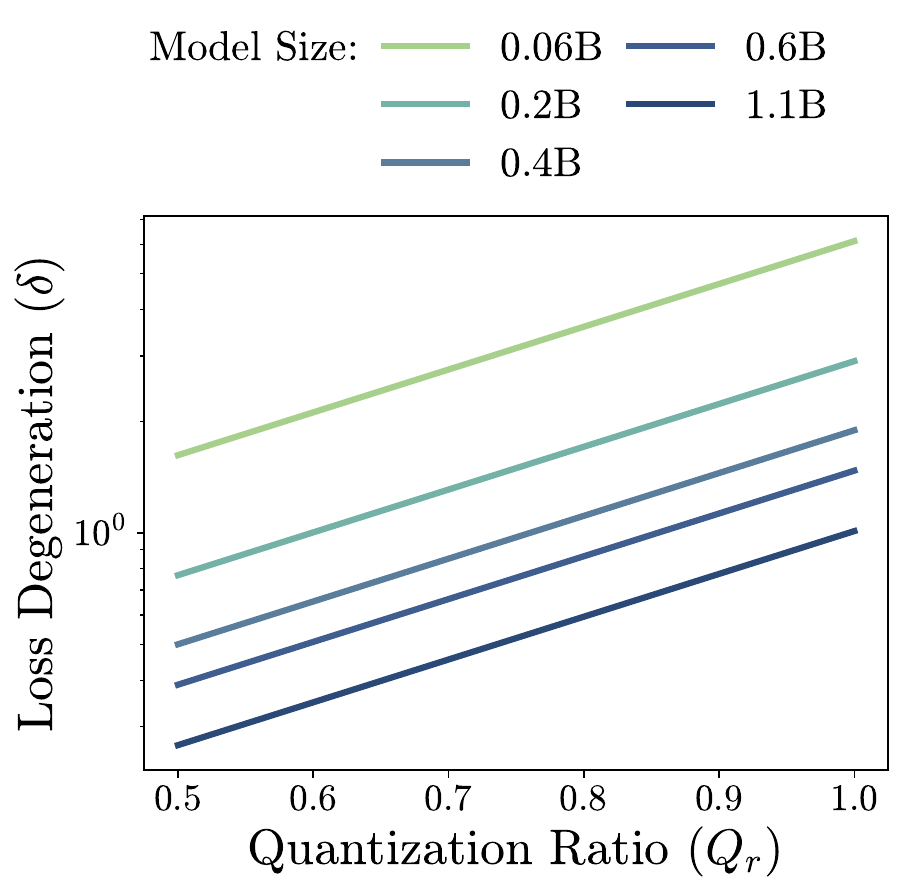}
    \captionsetup{font=scriptsize}
    \caption{CLM Predicted Loss}
    \end{subfigure}
\hfill
\begin{subfigure}[b]{0.3\textwidth} \centering
    \includegraphics[width=\textwidth]{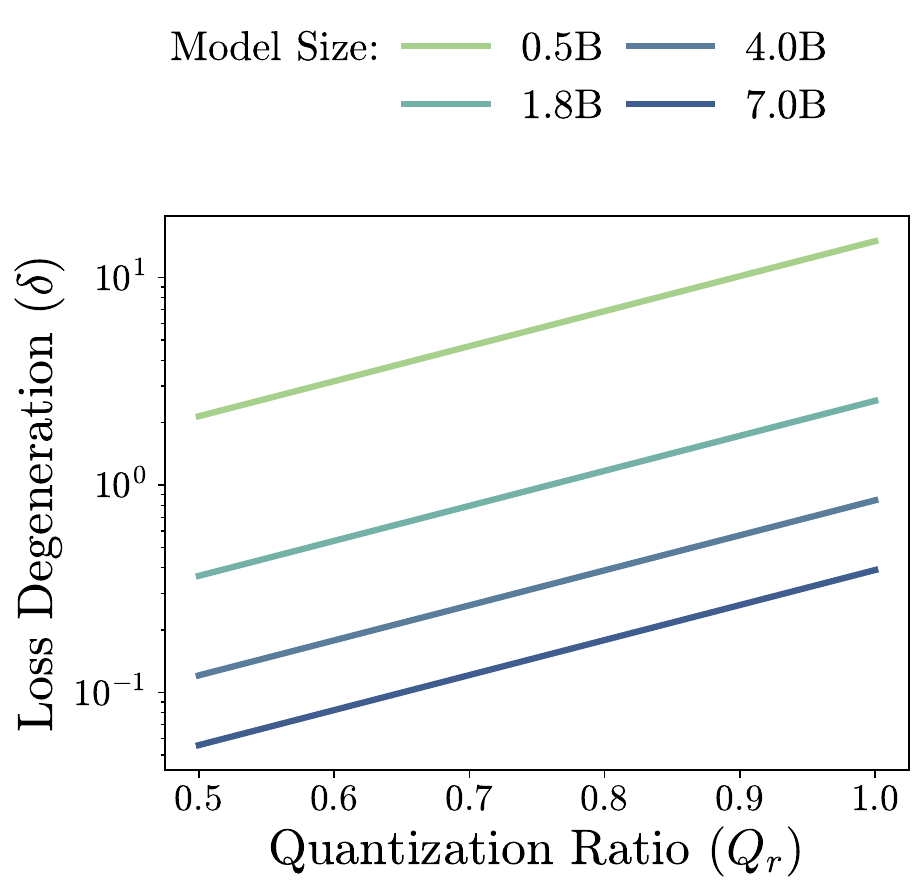}
    \captionsetup{font=scriptsize}
    \caption{Qwen1.5 Predicted Loss}
\end{subfigure}
\hfill
\begin{subfigure}[b]{0.3\textwidth} \centering
    \includegraphics[width=\textwidth]{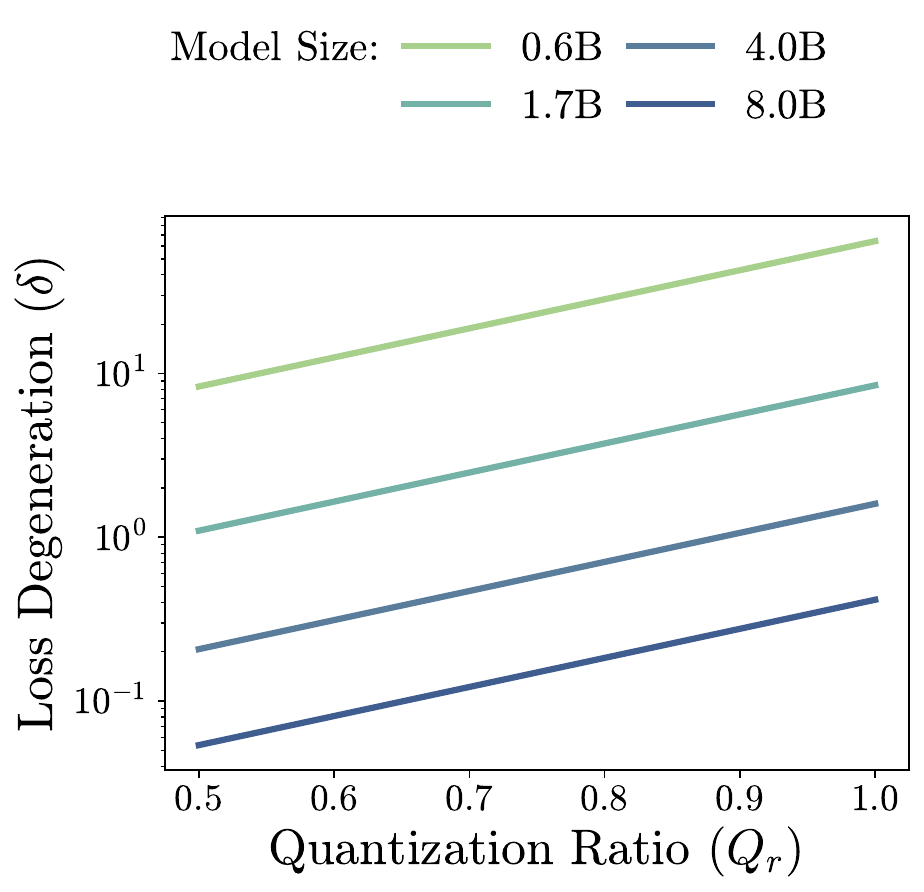}
    \captionsetup{font=scriptsize}
    \caption{Qwen3 Predicted Loss}
\end{subfigure}

\caption{\textbf{MXINT4 Weight-only ($\delta_\mu$)} (a,d,g,h) CLM MXINT4 Weight-only results; (b,e,h,k) Qwen-1.5 MXINT4 Weight-only results; (c,f,i,l) Qwen-3 MXINT4 Weight-only results.}
  \label{fig:appendix-w-only-mean}
\end{figure}

\clearpage
Figures~\ref{fig:appendix-w-only} and~\ref{fig:appendix-w-only-mean} display the loss contours for MXINT-4 weight-only quantization. This method shows moderate performance. It incurs a smaller degradation in loss than MXINT-2. The fitted contours once again align well with actual measurements, reinforcing the robustness of our scaling law across varying quantization granularities and levels of aggressiveness. Similar to the HQQ quantization, the effectiveness of MXINT-4 weight-only quantization and the distillation training process of Qwen-3 explains the outliers for Qwen-3 figures.

\section{Formula Fitting}
\label{app:formula-fitting}
\begin{figure}[H]
\centering
\begin{subfigure}[t]{0.22\textwidth}
\centering
\begin{tabular}{@{}l@{}}
\begin{minipage}[t][4cm][t]{\linewidth}
\centering
\begin{tabular}{ll}
\toprule
$C $ &  0.0028 \\
$A $ &  5.2055 \\
$\gamma_N $ &  0.7651 \\
$d $ &  13.6320 \\
$\gamma_c $ &  0.4741 \\
\bottomrule
\end{tabular}
\end{minipage}
\end{tabular}
\end{subfigure}
\hfill
\begin{subfigure}[t]{0.22\textwidth}
\centering
\includegraphics[width=\textwidth]{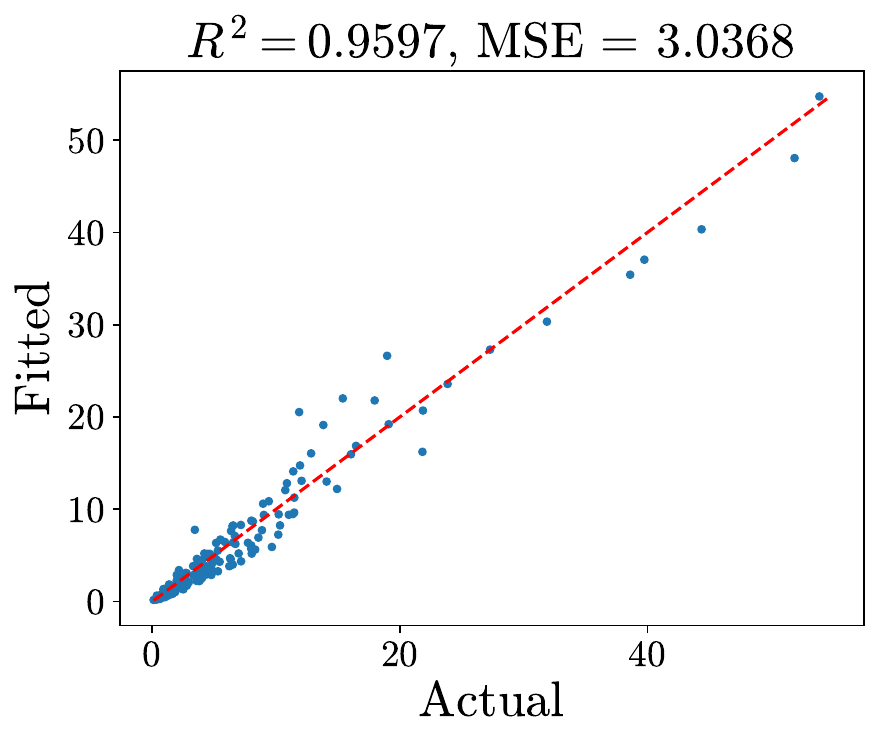}
\end{subfigure}
\hfill
\begin{subfigure}[t]{0.22\textwidth}
\centering
\begin{tabular}{@{}l@{}}
\begin{minipage}[t][4cm][t]{\linewidth}
\centering
\begin{tabular}{ll}
\toprule
$C $ &  0.0345 \\
$A $ &  3.2462 \\
$\gamma_N $ &  0.7150 \\
$d $ &  5.4028 \\
$\gamma_c $ &  0.4273 \\
\bottomrule
\end{tabular}
\end{minipage}
\end{tabular}
\end{subfigure}
\hfill
\begin{subfigure}[t]{0.22\textwidth}
\centering
\includegraphics[width=\textwidth]{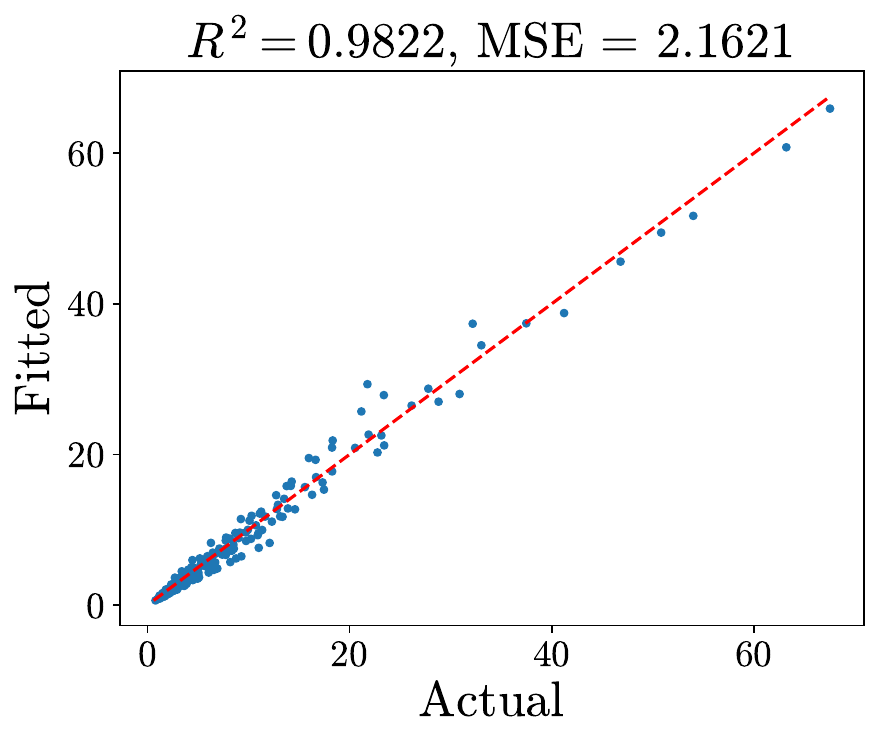}
\end{subfigure}
\vspace{-3.5em}
\caption{\textbf{CLM Strong Law.} The left two are for $\delta^{\text{opt}}$ and The right two are for $\delta_\mu$}.
  \label{fig:appendix-fitting-CLM Strong Law}
\end{figure}

\begin{figure}[H]
\centering
\begin{subfigure}[t]{0.22\textwidth}
\centering
\begin{tabular}{@{}l@{}}
\begin{minipage}[t][4cm][t]{\linewidth}
\centering
\begin{tabular}{ll}
\toprule
$C $ &  0.0001 \\
$A $ &  11.3016 \\
$\gamma_N $ &  1.2086 \\
$d $ &  -12.2589 \\
$\gamma_c $ &  0.5670 \\
\bottomrule
\end{tabular}
\end{minipage}
\end{tabular}
\end{subfigure}
\hfill
\begin{subfigure}[t]{0.22\textwidth}
\centering
\includegraphics[width=\textwidth]{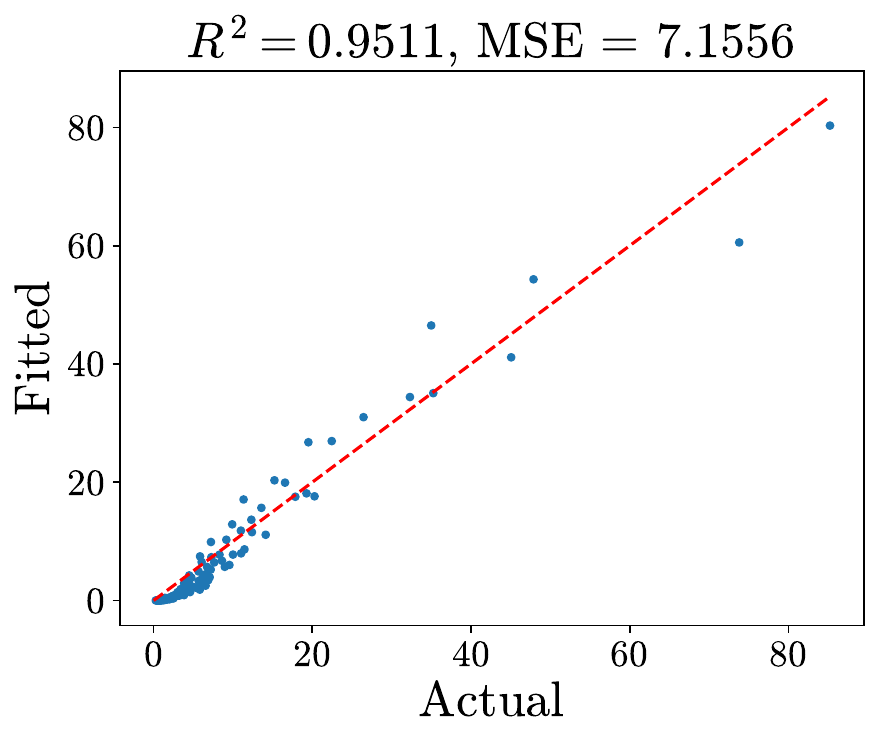}
\end{subfigure}
\hfill
\begin{subfigure}[t]{0.22\textwidth}
\centering
\begin{tabular}{@{}l@{}}
\begin{minipage}[t][4cm][t]{\linewidth}
\centering
\begin{tabular}{ll}
\toprule
$C $ &  0.0036 \\
$A $ &  6.5570 \\
$\gamma_N $ &  1.4627 \\
$d $ &  -5.4784 \\
$\gamma_c $ &  0.8237 \\
\bottomrule
\end{tabular}
\end{minipage}
\end{tabular}
\end{subfigure}
\hfill
\begin{subfigure}[t]{0.22\textwidth}
\centering
\includegraphics[width=\textwidth]{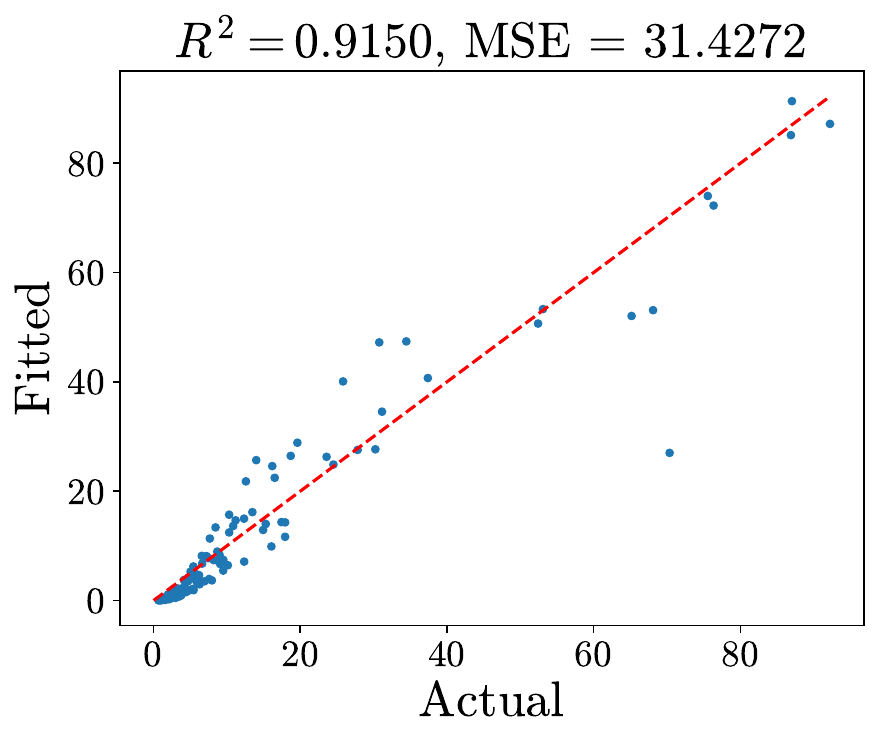}
\end{subfigure}
\vspace{-3.5em}
\caption{\textbf{Qwen-1.5 Strong Law.} The left two are for $\delta^{\text{opt}}$ and The right two are for $\delta_\mu$}.
  \label{fig:appendix-fitting-Qwen1.5 Strong Law}
\end{figure}

\begin{figure}[H]
\centering
\begin{subfigure}[t]{0.22\textwidth}
\centering
\begin{tabular}{@{}l@{}}
\begin{minipage}[t][4cm][t]{\linewidth}
\centering
\begin{tabular}{ll}
\toprule
$C $ &  0.0025 \\
$A $ &  8.1566 \\
$\gamma_N $ &  0.8181 \\
$d $ &  -13.1436 \\
$\gamma_c $ &  0.6208 \\
\bottomrule
\end{tabular}
\end{minipage}
\end{tabular}
\end{subfigure}
\hfill
\begin{subfigure}[t]{0.22\textwidth}
\centering
\includegraphics[width=\textwidth]{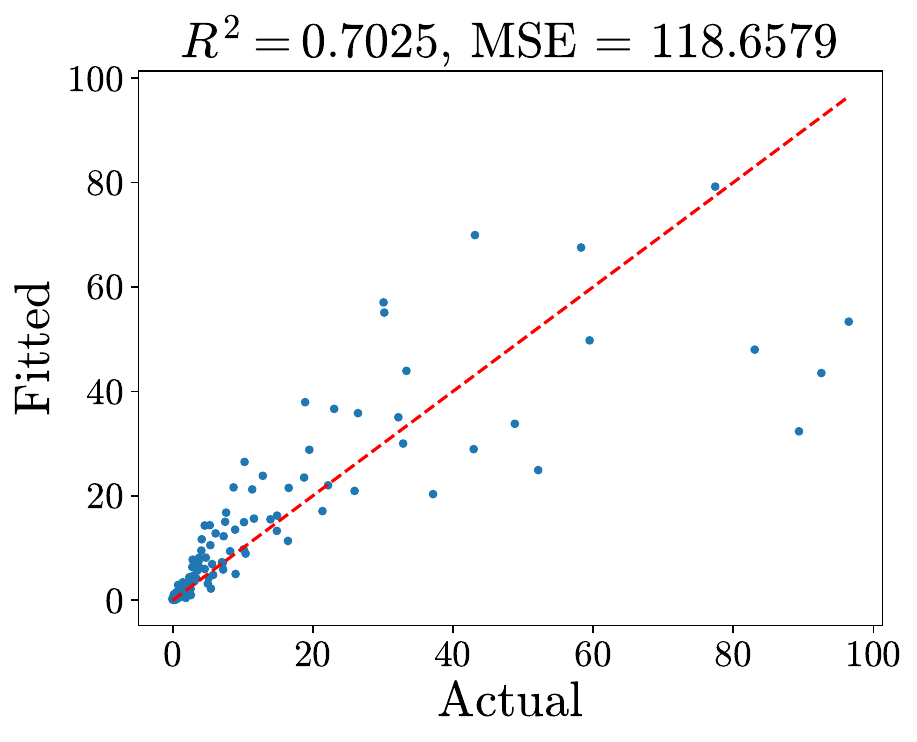}
\end{subfigure}
\hfill
\begin{subfigure}[t]{0.22\textwidth}
\centering
\begin{tabular}{@{}l@{}}
\begin{minipage}[t][4cm][t]{\linewidth}
\centering
\begin{tabular}{ll}
\toprule
$C $ &  0.0142 \\
$A $ &  4.1025 \\
$\gamma_N $ &  1.5909 \\
$d $ &  -3.7594 \\
$\gamma_c $ &  1.6736 \\
\bottomrule
\end{tabular}
\end{minipage}
\end{tabular}
\end{subfigure}
\hfill
\begin{subfigure}[t]{0.22\textwidth}
\centering
\includegraphics[width=\textwidth]{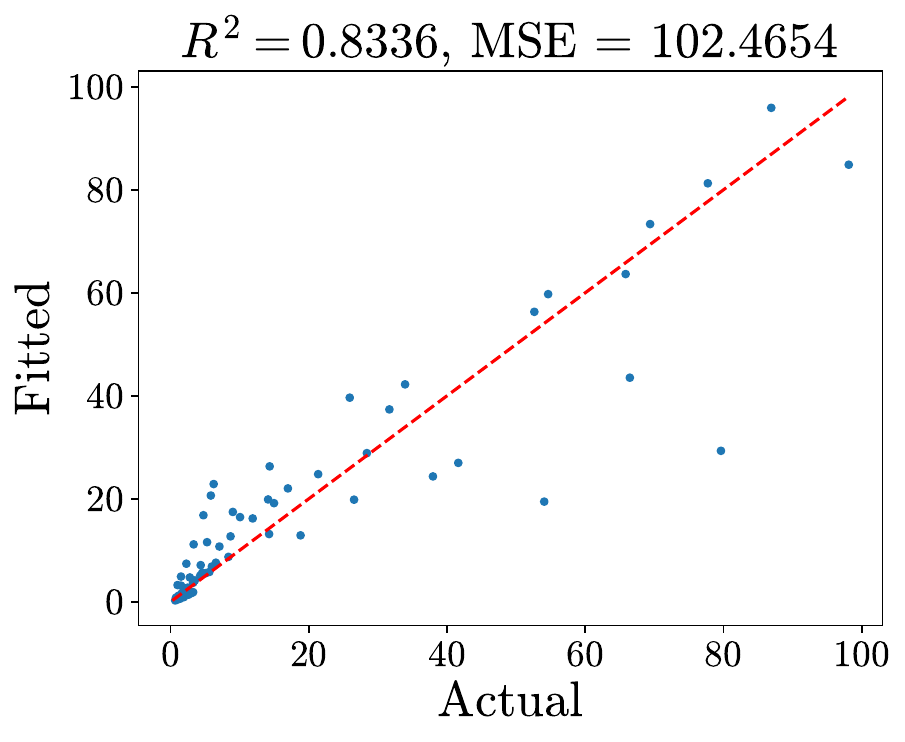}
\end{subfigure}
\vspace{-3.5em}
\caption{\textbf{Qwen-3 Strong Law.} The left two are for $\delta^{\text{opt}}$ and The right two are for $\delta_\mu$}.
  \label{fig:appendix-fitting-Qwen3 Strong Law}
\end{figure}

\begin{figure}[H]
\centering
\begin{subfigure}[t]{0.22\textwidth}
\centering
\begin{tabular}{@{}l@{}}
\begin{minipage}[t][4cm][t]{\linewidth}
\centering
\begin{tabular}{ll}
\toprule
$C $ &  0.0522 \\
$A $ &  2.1631 \\
$\gamma_N $ &  0.6471 \\
\bottomrule
\end{tabular}
\end{minipage}
\end{tabular}
\end{subfigure}
\hfill
\begin{subfigure}[t]{0.22\textwidth}
\centering
\includegraphics[width=\textwidth]{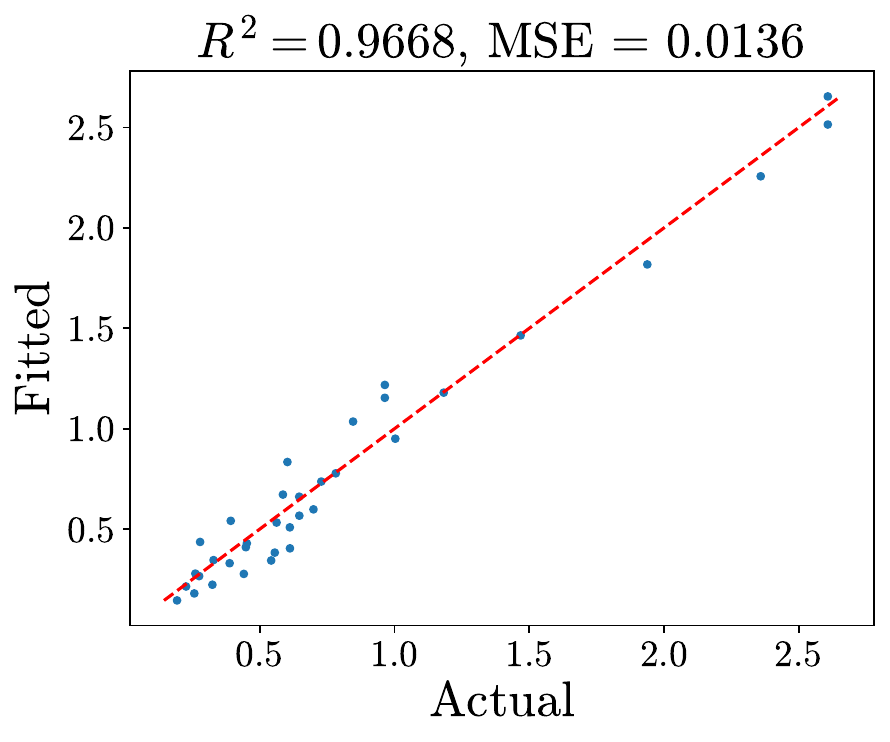}
\end{subfigure}
\hfill
\begin{subfigure}[t]{0.22\textwidth}
\centering
\begin{tabular}{@{}l@{}}
\begin{minipage}[t][4cm][t]{\linewidth}
\centering
\begin{tabular}{ll}
\toprule
$C $ &  0.1583 \\
$A $ &  1.3561 \\
$\gamma_N $ &  0.5508 \\
\bottomrule
\end{tabular}
\end{minipage}
\end{tabular}
\end{subfigure}
\hfill
\begin{subfigure}[t]{0.22\textwidth}
\centering
\includegraphics[width=\textwidth]{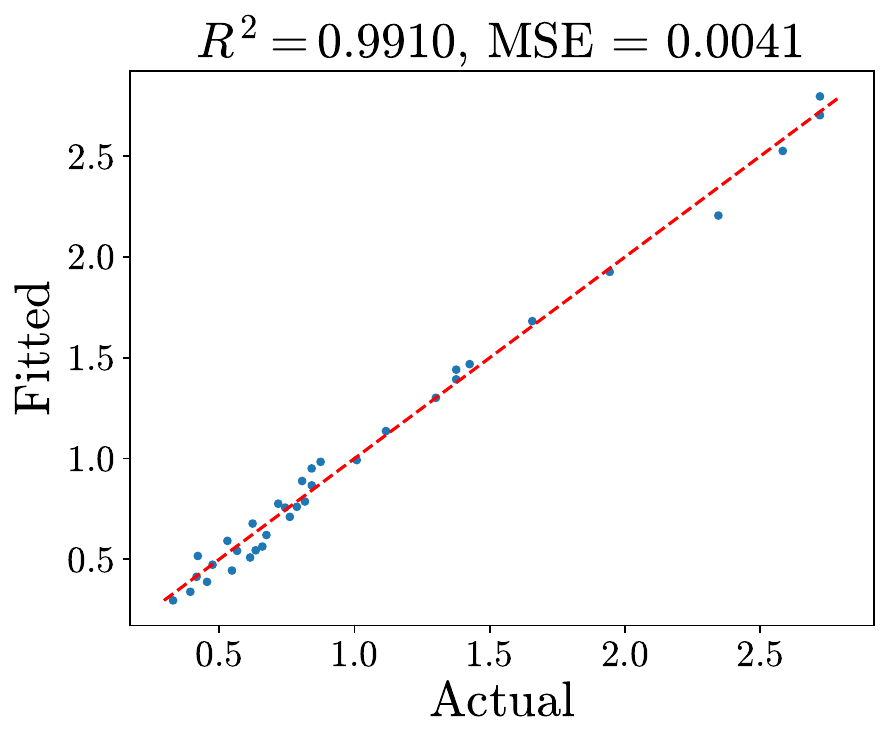}
\end{subfigure}
\vspace{-3.5em}
\caption{\textbf{HQQ CLM Weak Law.} The left two are for $\delta^{\text{opt}}$ and The right two are for $\delta_\mu$}.
  \label{fig:appendix-fitting-HQQ CLM Weak Law}
\end{figure}

\begin{figure}[H]
\centering
\begin{subfigure}[t]{0.22\textwidth}
\centering
\begin{tabular}{@{}l@{}}
\begin{minipage}[t][4cm][t]{\linewidth}
\centering
\begin{tabular}{ll}
\toprule
$C $ &  0.1140 \\
$A $ &  2.4336 \\
$\gamma_N $ &  0.5063 \\
\bottomrule
\end{tabular}
\end{minipage}
\end{tabular}
\end{subfigure}
\hfill
\begin{subfigure}[t]{0.22\textwidth}
\centering
\includegraphics[width=\textwidth]{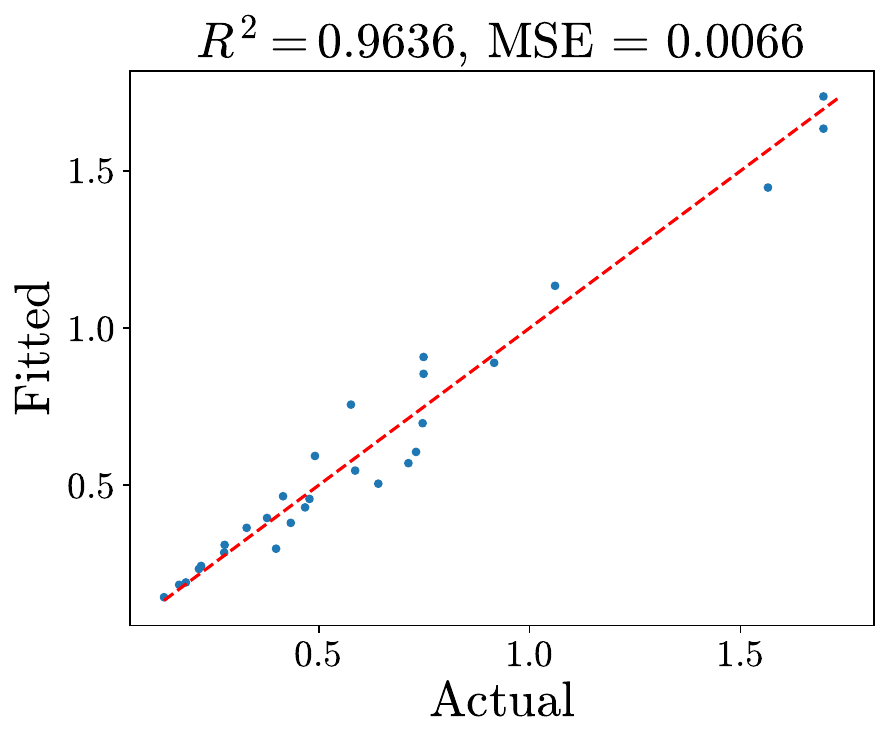}
\end{subfigure}
\hfill
\begin{subfigure}[t]{0.22\textwidth}
\centering
\begin{tabular}{@{}l@{}}
\begin{minipage}[t][4cm][t]{\linewidth}
\centering
\begin{tabular}{ll}
\toprule
$C $ &  0.3863 \\
$A $ &  1.4557 \\
$\gamma_N $ &  0.4341 \\
\bottomrule
\end{tabular}
\end{minipage}
\end{tabular}
\end{subfigure}
\hfill
\begin{subfigure}[t]{0.22\textwidth}
\centering
\includegraphics[width=\textwidth]{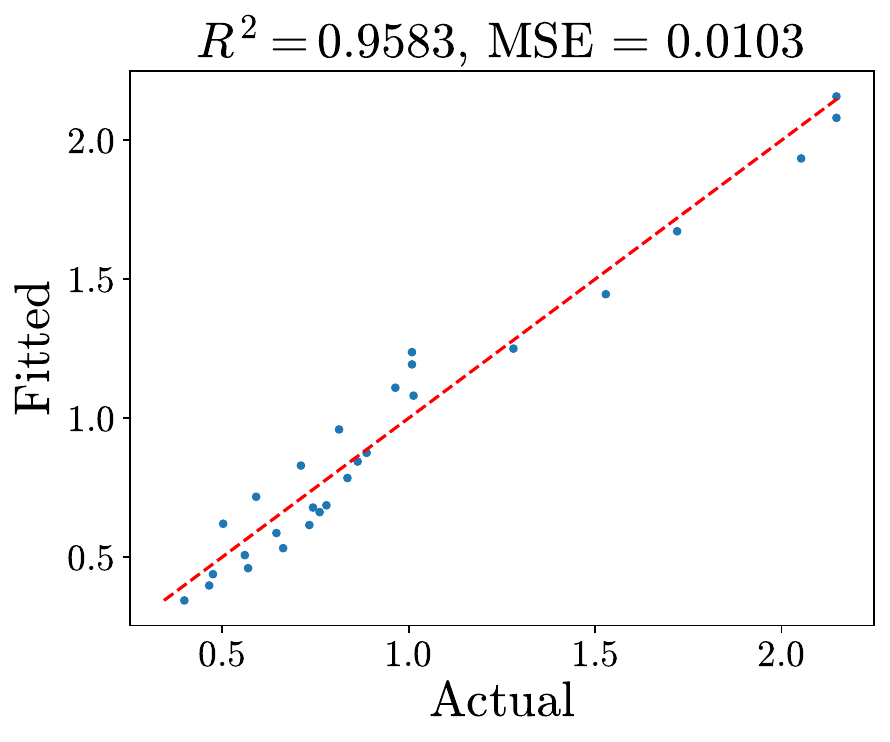}
\end{subfigure}
\vspace{-3.5em}
\caption{\textbf{HQQ Qwen-1.5 Weak Law.} The left two are for $\delta^{\text{opt}}$ and The right two are for $\delta_\mu$}.
  \label{fig:appendix-fitting-HQQ Qwen1.5 Weak Law}
\end{figure}

\begin{figure}[H]
\centering
\begin{subfigure}[t]{0.22\textwidth}
\centering
\begin{tabular}{@{}l@{}}
\begin{minipage}[t][4cm][t]{\linewidth}
\centering
\begin{tabular}{ll}
\toprule
$C $ &  0.1480 \\
$A $ &  2.7216 \\
$\gamma_N $ &  0.8841 \\
\bottomrule
\end{tabular}
\end{minipage}
\end{tabular}
\end{subfigure}
\hfill
\begin{subfigure}[t]{0.22\textwidth}
\centering
\includegraphics[width=\textwidth]{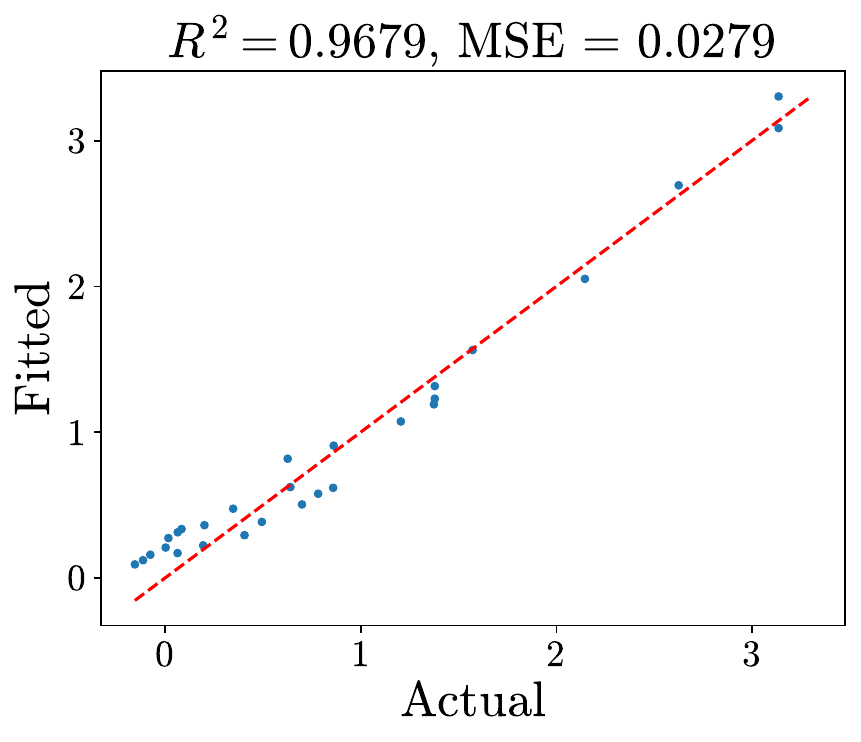}
\end{subfigure}
\hfill
\begin{subfigure}[t]{0.22\textwidth}
\centering
\begin{tabular}{@{}l@{}}
\begin{minipage}[t][4cm][t]{\linewidth}
\centering
\begin{tabular}{ll}
\toprule
$C $ &  0.6476 \\
$A $ &  1.5246 \\
$\gamma_N $ &  0.6990 \\
\bottomrule
\end{tabular}
\end{minipage}
\end{tabular}
\end{subfigure}
\hfill
\begin{subfigure}[t]{0.22\textwidth}
\centering
\includegraphics[width=\textwidth]{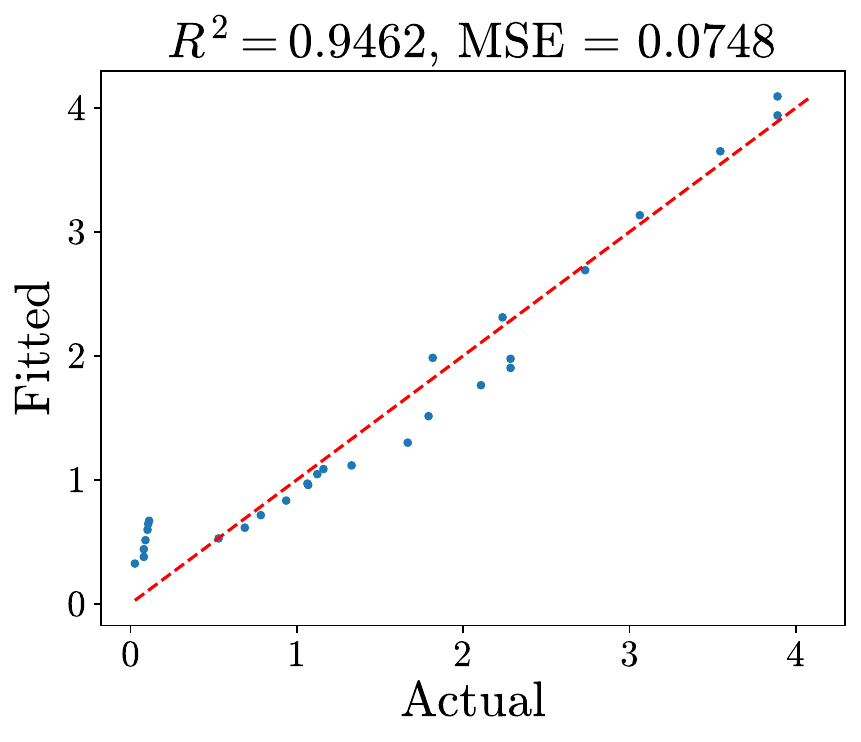}
\end{subfigure}
\vspace{-3.5em}
\caption{\textbf{HQQ Qwen-3 Weak Law.} The left two are for $\delta^{\text{opt}}$ and The right two are for $\delta_\mu$}.
  \label{fig:appendix-fitting-HQQ Qwen3 Weak Law}
\end{figure}

\begin{figure}[H]
\centering
\begin{subfigure}[t]{0.22\textwidth}
\centering
\begin{tabular}{@{}l@{}}
\begin{minipage}[t][4cm][t]{\linewidth}
\centering
\begin{tabular}{ll}
\toprule
$C $ &  0.1407 \\
$A $ &  9.1365 \\
$\gamma_N $ &  0.0992 \\
\bottomrule
\end{tabular}
\end{minipage}
\end{tabular}
\end{subfigure}
\hfill
\begin{subfigure}[t]{0.22\textwidth}
\centering
\includegraphics[width=\textwidth]{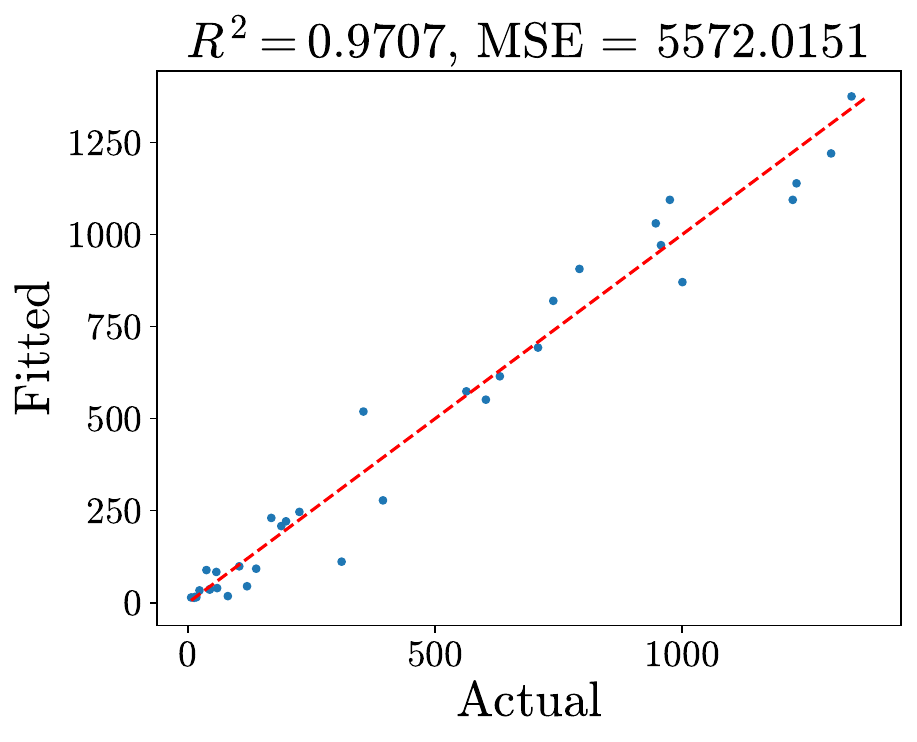}
\end{subfigure}
\hfill
\begin{subfigure}[t]{0.22\textwidth}
\centering
\begin{tabular}{@{}l@{}}
\begin{minipage}[t][4cm][t]{\linewidth}
\centering
\begin{tabular}{ll}
\toprule
$C $ &  6.1421 \\
$A $ &  5.7318 \\
$\gamma_N $ &  0.0821 \\
\bottomrule
\end{tabular}
\end{minipage}
\end{tabular}
\end{subfigure}
\hfill
\begin{subfigure}[t]{0.22\textwidth}
\centering
\includegraphics[width=\textwidth]{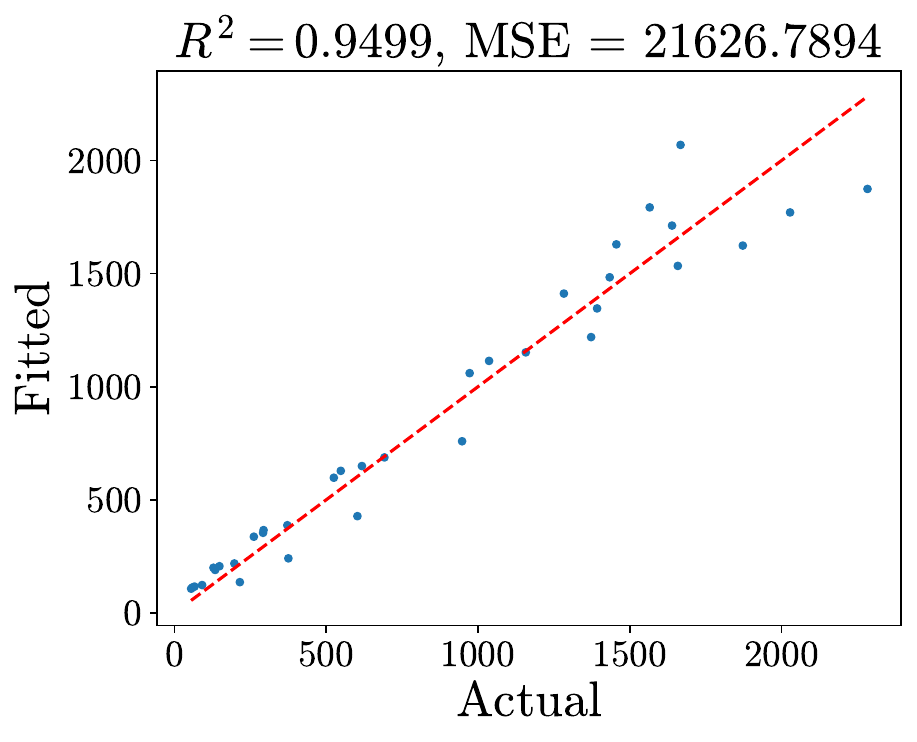}
\end{subfigure}
\vspace{-3.5em}
\caption{\textbf{MXINT-2 CLM Weak Law.} The left two are for $\delta^{\text{opt}}$ and The right two are for $\delta_\mu$}.
  \label{fig:appendix-fitting-MXINT2 CLM Weak Law}
\end{figure}

\begin{figure}[H]
\centering
\begin{subfigure}[t]{0.22\textwidth}
\centering
\begin{tabular}{@{}l@{}}
\begin{minipage}[t][4cm][t]{\linewidth}
\centering
\begin{tabular}{ll}
\toprule
$C $ &  0.0001 \\
$A $ &  21.7819 \\
$\gamma_N $ &  0.4173 \\
\bottomrule
\end{tabular}
\end{minipage}
\end{tabular}
\end{subfigure}
\hfill
\begin{subfigure}[t]{0.22\textwidth}
\centering
\includegraphics[width=\textwidth]{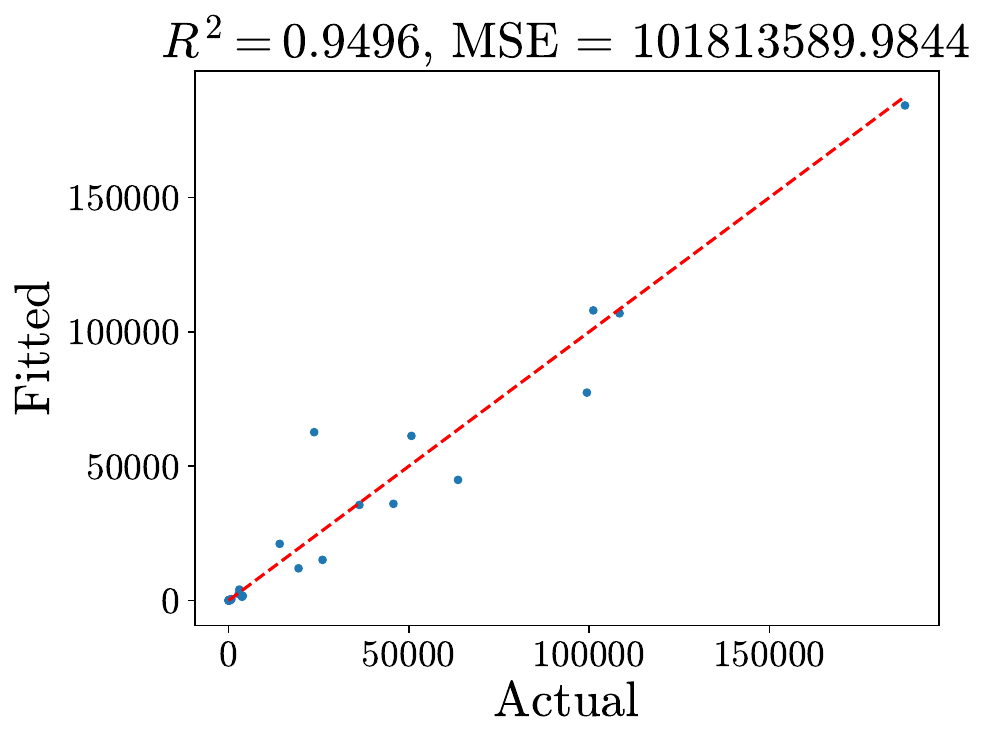}
\end{subfigure}
\hfill
\begin{subfigure}[t]{0.22\textwidth}
\centering
\begin{tabular}{@{}l@{}}
\begin{minipage}[t][4cm][t]{\linewidth}
\centering
\begin{tabular}{ll}
\toprule
$C $ &  1181.1395 \\
$A $ &  5.7997 \\
$\gamma_N $ &  0.3834 \\
\bottomrule
\end{tabular}
\end{minipage}
\end{tabular}
\end{subfigure}
\hfill
\begin{subfigure}[t]{0.22\textwidth}
\centering
\includegraphics[width=\textwidth]{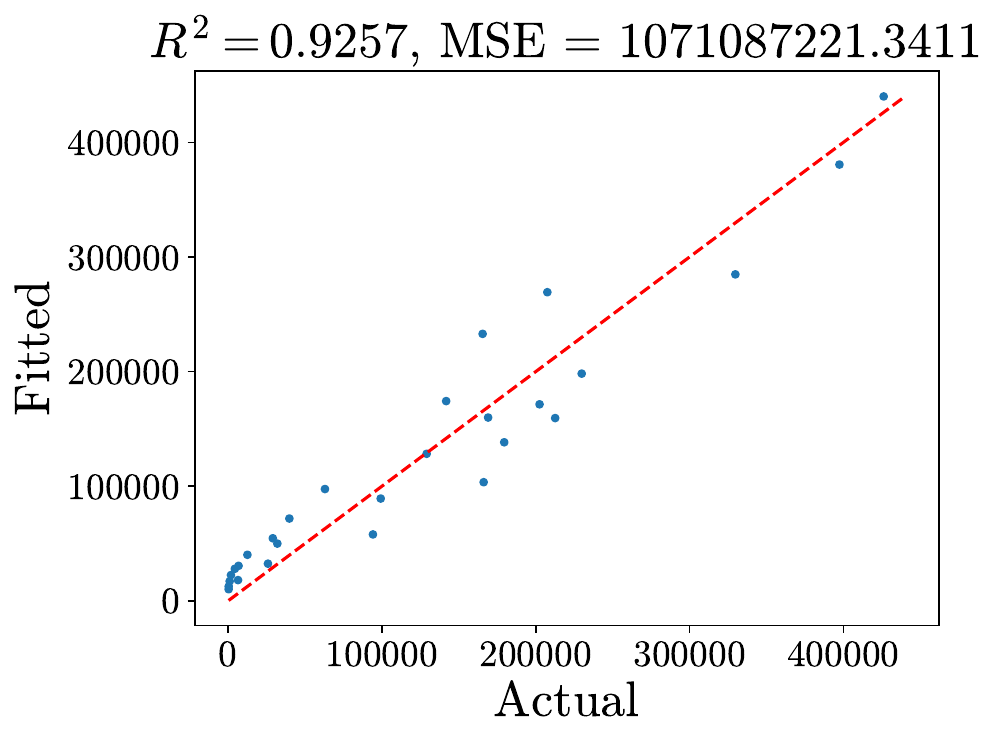}
\end{subfigure}
\vspace{-3.5em}
\caption{\textbf{MXINT-2 Qwen-1.5 Weak Law.} The left two are for $\delta^{\text{opt}}$ and The right two are for $\delta_\mu$}.
  \label{fig:appendix-fitting-MXINT2 Qwen1.5 Weak Law}
\end{figure}

\begin{figure}[H]
\centering
\begin{subfigure}[t]{0.22\textwidth}
\centering
\begin{tabular}{@{}l@{}}
\begin{minipage}[t][4cm][t]{\linewidth}
\centering
\begin{tabular}{ll}
\toprule
$C $ &  0.0000 \\
$A $ &  29.2508 \\
$\gamma_N $ &  0.5165 \\
\bottomrule
\end{tabular}
\end{minipage}
\end{tabular}
\end{subfigure}
\hfill
\begin{subfigure}[t]{0.22\textwidth}
\centering
\includegraphics[width=\textwidth]{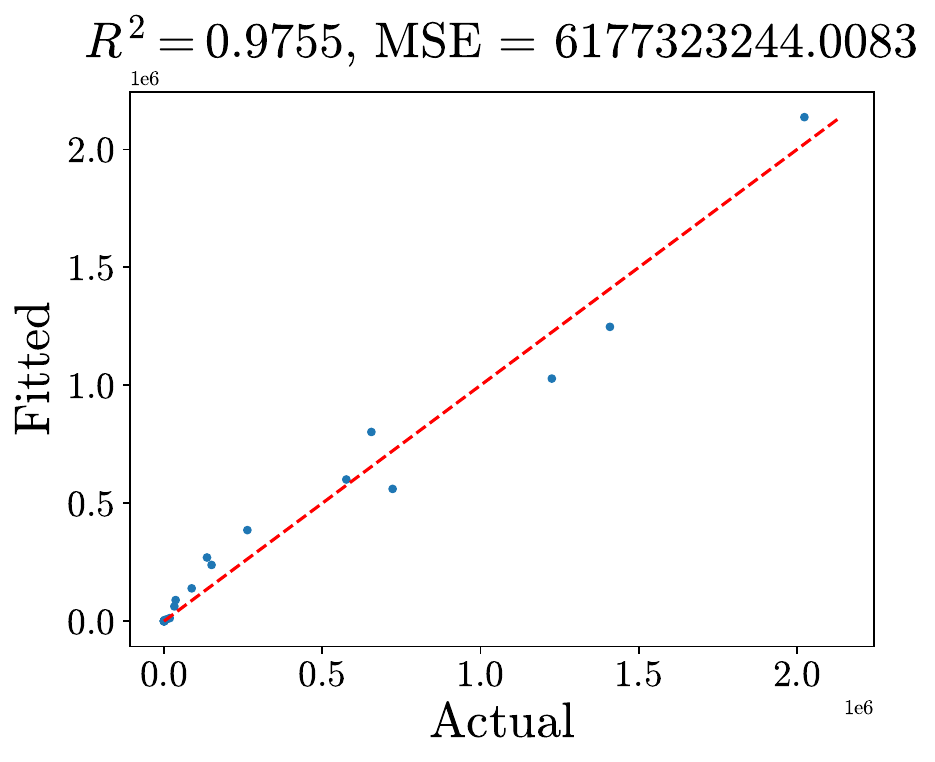}
\end{subfigure}
\hfill
\begin{subfigure}[t]{0.22\textwidth}
\centering
\begin{tabular}{@{}l@{}}
\begin{minipage}[t][4cm][t]{\linewidth}
\centering
\begin{tabular}{ll}
\toprule
$C $ &  0.0000 \\
$A $ &  21.2625 \\
$\gamma_N $ &  -4.6183 \\
\bottomrule
\end{tabular}
\end{minipage}
\end{tabular}
\end{subfigure}
\hfill
\begin{subfigure}[t]{0.22\textwidth}
\centering
\includegraphics[width=\textwidth]{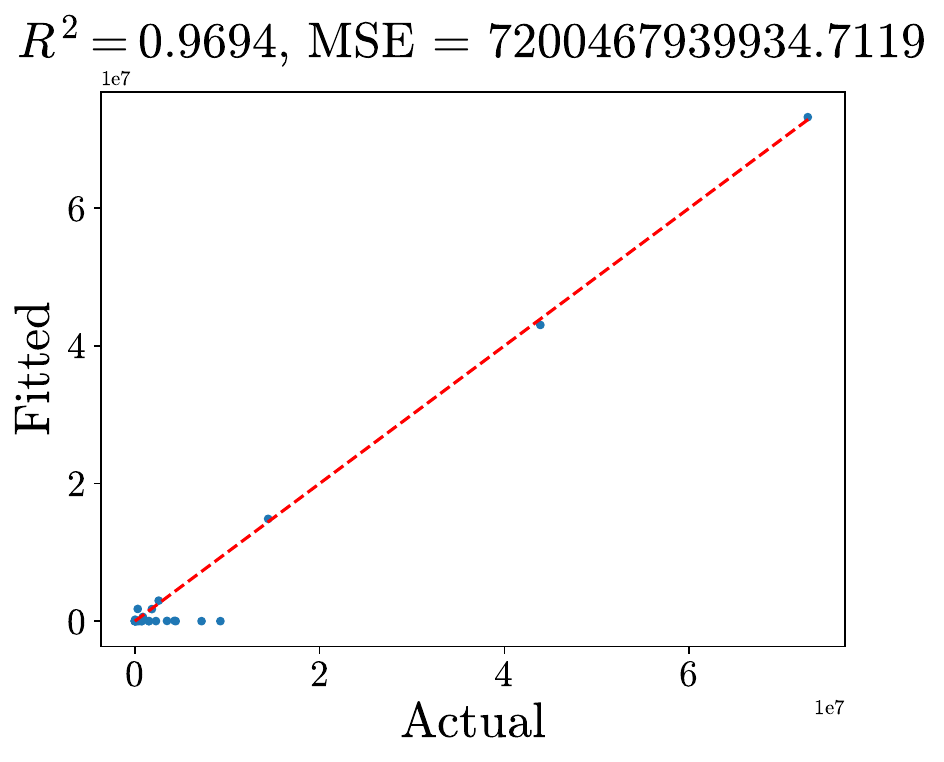}
\end{subfigure}
\vspace{-3.5em}
\caption{\textbf{MXINT-2 Qwen-3 Weak Law.} The left two are for $\delta^{\text{opt}}$ and The right two are for $\delta_\mu$}.
  \label{fig:appendix-fitting-MXINT2 Qwen3 Weak Law}
\end{figure}

\begin{figure}[H]
\centering
\begin{subfigure}[t]{0.22\textwidth}
\centering
\begin{tabular}{@{}l@{}}
\begin{minipage}[t][4cm][t]{\linewidth}
\centering
\begin{tabular}{ll}
\toprule
$C $ &  0.2187 \\
$A $ &  2.2312 \\
$\gamma_N $ &  0.8405 \\
\bottomrule
\end{tabular}
\end{minipage}
\end{tabular}
\end{subfigure}
\hfill
\begin{subfigure}[t]{0.22\textwidth}
\centering
\includegraphics[width=\textwidth]{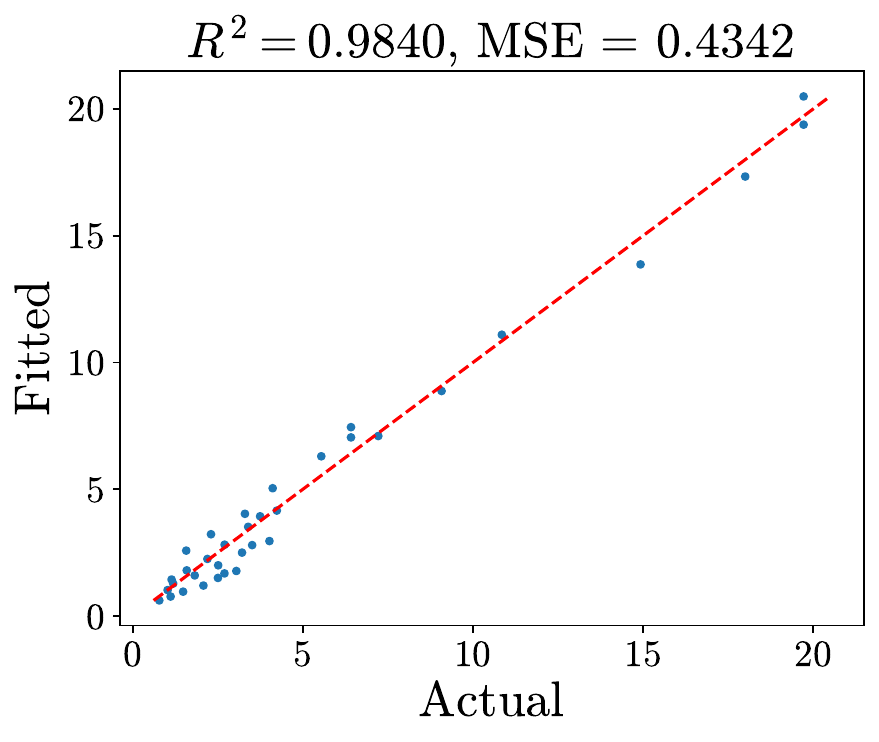}
\end{subfigure}
\hfill
\begin{subfigure}[t]{0.22\textwidth}
\centering
\begin{tabular}{@{}l@{}}
\begin{minipage}[t][4cm][t]{\linewidth}
\centering
\begin{tabular}{ll}
\toprule
$C $ &  0.6682 \\
$A $ &  1.4650 \\
$\gamma_N $ &  0.7259 \\
\bottomrule
\end{tabular}
\end{minipage}
\end{tabular}
\end{subfigure}
\hfill
\begin{subfigure}[t]{0.22\textwidth}
\centering
\includegraphics[width=\textwidth]{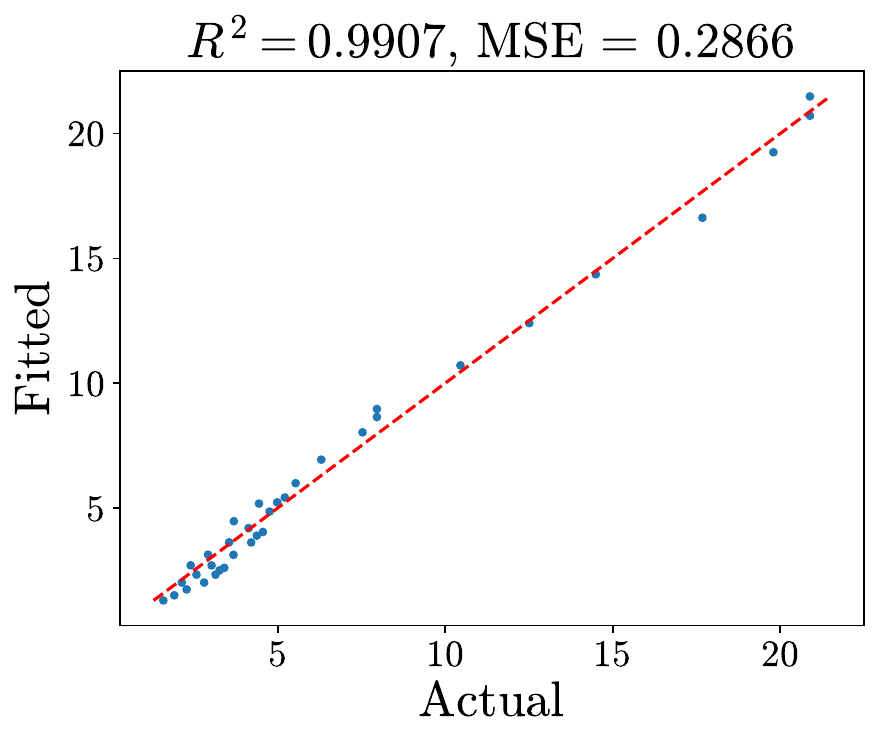}
\end{subfigure}
\vspace{-3.5em}
\caption{\textbf{MXINT-4 Layerwise CLM Weak Law. }The left two are for $\delta^{\text{opt}}$ and The right two are for $\delta_\mu$}.
  \label{fig:appendix-fitting-MXINT4 Layerwise CLM Weak Law}
\end{figure}

\begin{figure}[H]
\centering
\begin{subfigure}[t]{0.22\textwidth}
\centering
\begin{tabular}{@{}l@{}}
\begin{minipage}[t][4cm][t]{\linewidth}
\centering
\begin{tabular}{ll}
\toprule
$C $ &  0.0313 \\
$A $ &  4.2991 \\
$\gamma_N $ &  0.7828 \\
\bottomrule
\end{tabular}
\end{minipage}
\end{tabular}
\end{subfigure}
\hfill
\begin{subfigure}[t]{0.22\textwidth}
\centering
\includegraphics[width=\textwidth]{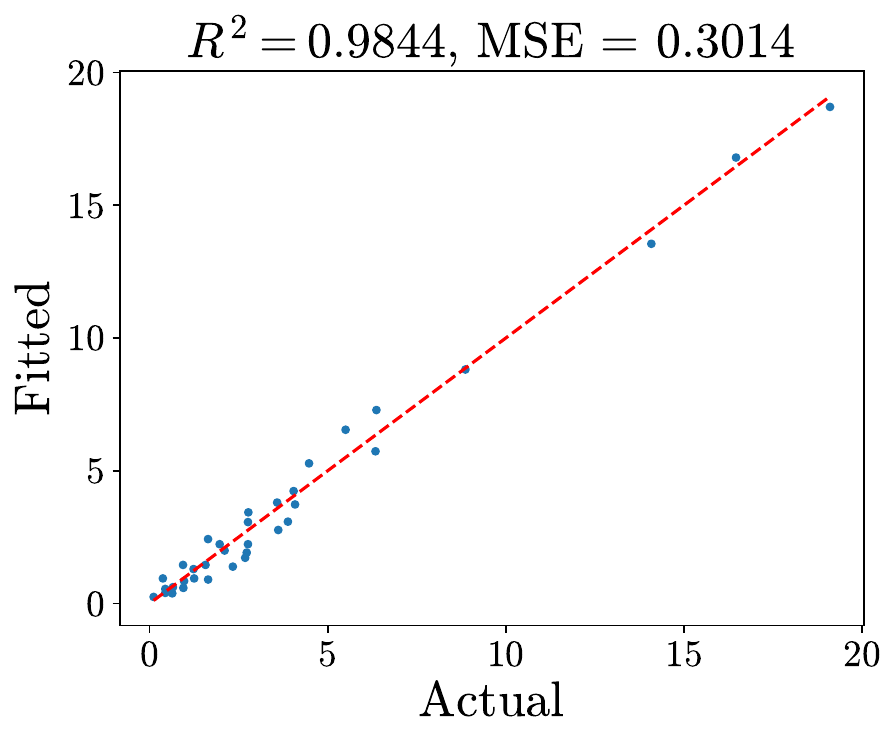}
\end{subfigure}
\hfill
\begin{subfigure}[t]{0.22\textwidth}
\centering
\begin{tabular}{@{}l@{}}
\begin{minipage}[t][4cm][t]{\linewidth}
\centering
\begin{tabular}{ll}
\toprule
$C $ &  0.1862 \\
$A $ &  2.8693 \\
$\gamma_N $ &  0.7030 \\
\bottomrule
\end{tabular}
\end{minipage}
\end{tabular}
\end{subfigure}
\hfill
\begin{subfigure}[t]{0.22\textwidth}
\centering
\includegraphics[width=\textwidth]{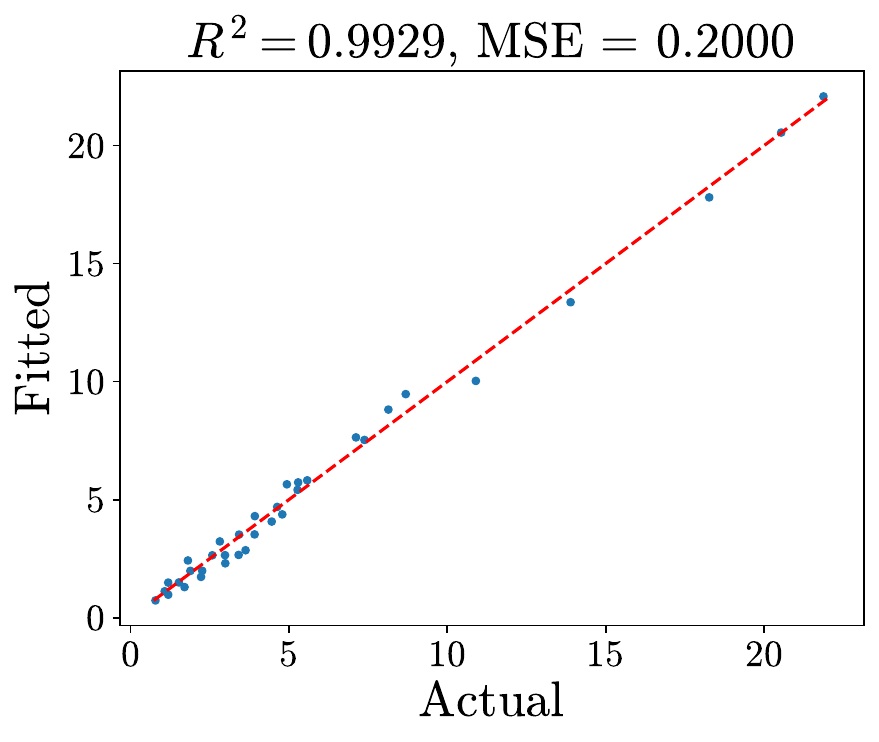}
\end{subfigure}
\vspace{-3.5em}
\caption{\textbf{MXINT-4 Matrix Multiplication-wise CLM Weak Law.} The left two are for $\delta^{\text{opt}}$ and The right two are for $\delta_\mu$}.
  \label{fig:appendix-fitting-MXINT4 Matrix Multiplication-wise CLM Weak Law}
\end{figure}

\begin{figure}[H]
\centering
\begin{subfigure}[t]{0.22\textwidth}
\centering
\begin{tabular}{@{}l@{}}
\begin{minipage}[t][4cm][t]{\linewidth}
\centering
\begin{tabular}{ll}
\toprule
$C $ &  0.5221 \\
$A $ &  2.7972 \\
$\gamma_N $ &  0.7188 \\
\bottomrule
\end{tabular}
\end{minipage}
\end{tabular}
\end{subfigure}
\hfill
\begin{subfigure}[t]{0.22\textwidth}
\centering
\includegraphics[width=\textwidth]{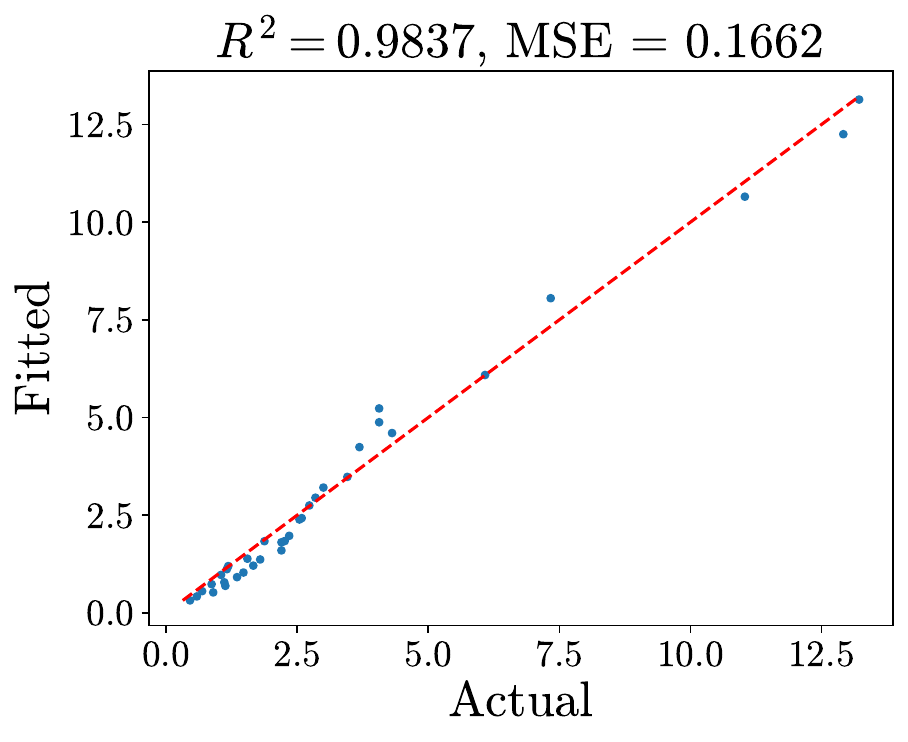}
\end{subfigure}
\hfill
\begin{subfigure}[t]{0.22\textwidth}
\centering
\begin{tabular}{@{}l@{}}
\begin{minipage}[t][4cm][t]{\linewidth}
\centering
\begin{tabular}{ll}
\toprule
$C $ &  1.0190 \\
$A $ &  2.2913 \\
$\gamma_N $ &  0.7626 \\
\bottomrule
\end{tabular}
\end{minipage}
\end{tabular}
\end{subfigure}
\hfill
\begin{subfigure}[t]{0.22\textwidth}
\centering
\includegraphics[width=\textwidth]{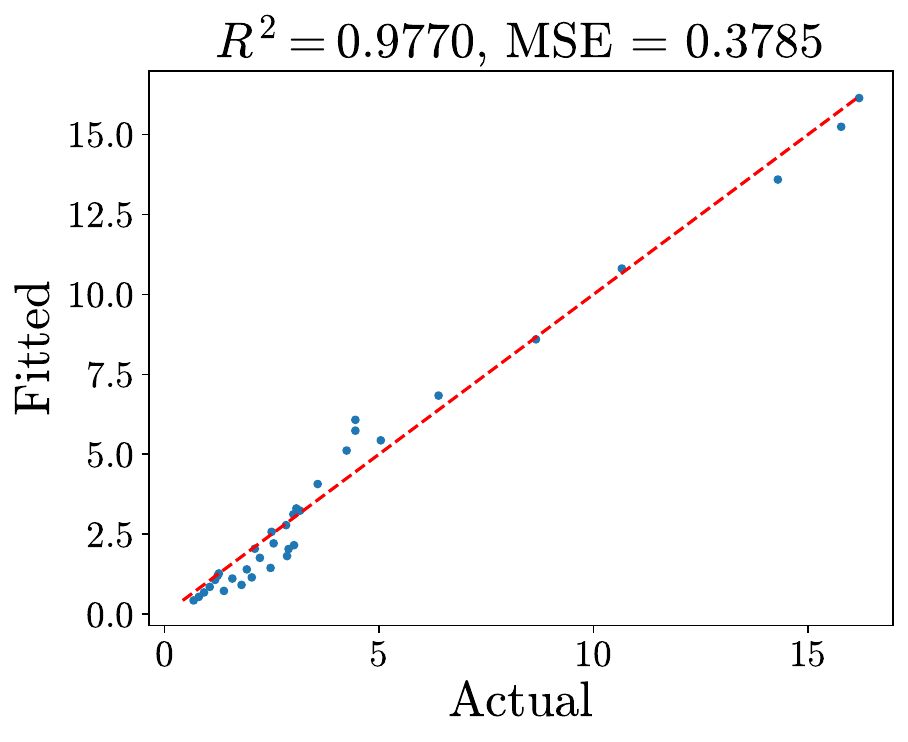}
\end{subfigure}
\vspace{-3.5em}
\caption{\textbf{MXINT-4 Layerwise Qwen-1.5 Weak Law.} The left two are for $\delta^{\text{opt}}$ and The right two are for $\delta_\mu$}.
  \label{fig:appendix-fitting-MXINT4 Layerwise Qwen1.5 Weak Law}
\end{figure}

\begin{figure}[H]
\centering
\begin{subfigure}[t]{0.22\textwidth}
\centering
\begin{tabular}{@{}l@{}}
\begin{minipage}[t][4cm][t]{\linewidth}
\centering
\begin{tabular}{ll}
\toprule
$C $ &  0.0286 \\
$A $ &  5.9666 \\
$\gamma_N $ &  0.8073 \\
\bottomrule
\end{tabular}
\end{minipage}
\end{tabular}
\end{subfigure}
\hfill
\begin{subfigure}[t]{0.22\textwidth}
\centering
\includegraphics[width=\textwidth]{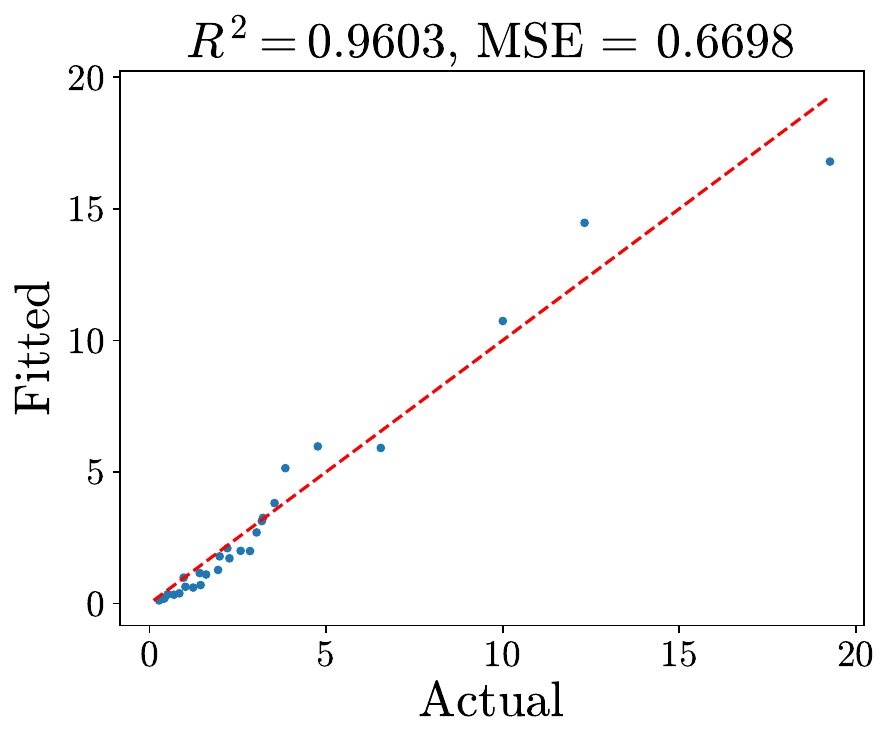}
\end{subfigure}
\hfill
\begin{subfigure}[t]{0.22\textwidth}
\centering
\begin{tabular}{@{}l@{}}
\begin{minipage}[t][4cm][t]{\linewidth}
\centering
\begin{tabular}{ll}
\toprule
$C $ &  0.1280 \\
$A $ &  4.9367 \\
$\gamma_N $ &  1.1931 \\
\bottomrule
\end{tabular}
\end{minipage}
\end{tabular}
\end{subfigure}
\hfill
\begin{subfigure}[t]{0.22\textwidth}
\centering
\includegraphics[width=\textwidth]{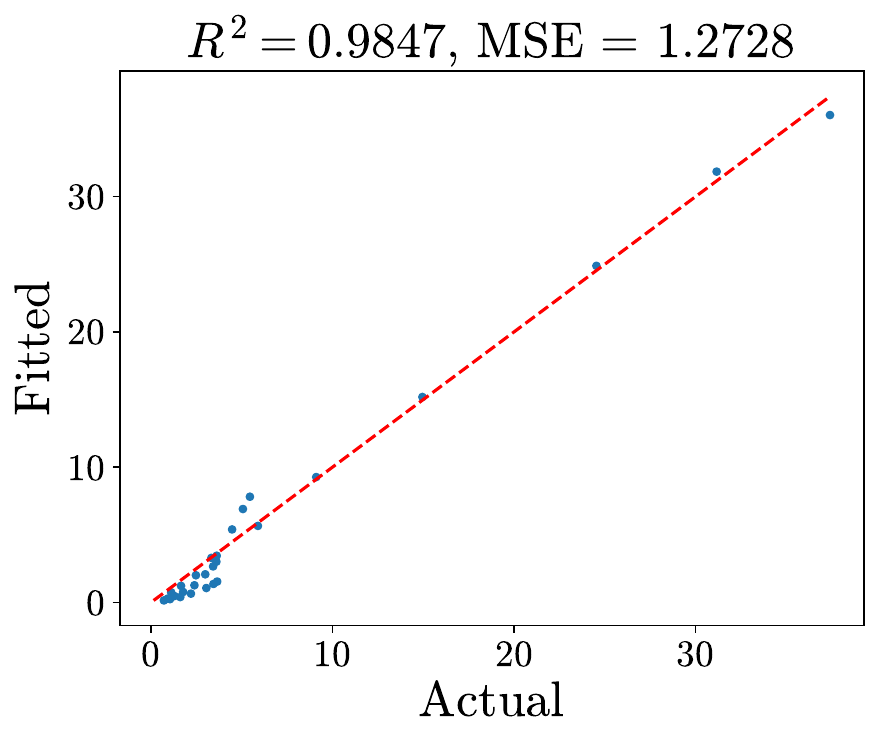}
\end{subfigure}
\vspace{-3.5em}
\caption{\textbf{MXINT-4 Matrix Multiplication-wise Qwen-1.5 Weak Law.} The left two are for $\delta^{\text{opt}}$ and The right two are for $\delta_\mu$}.
  \label{fig:appendix-fitting-MXINT4 Matrix Multiplication-wise Qwen1.5 Weak Law}
\end{figure}

\begin{figure}[H]
\centering
\begin{subfigure}[t]{0.22\textwidth}
\centering
\begin{tabular}{@{}l@{}}
\begin{minipage}[t][4cm][t]{\linewidth}
\centering
\begin{tabular}{ll}
\toprule
$C $ &  0.8209 \\
$A $ &  2.3712 \\
$\gamma_N $ &  0.9880 \\
\bottomrule
\end{tabular}
\end{minipage}
\end{tabular}
\end{subfigure}
\hfill
\begin{subfigure}[t]{0.22\textwidth}
\centering
\includegraphics[width=\textwidth]{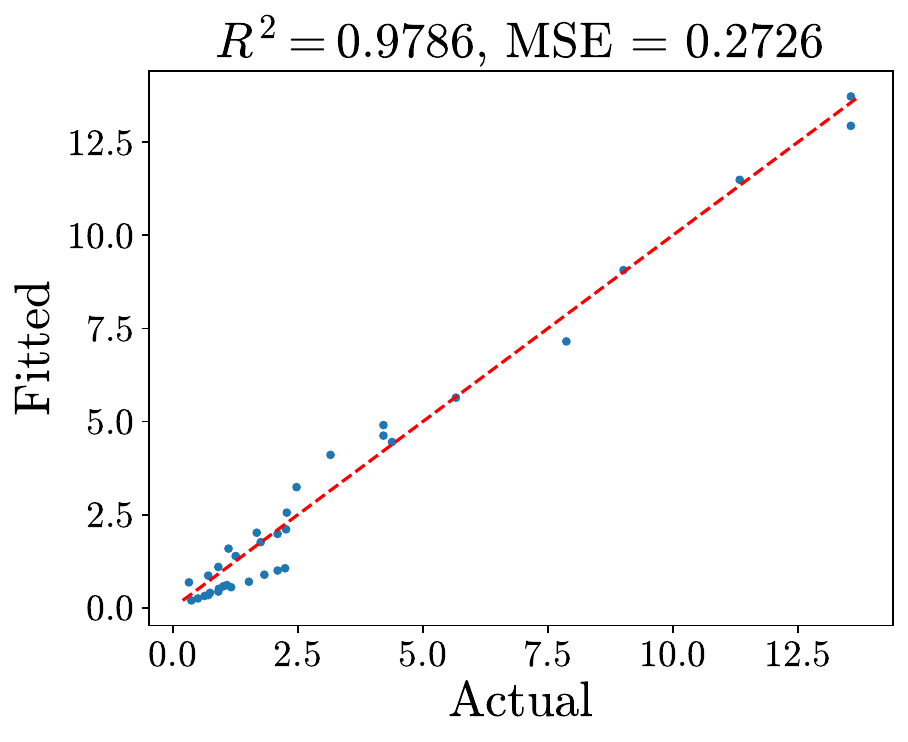}
\end{subfigure}
\hfill
\begin{subfigure}[t]{0.22\textwidth}
\centering
\begin{tabular}{@{}l@{}}
\begin{minipage}[t][4cm][t]{\linewidth}
\centering
\begin{tabular}{ll}
\toprule
$C $ &  1.5107 \\
$A $ &  1.9349 \\
$\gamma_N $ &  0.9487 \\
\bottomrule
\end{tabular}
\end{minipage}
\end{tabular}
\end{subfigure}
\hfill
\begin{subfigure}[t]{0.22\textwidth}
\centering
\includegraphics[width=\textwidth]{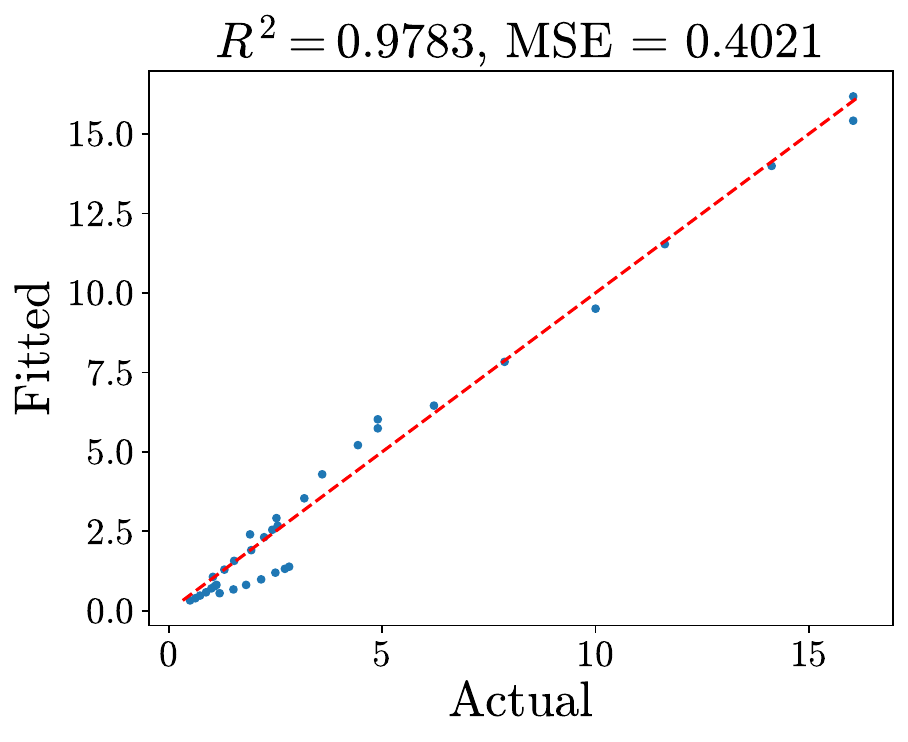}
\end{subfigure}
\vspace{-3.5em}
\caption{\textbf{MXINT-4 Layerwise Qwen-3 Weak Law.} The left two are for $\delta^{\text{opt}}$ and The right two are for $\delta_\mu$}.
  \label{fig:appendix-fitting-MXINT4 Layerwise Qwen3 Weak Law}
\end{figure}

\begin{figure}[H]
\centering
\begin{subfigure}[t]{0.22\textwidth}
\centering
\begin{tabular}{@{}l@{}}
\begin{minipage}[t][4cm][t]{\linewidth}
\centering
\begin{tabular}{ll}
\toprule
$C $ &  0.0412 \\
$A $ &  6.0802 \\
$\gamma_N $ &  0.9921 \\
\bottomrule
\end{tabular}
\end{minipage}
\end{tabular}
\end{subfigure}
\hfill
\begin{subfigure}[t]{0.22\textwidth}
\centering
\includegraphics[width=\textwidth]{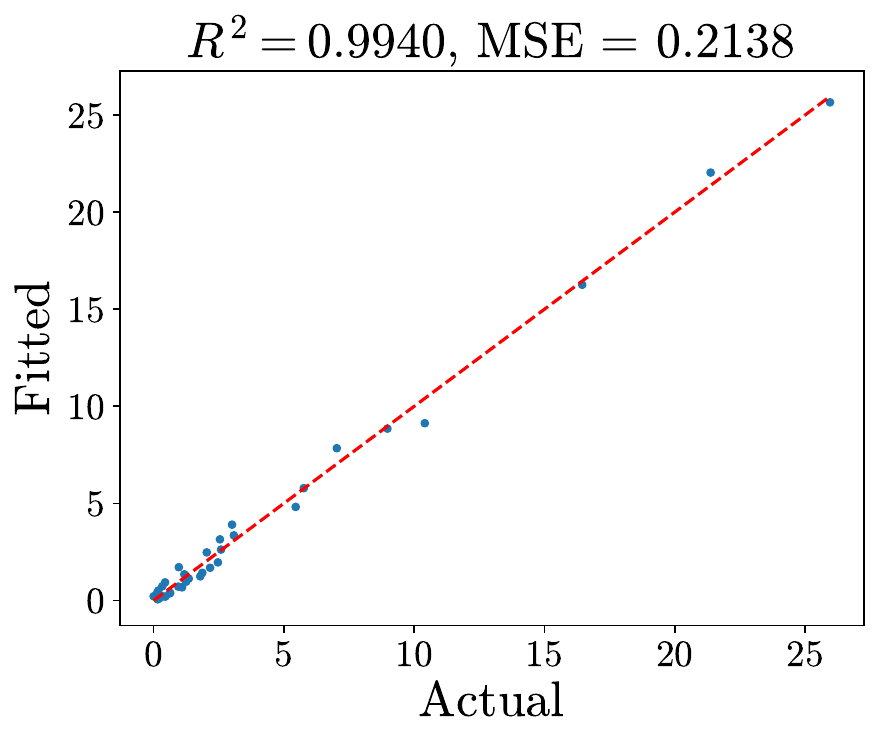}
\end{subfigure}
\hfill
\begin{subfigure}[t]{0.22\textwidth}
\centering
\begin{tabular}{@{}l@{}}
\begin{minipage}[t][4cm][t]{\linewidth}
\centering
\begin{tabular}{ll}
\toprule
$C $ &  0.3569 \\
$A $ &  5.0608 \\
$\gamma_N $ &  1.9352 \\
\bottomrule
\end{tabular}
\end{minipage}
\end{tabular}
\end{subfigure}
\hfill
\begin{subfigure}[t]{0.22\textwidth}
\centering
\includegraphics[width=\textwidth]{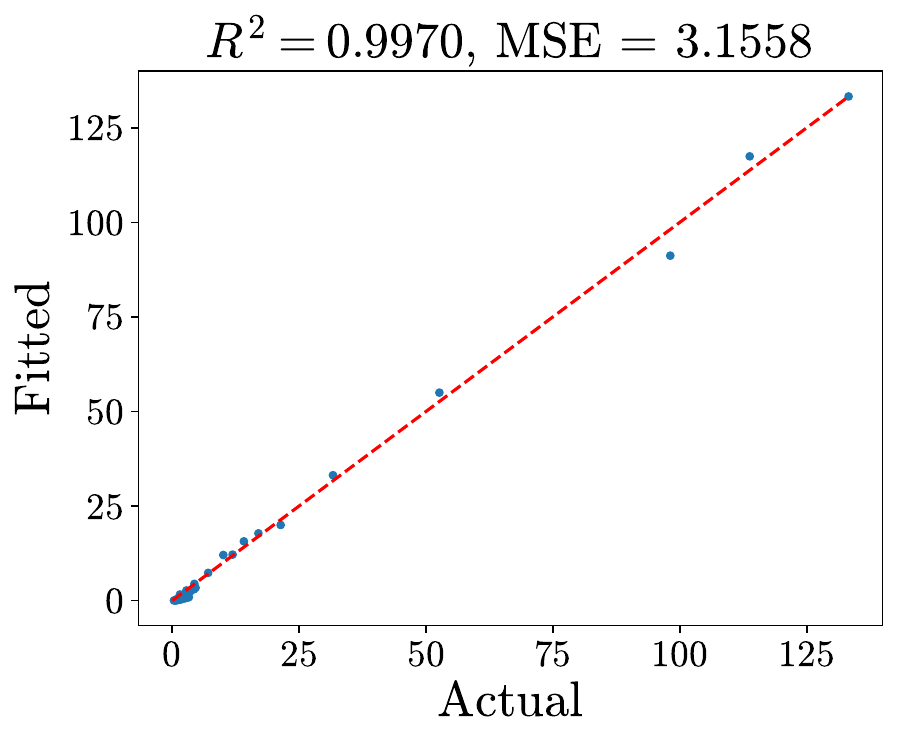}
\end{subfigure}
\vspace{-3.5em}
\caption{\textbf{MXINT-4 Matrix Multiplication-wise Qwen-3 Weak Law.} The left two are for $\delta^{\text{opt}}$ and The right two are for $\delta_\mu$}.
  \label{fig:appendix-fitting-MXINT4 Matrix Multiplication-wise Qwen3 Weak Law}
\end{figure}

\begin{figure}[H]
\centering
\begin{subfigure}[t]{0.22\textwidth}
\centering
\begin{tabular}{@{}l@{}}
\begin{minipage}[t][4cm][t]{\linewidth}
\centering
\begin{tabular}{ll}
\toprule
$C $ &  0.0089 \\
$A $ &  4.4940 \\
$\gamma_N $ &  0.6818 \\
\bottomrule
\end{tabular}
\end{minipage}
\end{tabular}
\end{subfigure}
\hfill
\begin{subfigure}[t]{0.22\textwidth}
\centering
\includegraphics[width=\textwidth]{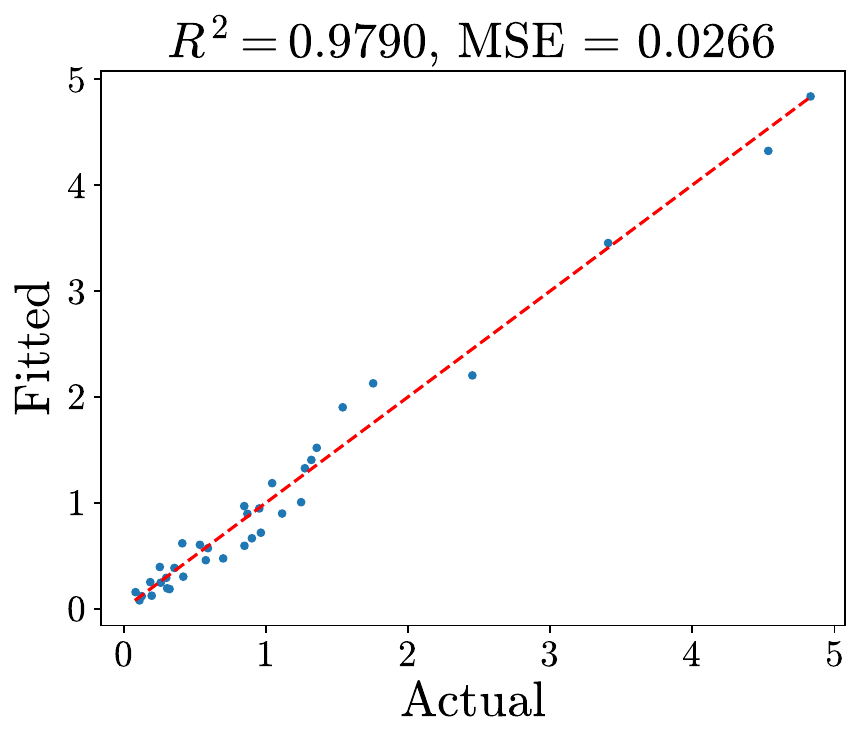}
\end{subfigure}
\hfill
\begin{subfigure}[t]{0.22\textwidth}
\centering
\begin{tabular}{@{}l@{}}
\begin{minipage}[t][4cm][t]{\linewidth}
\centering
\begin{tabular}{ll}
\toprule
$C $ &  0.0749 \\
$A $ &  2.6608 \\
$\gamma_N $ &  0.6197 \\
\bottomrule
\end{tabular}
\end{minipage}
\end{tabular}
\end{subfigure}
\hfill
\begin{subfigure}[t]{0.22\textwidth}
\centering
\includegraphics[width=\textwidth]{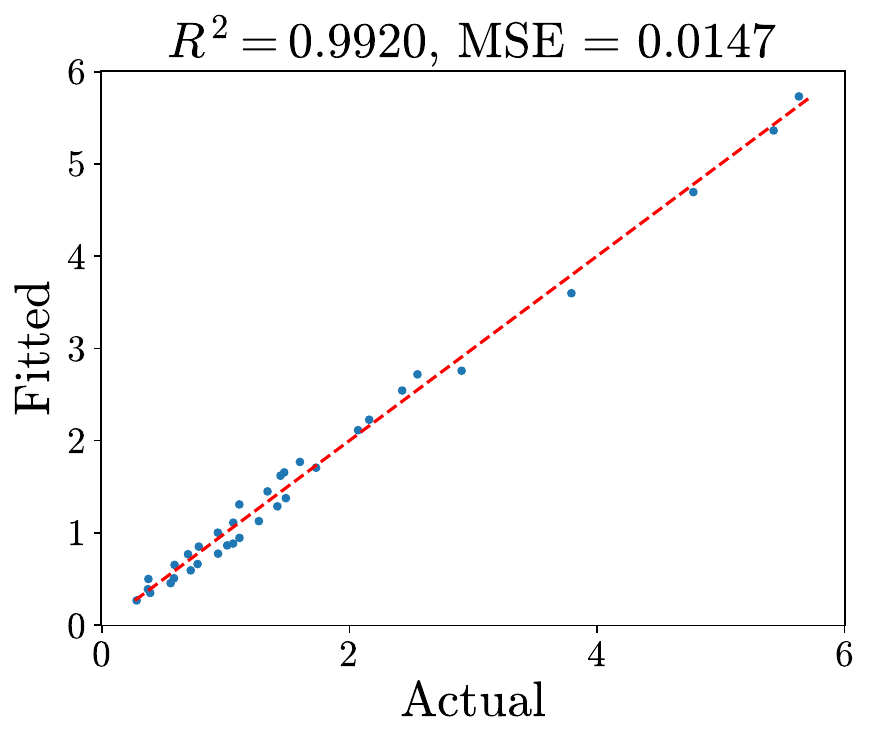}
\end{subfigure}
\vspace{-3.5em}
\caption{\textbf{Weight-only MXINT-4 CLM Weak Law. }The left two are for $\delta^{\text{opt}}$ and The right two are for $\delta_\mu$}.
  \label{fig:appendix-fitting-Weight-only MXINT4 CLM Weak Law}
\end{figure}

\begin{figure}[H]
\centering
\begin{subfigure}[t]{0.22\textwidth}
\centering
\begin{tabular}{@{}l@{}}
\begin{minipage}[t][4cm][t]{\linewidth}
\centering
\begin{tabular}{ll}
\toprule
$C $ &  0.0032 \\
$A $ &  7.1464 \\
$\gamma_N $ &  0.9663 \\
\bottomrule
\end{tabular}
\end{minipage}
\end{tabular}
\end{subfigure}
\hfill
\begin{subfigure}[t]{0.22\textwidth}
\centering
\includegraphics[width=\textwidth]{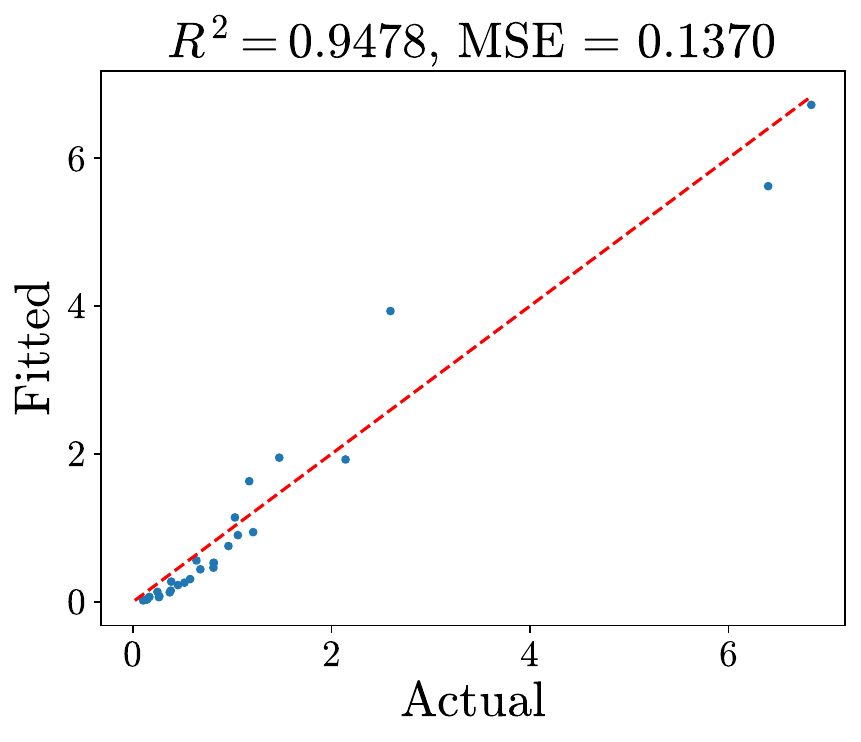}
\end{subfigure}
\hfill
\begin{subfigure}[t]{0.22\textwidth}
\centering
\begin{tabular}{@{}l@{}}
\begin{minipage}[t][4cm][t]{\linewidth}
\centering
\begin{tabular}{ll}
\toprule
$C $ &  0.1179 \\
$A $ &  3.8872 \\
$\gamma_N $ &  1.3831 \\
\bottomrule
\end{tabular}
\end{minipage}
\end{tabular}
\end{subfigure}
\hfill
\begin{subfigure}[t]{0.22\textwidth}
\centering
\includegraphics[width=\textwidth]{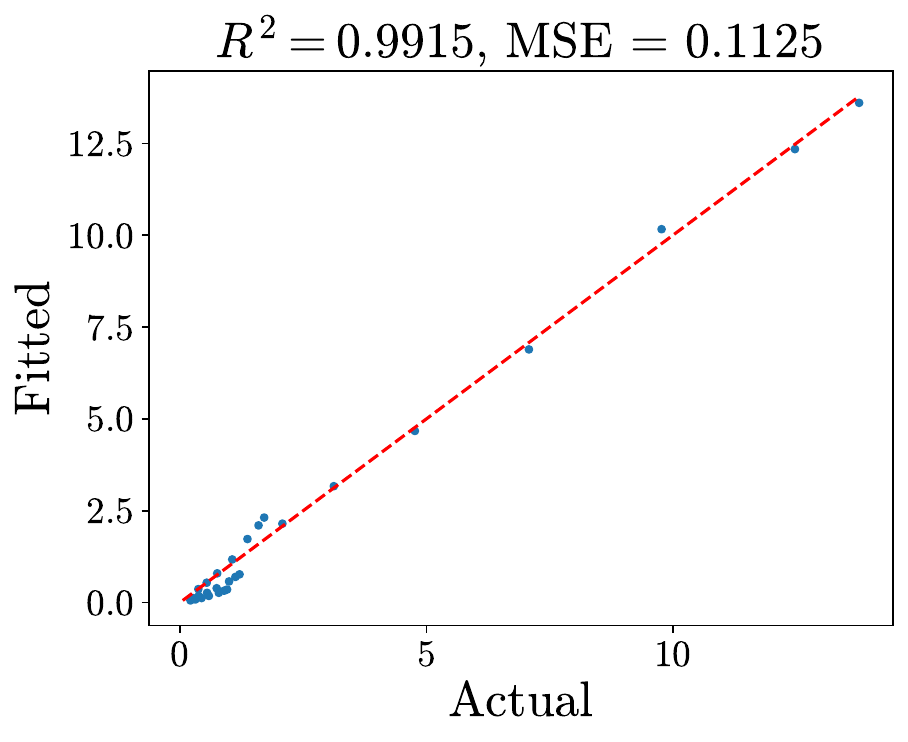}
\end{subfigure}
\vspace{-3.5em}
\caption{\textbf{Weight-only MXINT-4 Qwen-1.5 Weak Law. }The left two are for $\delta^{\text{opt}}$ and The right two are for $\delta_\mu$}.
  \label{fig:appendix-fitting-Weight-only MXINT4 Qwen1.5 Weak Law}
\end{figure}

\begin{figure}[H]
\centering
\begin{subfigure}[t]{0.22\textwidth}
\centering
\begin{tabular}{@{}l@{}}
\begin{minipage}[t][4cm][t]{\linewidth}
\centering
\begin{tabular}{ll}
\toprule
$C $ &  0.0188 \\
$A $ &  5.9150 \\
$\gamma_N $ &  1.1251 \\
\bottomrule
\end{tabular}
\end{minipage}
\end{tabular}
\end{subfigure}
\hfill
\begin{subfigure}[t]{0.22\textwidth}
\centering
\includegraphics[width=\textwidth]{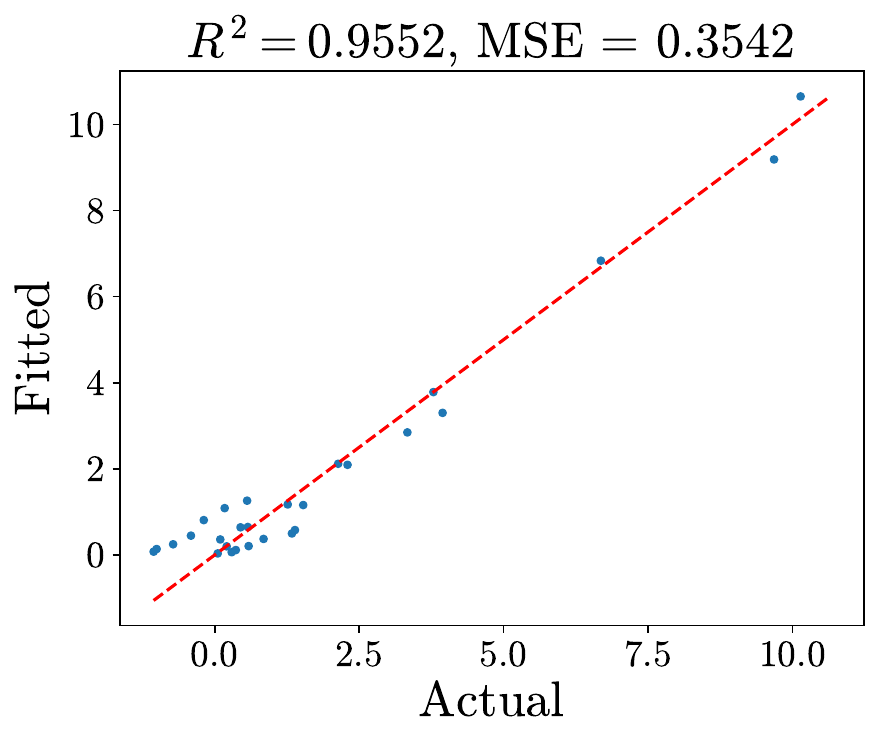}
\end{subfigure}
\hfill
\begin{subfigure}[t]{0.22\textwidth}
\centering
\begin{tabular}{@{}l@{}}
\begin{minipage}[t][4cm][t]{\linewidth}
\centering
\begin{tabular}{ll}
\toprule
$C $ &  0.3990 \\
$A $ &  4.0875 \\
$\gamma_N $ &  1.9460 \\
\bottomrule
\end{tabular}
\end{minipage}
\end{tabular}
\end{subfigure}
\hfill
\begin{subfigure}[t]{0.22\textwidth}
\centering
\includegraphics[width=\textwidth]{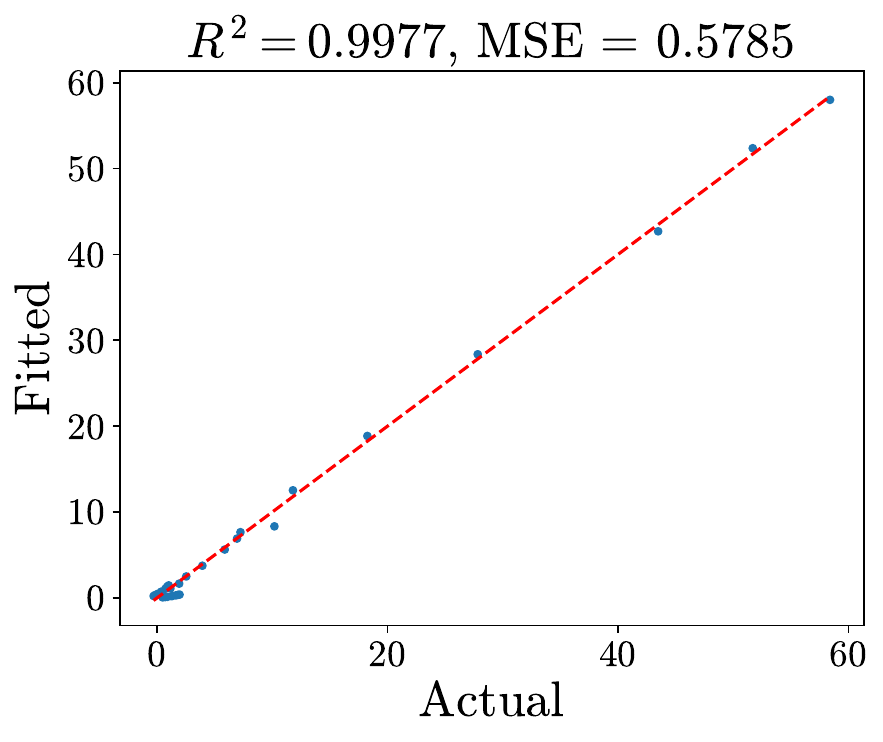}
\end{subfigure}
\vspace{-3.5em}
\caption{\textbf{Weight-only MXINT-4 Qwen-3 Weak Law.} The left two are for $\delta^{\text{opt}}$ and The right two are for $\delta_\mu$}.
  \label{fig:appendix-fitting-Weight-only MXINT4 Qwen3 Weak Law}
\end{figure}

We report all fitted coefficients and $R^2$ plots in Figures~\ref{fig:appendix-fitting-CLM Strong Law} to~\ref{fig:appendix-fitting-Weight-only MXINT4 Qwen3 Weak Law}. These include block-wise strong law fittings, as well as weak law fittings under layerwise, matrix-multiplication-wise, HQQ, MXINT2, and MXINT4 weight-only quantization. Each table reports the coefficients described in Section~\ref{sec:laws} from our scaling law equations, and the scatter plots compare predicted and actual PTQ losses to evaluate the fitting quality.

\paragraph{Overall Trends and Implications.}
Across all quantization levels and model families, the $R^2$ plots validate the predictive power of our scaling laws. Tight clustering along the diagonal confirms that the fitted models not only approximate empirical losses well but also generalize across compression levels and model sizes. Exceptions, including negative $\gamma_N$ or inflated constants, indicate model-instability or ill-posed settings such as distillation pre-training or over-compression.

\paragraph{Block-wise Strong Law Fitting}
Figures~\ref{fig:appendix-fitting-CLM Strong Law}–\ref{fig:appendix-fitting-Qwen3 Strong Law} display the strong law fits across model families. For CLM and Qwen1.5, we observe high-quality fits with consistent scaling exponents. However, Qwen3 shows greater variability, with large $\gamma_c$ values (e.g., $1.67$), indicating strong sensitivity to the quantization ratio $Q_r$, likely caused by its distillation-only training pipeline.

\paragraph{Layerwise and Matrix Multiplication-wise MXINT-4 Fitting}
Figures~\ref{fig:appendix-fitting-MXINT4 Layerwise CLM Weak Law}–\ref{fig:appendix-fitting-MXINT4 Matrix Multiplication-wise Qwen3 Weak Law} show the layerwise and matrix multiplication-wise MXINT4 quantization weak law results in strong, consistent fits across CLM and Qwen-1.5. 

\paragraph{HQQ Fitting}
Figures~\ref{fig:appendix-fitting-HQQ CLM Weak Law}–\ref{fig:appendix-fitting-HQQ Qwen3 Weak Law} demonstrate that HQQ quantization maintains consistently good fits across all model families. These weak law results further confirm the strength of HQQ in preserving model accuracy after quantization, and they validate that our scaling law generalizes even under high-performance quantizers.

\paragraph{MXINT-2 Fitting}
Figures~\ref{fig:appendix-fitting-MXINT2 CLM Weak Law}–\ref{fig:appendix-fitting-MXINT2 Qwen3 Weak Law} illustrate that MXINT2—our most aggressive quantization—yields degraded fits. This is especially evident for Qwen-1.5, where fitted constants reach extreme values (e.g., $C=1181.1$), and for Qwen-3, where $\gamma_N$ becomes negative ($-4.62$), indicating that MXINT-2 quantization is not ideal. Despite this, the $R^2$ plots still show approximate alignment, suggesting that the weak law scaling structure is partially preserved.

\paragraph{MXINT-4 Weight-only Fitting}
Figures~\ref{fig:appendix-fitting-Weight-only MXINT4 CLM Weak Law}-\ref{fig:appendix-fitting-Weight-only MXINT4 Qwen3 Weak Law} confirm that weight-only MXINT-4 quantization achieves a balanced tradeoff: it reduces parameter precision with less impact on performance compared to MXINT2. Fit quality remains strong with reasonable coefficients. These trends validate that restricting quantization to weights alone preserves much of the original loss landscape structure, and our weak law accurately tracks this behavior across CLM, Qwen-1.5, and Qwen-3.

\vspace{-3.1em}

\vspace{-3.1em}

\vspace{-3.1em}

\vspace{-3.1em}

\vspace{-3.1em}

\vspace{-3.1em}

\vspace{-3.1em}

\vspace{-3.1em}

\vspace{-3.1em}

\vspace{-3.1em}

\vspace{-3.1em}

\vspace{-3.1em}

\vspace{-3.1em}

\vspace{-3.1em}

\vspace{-3.1em}

\vspace{-3.1em}

\vspace{-3.1em}

\vspace{-3.1em}

\section{Further Explanation for our Estimation of $\mathbb{E}(\Delta)$ and $\min(\Delta)$}
\label{app:math-derivation}
As discussed in Section~\ref{sec:laws}, we are interested in two parameters of the discrete random variable $\Delta$: $\mathbb{E}(\Delta)$ and $\min(\Delta)$, which represent the mean and minimum (optimal) values of $\Delta$. In the weak law, they are denoted as $\delta^{\text{opt}}(N, Q_r)$ and $\mathbb{E}[\Delta(N, Q_r)]$. In the strong law, they are denoted as $\delta^{\text{opt}}(N, Q_r, Q_b)$ and $\mathbb{E}[\Delta(N, Q_r, Q_b)]$. 

We estimate those two parameters with the following two estimator. Given a list of i.i.d. discrete random variables $\{\Delta_i\}_{i=1}^n$. We denote the unbiased estimator of $\mathbb{E}(\Delta)$ as 
\begin{equation}
    E_n = \frac{\sum_{i=1}^n \Delta_i}{n}\,,
\end{equation}
and the biased estimator of $\min(\Delta)$ as
\begin{equation}
    M_n = \min(\{\Delta_i: i=1,\cdots, n\})\,.
\end{equation}
$E_n$ is an unbiased estimator as 
\begin{equation}
\begin{split}
    \mathbb{E}(E_n) - \mathbb{E}(\Delta) 
    &= \frac{\sum_{i=1}^n\mathbb{E}(\Delta_i)}{n} - \mathbb{E}(\Delta) \\
    &= \mathbb{E}(\Delta) - \mathbb{E}(\Delta) \\
    &= 0\,.
\end{split}
\end{equation}
On the other hand, $M_n$ is a biased estimator. Without loss of generality, we will calculate the bias of $M_n^\prime = \max(\{\Delta_i: i=1,\cdots, n\})$ with respect to $\max(\Delta)$. Since $\{\Delta_i\}_{i=1}^n$ are i.i.d, 
\begin{equation}
\begin{split}
    \Pr(M_n^\prime < x) 
    &= \prod_{i=1}^n \Pr(\Delta_i < x)\\
    &= F^{n}(x^-)\,,
\end{split}
\end{equation}
where $F$ is the cdf of $\Delta$ and $x^- = \max(\{x^\prime: x^\prime < x \land \Pr(\Delta=x^\prime)\neq 0\} \cup -\infty)$. Hence the expectation of $M_n^\prime$ is 
\begin{equation}
\begin{split}
    \mathbb{E}(M_n^\prime) 
    &= \sum_x x \Pr(M_n^\prime = x)\\
    &= \sum_x x \Pr(M_n^\prime \leq x \land M_n^\prime \nless x)\\
    &= \sum_x x (F^{n}(x) - F^{n}(x^-))\,.
\end{split}
\end{equation}
As 
\begin{equation}
    \sum_x (F^{n}(x) - F^{n}(x^-)) = F^{n}(\max(\Delta)) - F^{n}(-\infty) = 1 - 0 = 1\,, 
\end{equation}
$\mathbb{E}(M_n^\prime)$ is a re-weighting of different $x$-s and the weight for $\max(\Delta)$ is not $1$, hence $\mathbb{E}(M_n^\prime) < \max(\Delta)$. However, as $n\to \infty$, $(F^{n}(x) - F^{n}(x^-)) \to 0$ for all $x\neq\max(\Delta)$, so 
\begin{equation}
    \mathbb{E}(M_n^\prime) \xrightarrow{d} \max(\Delta) \text{ as } n\to\infty\,,
\end{equation}
making $\{M_n^\prime\}_{i=1}^\infty$ asymptotically unbiased. Recall notations from Section~\ref{sec:laws}, given $n$ realizations $\{\delta_i\}_{i=1}^n$ of the random variables $\{\Delta_i\}_{i=1}^n$, 
\begin{equation}
    \mu_{\delta} = \frac{\sum_{i=1}^n\delta_i}{n}
\end{equation}
and
\begin{equation}
    \delta_{\min} = \min(\{\delta_i: i=1,\cdots, n\})
\end{equation}
are our final estimation of the mean and minimum of $\Delta$. Our weak and strong law are designed to fit those estimations. To be specific, each fitting of the weak or the strong law tries to fit separately on $\mu_{\delta}$ and $\delta_{\min}$, which is also why each of our laws has two separate parts for expectation and minimum values.

\section{Additional Related Work}
\label{app:additional-related-work}
In addition to the backgrounds presented in~\Cref{sec:background}, we further discuss the related work on mixed precision. Specifically, we categorize the work under the following three topics: mixed-precision quantization, mixed-precision inference, and mixed-precision training.

\paragraph{Mixed-precision quantization}
Mixed precision approaches involve partitioning a model's parameters into both high-precision and low-precision components, which have been shown to better preserve model performance relative to uniform quantization. This is primarily seen in models that exhibit different sensitivities to quantization at various layers. Some mixed-precision LLM quantization work adopts the concept of weight salience to guide the search for fine-grained bit allocation. The first-order \citep{li2023llm} or second-order weight gradient \citep{huang2024slim} has been used to form such salience metrics, such that salient layers are left in higher precision while the rest are cast to low precision. There are also works performing the search in an end-to-end style with the quantized model performance as the objective, such as the accuracy on a downstream task \citep{zhang2023revisiting}. In both cases, mixed precision can be seen as a promising approach to provide a lossless reduction in LLM memory requirements, reducing average bit widths below levels achievable through uniform quantization.

\paragraph{Mixed-precision inference}
Mixed-precision inference methods targeting GPUs usually adopt regular mixed-precision strategies and computation patterns; The authors of GPTQ3.int8() \citep{dettmers2022gpt3} decompose the matrix multiplication in every linear layer into two sub-matrix-multiplications based on the activation magnitudes, achieving a 2-3$\times$ inference speedup by casting the low-magnitude sub-matrix to low precision. SpQR \citep{dettmers2023spqr} represents a weight matrix with grouped 3-bit integers and less than 1\% sensitive weight elements with FP16 values, achieving a 2$\times$ speedup compared to a quantized and sparse PyTorch baseline. These approaches enable reducing model size, but additional careful treatment is needed to improve inference throughput. For example, in \citep{li2023llm}, mixed precision LLM quantization at 2-bit and 3-bit showed no speedup compared to 4-bit, due to less efficient utilization of memory bandwidth. On the other hand, Any-Precision LLM \citep{park2024any} achieves throughput scaling at various precisions by providing CUDA kernels with a novel weight packing approach following a bitplane layout, achieving 1.3-1.8$\times$ speedup on mobile and edge devices. Additionally, works such as FlightLLM achieve high throughput by leveraging custom hardware designs \citep{zeng2024flightllm}.

\paragraph{Mixed-precision training}
Mixed-precision quantization has also been adopted in training to reduce the large memory footprint of gradient descent, which requires the storage of optimizer states and gradients in addition to forward activations. It has been shown that the training process can tolerate aggressive quantization and correct quantization noise in some components. \citep{micikevicius2017mixed} is the pioneering work proposing the storage of all weights, activations, and gradients in FP16, while updating a copy of weights in FP32. This work also proposed scaling up the forward pass loss and unscaling the gradient before the weight update to avoid under-utilization of the FP16 representable range, leading to half the memory requirement and speed-ups of 2-6$\times$ relative to FP32 training. Recently, more aggressive quantization has been studied for mixed-precision training. \citep{mellempudi2019mixed} trains models with E5M2 FP8, maintaining a master copy of the weights in FP16, and dynamically adjusting the scaling factor every few iterations. Hybrid-FP8~\citep{sun2019hybrid} improves FP8 training by using E4M3 for forward propagation and E5M2 for backward propagation, leading to matching performance to models trained with FP32. Popular implementations of FP8 mixed-precision training like TransformerEngine
\footnote{
    TransformerEngine: \url{https://github.com/NVIDIA/TransformerEngine}.}
have achieved a training acceleration of around 3-4$\times$ compared to FP16 mixed-precision training.

\end{document}